\documentclass[11pt]{article}

\usepackage[final]{acl}

\usepackage{times}
\usepackage{latexsym}

\usepackage[T1]{fontenc}

\usepackage[utf8]{inputenc}

\usepackage{microtype}

\usepackage{inconsolata}

\usepackage{graphicx}

\usepackage{titletoc}

\usepackage{xcolor}
\usepackage{colortbl}

\usepackage{booktabs}
\usepackage{multirow}
\usepackage{subcaption}
\usepackage{wrapfig}
\usepackage[most]{tcolorbox}
\usepackage{array}

\usepackage{algorithm}
\usepackage{algpseudocode}
\usepackage{amsmath}
\usepackage{amssymb}
\usepackage{fontawesome5}

\title{ReTrace: Visual Backtracking for Efficient Multimodal Large Language Models} 

\author{
\textbf{Guangzong Si}\textsuperscript{1,2,*},
\textbf{Dong Wang}\textsuperscript{3,*},
\textbf{Yifan Yu}\textsuperscript{4},
\\
\textbf{Miao Liu}\textsuperscript{3},
\textbf{Panwang Pan}\textsuperscript{5},
\textbf{Wentao Zhu}\textsuperscript{2,$\dagger$}
\\[3pt]
\textsuperscript{1}University of Science and Technology of China,
\textsuperscript{2}Eastern Institute of Technology, Ningbo,
\\
\textsuperscript{3}Tsinghua University,
\textsuperscript{4}ETH Zurich,
\textsuperscript{5}Xiamen University
\\[2pt]
{\small
\texttt{guangzongsi@mail.ustc.edu.cn}
\quad
\texttt{wangd16@tsinghua.org.cn}
\quad
\texttt{wtzhu@eitech.edu.cn}
}
\\[2pt]
{\small\textit{
* Equal contribution
\qquad
$\dagger$ Corresponding author
\qquad
\href{https://github.com/xufeiqiong/Retrace/tree/main}{\faGithub\ Project}
}}
}

\begin{document}
\maketitle

\begin{abstract}
Multimodal Large Language Models (MLLMs) encode images into dense visual token sequences, creating substantial attention and KV-cache overhead. Existing visual token compression methods reduce this cost by pruning low-saliency tokens in shallow layers, but this practice couples temporary computational deactivation with irreversible information deletion. We show that this coupling breaks an intrinsic inference pattern in MLLMs. Through layer-wise attention analysis, we identify \textbf{Visual Backtracking}: visual pivots remain spatially stable, yet in middle layers they actively retrieve context from tokens that appeared unimportant earlier. Thus, early low-saliency tokens can be inactive for computation while still being useful for later reasoning. Motivated by this finding, we propose \textbf{ReTrace}, a backtracking-aware compression framework that introduces a dormant token state. ReTrace caches non-pivot tokens outside expensive layer-wise computation, restores them during the backtracking window, and prunes only after their information has been absorbed. Experiments on Qwen2.5-VL and InternVL3 demonstrate nearly 90\% reductions in FLOPs and KV cache, a 2.83$\times$ prefilling speedup, and strong performance retention across complex vision-language benchmarks.

\end{abstract}

\section{Introduction}
\label{sec:intro}

\begin{figure}[h]
    \centering
    \includegraphics[width=\linewidth]{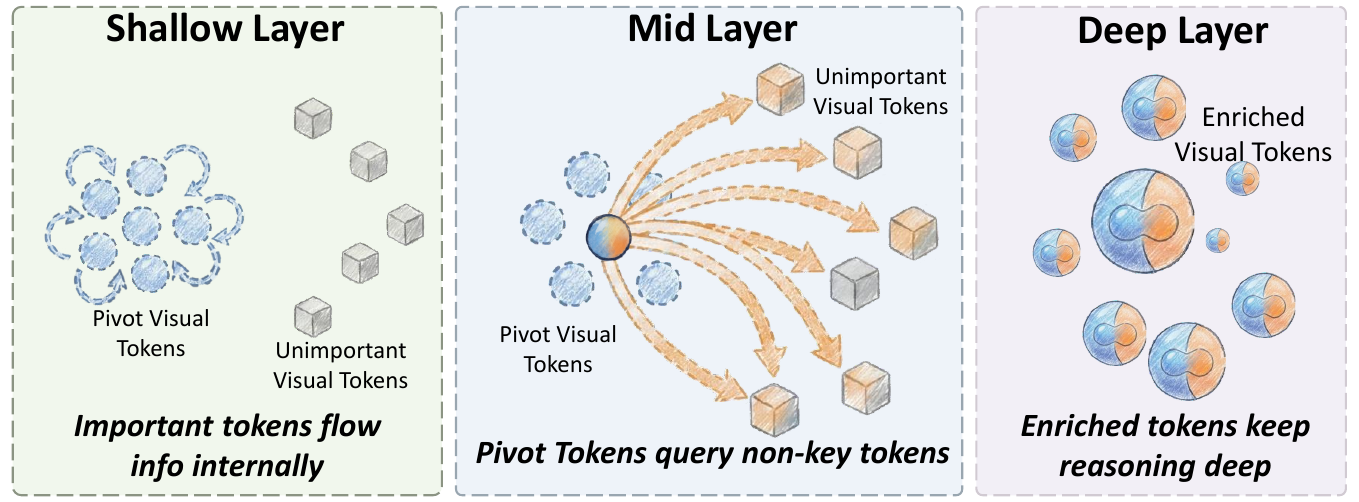}
    \caption{\textbf{Conceptual paradigm of the Visual Backtracking phenomenon.} In shallow layers, pivot tokens interact internally to build clear feature representations. In mid layers, a ``backtracking'' mechanism emerges, where pivot tokens strongly query unimportant visual tokens to absorb global context. Finally, in deep layers, forms \textit{Enriched Visual Tokens} essential for complex reasoning.}
    \label{fig:insight}
\end{figure}

Multimodal LLMs (MLLMs)~\cite{bai2025qwen2.5,internvl3,liu2023improvedllava,liu2023llava} encode visual inputs into dense token sequences. This enables fine-grained cross-modal reasoning~\cite{internvl25,internvl35,Qwen-VL,Qwen2-VL,vl1,vl3,vl5}, but the sequences can reach thousands of tokens per image as resolution and context grow~\cite{clip,dino,QwenTeam2025Qwen3VL,vit}. Processing all tokens is expensive, leading to quadratic attention cost and large KV cache~\cite{transformer,flashattention,flashattention2,vllm}.

A common solution is visual token pruning: tokens deemed unimportant in early layers are permanently removed or merged~\cite{fastv,fastvlm,visionzip,holov,dart,vlmpruner,vscan,himap,llava-prumerge,pyra,tome,plphp,tokencarve}. This approach assumes that early low-saliency tokens are unimportant for the rest of inference. In practice, this assumption can fail: early pruning degrades performance on complex reasoning tasks, even when the retained tokens seem sufficient to reconstruct the visual content. This raises an interesting question:

\begin{tcolorbox}[
    colback=cyan!3,
    colframe=cyan!45!blue,
    coltitle=black,
    colbacktitle=cyan!10,
    title=\textbf{\textit{Question}},
    boxrule=0.5pt,
    arc=2mm,
    width=\linewidth,
    left=2mm,
    right=2mm,
    top=1.2mm,
    bottom=1.2mm
]
\textit{Can a token be temporarily unnecessary for computation but still useful for later reasoning?}
\end{tcolorbox}

To answer this, we analyze not only which tokens are attended to, but also how attention flows across layers. We observe a three-phase pattern. In shallow layers, task-relevant visual anchors interact mainly among themselves. In middle layers, these anchors shift attention outward toward tokens that were previously ignored. They actively retrieve context needed for global reasoning. In deep layers, attention reconcentrates on enriched anchors for task-specific decisions. We call this middle-layer outward retrieval \textbf{Visual Backtracking}, as shown in Fig.~\ref{fig:insight}. It explains why early pruning fails: low-saliency tokens are not globally unimportant. They are phase-specific context sources accessed during a retrieval window that early pruning closes before it opens.

This finding reframes visual token compression. Instead of one-shot pruning, we treat it as \emph{staged information scheduling}. Non-anchor tokens do not participate in all layers, but they should not be permanently deleted before the backtracking window. Based on this, we propose \textbf{ReTrace}. ReTrace builds a pivot buffer as surrogate receivers, caches non-pivot tokens as dormant context bypassing computation, and restores this context at the backtracking layer to approximate the missing retrieval flow. Task-driven pruning is applied only after absorption completes, deferring permanent information loss.

Our contributions are:
\begin{itemize}
    \item We identify and quantify \textbf{Visual Backtracking}, showing that low-saliency tokens serve as phase-specific context sources during middle-layer inference. This provides a mechanistic explanation for why early pruning hurts performance.
    \item We formulate visual token compression as \textbf{staged information scheduling}, where tokens are cached, restored, and only later permanently pruned according to layer-wise demand.
    \item We implement this principle in \textbf{ReTrace} and show on Qwen2.5-VL and InternVL3 that it reduces FLOPs and KV cache by nearly 90\%, achieves a 2.83$\times$ prefilling speedup, and maintains strong performance across complex vision-language benchmarks.
\end{itemize}

\section{Related Work}
\label{sec:related_appendix}
\textbf{MLLMs.} MLLMs use Vision Transformers \cite{vit} to encode images or videos into token sequences, which are then projected into the LLM's semantic space \cite{bai2025qwen2.5,internvl3,liu2023llava}. Recent models like Qwen2.5-VL \cite{Qwen-VL,QwenTeam2025Qwen3VL,Qwen2-VL,bai2025qwen2.5} and InternVL3 \cite{internvl25,internvl3,internvl35} process high-resolution inputs to achieve fine-grained perception, resulting in long visual sequences. While this approach supports reasoning performance, the quadratic complexity of the self-attention mechanism increases computational overhead. Therefore, reducing visual context length without sacrificing performance is an important problem for MLLMs deployment.

\noindent \textbf{Visual token pruning.}
Visual token pruning \cite{fastv,fastvlm,visionzip,holov,dart,vlmpruner,vscan,himap} accelerates MLLM inference by reducing visual redundancy. Existing methods generally fall into vision-centric \cite{visionzip,dart,holov,fastvlm,vlmpruner} and text-centric \cite{fastv,himap,pyramiddrop2025} approaches. Vision-centric methods filter or merge tokens before the LLM layers \cite{visionzip}, while text-centric methods prune tokens in deeper layers based on cross-modal attention \cite{fastv}. Recent works also explore hybrid strategies that combine early visual reduction with late-stage text-guided pruning \cite{vscan}. G-Search \cite{zhao2025accelerating} further searches for layer-wise token retention schedules by exploiting the stability of token-importance rankings across layers. Our prior work, HiMAP \cite{himap}, exploits the layer-wise shift between visual-textual and intra-visual information flows to guide hierarchical pruning. However, these methods ultimately make irreversible pruning decisions once visual tokens are removed, preventing their information from being accessed again in later layers. In contrast, ReTrace reveals that early low-saliency tokens can still be revisited by visual pivots in middle layers. Rather than permanently discarding them, we cache these tokens as dormant context, restore their information when needed, and apply task-guided pruning only after context absorption.

\section{Motivation}
\label{sec:mov}
\begin{figure*}[t]
    \centering
    \includegraphics[width=\linewidth]{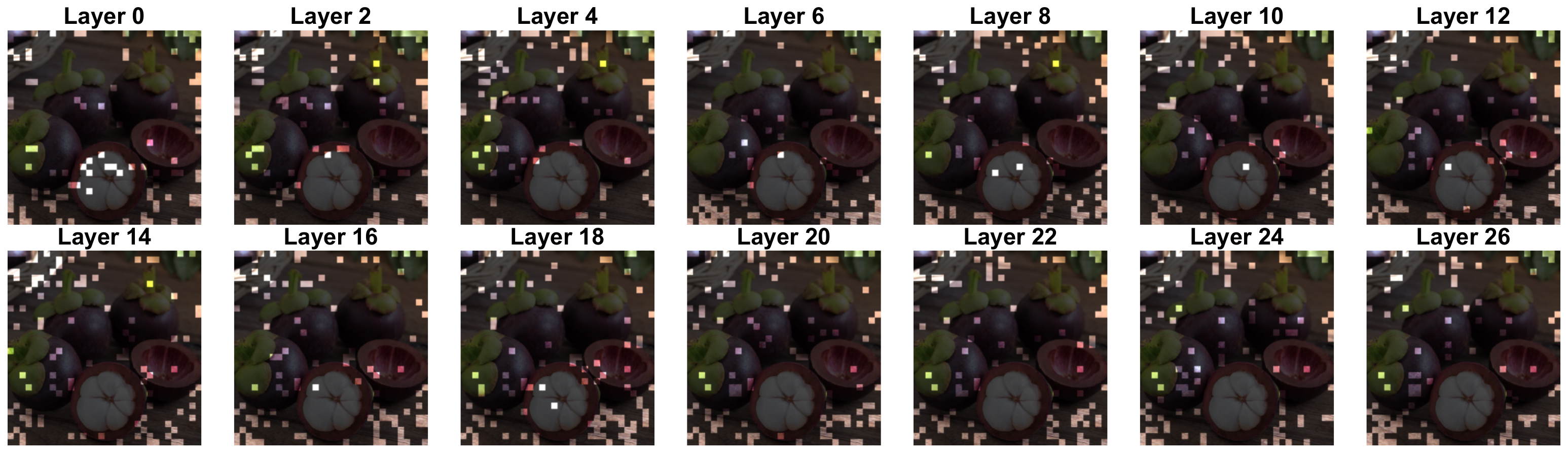}
\caption{\textbf{Visualization of pivot tokens across different layers.} We highlight the top 10\% visual tokens with the highest saliency scores. The selected regions remain highly consistent from shallow to deep layers, intuitively demonstrating the spatial stability of the model's visual attention.}
    \label{fig:vis_sample}
\end{figure*}

Early visual token pruning reduces computation by removing low-saliency tokens from the active sequence. Yet this operation makes two decisions at once. It not only stops these tokens from participating in expensive layer-wise computation, but also makes their information permanently inaccessible. This coupling is safe only if tokens ignored in shallow layers are never needed later.

We test this assumption by analyzing the layer-wise behavior of visual attention. Instead of only asking which tokens are important, we ask a more specific question: \textit{after the model forms visual anchors, where do these anchors look as inference proceeds?}

\begin{figure*}[t]
    \centering
    \begin{subfigure}{0.32\textwidth}
        \centering
        \includegraphics[width=\linewidth]{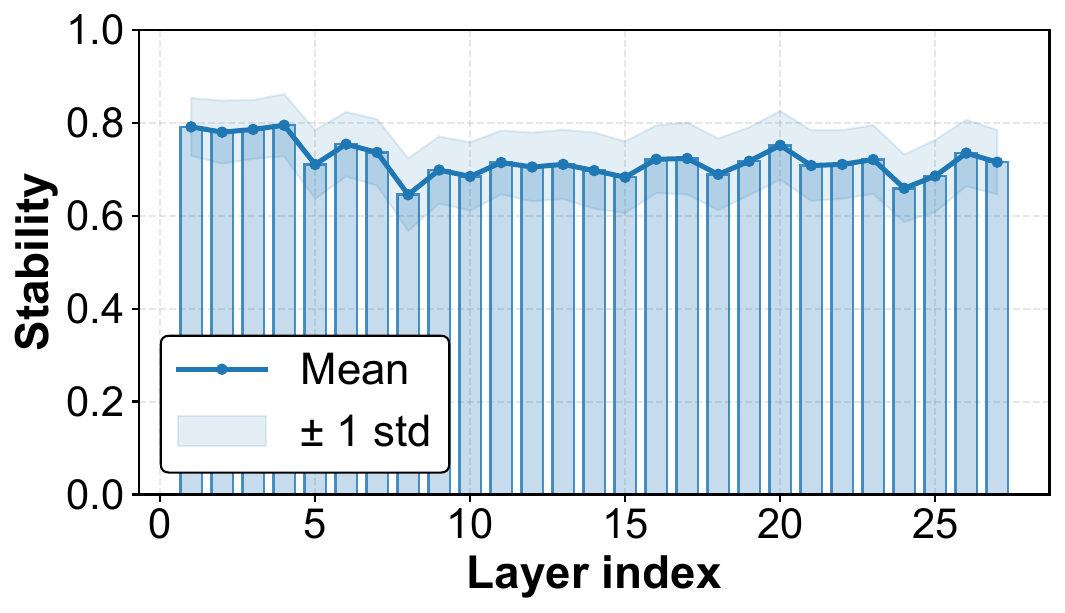}
        \caption{Cross-layer pivot stability.}
        \label{fig:retention_10}
    \end{subfigure}\hfill
    \begin{subfigure}{0.32\textwidth}
        \centering
        \includegraphics[width=\linewidth]{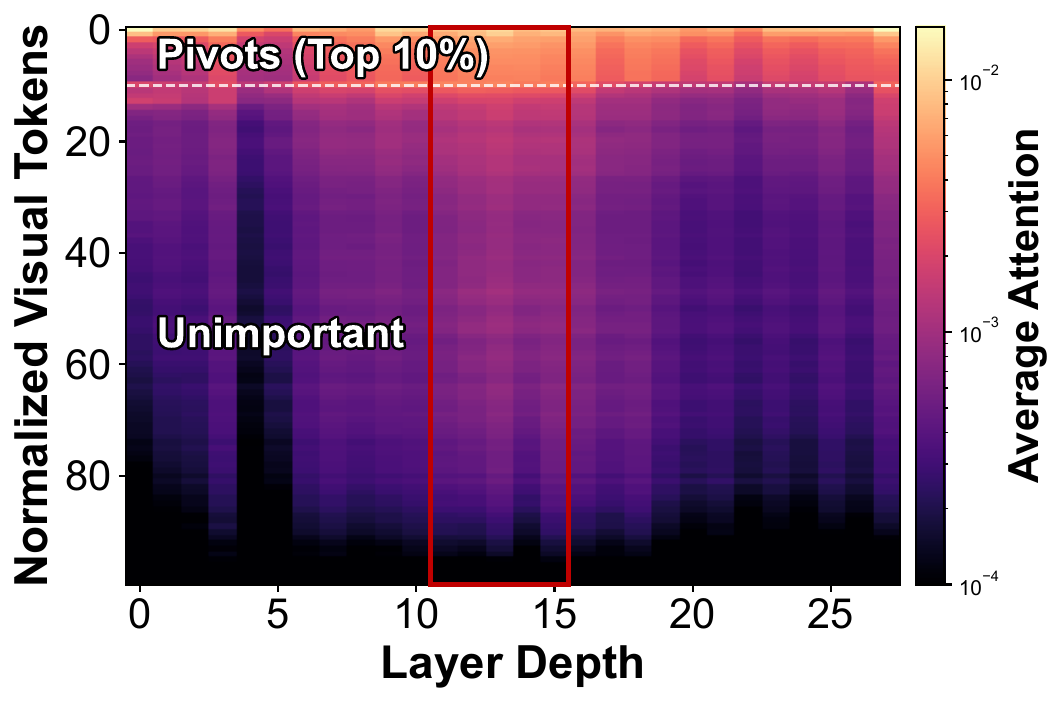}
        \caption{Pivot-to-visual attention heatmap.}
        \label{fig:waterfall_10}
    \end{subfigure}\hfill
    \begin{subfigure}{0.32\textwidth}
        \centering
        \includegraphics[width=\linewidth]{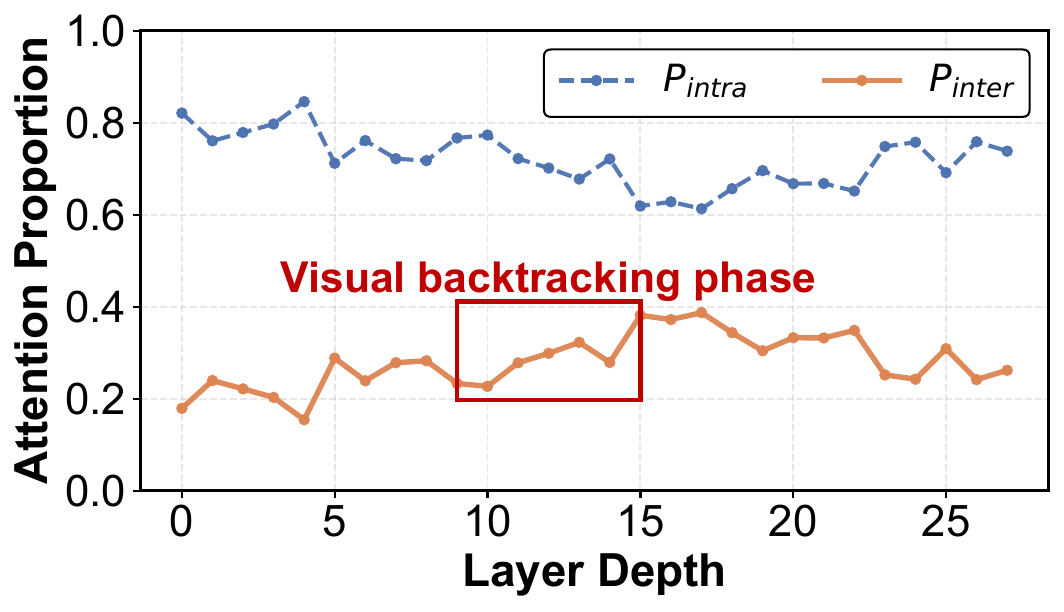}
        \caption{Intra-/inter-pivot attention flow.}
        \label{fig:flow_10}
    \end{subfigure}
\caption{\textbf{Quantitative evidence of visual backtracking.}
We analyze the layer-wise behavior of text-guided visual pivot tokens.
(a) The selected pivots remain spatially stable across layers, indicating that the model establishes persistent visual pivots early.
(b) Pivot-to-visual attention exhibits a distinct middle-layer expansion band, where pivots attend broadly to non-pivot visual tokens.
(c) The normalized attention flow shows that pivots shift from pivot-internal consolidation in shallow layers to pivot-to-context retrieval in middle layers, revealing the visual backtracking window.}
\label{fig:core_obs_quant}
\label{fig:core_obs_retention}
\end{figure*}
 
\noindent\textbf{Observation I: Visual pivots are spatially stable.}
We first examine whether the model forms stable visual anchors. If the identity of important visual tokens changes drastically across layers, then any structured analysis of attention flow would be unreliable.

Guided by textual instructions, we define the saliency score $s_l(v_i)$ of a visual token $v_i \in \mathcal{V}$ at layer $l$ by averaging the cross-attention weights it receives from all text tokens, where $\mathbf{C}_l$ denotes the text-to-visual attention matrix at layer $l$:
\begin{equation}
    s_l(v_i) = \frac{1}{|\mathcal{T}|} \sum_{t_{j} \in \mathcal{T}} \mathbf{C}_l(t_j, v_i)
\end{equation}
Based on this ranking, we select the top-$k$ tokens with the highest saliency as the pivot set $\mathcal{P}_l$, and treat the remaining tokens as the non-pivot set $\mathcal{U}_l$:
\begin{equation}
    \mathcal{P}_l = \text{Top-}k(\{s_l(v_i) \mid v_i \in \mathcal{V}\}), \quad \mathcal{U}_l = \mathcal{V} \setminus \mathcal{P}_l
\end{equation}
The cross-layer stability of the pivot set is defined as:
\begin{equation}
    \text{Stability}_l = \frac{|\mathcal{P}_l \cap \mathcal{P}_{l-1}|}{|\mathcal{P}_{l-1}|}
\end{equation}
 
As shown in Fig.~\ref{fig:vis_sample}, the top 10\% pivot tokens consistently focus on the main visual subjects from shallow to deep layers. Fig.~\ref{fig:retention_10} further quantifies this stability at a 10\% pivot ratio, showing that the selected pivots remain highly consistent across layers.

This result explains why early pruning can work to some extent: the model does form persistent visual anchors early. However, it also raises a sharper question. \textit{If the pivot tokens are so stable, why does removing non-pivot tokens still hurt complex reasoning?} The issue may not be who the pivots are, but what information they need later.

\noindent\textbf{Observation II: Stable pivots look back to non-pivot context.}
We next study the attention flow starting from pivot tokens. Within the visual self-attention mechanism, we partition the attention originating from pivot tokens $\mathcal{P}_l$ into two parts, where $\mathbf{A}_l$ denotes the visual self-attention matrix at layer $l$. The first part is the internal attention among pivots, denoted as $\mathcal{F}_{intra}^{(l)}$. The second part is the outward attention from pivots to non-pivot tokens $\mathcal{U}_l$, denoted as $\mathcal{F}_{inter}^{(l)}$:
\begin{equation}
\begin{aligned}
    \mathcal{F}_{intra}^{(l)} 
    &= \sum_{v_i \in \mathcal{P}_l} \sum_{v_j \in \mathcal{P}_l} 
    \mathbf{A}_l(v_i, v_j), \\
    \mathcal{F}_{inter}^{(l)} 
    &= \sum_{v_i \in \mathcal{P}_l} \sum_{v_k \in \mathcal{U}_l} 
    \mathbf{A}_l(v_i, v_k).
\end{aligned}
\end{equation}
We normalize these flows to obtain their relative proportions:
\begin{equation}
\begin{aligned}
    P_{intra}^{(l)}
    &= \frac{\mathcal{F}_{intra}^{(l)}}
    {\mathcal{F}_{intra}^{(l)} + \mathcal{F}_{inter}^{(l)}}, \\
    P_{inter}^{(l)}
    &= \frac{\mathcal{F}_{inter}^{(l)}}
    {\mathcal{F}_{intra}^{(l)} + \mathcal{F}_{inter}^{(l)}}.
\end{aligned}
\end{equation}
 
Fig.~\ref{fig:flow_10} shows a clear three-phase pattern. In shallow layers, $P_{intra}$ dominates. Pivot tokens mainly interact with each other, forming stable local visual anchors. In middle layers, $P_{inter}$ rises sharply. The pivots begin to attend to non-pivot tokens, including those that early saliency-based pruning would remove. This outward attention is also visible as the middle-layer expansion band in Fig.~\ref{fig:waterfall_10}. In deep layers, $P_{intra}$ increases again, suggesting that attention reconcentrates on enriched pivots for final task-specific reasoning.

We call this middle-layer outward retrieval \textbf{Visual Backtracking}. It shows that the identity of visual pivots can be stable while their information demand is not. Non-pivot tokens are not always useful enough to remain active across all layers, but they can still serve as context sources during a specific middle-layer window. In other words, a token can be computationally inactive in shallow layers without being informationally irrelevant.

This changes how we should view early pruning. Its main risk is not only that it may select the wrong tokens. It also deletes information before the model reaches the layer where that information would naturally be retrieved. Early pruning closes the retrieval route before it opens.

\noindent\textbf{Pilot Study: Restoration timing matters.}
\begin{figure}[t]
    \centering
    \includegraphics[width=\linewidth]{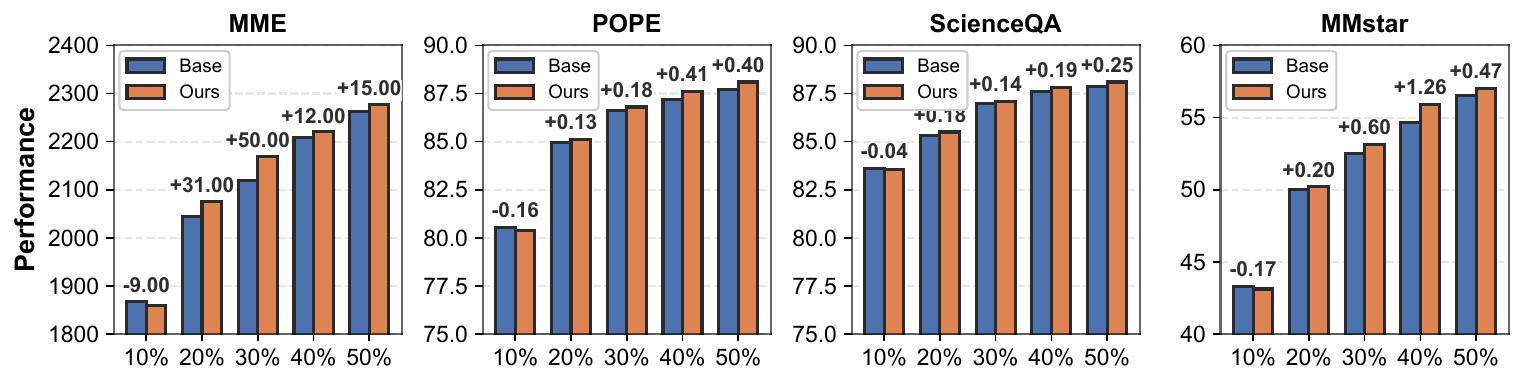}
    \caption{\textbf{Performance comparison of different feature integration paradigms.} Under identical retention ratios (10\%--50\%), backtracking-aware injection consistently outperforms rhythm-agnostic fusion at moderate ratios. At the extreme 10\% retention rate, injecting into an undersized buffer triggers \textit{Information Overload}, revealing the need for a multi-stage design.}
    \label{fig:mov_exper}
\end{figure}
 
If Visual Backtracking is important, then the timing of context restoration should matter. We test this with a controlled pilot study. We keep the same retention ratio and use the same non-pivot token features, but compare two ways of restoring them:
\begin{enumerate}
    \item \textbf{Rhythm-agnostic fusion (Base)}: Features of removed tokens are merged into retained pivots before any LLM processing begins.
    \item \textbf{Backtracking-aware injection (Ours)}: Pivot tokens first pass through shallow layers alone, and the cached context is injected only when the model enters the backtracking window.
\end{enumerate}
 
As shown in Fig.~\ref{fig:mov_exper}, backtracking-aware injection gives consistent gains at moderate retention ratios, such as +50 on MME at 30\% retention. Since both paradigms use the same token budget and the same non-pivot features, the gain does not come from keeping more information. It comes from restoring the information at the layer where the model naturally needs it.

The pilot study also shows a boundary case. At the extreme 10\% retention ratio, delayed injection can slightly underperform early fusion. When the pivot buffer is too small, too much cached context is forced into too few receivers. This can cause \textbf{information overload} instead of useful restoration.

This suggests that visual token compression should not prune directly to the final budget at the beginning. A better design should use an intermediate buffer large enough for context absorption, then prune to a smaller reasoning core after the backtracking window. This leads to the design of \textbf{ReTrace}: cache non-pivot tokens instead of deleting them, inject them when visual backtracking occurs, and prune only after their information has been absorbed.

\section{Method}

\begin{figure*}[t]
    \centering
    \includegraphics[width=\linewidth]{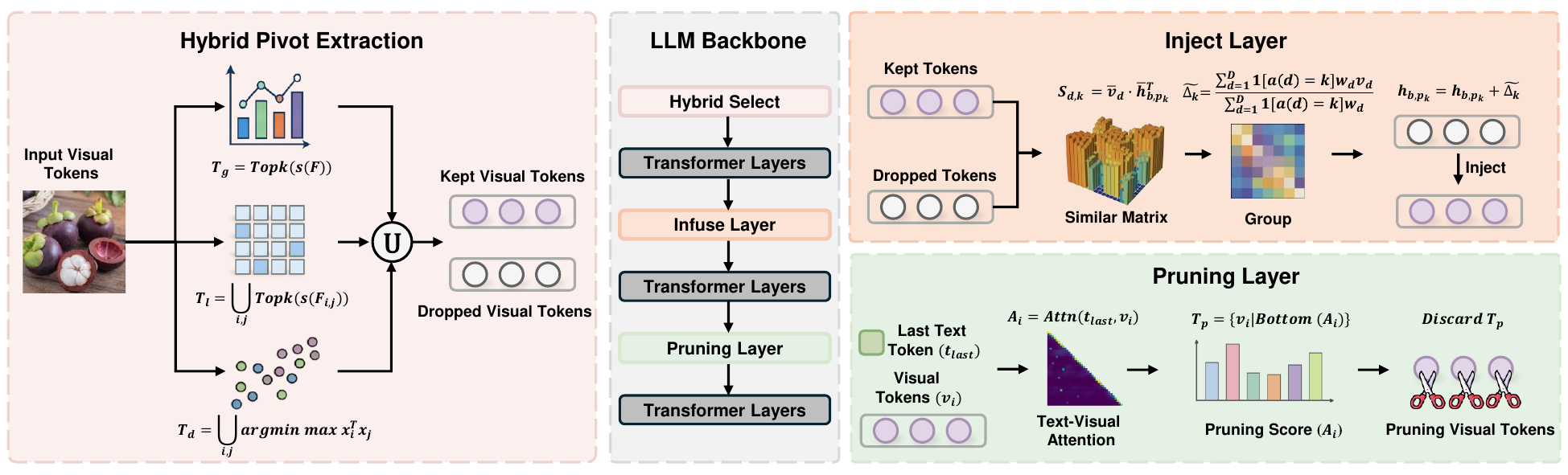}
    \caption{\textbf{Overall architecture of ReTrace.}
    ReTrace follows a \textit{cache--inject--prune} schedule.
    \textbf{(1) Pivot Buffer Construction} builds a compact active receiver buffer at the encoder output while caching the remaining tokens as dormant context.
    \textbf{(2) Backtracking-Aware Context Injection} restores the missing pivot-to-context retrieval route at the backtracking layer through similarity-based routing.
    \textbf{(3) Post-Absorption Task Pruning} removes task-irrelevant enriched pivots after context absorption.}
    \label{fig:over}
\end{figure*}

\subsection{Overall Architecture}

ReTrace decouples temporary computational deactivation from irreversible information deletion. Instead of making a one-shot keep/drop decision in shallow layers, it assigns visual tokens to three states: \textit{active}, \textit{dormant}, and \textit{deleted}. Active tokens enter the LLM backbone as a compact pivot buffer. Dormant tokens bypass expensive layer-wise computation but remain cached for later restoration. Deleted tokens are removed only after their information has been absorbed.

As shown in Fig.~\ref{fig:over}, ReTrace follows a \textit{cache--inject--prune} schedule. It first constructs a pivot buffer and caches the remaining tokens as dormant context. It then injects this context at the backtracking layer to approximate the missing pivot-to-context retrieval route. Finally, it prunes redundant enriched pivots for efficient final reasoning.

\subsection{Pivot Buffer Construction}
\label{sec:stage1}

The first stage builds an intermediate receiver buffer, not the final reasoning core. Given a visual feature sequence $\mathcal{V}$ of length $N$ and an intermediate retention ratio $\rho_1$, we set
\begin{equation}
    K_1 = \lfloor \rho_1 N \rfloor .
\end{equation}

The diagnostic pivots in Sec.~\ref{sec:mov} are computed from full-token LLM attention, which is useful for analysis but unavailable during compressed inference. We therefore construct a deployable encoder-level buffer using saliency and diversity cues. For implementation, we compute token saliency from selected vision-encoder attention layers. The buffer combines three subsets. 

\textbf{Global saliency} keeps tokens with high encoder-level saliency scores, denoted by $s_g(v_i)$:
\begin{equation}
    \mathcal{T}_g = \text{Top-}K_g(\{s_g(v_i) \mid v_i \in \mathcal{V}\}) .
\end{equation}

\textbf{Local saliency} preserves regional structures by selecting the most salient token in each spatial window $W_m \in \mathcal{W}$, using the local saliency score $s_l(v_i)$:
\begin{equation}
    \mathcal{T}_l = \{\arg\max_{v_i \in W_m} s_l(v_i) \mid W_m \in \mathcal{W}\} .
\end{equation}

\textbf{Spatial diversity} improves coverage through farthest-point sampling:

\begin{equation}
\begin{split}
\mathcal{T}_d
&= \left\{ v^* \;\middle|\;
v^* = \arg\max_{v_i \notin \mathcal{S}}
\min_{v_j \in \mathcal{S}} \right. \\
&\qquad \left.
\left(1 - \frac{v_i \cdot v_j^\top}
{\|v_i\| \|v_j\|}\right)
\right\}.
\end{split}
\end{equation}

The final pivot buffer is
\begin{equation}
    \mathcal{P} = \mathcal{T}_g \cup \mathcal{T}_l \cup \mathcal{T}_d, \quad |\mathcal{P}| = K_1 .
\end{equation}
Tokens outside the buffer are stored as dormant context,
\begin{equation}
    \mathcal{D} = \mathcal{V} \setminus \mathcal{P},
\end{equation}
and bypass the LLM backbone until the injection stage.

\subsection{Backtracking-Aware Context Injection}
\label{sec:stage2}

At the backtracking layer, ReTrace restores dormant context to approximate the pivot-to-context retrieval observed in full-token inference. Let $b$ denote the injection layer, $h_{b,p_k}$ the hidden state of the $k$-th pivot, and $v_d$ the cached state of the $d$-th dormant token (after the projector, i.e., in the LLM hidden space). We compute normalized similarities:
\begin{equation}
\begin{gathered}
\bar{h}_{b,p_k} = \frac{h_{b,p_k}}{\|h_{b,p_k}\|_2}, \quad
\bar{v}_d = \frac{v_d}{\|v_d\|_2}, \\[2pt]
S_{d,k} = \bar{v}_d \cdot \bar{h}_{b,p_k}^{\top}.
\end{gathered}
\end{equation}

Each dormant token is assigned to its closest pivot:
\begin{equation}
    a(d) = \arg\max_k S_{d,k}, \quad w_d = S_{d,a(d)} .
\end{equation}

For each pivot, assigned dormant features are aggregated into a contextual increment:
\begin{equation}
\begin{gathered}
\tilde{\Delta}_k =
\frac{
\sum_{d=1}^{|\mathcal{D}|} \mathbf{1}[a(d)=k]\, w_d v_d
}{
\sum_{d=1}^{|\mathcal{D}|} \mathbf{1}[a(d)=k]\, w_d + \epsilon
}, \\[2pt]
h_{b,p_k} \leftarrow h_{b,p_k} + \tilde{\Delta}_k .
\end{gathered}
\end{equation}
The updated tokens are enriched pivots. They carry the compact active sequence forward while recovering part of the context that early pruning would remove.

\subsection{Post-Absorption Task Pruning}
\label{sec:stage3}

After context injection, ReTrace prunes redundant enriched pivots to reach the final budget. We use the last text token $t_{last}$ as an instruction-aware query and score each enriched pivot $p_i$:
\begin{equation}
    A_i = \text{Attn}(t_{last}, p_i) .
\end{equation}

Given the final retention ratio $\rho_2$, the final active token number is
\begin{equation}
    K_2 = \lfloor \rho_2 K_1 \rfloor .
\end{equation}
The pruning set contains the bottom $K_1-K_2$ pivots:
\begin{equation}
    \mathcal{T}_p = \text{Bottom-}(K_1 - K_2)(\{p_i \mid A_i\}) .
\end{equation}

Only the remaining $K_2$ enriched pivots are used for final reasoning. In this way, ReTrace prunes after context restoration rather than deleting raw context sources before they can be used.

\section{Experiment}
\label{sec:experiments}
\subsection{Experimental Settings}
We evaluate our framework on four MLLMs (Qwen2.5-VL 7B/3B, InternVL3 8B/2B) across eight diverse vision-language benchmarks: ScienceQA \cite{sqa}, POPE \cite{pope}, MME \cite{mme}, MMBench \cite{mmbench}, MMBench-CN \cite{mmbench}, MMStar \cite{mmstar}, MMERealWorld (MRW) \cite{mmerealworld}, and MMERealWorld-CN (MRW\textsuperscript{CN}) \cite{mmerealworld}. To demonstrate superiority, we comprehensively compare our approach against six recent state-of-the-art visual token compression baselines: FastV \cite{fastv}, VisionZip \cite{visionzip}, Dart \cite{dart}, HoloV \cite{holov}, VLMPruner \cite{vlmpruner}, and Vscan \cite{vscan}.

Across all evaluated models, we set the injection layer index to 8, the pruning layer index to 19, and the hybrid extraction quota ratio ($T_g:T_l:T_d$) to $1:1:2$. $\rho_1$ adaptively scales to match the target overall compression, while $\rho_2$ is strictly fixed at $1/3$. Detailed hyperparameter settings are provided in the supplementary material.

\subsection{Main Results}

\begin{table*}[ht]
\centering
\resizebox{\textwidth}{!}{%
\begin{tabular}{l |c|cccccccc |c} 
\toprule
\rowcolor[HTML]{EAF3FF}
\textbf{Method} & \textbf{Venue} & \textbf{MMB} & \textbf{MMB\textsuperscript{CN}} & \textbf{MME} & \textbf{POPE} & \textbf{SQA\textsubscript{I}} & \textbf{MMStar} & \textbf{MRW} & \textbf{MRW\textsuperscript{CN}} & \textbf{Avg.} \\
\midrule

\rowcolor[HTML]{F6F6F6}
Qwen2.5VL-7B & - & 83.5 & 83.4 & 2347 & 87.7 & 88.1 & 63.9 & 57.4 & 50.2 & 100.0\% \\
\midrule

\rowcolor[HTML]{F1F6FB}
\multicolumn{11}{c}{\textit{Token Reduction Setting: Retain 33.3\% Tokens ($\downarrow$66.7\%)}} \\
\midrule
FastV \cite{fastv} & \textnormal{\textit{ECCV'24}} & 75.7 & 78.8 & 2072 & 86.3 & 85.1 & 49.4 & 46.1 & 42.5 & 88.9\% \\
VisionZip \cite{visionzip} & \textnormal{\textit{CVPR'25}} & 83.2 & 80.7 & 2160 & 86.7 & 86.0 & 53.8 & 48.4 & 44.2 & 92.7\% \\
Dart \cite{dart} & \textnormal{\textit{EMNLP'25}} & 80.1 & 78.3 & 2166 & 85.1 & 84.3 & 48.9 & 45.7 & 39.0 & 88.6\% \\
HoloV \cite{holov} & \textnormal{\textit{NeurIPS'25}} & 78.3 & 78.3 & 2093 & 84.9 & 85.8 & 48.1 & 35.3 & 34.7 & 84.7\% \\
VLMPruner \cite{vlmpruner} & \textnormal{\textit{CVPR'26}} & 77.2 & 76.8 & 2235 & 86.1 & 85.7 & 46.3 & 44.5 & 41.7 & 88.5\% \\
Vscan \cite{vscan} & \textnormal{\textit{TMLR'26}} & 80.6 & 78.6 & \textbf{2275} & 87.2 & 86.4 & 49.4 & 46.6 & 43.5 & 91.3\% \\
\rowcolor[HTML]{E8F2FF}
\textbf{Ours} & - & \textbf{83.5} & \textbf{81.7} & 2240 & \textbf{87.5} & \textbf{87.1} & \textbf{55.3} & \textbf{48.8} & \textbf{44.3} & \textbf{94.0\%} \\
\midrule

\rowcolor[HTML]{F1F6FB}
\multicolumn{11}{c}{\textit{Token Reduction Setting: Retain 22.2\% Tokens ($\downarrow$77.8\%)}} \\
\midrule
FastV \cite{fastv} & \textnormal{\textit{ECCV'24}} & 74.9 & 77.8 & 2036 & 84.9 & 83.2 & 45.0 & 41.3 & 40.3 & 85.5\% \\
VisionZip \cite{visionzip} & \textnormal{\textit{CVPR'25}} & 81.6 & 79.4 & 2038 & 85.5 & 83.8 & 51.2 & 43.7 & 40.2 & 88.6\% \\
Dart \cite{dart} & \textnormal{\textit{EMNLP'25}} & 79.2 & 76.8 & 2090 & 82.8 & 83.4 & 46.6 & 40.7 & 36.5 & 85.2\% \\
HoloV \cite{holov} & \textnormal{\textit{NeurIPS'25}} & 76.5 & 75.6 & 2043 & 82.0 & 81.4 & 46.0 & 31.7 & 31.4 & 80.7\% \\
VLMPruner \cite{vlmpruner} & \textnormal{\textit{CVPR'26}} & 76.3 & 75.2 & 2091 & 84.7 & 83.1 & 45.2 & 41.2 & 40.3 & 85.5\% \\
Vscan \cite{vscan} & \textnormal{\textit{TMLR'26}} & 80.5 & 76.3 & 2135 & 86.1 & 85.8 & 47.4 & 44.1 & 41.3 & 88.5\% \\
\rowcolor[HTML]{E8F2FF}
\textbf{Ours} & - & \textbf{81.7} & \textbf{80.8} & \textbf{2160} & \textbf{86.3} & \textbf{86.1} & \textbf{52.1} & \textbf{44.4} & \textbf{42.0} & \textbf{90.7\%} \\
\midrule

\rowcolor[HTML]{F1F6FB}
\multicolumn{11}{c}{\textit{Token Reduction Setting: Retain 11.1\% Tokens ($\downarrow$88.9\%)}} \\
\midrule
FastV \cite{fastv} & \textnormal{\textit{ECCV'24}} & 69.2 & 74.9 & 1940 & 79.7 & 80.1 & 40.0 & 34.0 & 35.7 & 78.8\% \\
VisionZip \cite{visionzip} & \textnormal{\textit{CVPR'25}} & \textbf{79.8} & 76.8 & 1828 & 82.2 & 81.9 & 44.8 & \textbf{36.5} & \textbf{37.5} & 82.6\% \\
Dart \cite{dart} & \textnormal{\textit{EMNLP'25}} & 73.4 & 72.9 & 1877 & 78.1 & 81.6 & 40.5 & 32.9 & 32.1 & 77.7\% \\
HoloV \cite{holov} & \textnormal{\textit{NeurIPS'25}} & 72.4 & 73.8 & 2006 & 77.5 & 81.1 & 41.8 & 23.8 & 26.1 & 75.0\% \\
VLMPruner \cite{vlmpruner} & \textnormal{\textit{CVPR'26}} & 75.1 & 73.7 & 2015 & 80.2 & 81.1 & 42.3 & 35.7 & 34.9 & 80.7\% \\
Vscan \cite{vscan} & \textnormal{\textit{TMLR'26}} & 76.8 & 74.4 & 1990 & 81.3 & \textbf{83.3} & 42.4 & 33.2 & 32.5 & 80.3\% \\
\rowcolor[HTML]{E8F2FF}
\textbf{Ours} & - & 79.6 & \textbf{77.4} & \textbf{2093} & \textbf{82.7} & 82.3 & \textbf{46.3} & 36.4 & 37.2 & \textbf{84.4\%} \\
\bottomrule
\end{tabular}%
}
\caption{\textbf{Performance comparisons on Qwen2.5VL-7B across 8 benchmarks.} The ``Avg.'' metric denotes the mean relative performance retention across all datasets compared to the baseline.}
\label{tab:qwenvl7b}
\end{table*}

\begin{table*}[!htbp]
\centering

\resizebox{\textwidth}{!}{%
\begin{tabular}{l |c|cccccccc |c} 
\toprule
\rowcolor[HTML]{EAF3FF}
\textbf{Method} & \textbf{Venue} & \textbf{MMB} & \textbf{MMB\textsuperscript{CN}} & \textbf{MME} & \textbf{POPE} & \textbf{SQA\textsubscript{I}} & \textbf{MMStar} & \textbf{MRW} & \textbf{MRW\textsuperscript{CN}} & \textbf{Avg.} \\
\midrule

\rowcolor[HTML]{F6F6F6}
InternVL3-8B & - & 83.5 & 83.4 & 2415 & 90.7 & 98.2 & 63.9 & 62.0 & 60.8 & 100.0\% \\
\midrule

\rowcolor[HTML]{F1F6FB}
\multicolumn{11}{c}{\textit{Token Reduction Setting: Retain 22.2\% Tokens ($\downarrow$77.8\%)}} \\
\midrule
FastV \cite{fastv} & \textnormal{\textit{ECCV'24}} & \textbf{83.8} & 83.0 & \textbf{2324} & 89.4 & 95.4 & 58.2 & 46.3 & 51.9 & 92.7\% \\
VisionZip \cite{visionzip} & \textnormal{\textit{CVPR'25}} & 80.3 & 82.0 & 2242 & 88.8 & 95.0 & 56.8 & 44.8 & 49.5 & 90.4\% \\
HoloV \cite{holov} & \textnormal{\textit{NeurIPS'25}} & 80.6 & 82.3 & 2201 & 88.7 & 95.4 & 57.0 & 44.7 & 50.0 & 90.5\% \\
\rowcolor[HTML]{E8F2FF}
\textbf{Ours} & - & 83.6 & \textbf{84.3} & 2319 & \textbf{90.2} & \textbf{96.7} & \textbf{59.7} & \textbf{48.7} & \textbf{53.3} & \textbf{94.4\%} \\
\midrule

\rowcolor[HTML]{F1F6FB}
\multicolumn{11}{c}{\textit{Token Reduction Setting: Retain 11.1\% Tokens ($\downarrow$88.9\%)}} \\
\midrule
FastV \cite{fastv} & \textnormal{\textit{ECCV'24}} & 79.5 & 78.7 & 2202 & 88.8 & 93.2 & 52.3 & 39.9 & 47.3 & 87.2\% \\
VisionZip \cite{visionzip} & \textnormal{\textit{CVPR'25}} & 78.7 & 76.8 & 1978 & 85.9 & 87.7 & 50.2 & 34.5 & 43.7 & 82.2\% \\
HoloV \cite{holov} & \textnormal{\textit{NeurIPS'25}} & 79.8 & 78.1 & 1964 & 87.9 & 89.4 & 51.5 & 36.1 & 44.2 & 83.9\% \\
\rowcolor[HTML]{E8F2FF}
\textbf{Ours} & - & \textbf{80.9} & \textbf{80.7} & \textbf{2210} & \textbf{89.0} & \textbf{94.1} & \textbf{55.9} & \textbf{41.5} & \textbf{48.6} & \textbf{89.4\%} \\
\bottomrule
\end{tabular}%
}
\caption{\textbf{Performance comparisons on InternVL3-8B across 8 benchmarks.} The ``Avg.'' metric denotes the mean relative performance retention across all datasets compared to the baseline.}
\label{tab:internvl3-8b}
\end{table*}

Tabs.~\ref{tab:qwenvl7b} - \ref{tab:internvl2b_compact} present the quantitative evaluation of our method against existing visual token compression baselines across different models. Overall, under various compression ratios, our framework demonstrates highly competitive performance across most benchmarks and maintains the highest average performance retention. For instance, as shown in Tab.~\ref{tab:qwenvl7b}, on Qwen2.5-VL-7B, under the 22.2\% and 11.1\% visual token retention settings, our method preserves 90.7\% and 84.4\% of the original model's comprehensive performance, respectively (calculated by averaging the normalized scores against the full-token baseline across all benchmarks). Although it does not achieve absolute superiority on every specific sub-task, the overall performance across multiple datasets objectively validates the strong generality of our method. More results are provided in the Appendix.\ref{sec:more_main_results}.

\begin{table}[!htbp]
\centering

\resizebox{\columnwidth}{!}{%
\begin{tabular}{l| cccc |c}
\toprule
\rowcolor[HTML]{EAF3FF}
\textbf{Method} & \textbf{MMB} & \textbf{MME} & \textbf{MMStar} & \textbf{MRW\textsuperscript{CN}} & \textbf{Avg.} \\
\midrule
\rowcolor[HTML]{F6F6F6}
Qwen2.5VL-3B & 79.1 & 2157 & 55.9 & 45.1 & 100.0\% \\
\midrule
FastV \cite{fastv} & 71.0 & 1698 & 36.6 & 32.2 & 76.3\% \\
VisionZip \cite{visionzip} & \textbf{73.2} & 1698 & 40.1 & 31.9 & 78.4\% \\
Dart \cite{dart} & 72.8 & 1640 & 36.8 & 29.4 & 74.8\% \\
HoloV \cite{holov} & 70.8 & 1673 & 38.6 & 25.1 & 72.9\% \\
VLMPruner \cite{vlmpruner} & 68.2 & 1735 & 36.0 & 26.7 & 72.6\% \\
Vscan \cite{vscan} & 69.6 & 1763 & 35.7 & 29.2 & 74.6\% \\
\rowcolor[HTML]{E8F2FF}
\textbf{Ours} & 72.3 & \textbf{1781} & \textbf{40.3} & \textbf{32.5} & \textbf{79.5\%} \\
\bottomrule
\end{tabular}%
}
\caption{\textbf{Compact comparison on Qwen2.5VL-3B at 11.1\% token retention.}}
\label{tab:qwenvl3b_compact}
\end{table}

\begin{figure}[ht]
    \centering
    \begin{subfigure}[t]{\linewidth}
        \centering
        \includegraphics[width=\linewidth]{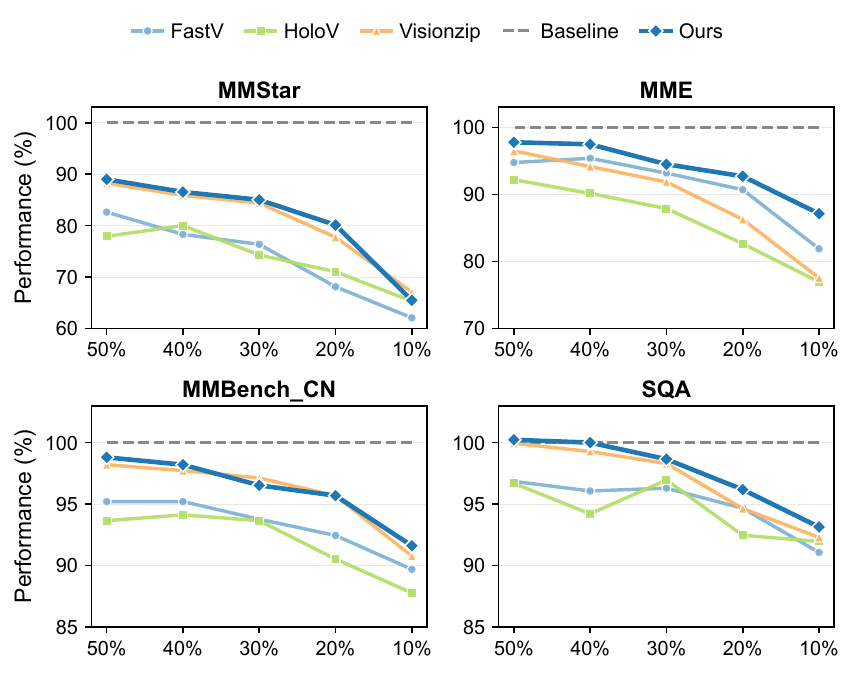}
        \caption{Qwen2.5-VL-7B.}
        \label{fig:qwenvl_retention}
    \end{subfigure}

    \begin{subfigure}[t]{\linewidth}
        \centering
        \includegraphics[width=\linewidth]{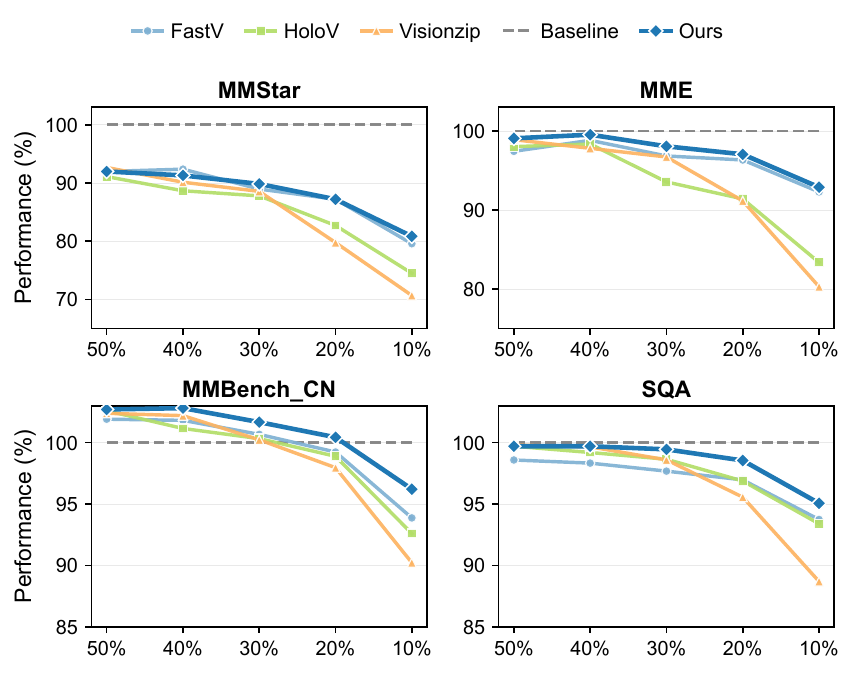}
        \caption{InternVL3-8B.}
        \label{fig:internvl_retention}
    \end{subfigure}

    \caption{\textbf{Performance degradation trends across different token retention ratios.}
    We compare the degradation curves on Qwen2.5-VL-7B and InternVL3-8B under different visual token retention ratios.}
    \label{fig:retention_trends}
\end{figure}

Furthermore, Fig.~\ref{fig:qwenvl_retention} and \ref{fig:internvl_retention} illustrate the performance degradation trends under continuous retention rate settings. Across multiple benchmarks, as the compression rate increases, our method maintains an overall leading position across all retention intervals. Compared to the significant performance drops observed in other baselines under extreme compression, our approach exhibits a smoother degradation trajectory. This further confirms the high practicality and robustness of our method in real-world deployments.

\begin{table}[t]
\centering

\resizebox{\columnwidth}{!}{%
\begin{tabular}{l |cccc |c}
\toprule
\rowcolor[HTML]{EAF3FF}
\textbf{Method} & \textbf{MMB} & \textbf{MME} & \textbf{MMStar} & \textbf{MRW\textsuperscript{CN}} & \textbf{Avg.} \\
\midrule
\rowcolor[HTML]{F6F6F6}
InternVL3-2B & 80.8 & 2221 & 60.7 & 52.1 & 100.0\% \\
\midrule
FastV \cite{fastv} & 54.0 & 1398 & 36.4 & 25.9 & 59.9\% \\
VisionZip \cite{visionzip} & 67.7 & 1747 & 42.5 & 32.9 & 73.9\% \\
HoloV \cite{holov} & 70.1 & 1743 & 44.8 & 34.8 & 76.5\% \\
\rowcolor[HTML]{E8F2FF}
\textbf{Ours} & \textbf{74.3} & \textbf{1965} & \textbf{47.9} & \textbf{37.1} & \textbf{82.6\%} \\
\bottomrule
\end{tabular}%
}
\caption{\textbf{Compact comparison on InternVL3-2B at 11.1\% token retention.}}
\label{tab:internvl2b_compact}
\end{table}

\subsection{Ablation Results}

\begin{figure}[t]
    \centering
    \includegraphics[width=\linewidth]{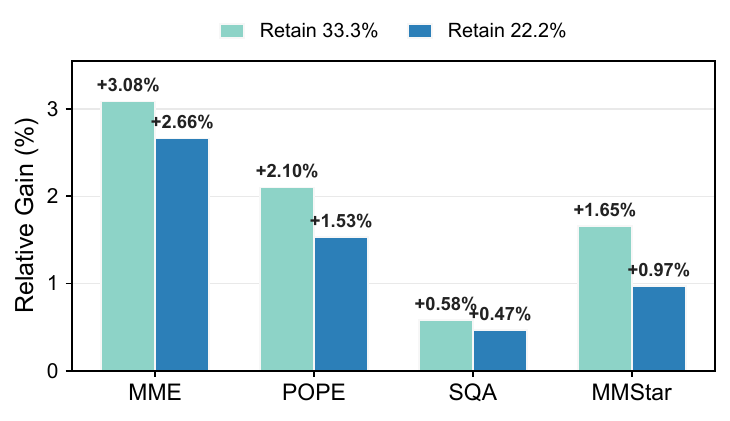}
    \caption{\textbf{Effect of backtracking-aware injection.}
    We report the relative gains brought by context injection under different retention ratios.}
    \label{fig:ablation_injection}
\end{figure}

To validate the core designs of \textbf{ReTrace}, we present two main ablations in the main paper: the effect of backtracking-aware injection and the choice of injection layer. Additional ablations on pivot buffer construction are provided in the supplementary material.

First, we isolate the contribution of the injection mechanism. As shown in Fig.~\ref{fig:ablation_injection}, adding backtracking-aware injection consistently improves performance over the no-injection variant across different benchmarks and retention ratios. This confirms that dormant context should not merely be cached, but restored when the model enters the backtracking window.

\begin{figure}[!htbp]
    \centering
    \includegraphics[width=\linewidth]{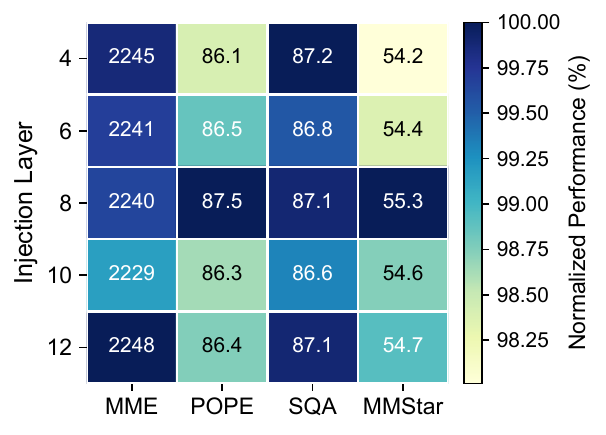}
    \caption{\textbf{Ablation on injection layer placement.}
    We evaluate different injection layers and observe the most balanced performance around the middle-layer backtracking window.}
    \label{fig:ablation_layer}
\end{figure}

Second, we study where the injection should be placed. As shown in Fig.~\ref{fig:ablation_layer}, injecting around layer 8 gives the most robust overall performance across benchmarks. Earlier injection provides context before the pivots have sufficiently formed, while later injection may miss the middle-layer retrieval window. This result supports our central claim that ReTrace benefits from restoring context at the layer where the model naturally looks back.

\subsection{Efficiency Results}

\begin{table}[t]
\centering
\small

\setlength{\tabcolsep}{4pt}
\resizebox{\columnwidth}{!}{%
\begin{tabular}{l |c c c c |c}
\toprule
\textbf{Method} 
& \textbf{Tokens $\downarrow$} 
& \textbf{FLOPs $\downarrow$} 
& \textbf{KV $\downarrow$} 
& \textbf{MME $\uparrow$} 
& \textbf{Prefill Speedup $\uparrow$} \\
\midrule

Qwen2.5VL-7B 
& 100\% 
& 1043.83 
& 211.4 MB 
& \textbf{2347} 
& 1.00$\times$ \\

\rowcolor[HTML]{E8F2FF}
+ Ours 
& 22.2\% 
& 279.82 
& 37.0 MB 
& 2160 
& 2.28$\times$ \\

\rowcolor[HTML]{E8F2FF}
+ Ours 
& \textbf{11.1\%} 
& \textbf{148.06} 
& \textbf{26.7 MB} 
& 2093 
& \textbf{2.83}$\times$ \\

\bottomrule
\end{tabular}%
}
\caption{\textbf{Compact efficiency results on Qwen2.5VL-7B.} Runtime is evaluated on a single NVIDIA A100-80GB GPU with FlashAttention.}
\label{tab:efficiency_compact}
\end{table}

Tab.~\ref{tab:efficiency_compact} details the hardware efficiency on Qwen2.5-VL-7B evaluated with FlashAttention. At the extreme 11.1\% retention rate, our method reduces FLOPs and KV Cache by approximately 87\%, achieving a 2.83$\times$ prefilling speedup. Despite this nearly 90\% resource reduction, the model maintains a competitive MME score of 2093, demonstrating a highly practical accuracy-efficiency trade-off for MLLMs deployment.

\subsection{Qualitative Results}
\begin{figure}[ht]
\centering
\scriptsize

\includegraphics[width=0.7\linewidth]{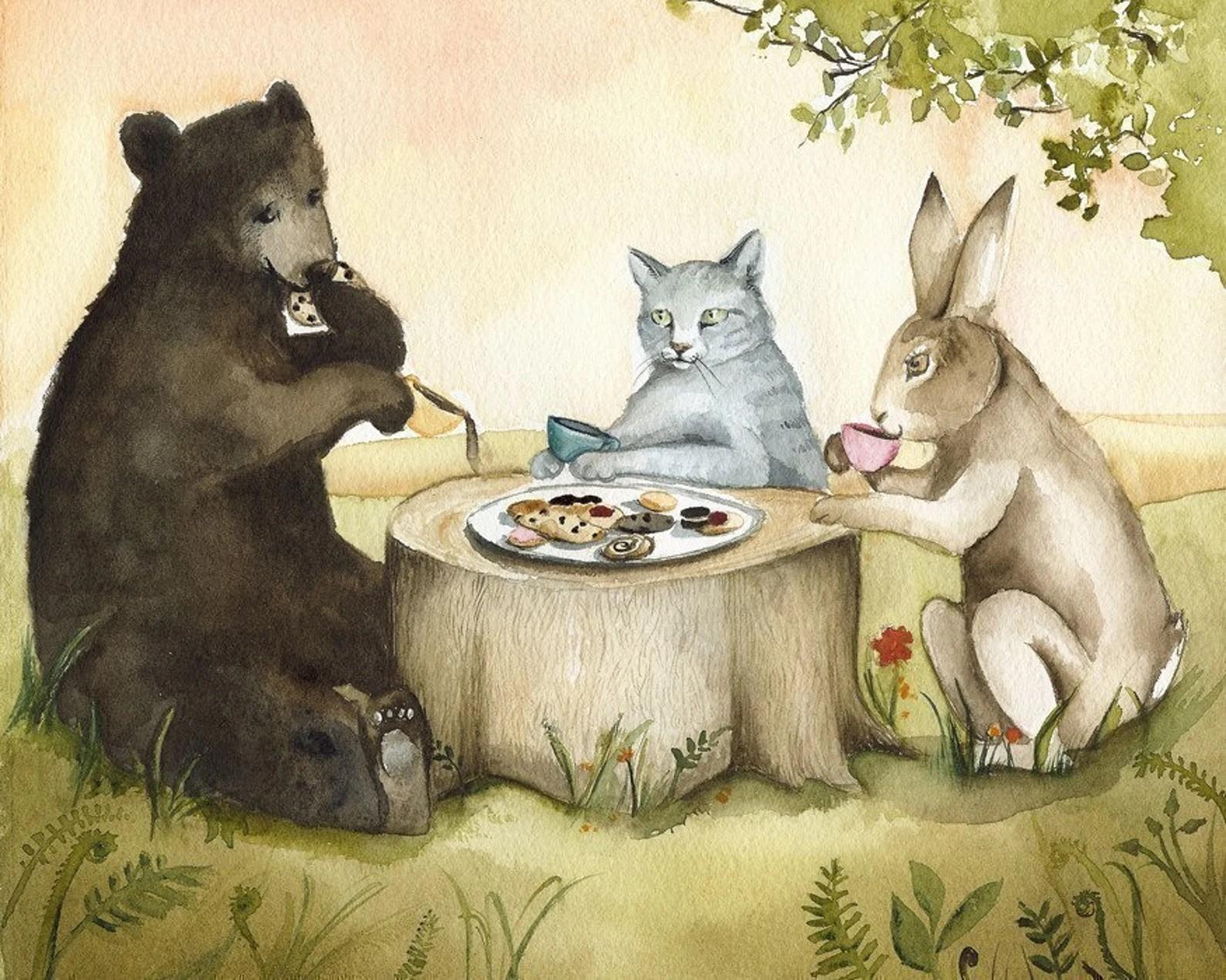}

\begin{tcolorbox}[
    enhanced,
    colback=cyan!3,
    colframe=cyan!45!blue!45!gray,
    boxrule=0.5pt,
    arc=1pt,
    left=3pt, right=3pt, top=3pt, bottom=3pt,
    width=\linewidth,
]
\setlength{\baselineskip}{9pt}
\textbf{\textcolor{cyan!55!blue!80!black}{User:}} \textit{Describe this image.} \\[2pt]
\textbf{\textcolor{cyan!55!blue!80!black}{Ours:}} {The image depicts a whimsical scene of three anthropomorphic animals sitting at a table, enjoying a tea party in a forest setting. The animal on the left appears to be a bear, the one in the middle is a cat, and the one on the right is a rabbit. They are all holding cups and saucers, and there are plates of cookies in front of them. The background features greenery, including grass and trees, suggesting they are in a natural, outdoor environment.}
\end{tcolorbox}

\caption{\textbf{Qualitative result at 11.1\% visual-token retention.}}
\label{fig:qualitative_comparison}
\end{figure}

As illustrated in Fig.~\ref{fig:qualitative_comparison}, the qualitative results demonstrate that our method maintains highly robust reasoning capabilities even under significant token compression. More diverse qualitative examples are provided in the supplementary material.

\section{Conclusion}

This paper identifies \textbf{Visual Backtracking} in MLLMs, showing that early low-saliency tokens can remain useful as middle-layer context sources. This finding explains why early pruning can harm complex reasoning: it treats temporary computational deactivation as irreversible information deletion. To address this, we propose \textbf{ReTrace}, a multi-stage compression framework that caches non-pivot tokens as dormant context, restores them when backtracking occurs, and prunes redundant enriched pivots only after context absorption. By reducing compute and memory overhead by nearly 90\% while maintaining strong reasoning performance, ReTrace provides a practical solution for efficient long-sequence multimodal inference. This highlights the broader importance of adaptive token compression that follows the model's layer-wise information demand, rather than relying on fixed early-layer pruning decisions.

\section*{Limitations}
\label{sec:lim}
ReTrace currently uses fixed layer indices for context injection and late-stage pruning. This fixed schedule can be suboptimal because the optimal restoration and pruning points may vary with input visual complexity and architecture-specific layer-wise dynamics. In addition, extreme compression can cause information overload when too much dormant context is injected into a small pivot buffer.

\section*{Ethical Considerations and Potential Risks}
\label{sec:ethics_risks}

This work studies efficient inference for MLLMs and does not introduce new data involving human subjects, private information, or personally identifiable content. Our experiments use public benchmarks and publicly available model checkpoints. The main risk lies in deployment. Since ReTrace compresses visual tokens, it may change model outputs in some cases, especially under aggressive compression. Such changes should be carefully evaluated before using compressed MLLMs in high-stakes domains such as medical, legal, financial, or safety-critical applications. ReTrace is an efficiency technique and should not replace safety alignment, factuality checking, or human oversight. Faster inference may also reduce the cost of large-scale automated generation or visual analysis. This risk is shared by many model acceleration methods. Users should follow the licenses of the underlying models and datasets and apply proper safeguards in real-world deployment.

\section*{Acknowledgments}

We thank the High-Performance Computing Center at Eastern Institute of Technology and Ningbo Institute of Digital Twin for providing the computing resources that supported this work. We also thank the anonymous reviewers for their constructive comments and valuable suggestions.

\bibliography{custom}
\newpage
\appendix
\raggedbottom

\section{Overview}

This supplementary material provides additional details and results to support the main manuscript. The contents are organized as follows.

We first describe the experimental configurations in Sec.~\ref{sec:setting} and present the pseudocode of our method in Sec.~\ref{sec:algo}. Next, we include additional motivation experiments in Sec.~\ref{sec:more_mov}, followed by more experimental results in Sec.~\ref{sec:more_main}, including extended quantitative evaluations across different model scales. Further ablation studies are provided in Sec.~\ref{sec:more_abla}.

We also analyze the theoretical complexity and empirical latency overhead of our method in Sec.~\ref{sec:eff}. In addition, we visualize the pivot buffer construction in Sec.~\ref{sec:vis_hpe} and provide more qualitative examples in Sec.~\ref{sec:more_qualitative}. Finally, we report details of the benchmarks and MLLMs in Secs.~\ref{sec:bench} and~\ref{sec:model}, declare the use of AI assistance in Sec.~\ref{sec:ai}.

\section{Settings Details}
\label{sec:setting}
Across InternVL-8B and Qwen2.5VL-7B, the injection and pruning layer indices are empirically set to 8 and 19, respectively. The hybrid extraction quota ratio ($T_g:T_l:T_d$) is configured to $1:1:2$. To match the target overall compression, $\rho_1$ scales adaptively while $\rho_2$ remains strictly fixed at $1/3$. Additionally, for local saliency extraction, we employ a $4 \times 4$ grid window to effectively capture regional geometric variations without introducing excessive computational overhead. The specific combinations of $\rho_1$ and $\rho_2$ for various target average retention ratios are detailed in Tab. \ref{tab:retention_ratios}.

Global saliency is computed from the final vision-encoder layer for all models. For local saliency, we use a model-specific layer choice. In Qwen2.5-VL, local saliency is computed from the 6-th vision-encoder layer to better preserve regional structures. In InternVL3, we use the final vision-encoder layer, since its tiling-based high-resolution processing already maintains local details within each tile. All models are evaluated with single-run greedy decoding, with sampling disabled.

\begin{table}
\centering

\begin{tabular}{l|cc}
\toprule
\textbf{Average Retention Ratio} & \textbf{$\rho_1$} & \textbf{$\rho_2$} \\
\midrule
11.1\% & 13.7\% & 33.3\%  \\
22.2\% & 27.4\% & 33.3\%  \\
33.3\% & 41.17\% & 33.3\% \\
\bottomrule
\end{tabular}
\caption{Hyperparameter configurations for different target compression levels. The late-stage retention $\rho_2$ is strictly fixed, while the mid-stage retention $\rho_1$ adaptively scales to meet the overall average retention ratio.}
\label{tab:retention_ratios}
\end{table}

\section{Algorithm}
\label{sec:algo}

We provide the detailed pseudocode of \textbf{ReTrace} in Alg.~\ref{alg:retrace}.

\begin{algorithm}[t]
\caption{ReTrace}
\label{alg:retrace}
\begin{algorithmic}[1]
\Require 
    Visual tokens $\mathcal{V} = \{v_1, v_2, \dots, v_N\}$; grid windows $\mathcal{W}$;
    retention ratios $\rho_{1}, \rho_{2}$; injection layer $b$; pruning layer $p$;
    last text token $t_{last}$.
\Ensure 
    Compact visual token set $\mathcal{P}_{final}$.

\Statex \textbf{\% Stage 1: Cache -- Pivot Buffer Construction}
\State $K_1 \gets \lfloor \rho_{1}N \rfloor, \quad K_2 \gets \lfloor \rho_{2}K_{1} \rfloor$
\State Compute global saliency scores $s_g(v_i)$ from a selected encoder attention layer
\State Compute local saliency scores $s_l(v_i)$ from a selected encoder attention layer

\State $T_g \gets \text{Top-}K_g(\{s_g(v_i) \mid v_i \in \mathcal{V}\})$
\Comment{Global saliency}

\State $T_l \gets \{\arg\max_{v_i \in W_m} s_l(v_i) \mid W_m \in \mathcal{W}\}$
\Comment{Local saliency}

\State $T_d \gets \text{FarthestPointSample}(\mathcal{V} \setminus (T_g \cup T_l), \text{metric}=\text{cosine})$
\Comment{Spatial diversity}

\State $\mathcal{P} \gets T_g \cup T_l \cup T_d$
\Comment{Active pivot buffer, $|\mathcal{P}| = K_1$}

\State $\mathcal{D} \gets \mathcal{V} \setminus \mathcal{P}$
\Comment{Dormant context}

\State $\hat{\mathcal{P}}, \hat{\mathcal{D}} \gets \text{Proj}(\mathcal{P}), \text{Proj}(\mathcal{D})$
\Comment{Map visual features to the LLM hidden space}

\State $\mathcal{P} \gets \text{LLM}_{1 \to b-1}(\hat{\mathcal{P}})$
\Comment{Forward projected active pivots}

\Statex \textbf{\% Stage 2: Inject -- Backtracking-Aware Context Restoration}
\For{each $\hat{v}_d \in \hat{\mathcal{D}}$}
    \State $S_{d,k} \gets 
    \frac{\hat{v}_d}{\|\hat{v}_d\|_2}
    \cdot
    \left(
    \frac{h_{b,p_k}}{\|h_{b,p_k}\|_2}
    \right)^\top
    \quad \forall p_k \in \mathcal{P}$
    \State $a(d) \gets \arg\max_k S_{d,k}$
    \State $w_d \gets S_{d,a(d)}$
\EndFor

\For{each $p_k \in \mathcal{P}$}
    \State $\tilde{\Delta}_k \gets
    \frac{
    \sum_{\hat{v}_d \in \hat{\mathcal{D}}} \mathbf{1}[a(d)=k]\, w_d \hat{v}_d
    }{
    \sum_{\hat{v}_d \in \hat{\mathcal{D}}} \mathbf{1}[a(d)=k]\, w_d + \epsilon
    }$
    \State $h_{b,p_k} \gets h_{b,p_k} + \tilde{\Delta}_k$
\EndFor

\State $\mathcal{P} \gets \text{LLM}_{b \to p-1}(\mathcal{P})$
\Comment{Forward enriched pivots}

\Statex \textbf{\% Stage 3: Prune -- Post-Absorption Task Pruning}
\State $A_i \gets \text{Attn}(t_{last}, p_i) \quad \forall p_i \in \mathcal{P}$
\State $T_p \gets \text{Bottom-}(K_1 - K_2)(\{p_i \mid A_i\})$
\State $\mathcal{P}_{final} \gets \mathcal{P} \setminus T_p$
\State $\mathcal{P}_{final} \gets \text{LLM}_{p \to L}(\mathcal{P}_{final})$
\Comment{Forward to final layer}

\State \Return $\mathcal{P}_{final}$
\end{algorithmic}
\end{algorithm}
\section{More Motivation Experiment}
\label{sec:more_mov}

We include several additional analyses to better understand Visual Backtracking. The first part checks whether the pattern remains stable when the pivot ratio changes. The second part gives more visual examples of the backtracking process. The last part uses 3D feature visualization to show how visual tokens are organized across layers. Together, these results support the main observation: MLLMs do not simply keep using the same visual anchors, but revisit broader visual context in the middle layers.

\subsection{Robustness Across Pivot Ratios}
\label{sec:app_pivot_ratio_robustness}

We further examine whether the observed backtracking pattern depends on the choice of pivot ratio. In the main analysis, pivots are defined by selecting the top-ranked visual tokens according to text-guided saliency. Here, we repeat the same analysis under top 10\%, 15\%, and 20\% pivot ratios.

\begin{figure*}[t]
    \centering
    \begin{subfigure}{0.32\textwidth}
        \centering
        \includegraphics[width=\linewidth]{figs/retention_from_csv_1_zhuzi.pdf}
        \caption{Top 10\%}
        \label{fig:retention_10_app}
    \end{subfigure}\hfill
    \begin{subfigure}{0.32\textwidth}
        \centering
        \includegraphics[width=\linewidth]{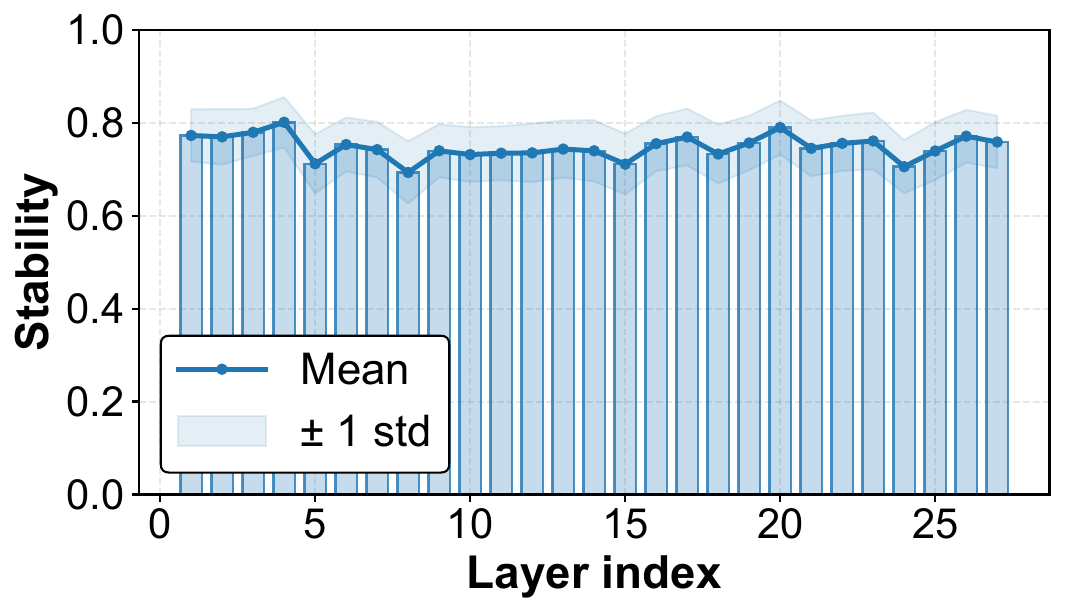}
        \caption{Top 15\%} 
        \label{fig:retention_15}
    \end{subfigure}\hfill
    \begin{subfigure}{0.32\textwidth}
        \centering
        \includegraphics[width=\linewidth]{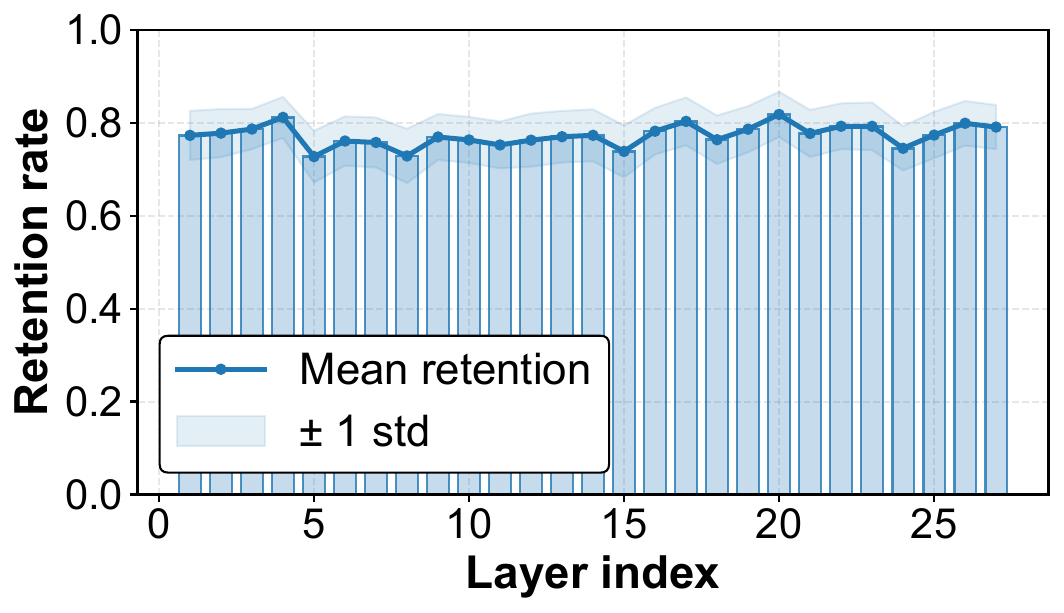}
        \caption{Top 20\%} 
        \label{fig:retention_20}
    \end{subfigure}
\caption{\textbf{Cross-layer spatial stability of pivot tokens under varying pivot ratios.} The cross-layer stability of pivot tokens remains consistently high across different top-$k$ ratio.}
    \label{fig:core_obs_retention_app}
\end{figure*}

Fig.~\ref{fig:core_obs_retention_app} shows that pivot tokens remain highly stable across layers for all three settings. This indicates that the model forms persistent visual anchors early, and this conclusion is not tied to a specific top-$k$ threshold.

\begin{figure*}[t]
    \centering
    \begin{subfigure}{0.32\textwidth}
        \centering
        \includegraphics[width=\linewidth]{./figs/waterfall_from_csv_1.pdf}
        \caption{Top 10\%}
        \label{fig:waterfall_10_app}
    \end{subfigure}\hfill
    \begin{subfigure}{0.32\textwidth}
        \centering
        \includegraphics[width=\linewidth]{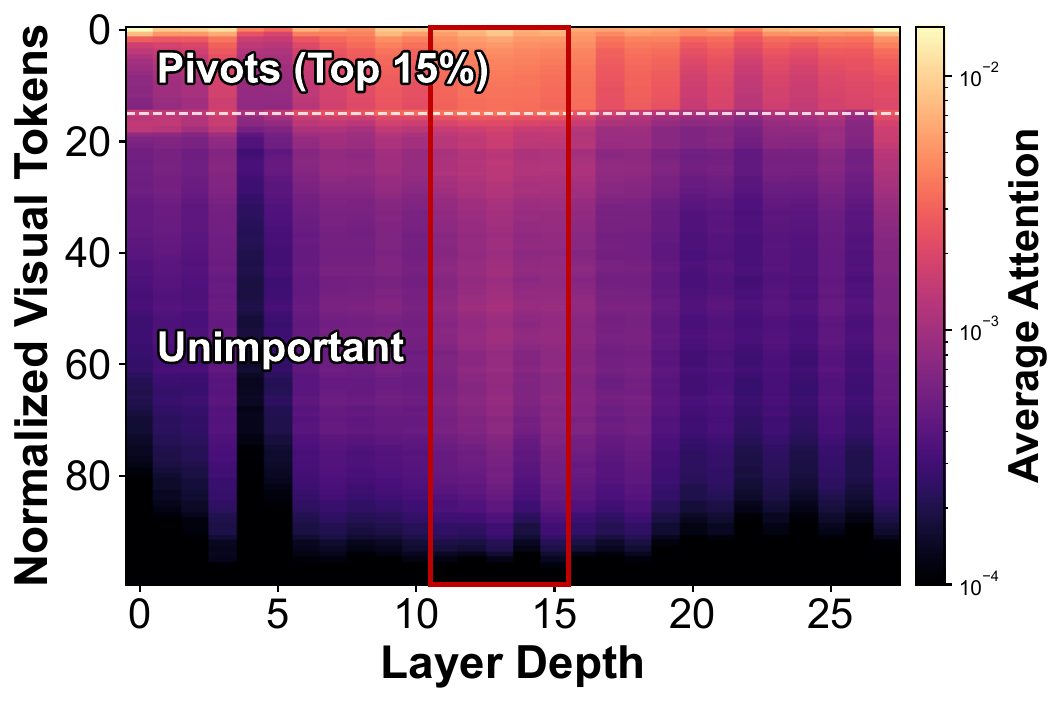}
        \caption{Top 15\%}
        \label{fig:waterfall_15}
    \end{subfigure}\hfill
    \begin{subfigure}{0.32\textwidth}
        \centering
        \includegraphics[width=\linewidth]{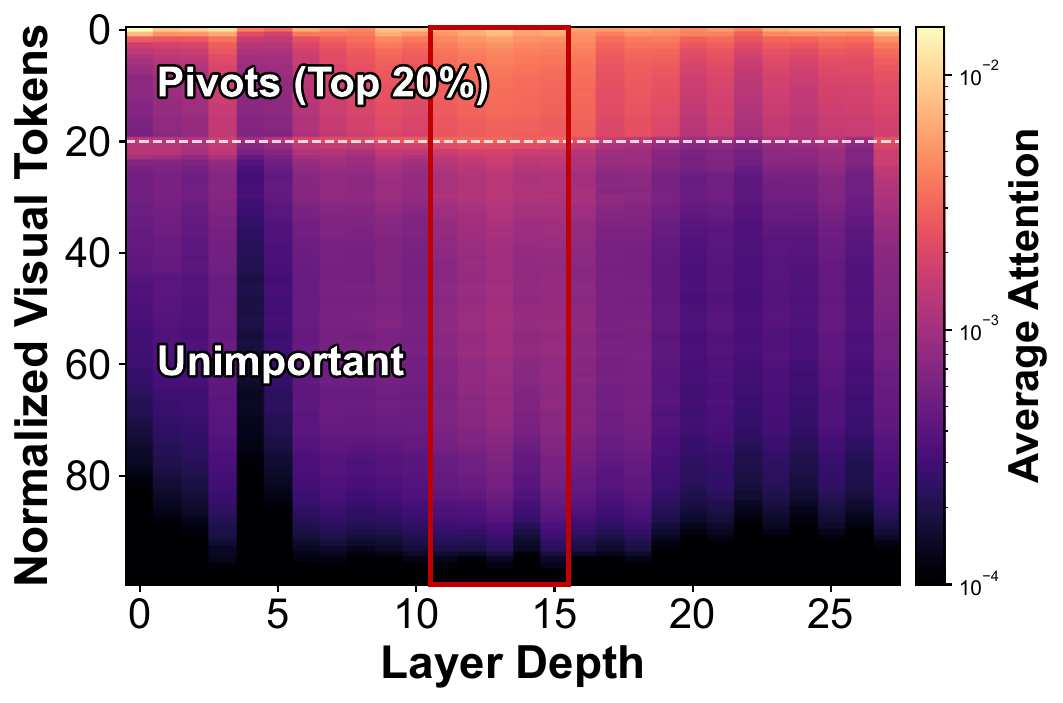}
        \caption{Top 20\%}
        \label{fig:waterfall_20}
    \end{subfigure}
    \caption{\textbf{Attention heatmap depicting the backtracking band.} The attention heatmap from pivot tokens to all visual tokens consistently reveals a distinct vertical bright band in the middle layers, regardless of the \textbf{pivot ratio}.}
    \label{fig:core_obs_waterfall}
\end{figure*}

\begin{figure*}[t]
    \centering
    \begin{subfigure}{0.32\textwidth}
        \centering
        \includegraphics[width=\linewidth]{./figs/flow_from_csv_1.pdf}
        \caption{Top 10\%}
        \label{fig:flow_10_app}
    \end{subfigure}\hfill
    \begin{subfigure}{0.32\textwidth}
        \centering
        \includegraphics[width=\linewidth]{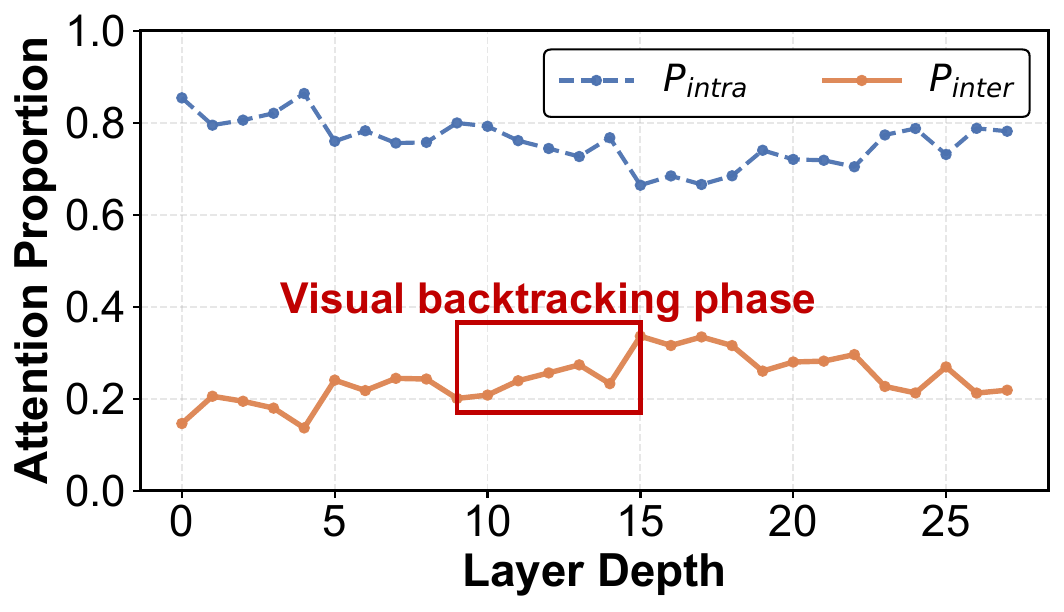}
        \caption{Top 15\%}
        \label{fig:flow_15}
    \end{subfigure}\hfill
    \begin{subfigure}{0.32\textwidth}
        \centering
        \includegraphics[width=\linewidth]{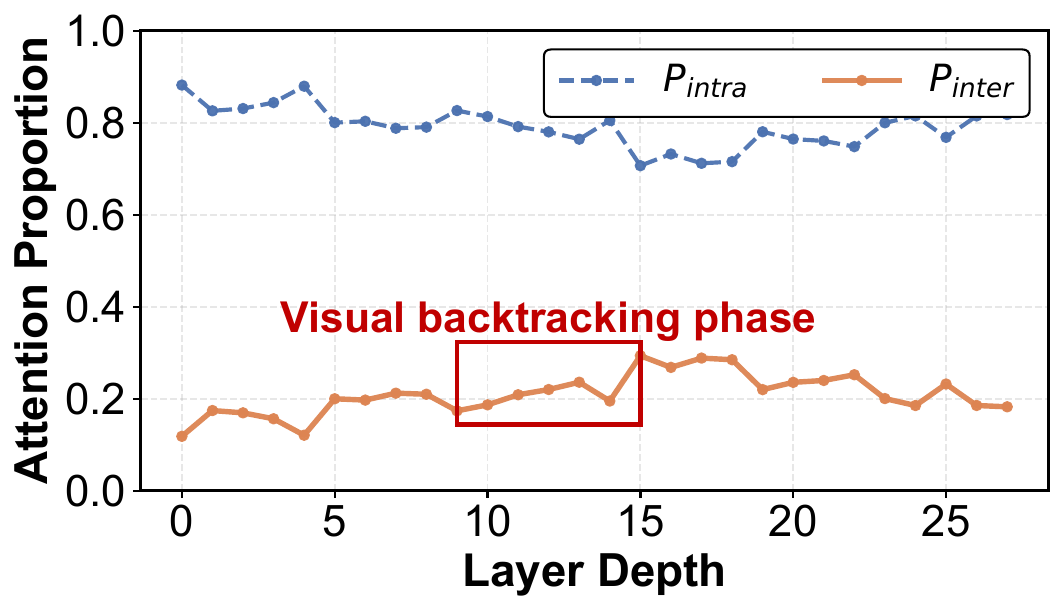}
        \caption{Top 20\%}
        \label{fig:flow_20}
    \end{subfigure}
    \caption{\textbf{Temporal evolution of normalized attention proportions.} The temporal evolution shows an abrupt ``Visual Backtracking'' peak (orange curve) where pivot tokens strongly query the background across different \textbf{top-$k$ ratio settings}.}
    \label{fig:core_obs_proportions}
\end{figure*}

Fig.~\ref{fig:core_obs_waterfall} visualizes the attention from pivot tokens to all visual tokens. Across different pivot ratios, we consistently observe a middle-layer expansion band, where pivots attend more broadly to non-pivot visual regions. Fig.~\ref{fig:core_obs_proportions} further confirms this trend: pivot-to-non-pivot attention peaks in the middle layers and then decreases in deeper layers.

These results show that Visual Backtracking is robust to the pivot ratio. The phenomenon is therefore unlikely to be an artifact of a particular top-$k$ choice. Instead, it reflects a stable layer-wise pattern: visual pivots first consolidate, then retrieve context from non-pivot regions, and finally reconcentrate for task-specific reasoning.

\subsection{Visualization of the Visual Backtracking Phenomenon} 
\begin{figure*}
    \centering
    \includegraphics[width=\linewidth]{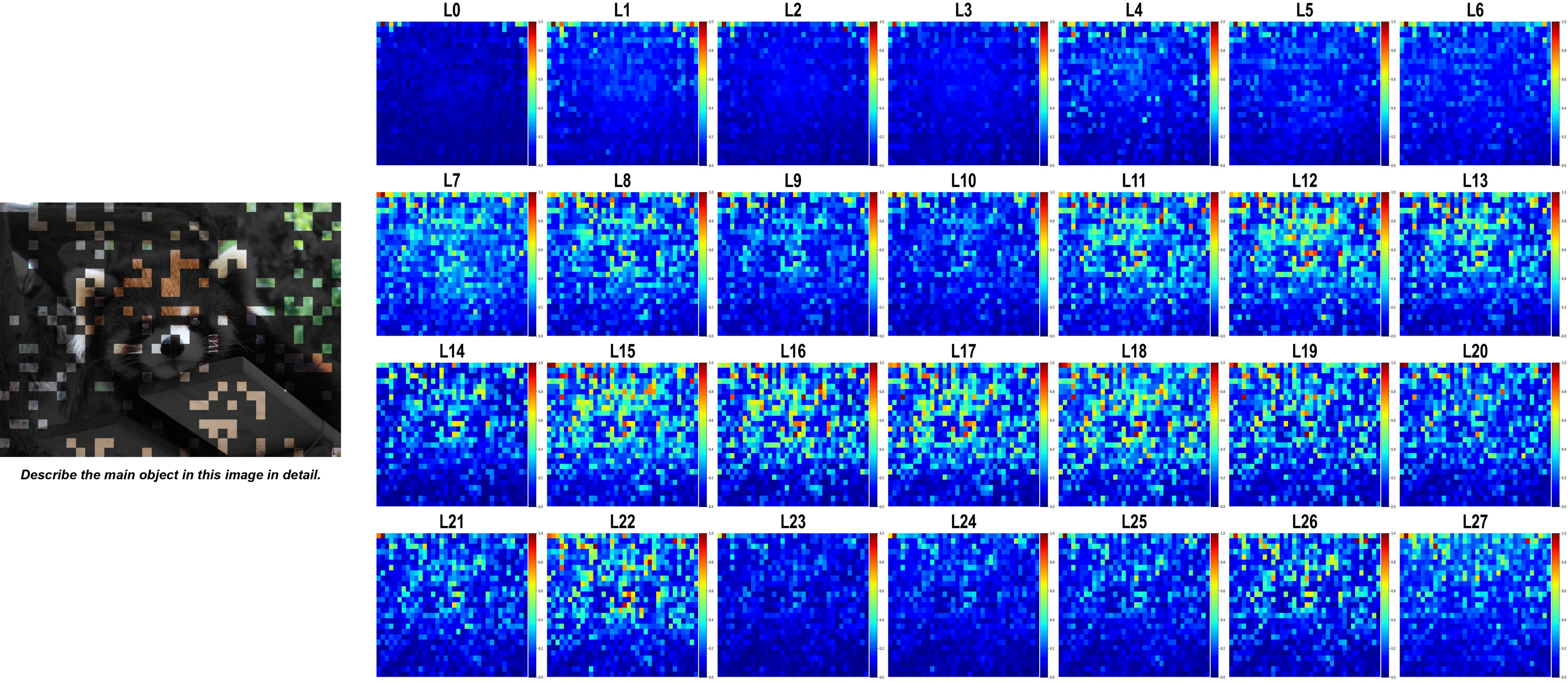}
    \caption{\textbf{Layer-wise visualization of the visual backtracking phenomenon.} We present the attention heatmaps from pivot tokens to all visual tokens across all transformer layers. The visualization clearly demonstrates a distinct dynamic: while attention remains highly localized on main subjects in the shallow and deep layers, it undergoes a dramatic spatial expansion in the mid-layers to actively query seemingly unimportant background tokens.}
    \label{fig:vis_backtracking_case1}
\end{figure*}

\begin{figure*}
    \centering
    \includegraphics[width=\linewidth]{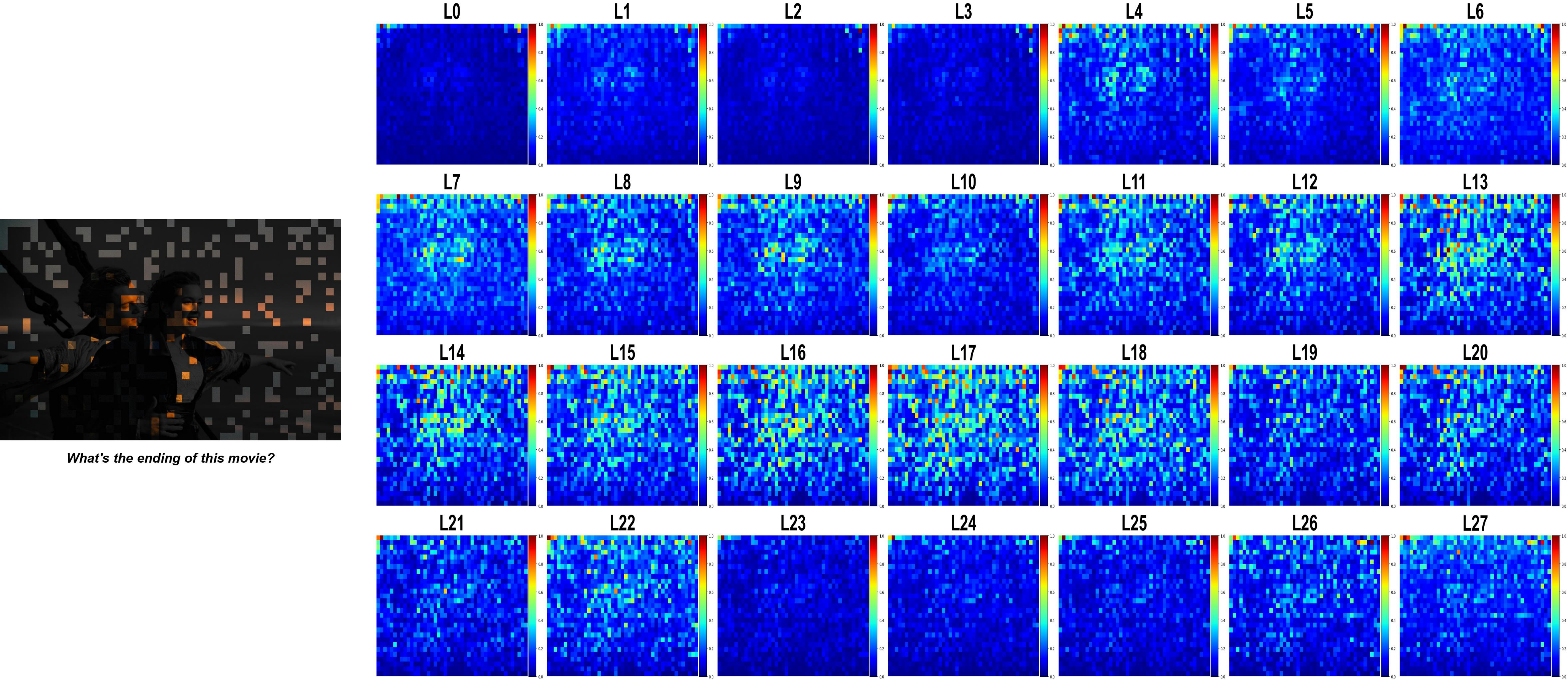}
    \caption{\textbf{Layer-wise visualization of the visual backtracking phenomenon.}}
    \label{fig:vis_backtracking_case2}
\end{figure*}

To provide a concrete illustration of the visual backtracking mechanism, we visualize the layer-wise attention dynamics. In \ref{fig:vis_backtracking_case1} and \ref{fig:vis_backtracking_case2}, the left panels highlight the identified top-$k$ pivot tokens on the input images, while the right matrices display the attention heatmaps from these pivots to all visual tokens across 28 transformer layers. A distinct structural shift is observable throughout the network depth. In both the shallow and deep layers, the attention remains highly concentrated on the highly salient tokens. However, in the intermediate layers, the attention distribution undergoes a drastic spatial expansion, actively querying a broader set of seemingly unimportant tokens. This layer-wise observation directly validates that mid-layer computations heavily rely on broad contextual layouts for visual reasoning, structurally precluding the aggressive pruning of these less salient tokens at early stages.

\subsection{3D Visualization of Feature Manifold} 
\begin{figure*}[t] 
    \centering

    \newcommand{\subfigwidth}{0.24\textwidth}

    \begin{subfigure}[c]{\subfigwidth}
        \centering
        \small \textit{Layer 10}\\[1mm]
        \includegraphics[width=\linewidth]{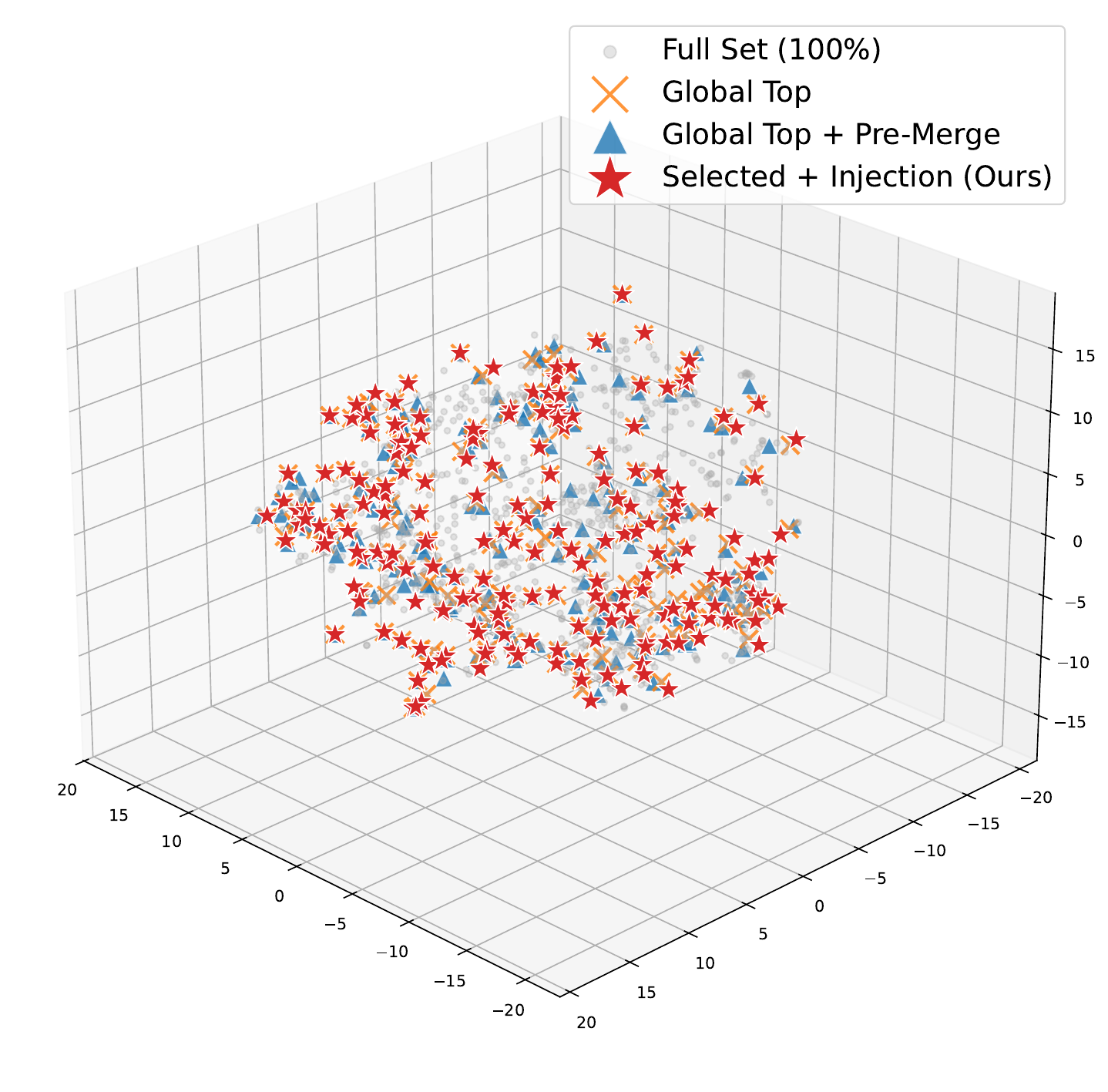}
    \end{subfigure}\hfill
    \begin{subfigure}[c]{\subfigwidth}
        \centering
        \small \textit{Layer 11}\\[1mm]
        \includegraphics[width=\linewidth]{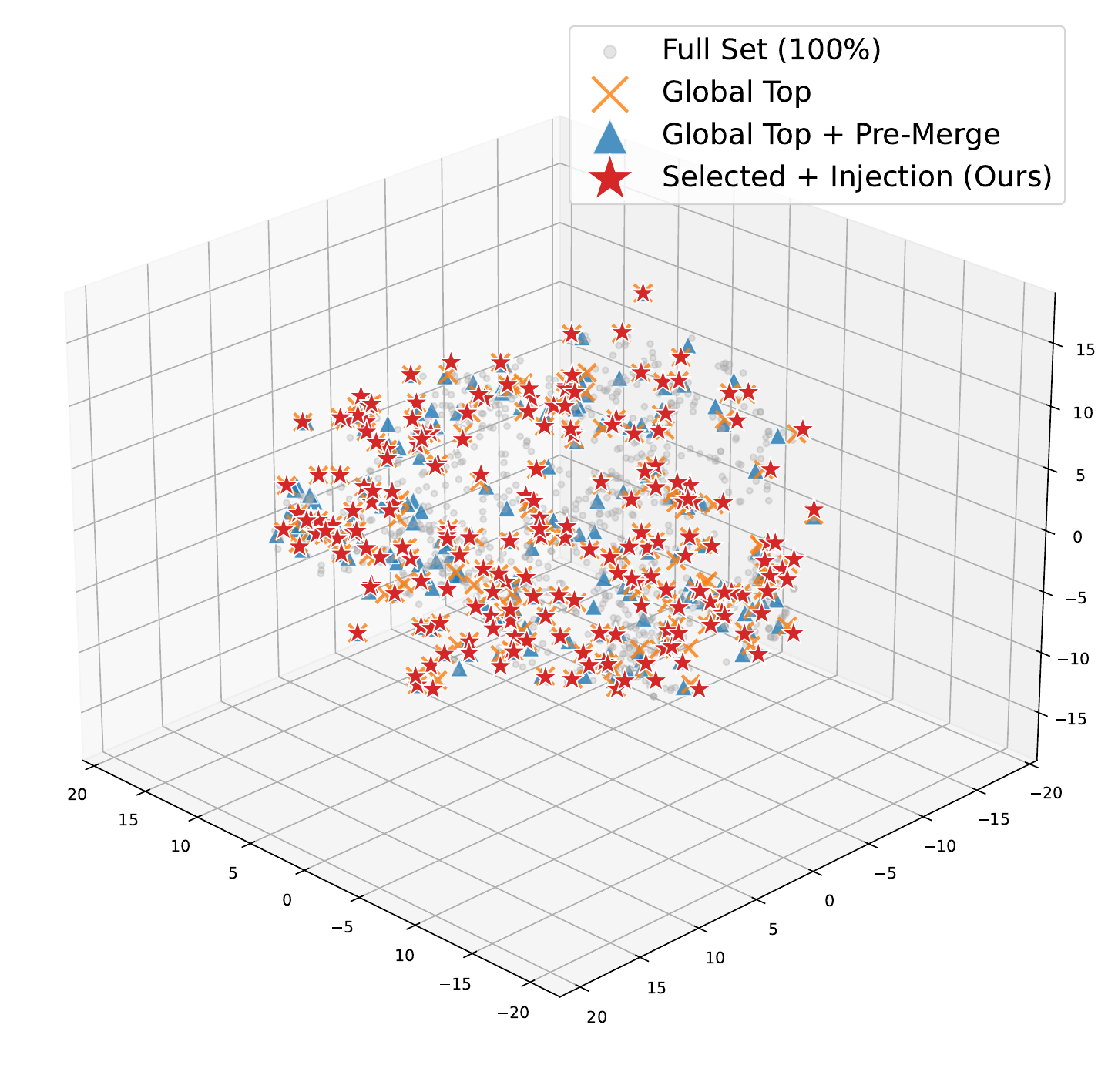}
    \end{subfigure}\hfill
    \begin{subfigure}[c]{\subfigwidth}
        \centering
        \small \textit{Layer 12}\\[1mm]
        \includegraphics[width=\linewidth]{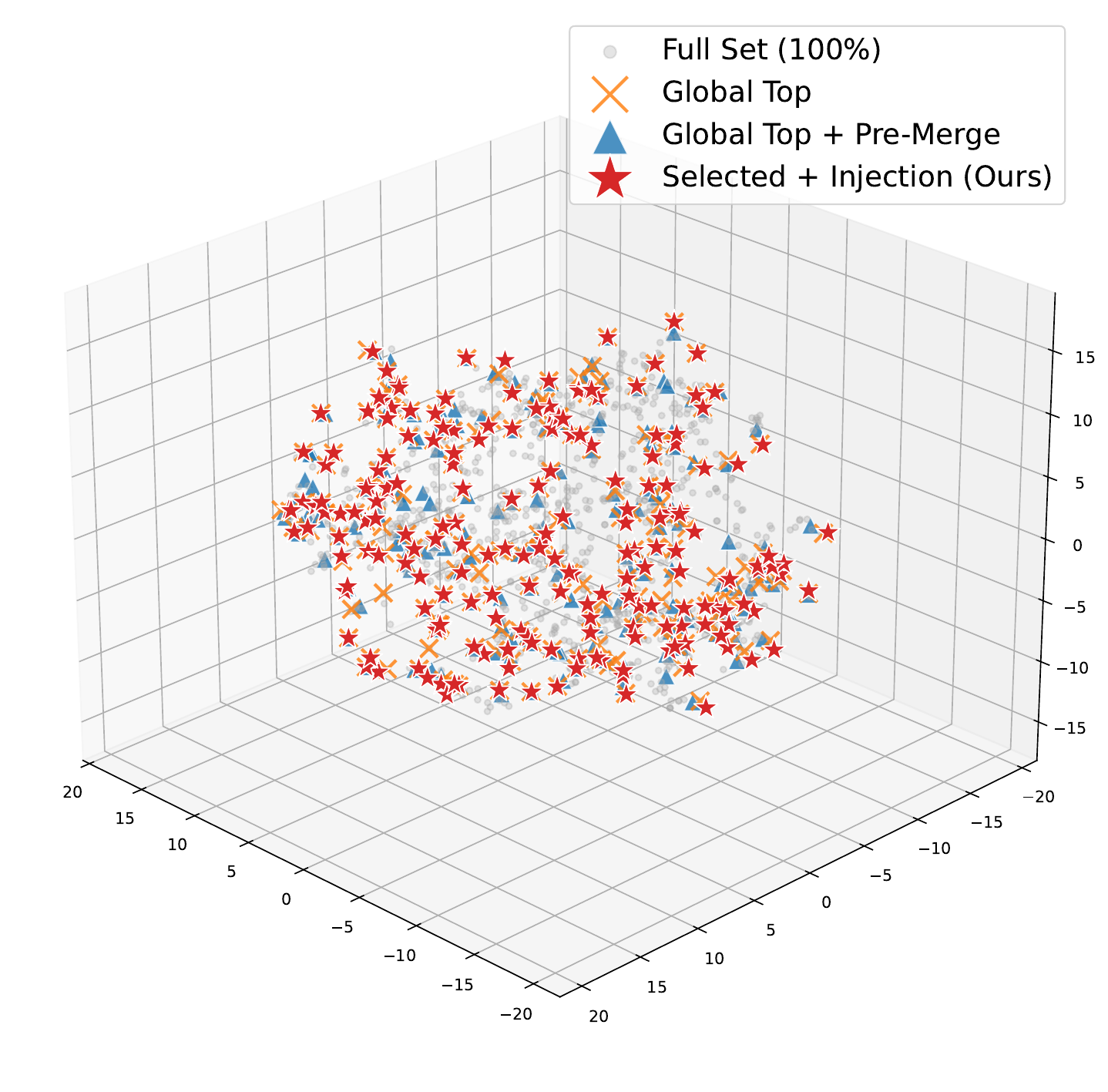}
    \end{subfigure}\hfill
    \begin{subfigure}[c]{\subfigwidth}
        \centering
        \small \textit{Layer 13}\\[1mm]
        \includegraphics[width=\linewidth]{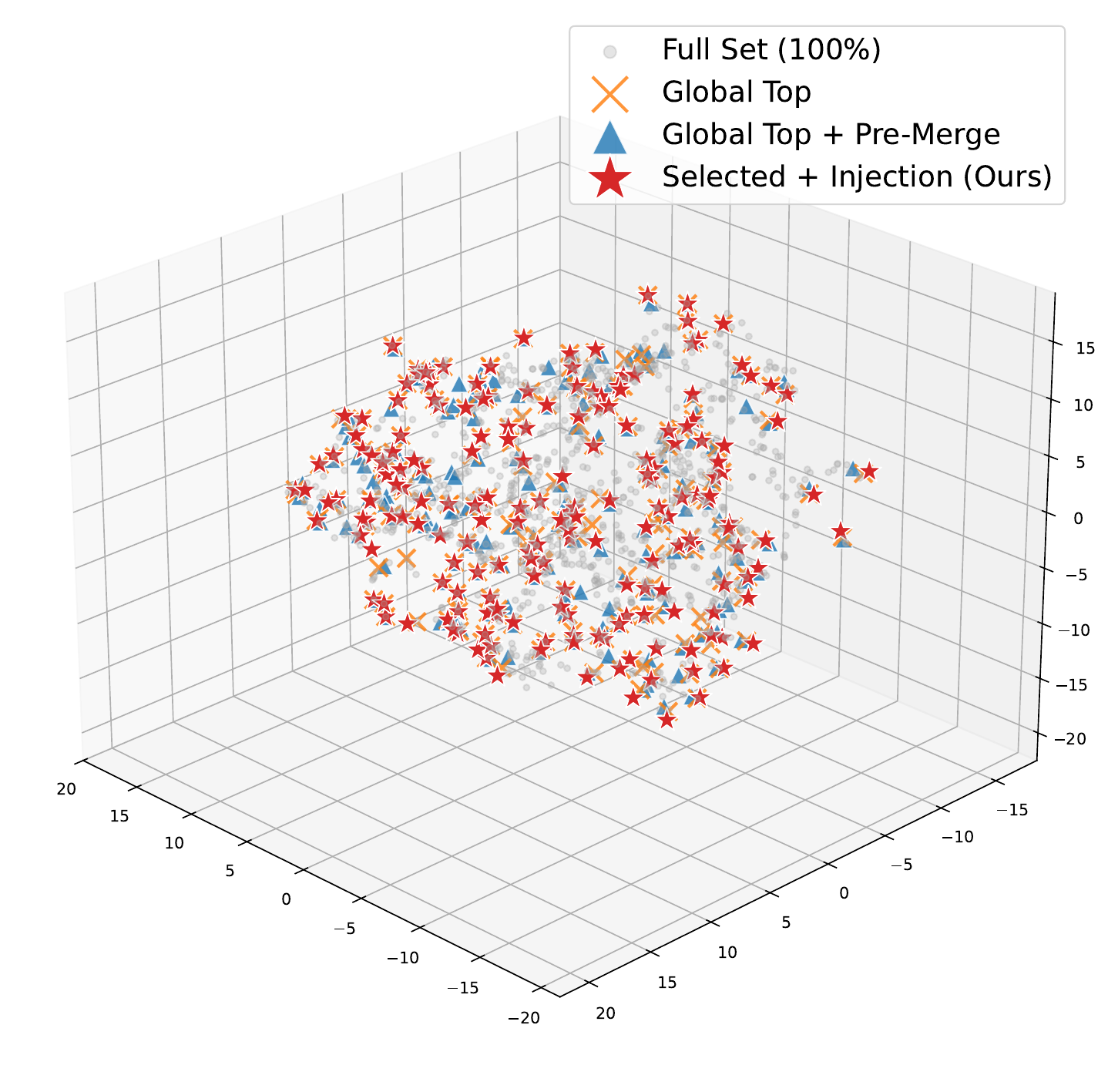}
    \end{subfigure}\\[1.5mm]

    \begin{subfigure}[c]{\subfigwidth}
        \centering
        \small \textit{Layer 14}\\[1mm]
        \includegraphics[width=\linewidth]{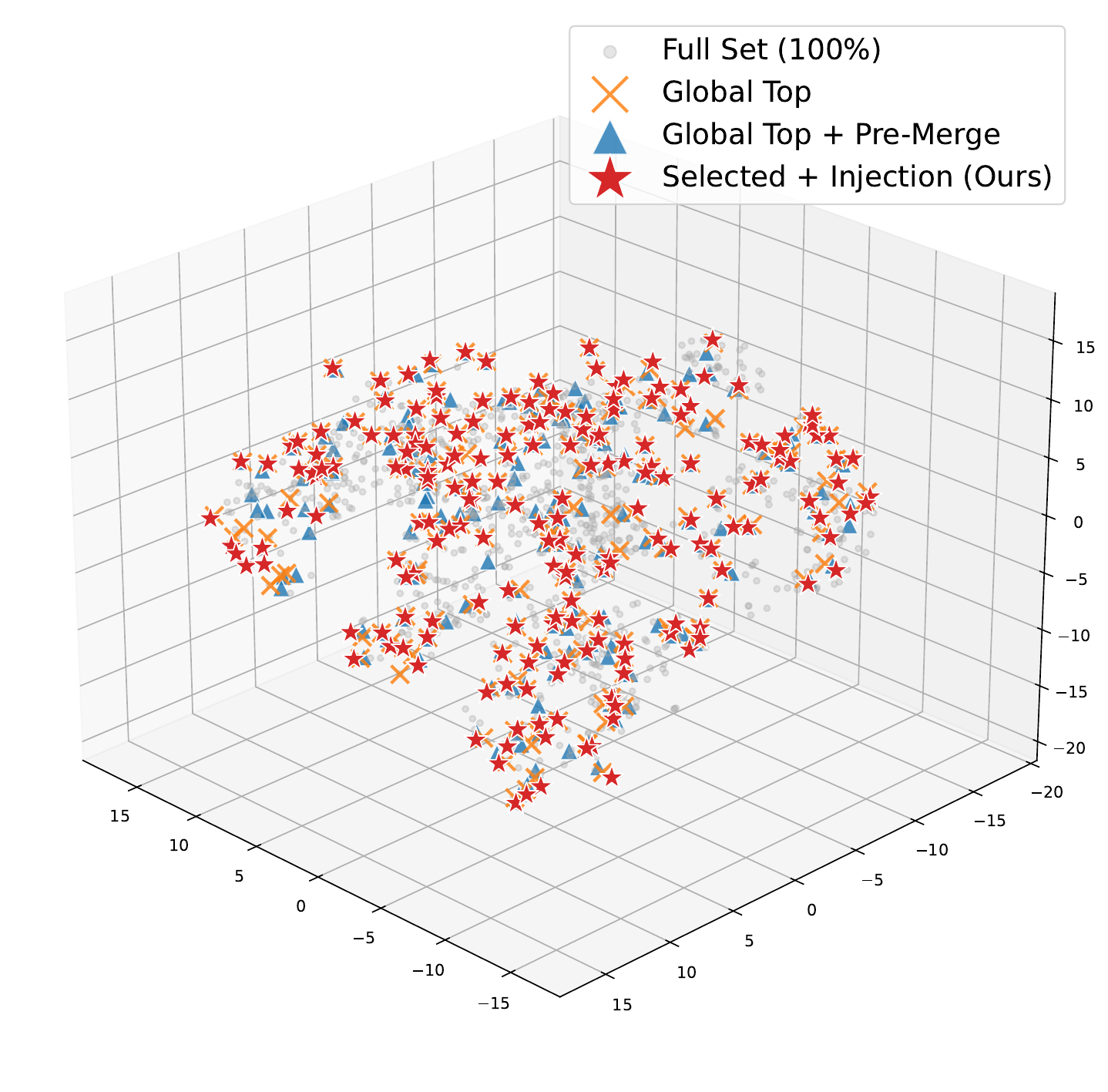}
    \end{subfigure}\hfill
    \begin{subfigure}[c]{\subfigwidth}
        \centering
        \small \textit{Layer 15}\\[1mm]
        \includegraphics[width=\linewidth]{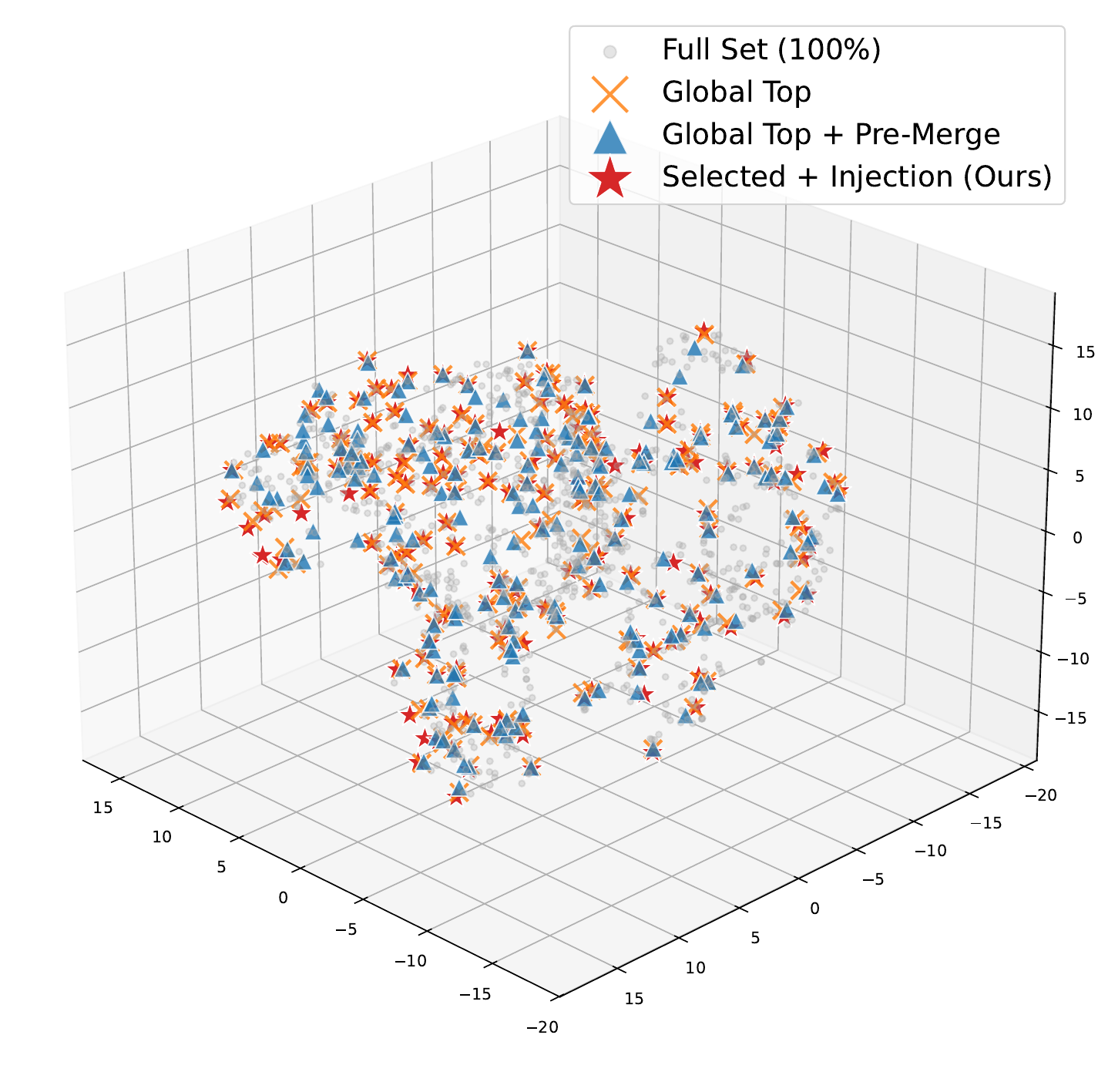}
    \end{subfigure}\hfill
    \begin{subfigure}[c]{\subfigwidth}
        \centering
        \small \textit{Layer 16}\\[1mm]
        \includegraphics[width=\linewidth]{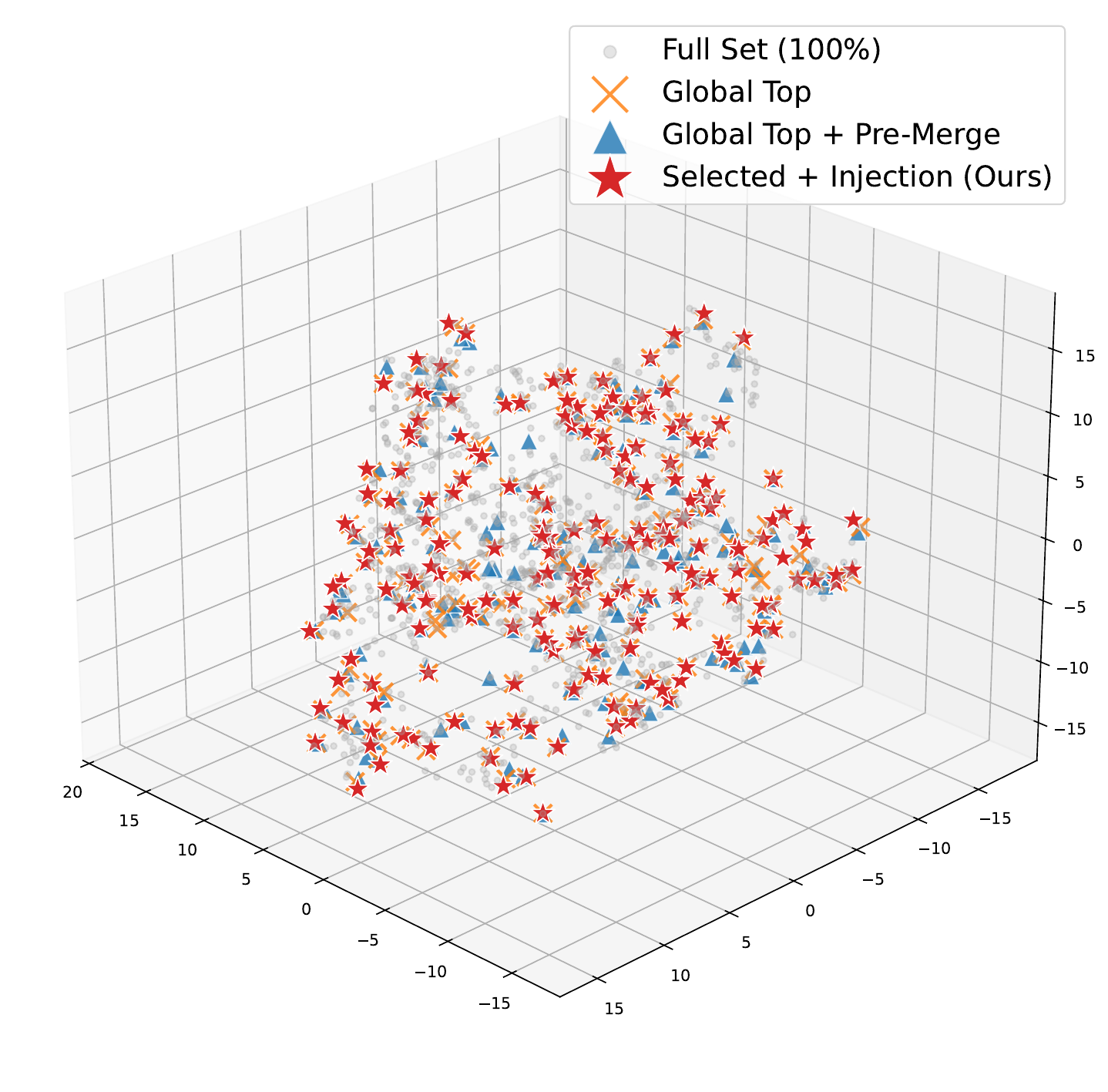}
    \end{subfigure}\hfill
    \begin{subfigure}[c]{\subfigwidth}
        \centering
        \small \textit{Layer 17}\\[1mm]
        \includegraphics[width=\linewidth]{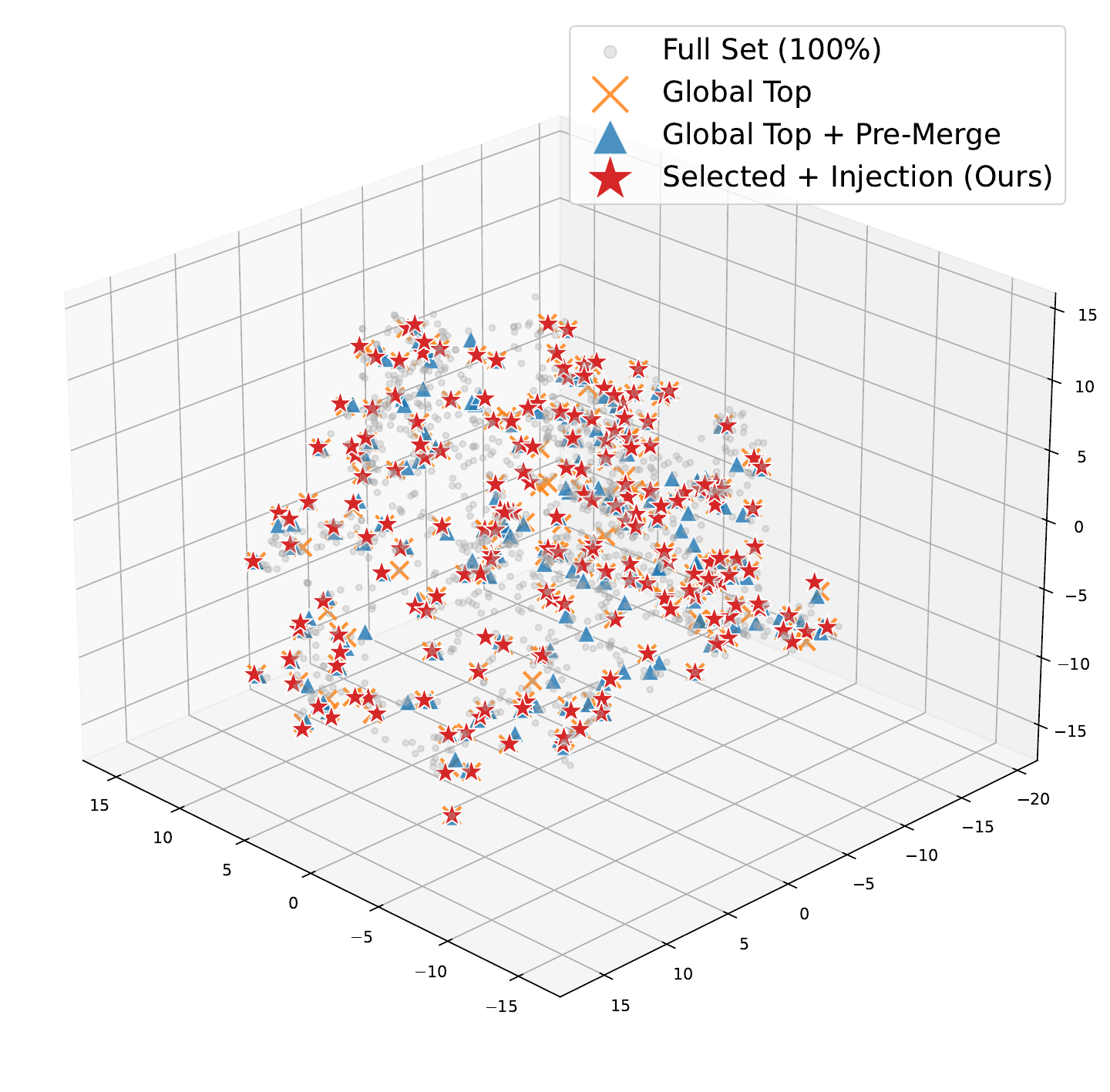}
    \end{subfigure}\\[1.5mm]

    \begin{subfigure}[c]{\subfigwidth}
        \centering
        \small \textit{Layer 18}\\[1mm]
        \includegraphics[width=\linewidth]{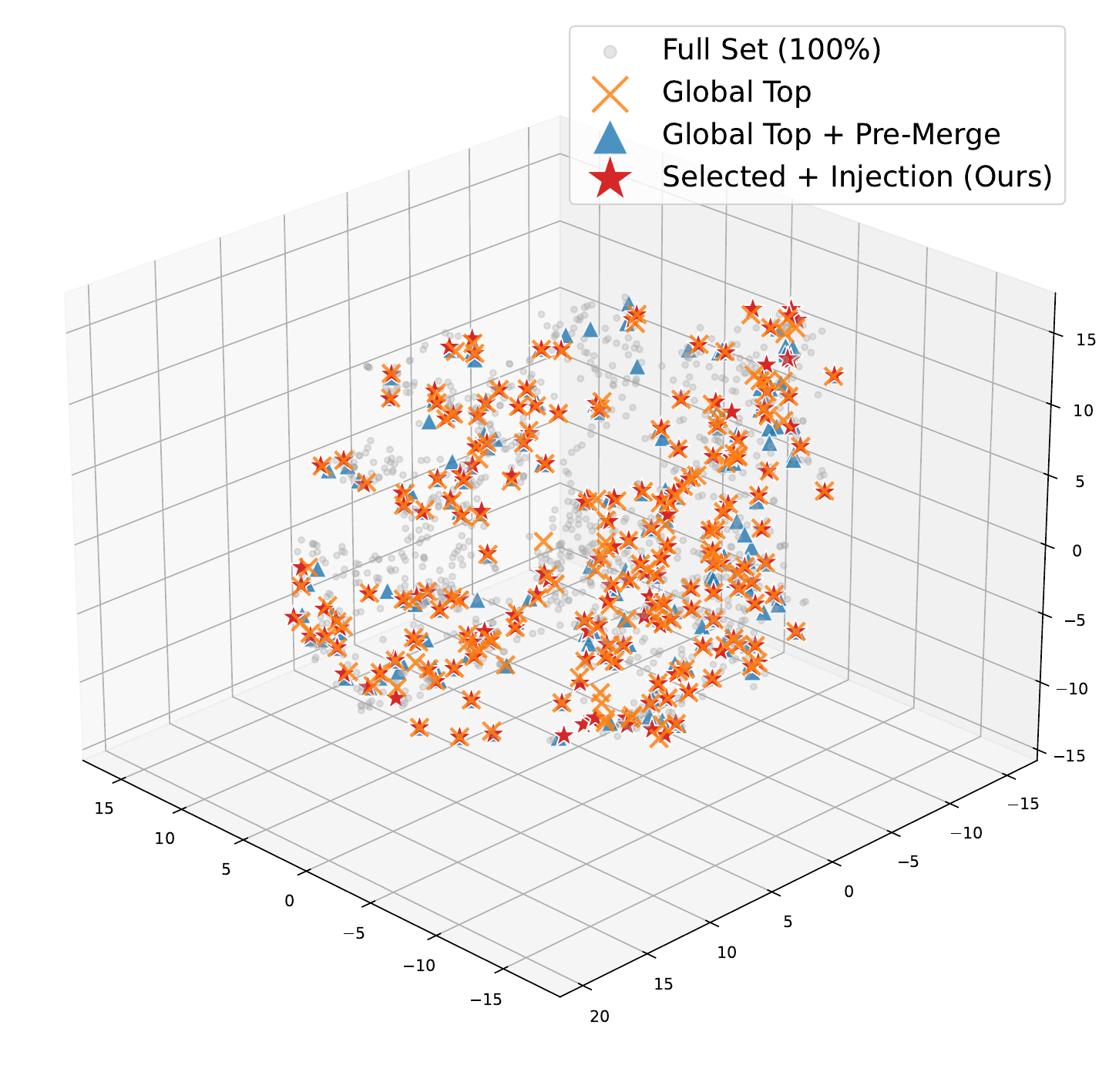}
    \end{subfigure}\hfill
    \begin{subfigure}[c]{\subfigwidth}
        \centering
        \small \textit{Layer 19}\\[1mm]
        \includegraphics[width=\linewidth]{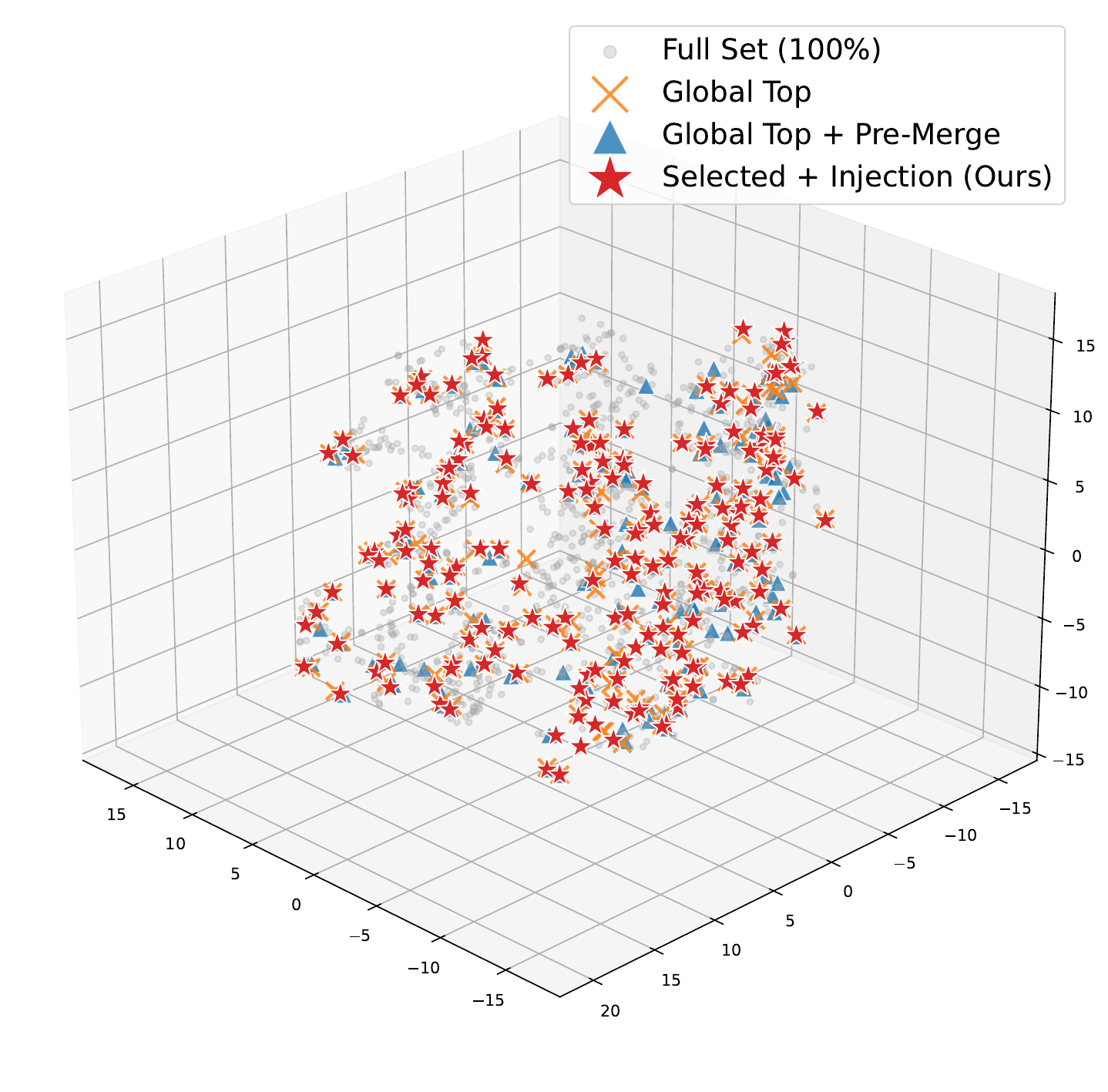}
    \end{subfigure}\hfill
    \begin{subfigure}[c]{\subfigwidth}
        \centering
        \small \textit{Layer 20}\\[1mm]
        \includegraphics[width=\linewidth]{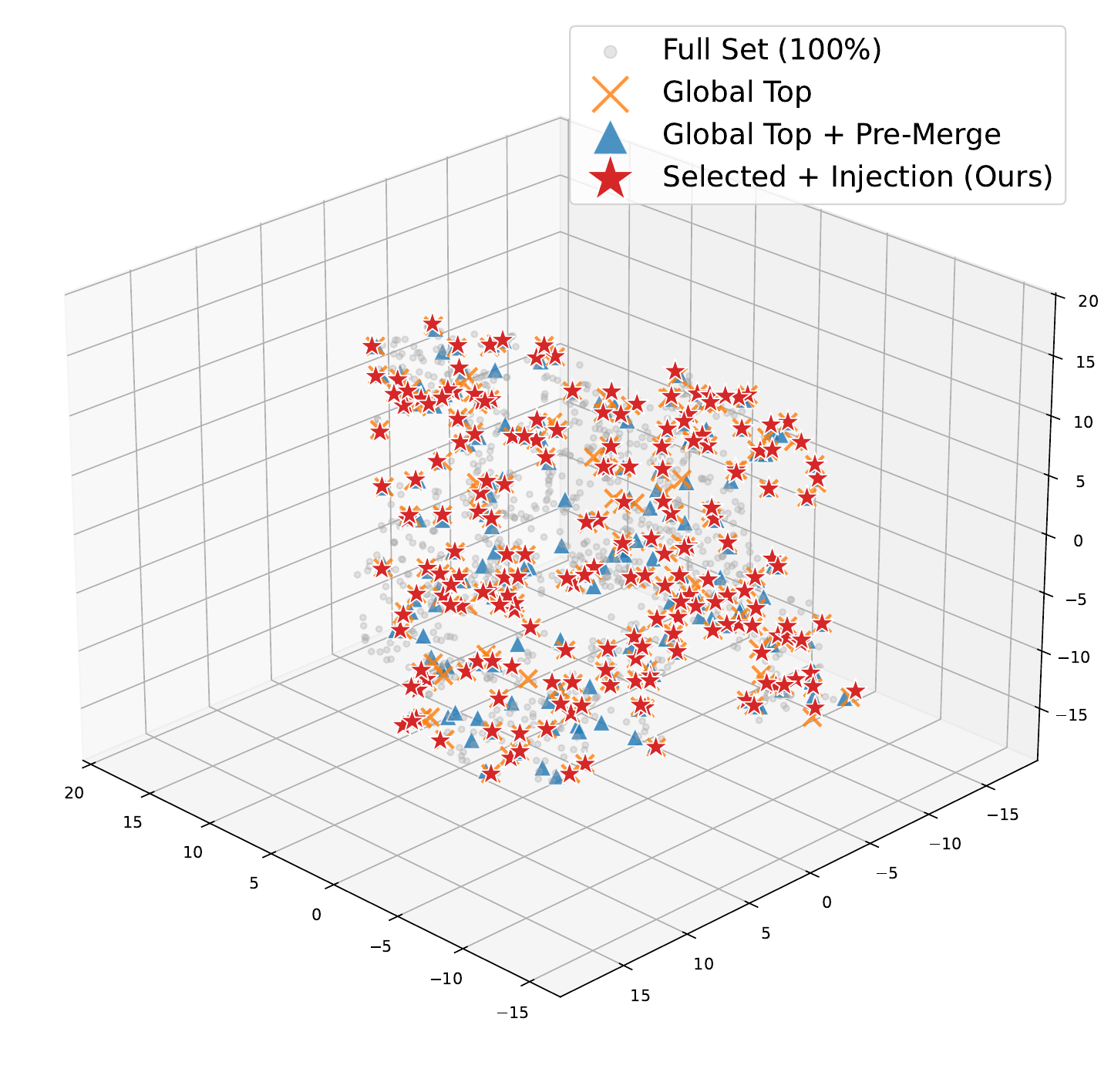}
    \end{subfigure}\hfill
    \begin{subfigure}[c]{\subfigwidth}
        \centering
        \small \textit{Layer 21}\\[1mm]
        \includegraphics[width=\linewidth]{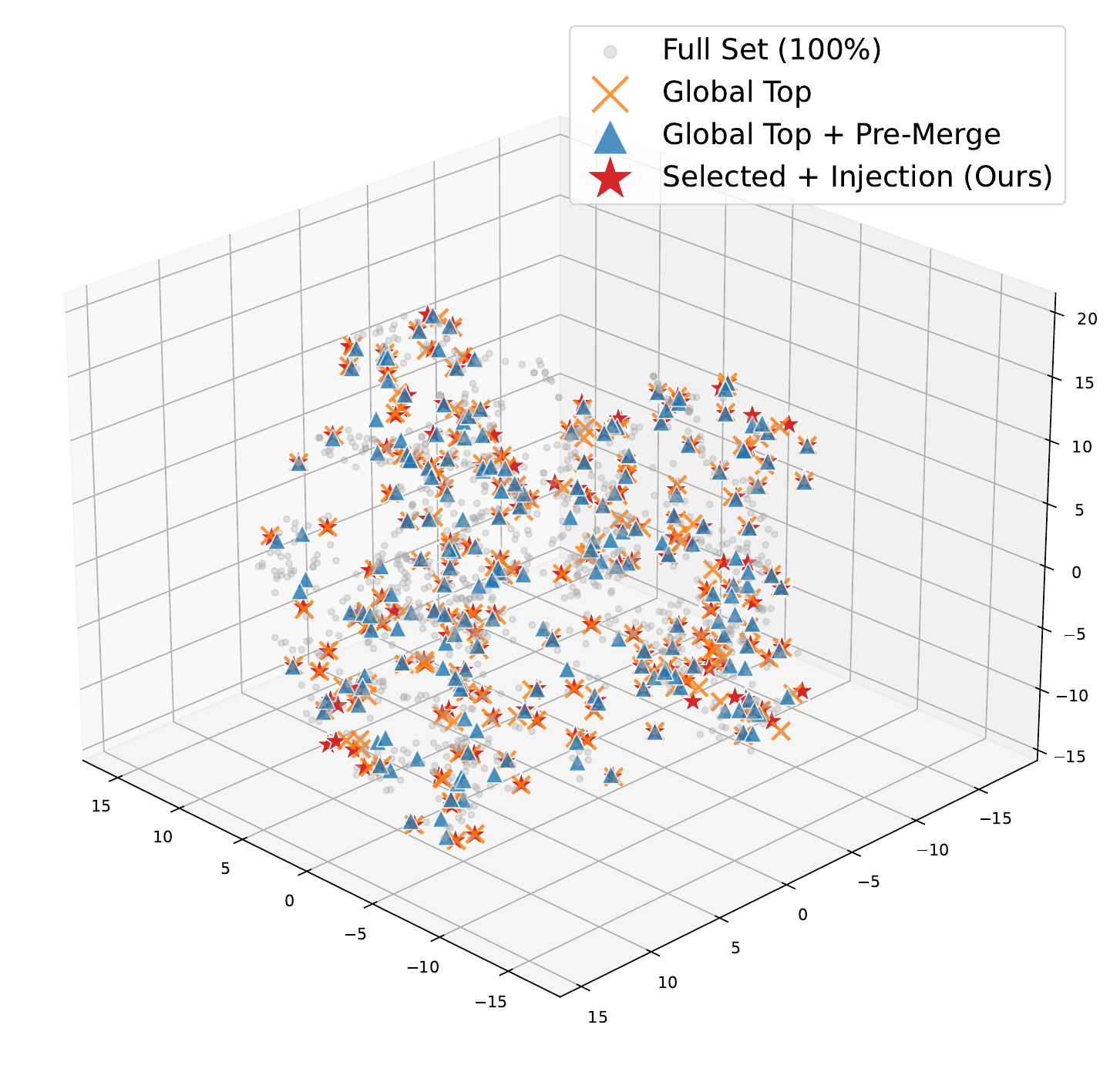}
    \end{subfigure}

    \caption{\textbf{Evolution of 3D feature representations across intermediate layers.}}
    \label{fig:layer_evolution_3d}
\end{figure*}

To evaluate the representational integrity of different token reduction strategies, we project the hidden states of visual tokens from intermediate layers into a 3D space using t-SNE. As shown in Fig.~\ref{fig:layer_evolution_3d}, the full token set establishes the reference distribution manifold. Baseline methods, including global top selection and pre-merging, exhibit structural deviations from this reference, frequently resulting in localized clusters that fail to span the original feature space. In contrast, the application of intermediate token injection yields a spatial distribution that consistently covers the reference manifold across most observed layers. This indicates that the injection mechanism preserves the topological structure of the visual features, preventing the representational collapse observed in standard pruning approaches.

\section{More Results}
\label{sec:more_main}
This section extends our quantitative evaluations across diverse parameter scales, presenting detailed performance metrics for both smaller MLLMs (2B/3B) and a larger 32B regime.
\subsection{More Main Results}
\label{sec:more_main_results}
We provide here the results of the experimental models used in the main text under additional compression-ratio settings, as shown in the Tab.~\ref{tab:internvl3-8b_app}, Tab.~\ref{tab:qwenvl3b_more} and Tab.~\ref{tab:internvl2b_more}.
\begin{table*}[t]
\centering

\resizebox{\textwidth}{!}{%
\begin{tabular}{l |c|cccccccc |c} 
\toprule
\rowcolor[HTML]{EAF3FF}
\textbf{Method} & \textbf{Venue} & \textbf{MMB} & \textbf{MMB\textsuperscript{CN}} & \textbf{MME} & \textbf{POPE} & \textbf{SQA\textsubscript{I}} & \textbf{MMStar} & \textbf{MRW} & \textbf{MRW\textsuperscript{CN}} & \textbf{Avg.} \\
\midrule

\rowcolor[HTML]{F6F6F6}
InternVL3-8B & - & 83.5 & 83.4 & 2415 & 90.7 & 98.2 & 63.9 & 62.0 & 60.8 & 100.0\% \\
\midrule

\rowcolor[HTML]{F1F6FB}
\multicolumn{11}{c}{\textit{Token Reduction Setting: Retain 33.3\% Tokens ($\downarrow$66.7\%)}} \\
\midrule
FastV \cite{fastv} & \textnormal{\textit{ECCV'24}} & 84.2 & 84.3 & 2347 & 90.0 & 96.2 & 59.2 & 50.2 & 54.7 & 95.1\% \\
VisionZip \cite{visionzip} & \textnormal{\textit{CVPR'25}} & 81.3 & 84.3 & 2290 & 89.4 & 97.1 & 60.5 & 50.8 & 53.8 & 94.2\% \\
HoloV \cite{holov} & \textnormal{\textit{NeurIPS'25}} & 82.4 & 84.2 & 2285 & 89.7 & 97.3 & 60.6 & 50.2 & 53.8 & 94.4\% \\
\rowcolor[HTML]{E8F2FF}
\textbf{Ours} & - & \textbf{85.1} & \textbf{85.7} & \textbf{2351} & \textbf{90.5} & \textbf{97.5} & \textbf{61.2} & \textbf{53.0} & \textbf{55.9} & \textbf{96.8\%} \\
\midrule

\rowcolor[HTML]{F1F6FB}
\multicolumn{11}{c}{\textit{Token Reduction Setting: Retain 22.2\% Tokens ($\downarrow$77.8\%)}} \\
\midrule
FastV \cite{fastv} & \textnormal{\textit{ECCV'24}} & \textbf{83.8} & 83.0 & \textbf{2324} & 89.4 & 95.4 & 58.2 & 46.3 & 51.9 & 92.7\% \\
VisionZip \cite{visionzip} & \textnormal{\textit{CVPR'25}} & 80.3 & 82.0 & 2242 & 88.8 & 95.0 & 56.8 & 44.8 & 49.5 & 90.4\% \\
HoloV \cite{holov} & \textnormal{\textit{NeurIPS'25}} & 80.6 & 82.3 & 2201 & 88.7 & 95.4 & 57.0 & 44.7 & 50.0 & 90.5\% \\
\rowcolor[HTML]{E8F2FF}
\textbf{Ours} & - & 83.6 & \textbf{84.3} & 2319 & \textbf{90.2} & \textbf{96.7} & \textbf{59.7} & \textbf{48.7} & \textbf{53.3} & \textbf{94.4\%} \\
\midrule

\rowcolor[HTML]{F1F6FB}
\multicolumn{11}{c}{\textit{Token Reduction Setting: Retain 11.1\% Tokens ($\downarrow$88.9\%)}} \\
\midrule
FastV \cite{fastv} & \textnormal{\textit{ECCV'24}} & 79.5 & 78.7 & 2202 & 88.8 & 93.2 & 52.3 & 39.9 & 47.3 & 87.2\% \\
VisionZip \cite{visionzip} & \textnormal{\textit{CVPR'25}} & 78.7 & 76.8 & 1978 & 85.9 & 87.7 & 50.2 & 34.5 & 43.7 & 82.2\% \\
HoloV \cite{holov} & \textnormal{\textit{NeurIPS'25}} & 79.8 & 78.1 & 1964 & 87.9 & 89.4 & 51.5 & 36.1 & 44.2 & 83.9\% \\
\rowcolor[HTML]{E8F2FF}
\textbf{Ours} & - & \textbf{80.9} & \textbf{80.7} & \textbf{2210} & \textbf{89.0} & \textbf{94.1} & \textbf{55.9} & \textbf{41.5} & \textbf{48.6} & \textbf{89.4\%} \\
\bottomrule
\end{tabular}%
}
\caption{\textbf{Performance comparisons on InternVL3-8B across 8 benchmarks.}}
\label{tab:internvl3-8b_app}
\end{table*}
\subsection{More Quantitative Results on Smaller MLLMs}
\label{sec:more_quantitative}

In this section, we provide extended quantitative evaluations on Qwen2.5VL-3B and InternVL3-2B to complement the main manuscript. As detailed in Tab.~\ref{tab:qwenvl3b_more} and Tab.~\ref{tab:internvl2b_more}, we report the performance across multiple token retention settings.

The comprehensive results further demonstrate the robust superiority of ReTrace. Across all compression scales, our method consistently outperforms recent state-of-the-art baselines including FastV, VisionZip, and Vscan. The performance gap becomes particularly evident under highly restricted token budgets. For instance, on the InternVL3-2B model at an 11.1\% retention rate, ReTrace achieves an average score of 82.5\%, exceeding the strongest baseline (HoloV) by +5.9\%. This consistent margin across diverse architectures and sparsity levels confirms that our hybrid extraction strategy effectively balances inference acceleration and reasoning capacity, independent of the model's parameter scale.

\begin{table*}[t]
\centering

\resizebox{\textwidth}{!}{%
\begin{tabular}{l |c|cccccccc |c} 
\toprule
\rowcolor[HTML]{EAF3FF}
\textbf{Method} & \textbf{Venue} & \textbf{MMB} & \textbf{MMB\textsuperscript{CN}} & \textbf{MME} & \textbf{POPE} & \textbf{SQA\textsubscript{I}} & \textbf{MMStar} & \textbf{MRW} & \textbf{MRW\textsuperscript{CN}} & \textbf{Avg.} \\
\midrule

\rowcolor[HTML]{F6F6F6}
Qwen2.5VL-3B & - & 79.1 & 78.1 & 2157 & 87.3 & 81.0 & 55.9 & 53.1 & 45.1 & 100.0\% \\
\midrule

\rowcolor[HTML]{F1F6FB}
\multicolumn{11}{c}{\textit{Token Reduction Setting: Retain 33.3\% Tokens ($\downarrow$66.7\%)}} \\
\midrule
FastV \cite{fastv} & \textnormal{\textit{ECCV'24}} & 73.7 & 73.4 & 2026 & 84.2 & 80.3 & 44.4 & 39.2 & 36.7 & 88.9\% \\
VisionZip \cite{visionzip} & \textnormal{\textit{CVPR'25}} & 76.0 & 73.6 & 1964 & 84.2 & 79.8 & \textbf{48.0} & 40.2 & 37.7 & 90.2\% \\
Dart \cite{dart} & \textnormal{\textit{EMNLP'25}} & 75.3 & 73.7 & 1950 & 82.8 & 78.9 & 36.8 & 40.3 & 36.0 & 86.7\% \\
HoloV \cite{holov} & \textnormal{\textit{NeurIPS'25}} & 76.1 & 74.0 & 2015 & 83.5 & 81.2 & 44.1 & 36.3 & 33.5 & 87.7\% \\
VLMPruner \cite{vlmpruner} & \textnormal{\textit{CVPR'26}} & 75.8 & 73.2 & 1991 & 83.9 & 81.1 & 45.7 & 39.7 & 35.2 & 89.1\% \\
Vscan \cite{vscan} & \textnormal{\textit{TMLR'26}} & 72.8 & 72.5 & 1977 & 82.7 & 80.1 & 41.2 & 34.3 & 33.5 & 85.3\% \\
\rowcolor[HTML]{E8F2FF}
\textbf{Ours} & - & \textbf{76.4} & \textbf{74.2} & \textbf{2091} & \textbf{85.5} & \textbf{82.4} & 47.5 & \textbf{41.4} & \textbf{39.0} & \textbf{92.2\%} \\
\midrule

\rowcolor[HTML]{F1F6FB}
\multicolumn{11}{c}{\textit{Token Reduction Setting: Retain 22.2\% Tokens ($\downarrow$77.8\%)}} \\
\midrule
FastV \cite{fastv} & \textnormal{\textit{ECCV'24}} & 72.8 & 71.9 & 1918 & 82.3 & 80.2 & 42.1 & 34.7 & 33.4 & 85.1\% \\
VisionZip \cite{visionzip} & \textnormal{\textit{CVPR'25}} & 75.1 & 71.8 & 1795 & 83.4 & \textbf{81.2} & 44.0 & 37.7 & 35.2 & 86.7\% \\
Dart \cite{dart} & \textnormal{\textit{EMNLP'25}} & 74.9 & 70.6 & 1804 & 80.0 & 78.9 & 36.8 & 36.3 & 33.6 & 83.3\% \\
HoloV \cite{holov} & \textnormal{\textit{NeurIPS'25}} & 74.7 & 71.9 & 1795 & 81.1 & 79.6 & 41.7 & 32.7 & 29.7 & 82.9\% \\
VLMPruner \cite{vlmpruner} & \textnormal{\textit{CVPR'26}} & 73.2 & 71.1 & 1935 & 79.2 & 79.7 & 41.2 & 34.1 & 32.7 & 84.1\% \\
Vscan \cite{vscan} & \textnormal{\textit{TMLR'26}} & 70.6 & 68.5 & 1838 & 80.9 & 79.7 & 40.4 & 32.6 & 32.4 & 82.3\% \\
\rowcolor[HTML]{E8F2FF}
\textbf{Ours} & - & \textbf{75.7} & \textbf{72.5} & \textbf{2003} & \textbf{83.6} & 81.1 & \textbf{45.7} & \textbf{38.0} & \textbf{36.6} & \textbf{89.0\%} \\
\midrule

\rowcolor[HTML]{F1F6FB}
\multicolumn{11}{c}{\textit{Token Reduction Setting: Retain 11.1\% Tokens ($\downarrow$88.9\%)}} \\
\midrule
FastV \cite{fastv} & \textnormal{\textit{ECCV'24}} & 71.0 & 69.9 & 1698 & 79.0 & 79.0 & 36.6 & 30.8 & 32.2 & 80.1\% \\
VisionZip \cite{visionzip} & \textnormal{\textit{CVPR'25}} & \textbf{73.2} & 70.0 & 1698 & \textbf{81.0} & 79.8 & 40.1 & 31.3 & 31.9 & 81.7\% \\
Dart \cite{dart} & \textnormal{\textit{EMNLP'25}} & 72.8 & 65.8 & 1640 & 73.8 & 78.9 & 36.8 & 29.6 & 29.4 & 77.6\% \\
HoloV \cite{holov} & \textnormal{\textit{NeurIPS'25}} & 70.8 & 69.0 & 1673 & 76.4 & 77.5 & 38.6 & 26.1 & 25.1 & 76.6\% \\
VLMPruner \cite{vlmpruner} & \textnormal{\textit{CVPR'26}} & 68.2 & 66.7 & 1735 & 74.1 & 73.9 & 36.0 & 27.5 & 26.7 & 75.4\% \\
Vscan \cite{vscan} & \textnormal{\textit{TMLR'26}} & 69.6 & 63.6 & 1763 & 76.2 & 78.1 & 35.7 & 28.8 & 29.2 & 77.2\% \\
\rowcolor[HTML]{E8F2FF}
\textbf{Ours} & - & 72.3 & \textbf{70.6} & \textbf{1781} & 77.4 & \textbf{80.0} & \textbf{40.3} & \textbf{32.2} & \textbf{32.5} & \textbf{82.1\%} \\
\bottomrule
\end{tabular}%
}
\caption{\textbf{Performance comparisons on Qwen2.5VL-3B across 8 benchmarks.} The ``Avg.'' metric denotes the mean relative performance retention across all datasets compared to the baseline.}
\label{tab:qwenvl3b_more}
\end{table*}

\begin{table*}[t]
\centering

\resizebox{\textwidth}{!}{%
\begin{tabular}{l |c|cccccccc |c} 
\toprule
\rowcolor[HTML]{EAF3FF}
\textbf{Method} & \textbf{Venue} & \textbf{MMB} & \textbf{MMB\textsuperscript{CN}} & \textbf{MME} & \textbf{POPE} & \textbf{SQA\textsubscript{I}} & \textbf{MMStar} & \textbf{MRW} & \textbf{MRW\textsuperscript{CN}} & \textbf{Avg.} \\
\midrule

\rowcolor[HTML]{F6F6F6}
InternVL3-2B & - & 80.8 & 79.0 & 2221 & 89.9 & 95.7 & 60.7 & 53.8 & 52.1 & 100.0\% \\
\midrule

\rowcolor[HTML]{F1F6FB}
\multicolumn{11}{c}{\textit{Token Reduction Setting: Retain 33.3\% Tokens ($\downarrow$66.7\%)}} \\
\midrule
FastV \cite{fastv} & \textnormal{\textit{ECCV'24}} & 67.2 & 65.5 & 1570 & 75.3 & 79.6 & 43.5 & 25.5 & 33.2 & 73.3\% \\
VisionZip \cite{visionzip} & \textnormal{\textit{CVPR'25}} & 78.9 & 76.3 & 2041 & 87.2 & 93.2 & 51.8 & 33.5 & 41.9 & 88.6\% \\
HoloV \cite{holov} & \textnormal{\textit{NeurIPS'25}} & 79.2 & 76.0 & 1997 & 88.0 & 92.9 & 52.8 & 34.2 & 42.6 & 88.9\% \\
\rowcolor[HTML]{E8F2FF}
\textbf{Ours} & - & \textbf{80.6} & \textbf{76.9} & \textbf{2085} & \textbf{89.2} & \textbf{94.2} & \textbf{53.7} & \textbf{37.1} & \textbf{44.9} & \textbf{91.5\%} \\
\midrule

\rowcolor[HTML]{F1F6FB}
\multicolumn{11}{c}{\textit{Token Reduction Setting: Retain 22.2\% Tokens ($\downarrow$77.8\%)}} \\
\midrule
FastV \cite{fastv} & \textnormal{\textit{ECCV'24}} & 63.7 & 61.8 & 1472 & 70.6 & 77.6 & 40.5 & 21.5 & 29.5 & 68.3\% \\
VisionZip \cite{visionzip} & \textnormal{\textit{CVPR'25}} & 75.6 & 73.2 & 1949 & 85.2 & 90.5 & 49.2 & 28.7 & 38.4 & 83.9\% \\
HoloV \cite{holov} & \textnormal{\textit{NeurIPS'25}} & 76.0 & 66.7 & 1934 & 85.9 & 90.9 & 50.0 & 29.4 & 39.2 & 83.5\% \\
\rowcolor[HTML]{E8F2FF}
\textbf{Ours} & - & \textbf{78.9} & \textbf{75.5} & \textbf{2072} & \textbf{88.1} & \textbf{93.5} & \textbf{53.5} & \textbf{32.2} & \textbf{41.2} & \textbf{88.7\%} \\
\midrule

\rowcolor[HTML]{F1F6FB}
\multicolumn{11}{c}{\textit{Token Reduction Setting: Retain 11.1\% Tokens ($\downarrow$88.9\%)}} \\
\midrule
FastV \cite{fastv} & \textnormal{\textit{ECCV'24}} & 54.0 & 50.8 & 1398 & 64.7 & 74.5 & 36.4 & 15.6 & 25.9 & 60.3\% \\
VisionZip \cite{visionzip} & \textnormal{\textit{CVPR'25}} & 67.7 & 66.3 & 1747 & 79.4 & 84.4 & 42.5 & 19.7 & 32.9 & 74.1\% \\
HoloV \cite{holov} & \textnormal{\textit{NeurIPS'25}} & 70.1 & 66.7 & 1743 & 81.8 & 85.8 & 44.8 & 22.6 & 34.8 & 76.6\% \\
\rowcolor[HTML]{E8F2FF}
\textbf{Ours} & - & \textbf{74.3} & \textbf{71.7} & \textbf{1965} & \textbf{84.4} & \textbf{89.8} & \textbf{47.9} & \textbf{27.5} & \textbf{37.1} & \textbf{82.5\%} \\
\bottomrule
\end{tabular}%
}
\caption{\textbf{Performance comparisons on InternVL3-2B across 8 benchmarks.} The ``Avg.'' metric denotes the mean relative performance retention across all datasets compared to the baseline.}
\label{tab:internvl2b_more}
\end{table*}

\subsection{Scalability to Larger Regimes (32B)}

To validate the scalability of our approach, we further extend the evaluation to the 32B-parameter regime using Qwen2.5-VL-32B. As reported in Tab.~\ref{tab:mmbench_accuracy_32b}, Tab.~\ref{tab:mmbench_cn_32b}, and Tab.~\ref{tab:mme_scores_32b}, our method consistently outperforms recent visual-token reduction baselines, including FastV~\cite{fastv}, PDrop~\cite{pyramiddrop2025}, and Vscan~\cite{vscan}, across all tested retention ratios and benchmarks. The advantage is especially clear under aggressive compression. At the extreme 11\% retention ratio, our method achieves 82.21 on MMBench, 80.67 on MMBench-CN, and 2037 on MME, outperforming the strongest baseline Vscan by +4.12, +3.44, and +22 points, respectively. The gains become even more pronounced on MME at moderate compression levels: at 22\% and 33\% retention, our method surpasses Vscan by +117 and +85 points, indicating that the proposed cache--inject--prune schedule better preserves the visual evidence needed by large-scale MLLMs. Moreover, when the retention ratio reaches 50\%, our method nearly matches the uncompressed 100\% upper bound, with only a 0.52-point drop on MMBench, a 1.04-point drop on MMBench-CN, and an 18-point drop on MME. These results suggest that the proposed multi-stage mechanism remains effective and robust when scaled to massive MLLMs, maintaining strong accuracy while substantially reducing the visual token budget.

\subsection{Comparison with Additional Compression Methods}

The baselines selected in the main experiments already cover representative visual-token compression paradigms. For completeness, we further include additional methods under different retention ratios. As shown in Tab.~\ref{tab:retention_comparison}, ReTrace continues to maintain strong overall performance across diverse benchmarks.

\begin{table*}[h]
\centering

\begin{tabular}{l|l| ccccc}
\toprule
\textbf{Model} & \textbf{Method} & \textbf{11\%} & \textbf{22\%} & \textbf{33\%} & \textbf{50\%} & \textbf{100\%} \\
\midrule
\multirow{4}{*}{\textbf{32B}} & FastV & 70.13 & 78.09 & 81.01 & 85.40 & 87.37 \\
 & PDrop & 73.22 & 79.04 & 82.74 & 86.03 & 87.37 \\
 & Vscan & 78.09 & 82.47 & 85.31 & 86.17 & 87.37 \\
 \rowcolor[HTML]{E8F2FF}
 & Ours & \textbf{82.21} & \textbf{85.48} & \textbf{86.42} & \textbf{86.85} & \textbf{87.37} \\
\bottomrule
\end{tabular}
\caption{\textbf{MMBench accuracy} under different visual token retention ratios. Higher is better.}
\label{tab:mmbench_accuracy_32b}
\end{table*}

\begin{table*}[h]
\centering

\begin{tabular}{l|l| ccccc}
\toprule
\textbf{Model} & \textbf{Method} & \textbf{11\%} & \textbf{22\%} & \textbf{33\%} & \textbf{50\%} & \textbf{100\%} \\
\midrule
\multirow{4}{*}{\textbf{32B}} & FastV & 67.61 & 76.12 & 79.81 & 83.42 & 86.00 \\
 & PDrop & 71.31 & 77.65 & 80.72 & 84.28 & 86.00 \\
 & Vscan & 77.23 & 80.76 & 84.45 & 84.78 & 86.00 \\
 \rowcolor[HTML]{E8F2FF}
 & Ours & \textbf{80.67} & \textbf{83.93} & \textbf{84.53} & \textbf{84.96} & \textbf{86.00} \\
\bottomrule
\end{tabular}
\caption{\textbf{MMBench-CN accuracy} under different visual token retention ratios. Higher is better.}
\label{tab:mmbench_cn_32b}
\end{table*}

\begin{table*}[h]
\centering

\begin{tabular}{l|l| ccccc}
\toprule
\textbf{Model} & \textbf{Method} & \textbf{11\%} & \textbf{22\%} & \textbf{33\%} & \textbf{50\%} & \textbf{100\%} \\
\midrule
\multirow{4}{*}{\textbf{32B}} & FastV & 1536 & 1924 & 2124 & 2316 & 2432 \\
 & PDrop & 1536 & 1797 & 2078 & 2279 & 2432 \\
 & Vscan & 2015 & 2189 & 2288 & 2378 & 2432 \\
 \rowcolor[HTML]{E8F2FF}
 & Ours & \textbf{2037} & \textbf{2306} & \textbf{2373} & \textbf{2414} & \textbf{2432} \\
\bottomrule
\end{tabular}
\caption{\textbf{MME total scores} under different visual token retention ratios. Higher is better.}
\label{tab:mme_scores_32b}
\end{table*}

\begin{table*}[t]
\centering
\setlength{\tabcolsep}{4.2pt}
\renewcommand{\arraystretch}{1.08}

\resizebox{\linewidth}{!}{%
\begin{tabular}{@{}clcccccccccc@{}}
\toprule
\textbf{Retention} & \textbf{Method}
& \textbf{MMB} & \textbf{MMB-CN} & \textbf{MME}
& \textbf{POPE} & \textbf{SQA} & \textbf{MMStar}
& \textbf{MRW} & \textbf{MRW-CN} & \textbf{Avg.} \\
\midrule

22.2\% & HiMap
& 78.3 & 77.2 & 1935 & 83.7
& 84.2 & 48.3 & 40.9 & 39.5
& 85.7\% \\

22.2\% & PDrop
& 79.7 & 79.4 & 2017 & 83.3 & 85.3 & 49.3 & 42.3 & 39.4 & 87.2\% \\

22.2\% & G-Search
& 80.3 & 78.9 & 2103 & 84.6 & 83.9 & 50.5 & 41.8 & 39.9 & 87.9\% \\

\rowcolor[HTML]{EAF3FF}
22.2\% & \textbf{ReTrace}
& \textbf{81.7} & \textbf{80.8} & \textbf{2160}
& \textbf{86.3} & \textbf{86.1} & \textbf{52.1}
& \textbf{44.4} & \textbf{42.0} & \textbf{90.7\%} \\

\midrule

11.1\% & HiMap
& 77.2 & 75.1 & 1847 & 79.9
& 81.1 & 43.3 & 34.2 & 33.1
& 79.7\% \\

11.1\% & PDrop
& 76.9 & 75.5 & 1932 & 80.2 & 81.2 & 44.0 & 35.0 & 34.9 & 81.0\% \\

11.1\% & G-Search
& 77.5 & 74.9 & 2021 & 80.1 & 80.7 & 44.6 & 35.4 & 34.5 & 81.5\% \\

\rowcolor[HTML]{EAF3FF}
11.1\% & \textbf{ReTrace}
& \textbf{79.6} & \textbf{77.4} & \textbf{2093}
& \textbf{82.7} & \textbf{82.3} & \textbf{46.3}
& \textbf{36.4} & \textbf{37.2} & \textbf{84.4\%} \\

\bottomrule
\end{tabular}%
}

\caption{\textbf{Comparison under different visual-token retention ratios.}
ReTrace consistently outperforms representative visual-token compression methods across both retention settings.}
\label{tab:retention_comparison}
\end{table*}

\section{More Ablation Results}
\label{sec:more_abla}

\begin{table*}[t]
\centering

\begin{subtable}{0.48\textwidth}
\centering

\resizebox{\linewidth}{!}{%
\begin{tabular}{l| cccc} 
\toprule
\rowcolor[HTML]{EAF3FF}
\textbf{\makebox[2.8cm][l]{Layer Index}} & \textbf{MME} & \textbf{POPE} & \textbf{SQA} & \textbf{MMStar} \\
\midrule
4  & 2245 & 86.1 & \textbf{87.2} & 54.2 \\
6  & 2241 & 86.5 & 86.8 & 54.4 \\
\rowcolor[HTML]{E8F2FF}
\textbf{8 (Ours)} & 2240 & \textbf{87.5} & 87.1 & \textbf{55.3} \\
10 & 2229 & 86.3 & 86.6 & 54.6 \\
12 & \textbf{2248} & 86.4 & 87.1 & 54.7 \\
\bottomrule
\end{tabular}%
}
\caption{Injection layer index.}
\label{tab:sub_fusion_layers}
\end{subtable}
\hfill
\begin{subtable}{0.48\textwidth}
\centering

\resizebox{\linewidth}{!}{%
\begin{tabular}{l| cccc} 
\toprule
\rowcolor[HTML]{EAF3FF}
\textbf{\makebox[2.8cm][l]{$T_g:T_l:T_d$}} & \textbf{MME} & \textbf{POPE} & \textbf{SQA} & \textbf{MMStar} \\
\midrule
1:1:0 & 2232 & 87.4 & 86.9 & \textbf{55.5} \\
1:0:1 & 2249 & 85.8 & 86.5 & 54.4 \\
\rowcolor[HTML]{E8F2FF}
\textbf{1:1:2 (Ours)} & 2240 & \textbf{87.5} & 87.1 & 55.3 \\
1:0:0 & 2244 & 87.4 & 87.2 & 54.4 \\
0:0:1 & \textbf{2279} & 87.4 & \textbf{87.5} & 54.9 \\
\bottomrule
\end{tabular}%
}
\caption{Pivot extraction strategies.}
\label{tab:sub_selection}
\end{subtable}
\caption{\textbf{Ablation studies on injection layer placement and pivot extraction strategies (Qwen2.5-VL-7B).}}
\label{tab:combined_ablations}
\end{table*}

\begin{table*}[t]
\centering

\resizebox{0.9\textwidth}{!}{%
\setlength{\tabcolsep}{3pt}
\begin{tabular}{@{} l| cccc cccc @{}}
\toprule
\multirow{2}{*}{\textbf{Method}} 
& \multicolumn{4}{c}{\textbf{Retention: 33.3\%}} 
& \multicolumn{4}{c}{\textbf{Retention: 22.2\%}} \\
\cmidrule(lr){2-5} \cmidrule(l){6-9}
& \textbf{MME} & \textbf{POPE} & \textbf{SQA} & \textbf{MMStar} 
& \textbf{MME} & \textbf{POPE} & \textbf{SQA} & \textbf{MMStar} \\
\midrule

\textit{w/o} inject 
& 2173 & 85.7 & 86.6 & 54.4 
& 2104 & 85.0 & 85.7 & 51.6 \\

\rowcolor[HTML]{E8F2FF}
\textbf{\textit{w/} inject (Ours)} 
& \textbf{2240} & \textbf{87.5} & \textbf{87.1} & \textbf{55.3} 
& \textbf{2160} & \textbf{86.3} & \textbf{86.1} & \textbf{52.1} \\

\bottomrule
\end{tabular}%
}
\caption{\textbf{Ablation study on the injection mechanism.}}
\label{tab:ablation_qwen_fuse}
\end{table*}
We present detailed ablation studies analyzing the optimal layer selection and $p_2$.
\subsection{Effect of Layer Selection.} 
To determine the optimal stage for executing intermediate layer pruning, as shown in Tab.~\ref{tab:layer_ablation}, we conduct an ablation study across different layers. Pruning too early results in a notable performance drop. This occurs because the model requires a sufficient number of layers to digest and model the initial injection of visual tokens; truncating the sequence before these features are fully processed leads to the loss of critical contextual clues. Conversely, delaying the pruning to deeper layers yields sub-optimal outcomes, which can be attributed to the over-homogenization of spatial features. The results indicate that layer 19 provides the optimal balance, ensuring adequate digestion of injected visual information while avoiding feature homogenization.

\begin{table}[t]
\centering

\resizebox{\columnwidth}{!}{%
\begin{tabular}{c|cc}
\toprule
\textbf{Layer Index} & \textbf{MME} & \textbf{MMStar} \\
\midrule
9 & 1786 & 41.9 \\
11 & 1789 & 41.8 \\
13 & 1800 & 41.6 \\
15 & 2006 & 44.3 \\
17 & 2053 & 45.7 \\
\rowcolor[HTML]{E8F2FF}
\textbf{19 (Ours)} & \textbf{2093} & \textbf{46.3} \\
21 & 2064 & 43.4 \\
\bottomrule
\end{tabular}%
}
\caption{\textbf{Ablation study on layer selection.} Performance on MME and MMStar benchmarks when applying our extraction strategy at different intermediate layers.}
\label{tab:layer_ablation}
\end{table}
\subsection{Impact of $p_2$.}
We further investigate the intermediate pruning coefficient $p_2$, which dictates the proportion of tokens retained at the intermediate stage. In this ablation, the global token retention budget is strictly fixed at 11.1\%. Setting $p_2 = 1/3$ yields the optimal performance. A smaller coefficient drops too many tokens at the intermediate stage, causing an irreversible loss of spatial and structural details. Conversely, a larger coefficient retains an excessive number of tokens intermediately.

\begin{table}[h]
\centering

\begin{tabular}{l|ccc}
\toprule
\textbf{$p_2$} & \textbf{MME} & \textbf{SQA\textsubscript{I}} & \textbf{MMStar} \\
\midrule
1/5 & 2054 & 80.9 & 44.9 \\
1/4 & 2069 & 81.4 & 43.3 \\
\rowcolor[HTML]{E8F2FF}
1/3 & \textbf{2093} & \textbf{82.3} & \textbf{46.3} \\
1/2 & 2050 & 82.2 & 43.4 \\
\bottomrule
\end{tabular}
\caption{\textbf{Ablation study on the intermediate layer pruning $p_2$.} The total token retention budget is strictly fixed at 11.1\%. Setting the coefficient $p_2$ to 1/3 yields the optimal balance across multiple benchmarks.}
\label{tab:coefficient_ablation}
\end{table}

\section{Efficiency}
\label{sec:eff}
This section comprehensively evaluates the computational efficiency of our method, providing both a theoretical complexity analysis and an empirical latency profiling of the key algorithmic components.

\begin{table*}[t]
\centering

\resizebox{\textwidth}{!}{%
\begin{tabular}{l| c c c c c c |c |c}
\toprule
\multirow{2}{*}{\textbf{Method}} 
& \multirow{2}{*}{\textbf{Tokens $\downarrow$}} 
& \textbf{Total Time $\downarrow$} 
& \textbf{Prefilling Time $\downarrow$} 
& \multirow{2}{*}{\textbf{FLOPs $\downarrow$}} 
& \textbf{KV Cache $\downarrow$} 
& \textbf{MME $\uparrow$}
& \multicolumn{2}{c}{\textbf{Speedup $\uparrow$}} \\
\cmidrule(lr){8-9}
& & \textbf{(Sec)} & \textbf{(Sec)} & & \textbf{(MB)} & \textbf{(Score)} & \textbf{Total} & \textbf{Prefilling} \\
\midrule

Qwen2.5VL-7B 
& 100\% 
& 557.443 
& 239.833 
& 1043.83 
& 211.4 
& \textbf{2347} 
& 1.00$\times$ 
& 1.00$\times$ \\

\rowcolor[HTML]{E8F2FF}
+ Ours 
& 22.2\% 
& 401.919 
& 105.234 
& 279.82 
& 37.0 
& 2160 
& 1.38$\times$ 
& 2.28$\times$ \\

\rowcolor[HTML]{E8F2FF}
+ Ours 
& \textbf{11.1\%} 
& \textbf{373.961} 
& \textbf{84.809} 
& \textbf{148.06} 
& \textbf{26.7} 
& 2093 
& \textbf{1.49}$\times$ 
& \textbf{2.83}$\times$ \\

\bottomrule
\end{tabular}%
}
\caption{\textbf{Efficiency results on Qwen2.5VL-7B.} Performance, runtime, FLOPs, and memory overhead are evaluated on a single NVIDIA A100-80GB GPU with FlashAttention.}
\label{tab:efficiency_comparison}
\end{table*}

\subsection{Theoretical Efficiency Analysis}
\label{sec:efficiency_analysis}

We provide a theoretical analysis to show the efficiency gain of \textbf{ReTrace}. Let $N$ be the number of input visual tokens, $L$ the number of LLM layers, and $D$ the hidden dimension. Dense inference processes all visual tokens through all layers. Its attention and MLP cost can be written as
\begin{equation}
    \mathcal{O}_{\text{base}}
    =
    L \cdot \mathcal{O}(N^2D + ND^2).
\end{equation}
The KV cache memory scales as
\begin{equation}
    \mathcal{M}_{\text{base}}
    =
    \mathcal{O}(LND).
\end{equation}

ReTrace reduces the active visual sequence in two stages. Given the intermediate retention ratio $\rho_1$, the first active length is
\begin{equation}
    K_1 = \rho_1 N.
\end{equation}
After post-absorption pruning with retention ratio $\rho_2$, the final active length becomes
\begin{equation}
    K_2 = \rho_2 K_1 = \rho_1 \rho_2 N.
\end{equation}
Let $p$ denote the late pruning layer. From layer $1$ to $p-1$, ReTrace uses $K_1$ active visual tokens. From layer $p$ to $L$, it uses $K_2$ active visual tokens. The total computational cost is therefore
\begin{equation}
\begin{aligned}
    \mathcal{O}_{\text{ReTrace}}
    &=
    (p-1)\cdot \mathcal{O}(K_1^2D + K_1D^2) \\
    &\quad +
    (L-p+1)\cdot \mathcal{O}(K_2^2D + K_2D^2) \\
    &\quad +
    \mathcal{O}_{\text{overhead}} .
\end{aligned}
\end{equation}

The overhead comes from three one-time operations. Pivot buffer construction is performed at the encoder output. Context injection computes similarities between dormant tokens and active pivots, with cost
\begin{equation}
    \mathcal{O}\big((N-K_1)K_1D\big).
\end{equation}
Post-absorption pruning uses a single text-to-visual query, with cost
\begin{equation}
    \mathcal{O}(K_1D).
\end{equation}
These operations are executed once rather than at every LLM layer, so they are small compared with repeated dense self-attention in deep models.

When the quadratic attention term dominates, the theoretical FLOPs ratio can be approximated as
\begin{equation}
\begin{aligned}
    \eta_{\text{FLOPs}}
    &\approx
    \frac{
    (p-1)K_1^2
    +
    (L-p+1)K_2^2
    }{
    LN^2
    } \\
    &=
    \rho_1^2
    \left[
    \frac{p-1}{L}
    +
    \frac{L-p+1}{L}\rho_2^2
    \right].
\end{aligned}
\end{equation}
Similarly, the KV cache ratio is
\begin{equation}
\begin{aligned}
    \eta_{\text{KV}}
    &=
    \frac{
    (p-1)K_1
    +
    (L-p+1)K_2
    }{
    LN
    } \\
    &=
    \rho_1
    \left[
    \frac{p-1}{L}
    +
    \frac{L-p+1}{L}\rho_2
    \right].
\end{aligned}
\end{equation}

This analysis shows that ReTrace obtains a compounding reduction from both $\rho_1$ and $\rho_2$. In particular, the attention cost is suppressed by the quadratic factor $\rho_1^2$, and further reduced after late-stage pruning through $\rho_2^2$. This matches our empirical efficiency results, where aggressive compression yields about $87\%$ FLOPs reduction and a large KV-cache saving.

\subsection{Overhead of Buffer Construction and Context Injection}

We profile the additional runtime cost introduced by \textbf{ReTrace}, including spatial-diversity buffer construction and backtracking-aware context injection. The profiling is conducted on the MME benchmark with 2,374 test samples. As shown in Tab.~\ref{tab:efficiency_overhead}, under the 11.1\% retention setting, the two operations take only 0.7 ms and 0.1 ms per sample, accounting for 0.42\% and 0.06\% of the total latency. Across all evaluated sparsity levels, their combined overhead stays below 0.5\%. This confirms that ReTrace adds negligible runtime overhead while preserving the practical speedup from visual token compression.

\begin{table*}[t]
\centering

\resizebox{\textwidth}{!}{
\begin{tabular}{l|c|cccccc}
\toprule
\multirow{2}{*}{\textbf{Tokens} $\downarrow$} 
& \textbf{Total Time} $\downarrow$ 
& \multicolumn{2}{c}{\textbf{$T_d$ Extraction}} 
& \multicolumn{2}{c}{\textbf{Mid-stage Injection}} 
& \multicolumn{2}{c}{\textbf{Overhead Proportion} $\downarrow$} \\
\cmidrule(lr){3-4} \cmidrule(lr){5-6} \cmidrule(lr){7-8}
& (Sec) & Total (s) & Avg. (ms) & Total (s) & Avg. (ms) & $T_d$ Ext. & Inject \\
\midrule
\rowcolor[HTML]{E8F2FF}
33.3\% & 426.85 & 1.78 & 0.8 & 0.20 & 0.1 & 0.42\% & 0.05\% \\
\rowcolor[HTML]{E8F2FF}
22.2\% & 401.92 & 1.66 & 0.7 & 0.21 & 0.1 & 0.41\% & 0.05\% \\
\rowcolor[HTML]{E8F2FF}
11.1\% & 373.96 & 1.58 & 0.7 & 0.23 & 0.1 & 0.42\% & 0.06\% \\
\bottomrule
\end{tabular}
}
\caption{Algorithmic overhead analysis under different token retention settings on Qwen2.5-VL-7B. The absolute execution times and their relative proportions explicitly demonstrate that the spatial diversity extraction ($T_d$) and mid-stage injection introduce negligible latency to the overall inference.}
\label{tab:efficiency_overhead}
\end{table*}

\section{Visualization of Pivot Buffer Construction}
\label{sec:vis_hpe}

\begin{figure*}[p] 
    \centering

    \begin{subfigure}{0.16\textwidth}
        \centering
        \small \textit{Original}\\[1mm]
        \includegraphics[width=\linewidth]{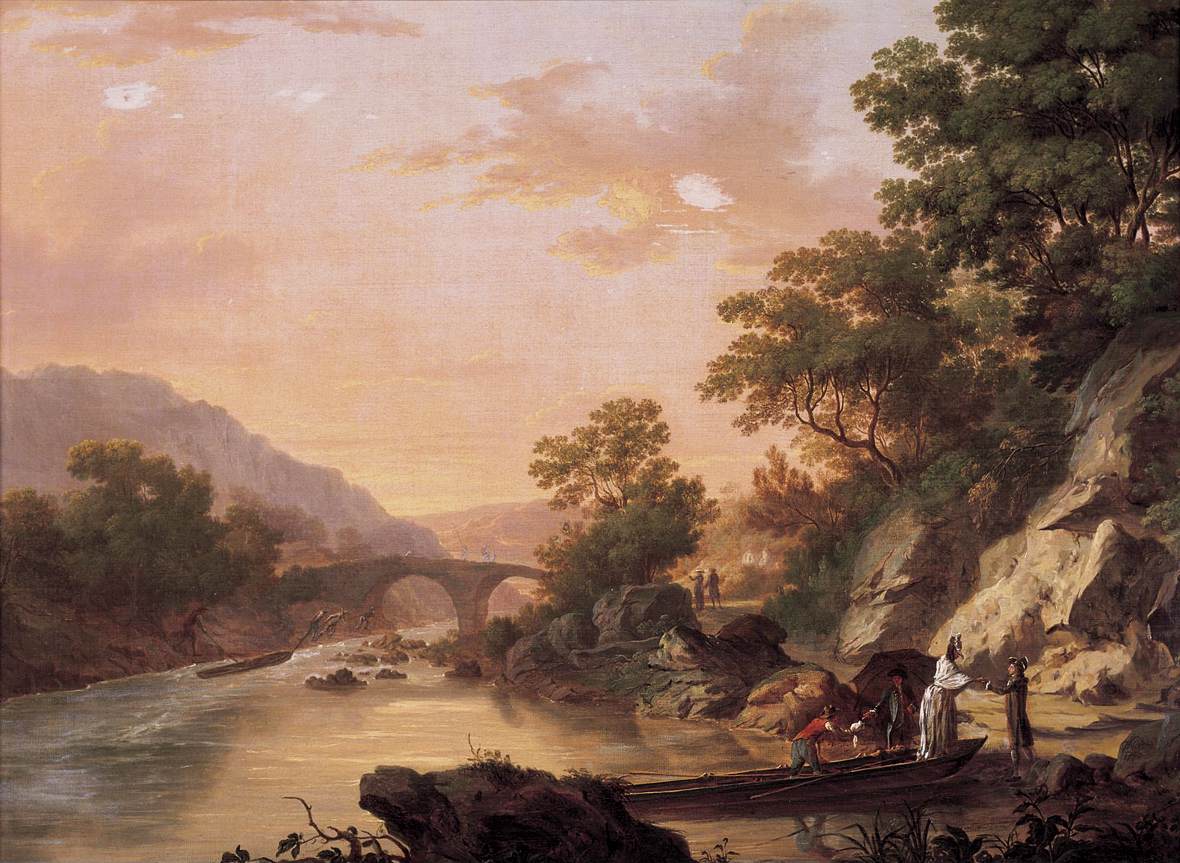}
    \end{subfigure}\hfill
    \begin{subfigure}{0.16\textwidth}
        \centering
        \small \textit{Retain 10\%}\\[1mm]
        \includegraphics[width=\linewidth]{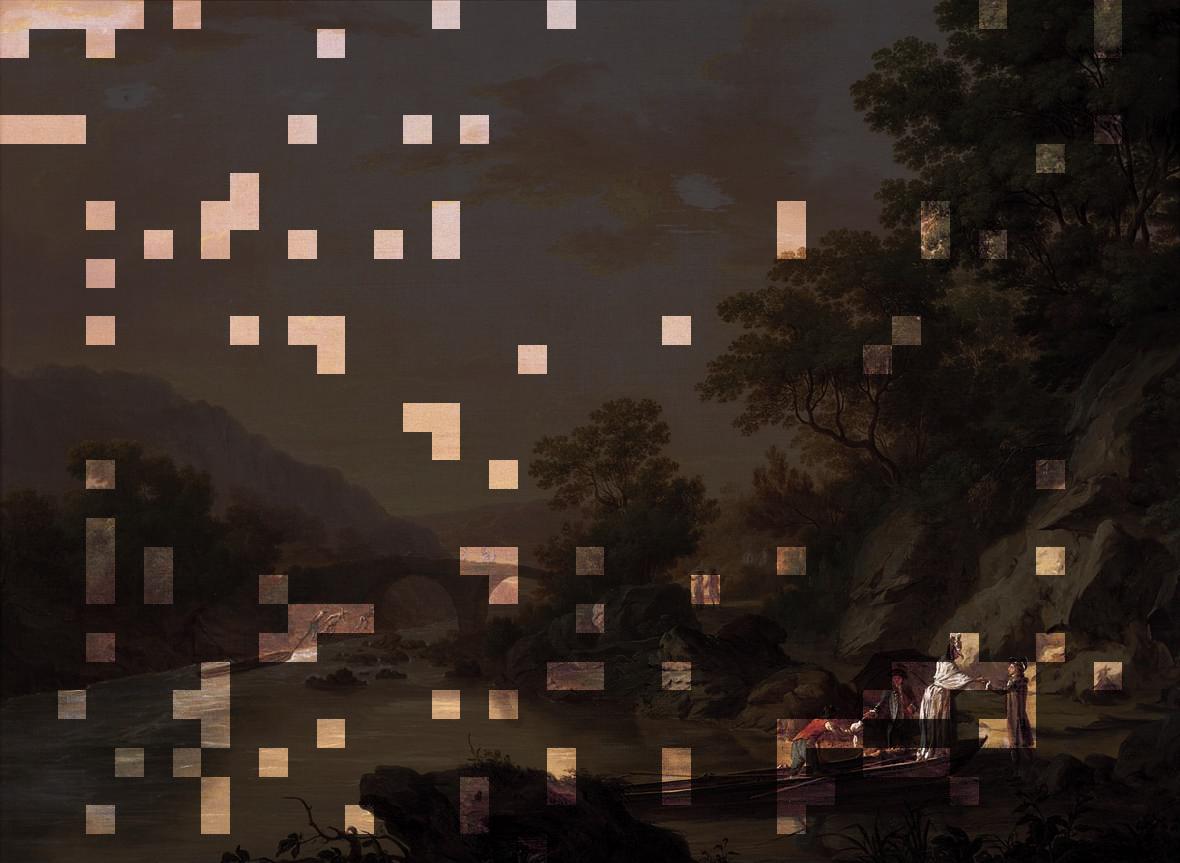}
    \end{subfigure}\hfill
    \begin{subfigure}{0.16\textwidth}
        \centering
        \small \textit{Retain 20\%}\\[1mm]
        \includegraphics[width=\linewidth]{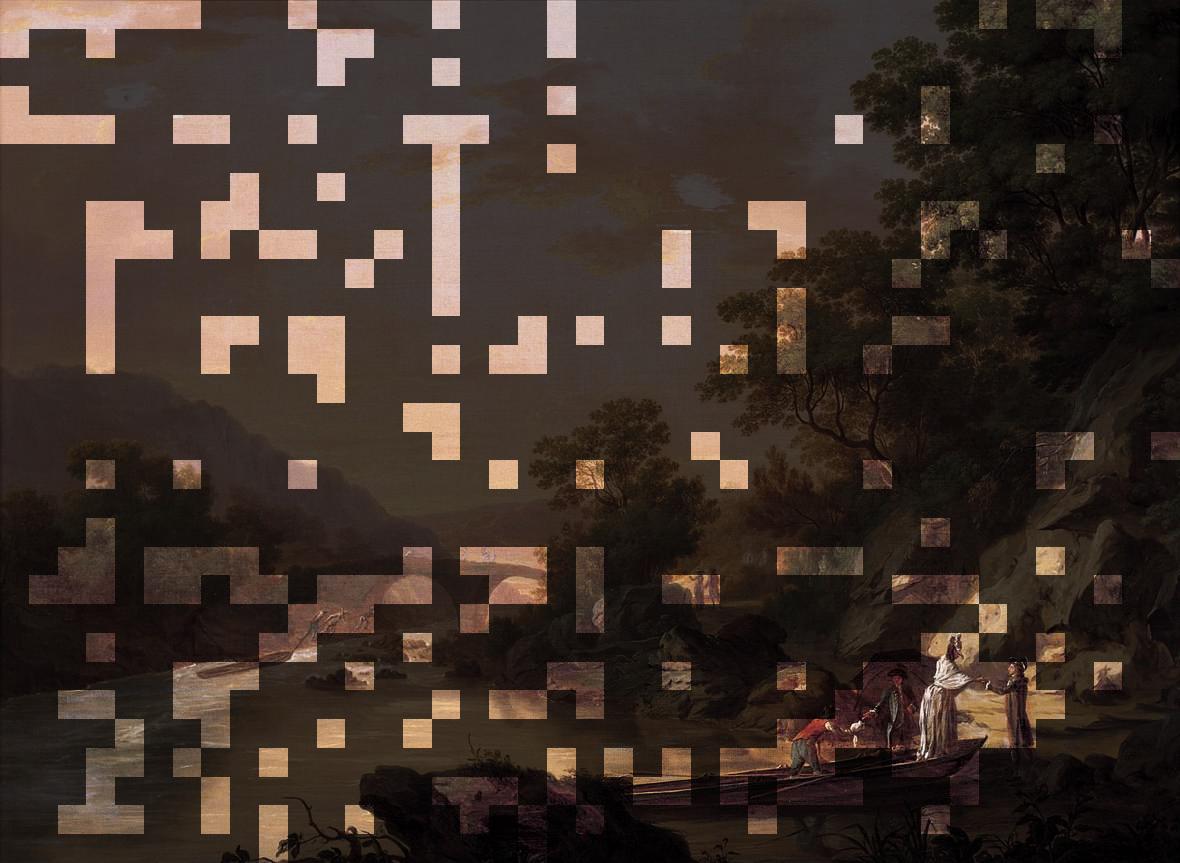}
    \end{subfigure}\hfill
    \begin{subfigure}{0.16\textwidth}
        \centering
        \small \textit{Retain 30\%}\\[1mm]
        \includegraphics[width=\linewidth]{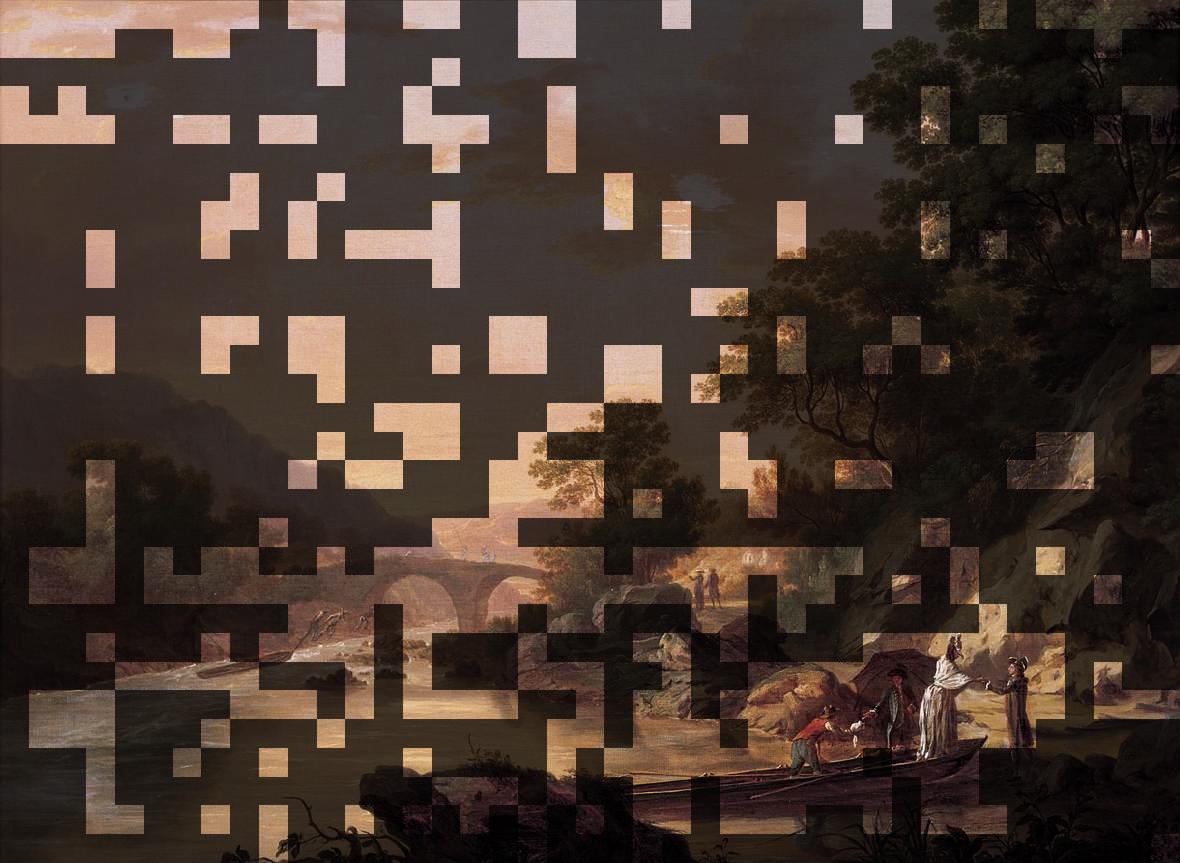}
    \end{subfigure}\hfill
    \begin{subfigure}{0.16\textwidth}
        \centering
        \small \textit{Retain 40\%}\\[1mm]
        \includegraphics[width=\linewidth]{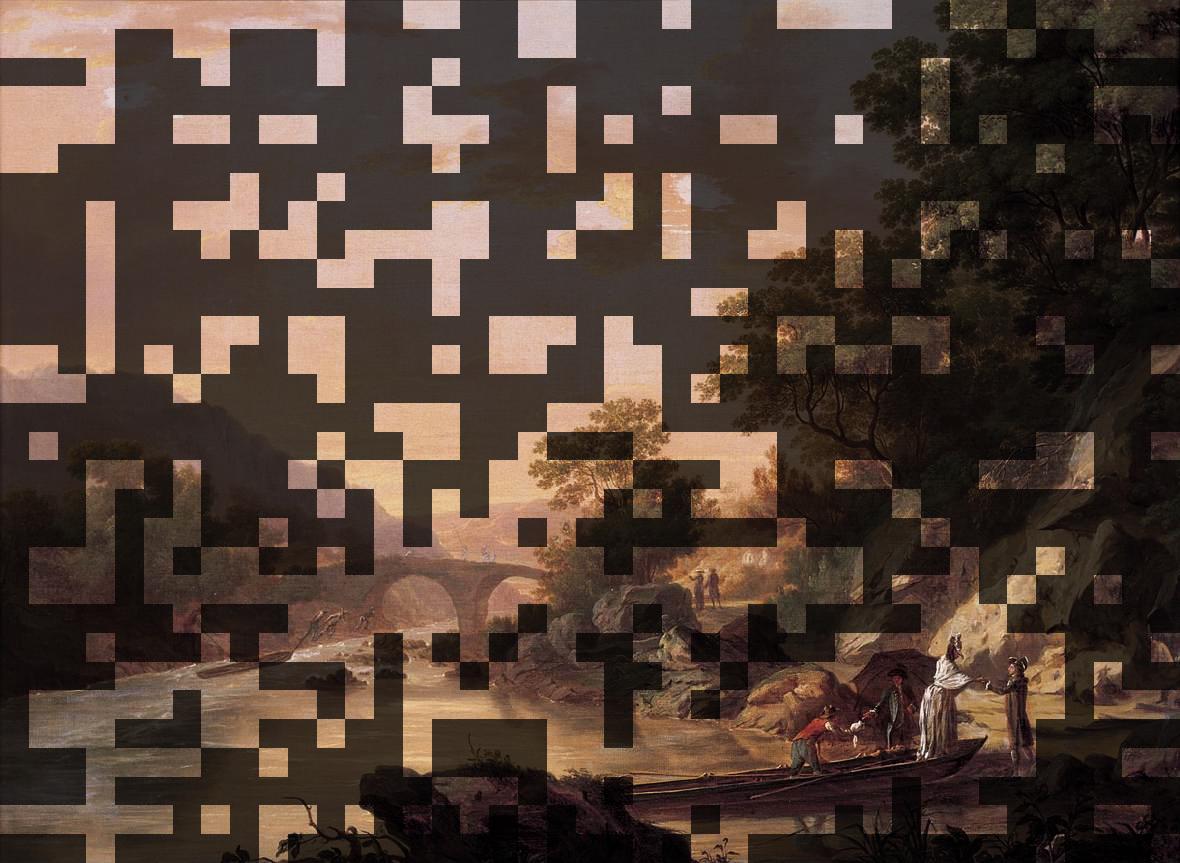}
    \end{subfigure}\hfill
    \begin{subfigure}{0.16\textwidth}
        \centering
        \small \textit{Retain 50\%}\\[1mm]
        \includegraphics[width=\linewidth]{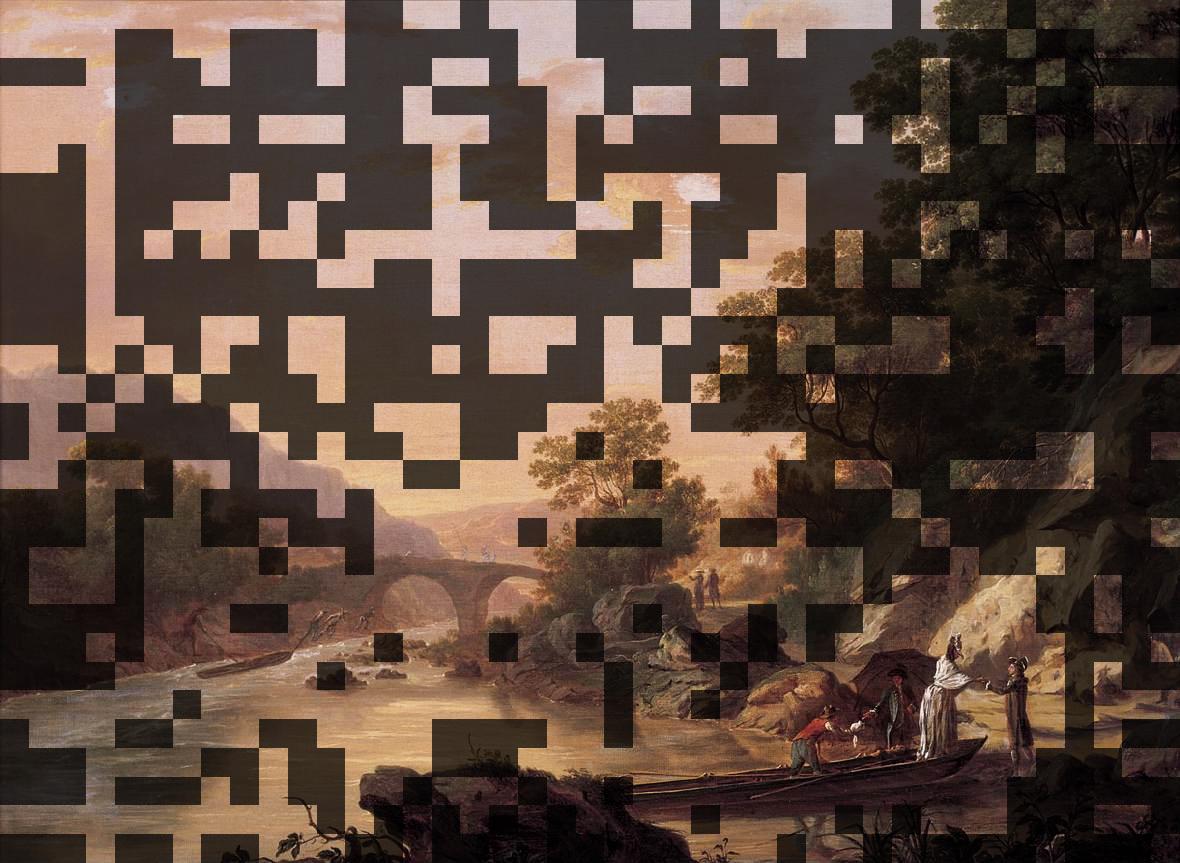}
    \end{subfigure}

    \vspace{1.5mm}

    \begin{subfigure}{0.16\textwidth}
        \centering
        \includegraphics[width=\linewidth]{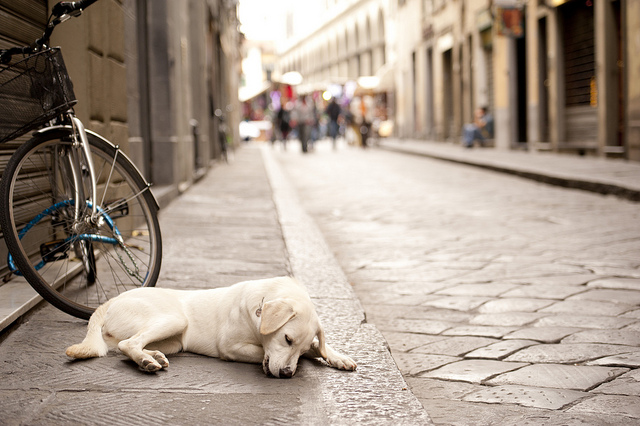}
    \end{subfigure}\hfill
    \begin{subfigure}{0.16\textwidth}
        \centering
        \includegraphics[width=\linewidth]{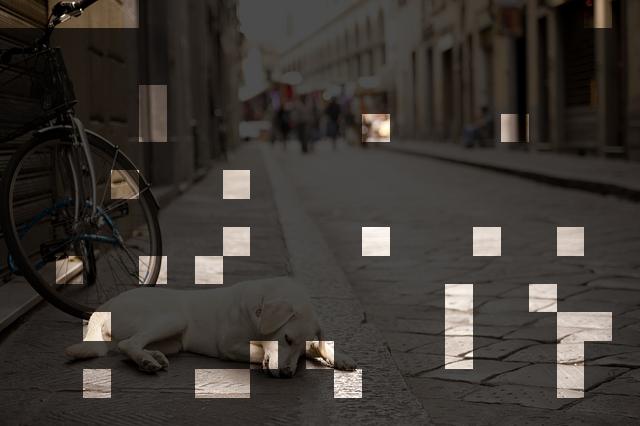}
    \end{subfigure}\hfill
    \begin{subfigure}{0.16\textwidth}
        \centering
        \includegraphics[width=\linewidth]{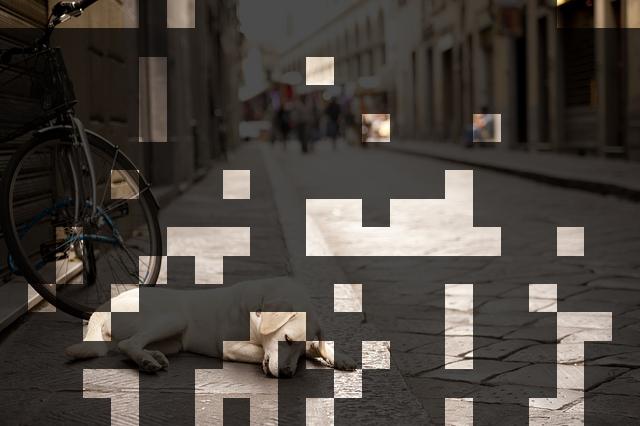}
    \end{subfigure}\hfill
    \begin{subfigure}{0.16\textwidth}
        \centering
        \includegraphics[width=\linewidth]{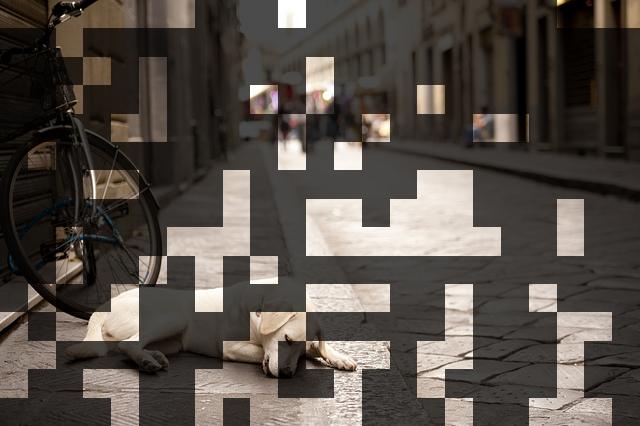}
    \end{subfigure}\hfill
    \begin{subfigure}{0.16\textwidth}
        \centering
        \includegraphics[width=\linewidth]{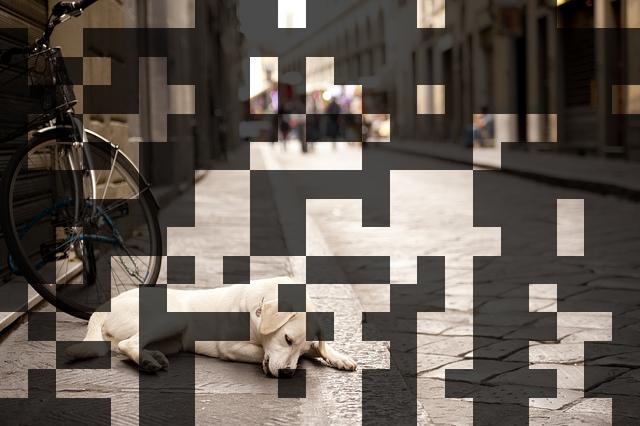}
    \end{subfigure}\hfill
    \begin{subfigure}{0.16\textwidth}
        \centering
        \includegraphics[width=\linewidth]{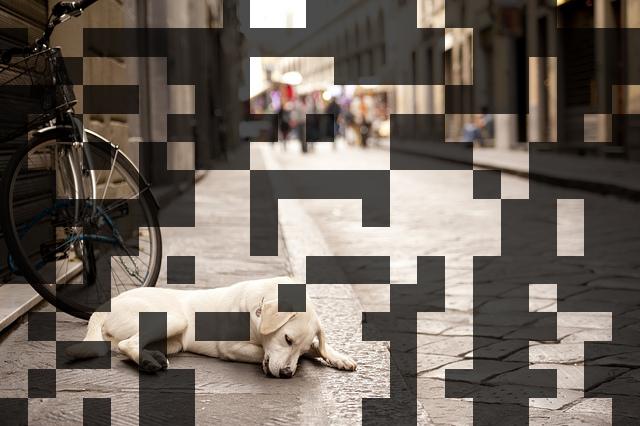}
    \end{subfigure}

    \vspace{1.5mm}

    \begin{subfigure}{0.16\textwidth}
        \centering
        \includegraphics[width=\linewidth]{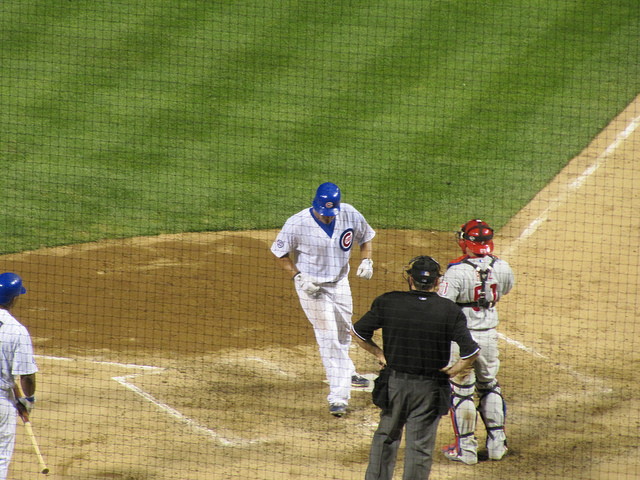}
    \end{subfigure}\hfill
    \begin{subfigure}{0.16\textwidth}
        \centering
        \includegraphics[width=\linewidth]{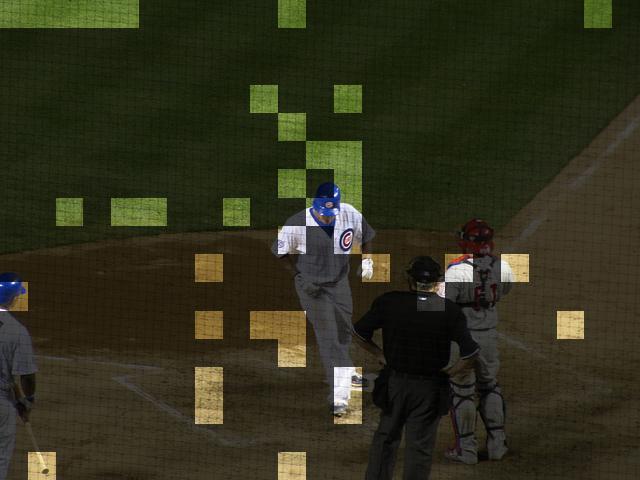}
    \end{subfigure}\hfill
    \begin{subfigure}{0.16\textwidth}
        \centering
        \includegraphics[width=\linewidth]{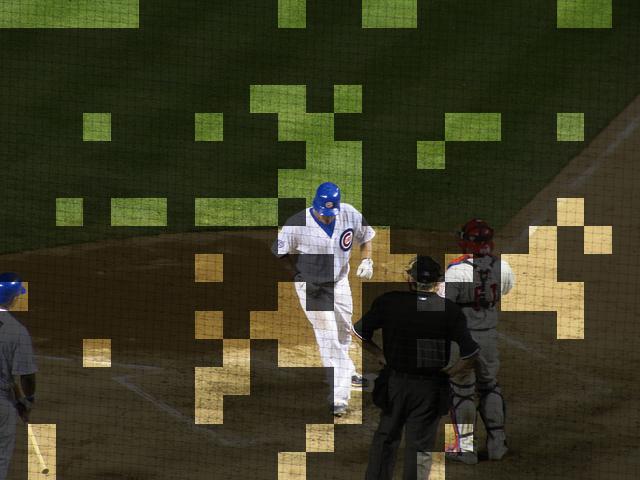}
    \end{subfigure}\hfill
    \begin{subfigure}{0.16\textwidth}
        \centering
        \includegraphics[width=\linewidth]{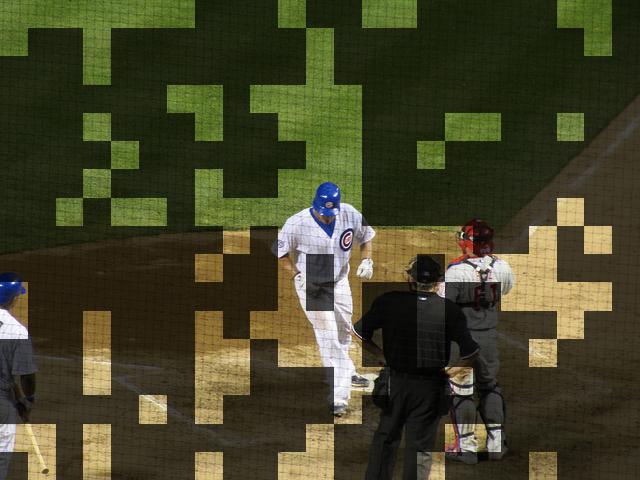}
    \end{subfigure}\hfill
    \begin{subfigure}{0.16\textwidth}
        \centering
        \includegraphics[width=\linewidth]{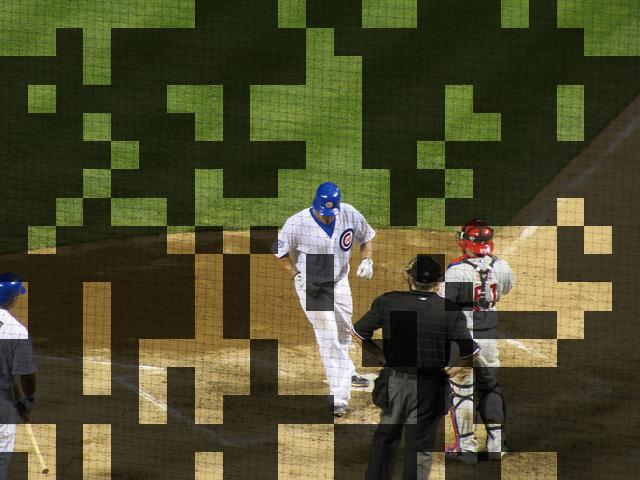}
    \end{subfigure}\hfill
    \begin{subfigure}{0.16\textwidth}
        \centering
        \includegraphics[width=\linewidth]{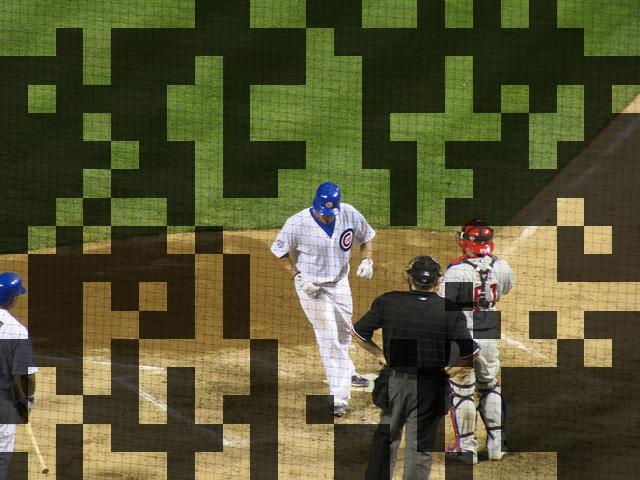}
    \end{subfigure}

    \vspace{1.5mm}

    \begin{subfigure}{0.16\textwidth}
        \centering
        \includegraphics[width=\linewidth]{figs/vis_hpe/input_images/COCO_val2014_000000580975.jpg}
    \end{subfigure}\hfill
    \begin{subfigure}{0.16\textwidth}
        \centering
        \includegraphics[width=\linewidth]{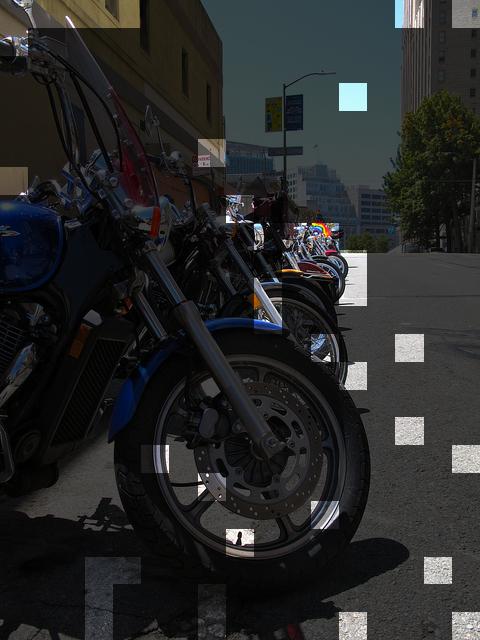}
    \end{subfigure}\hfill
    \begin{subfigure}{0.16\textwidth}
        \centering
        \includegraphics[width=\linewidth]{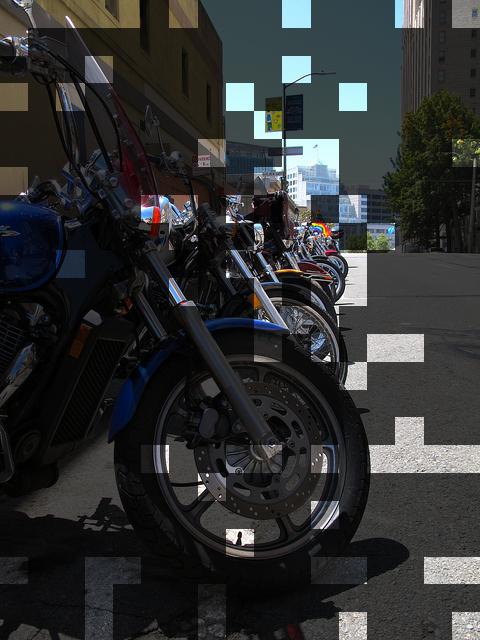}
    \end{subfigure}\hfill
    \begin{subfigure}{0.16\textwidth}
        \centering
        \includegraphics[width=\linewidth]{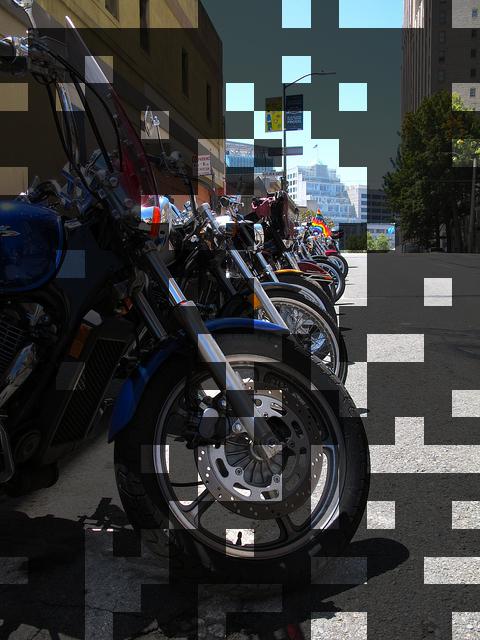}
    \end{subfigure}\hfill
    \begin{subfigure}{0.16\textwidth}
        \centering
        \includegraphics[width=\linewidth]{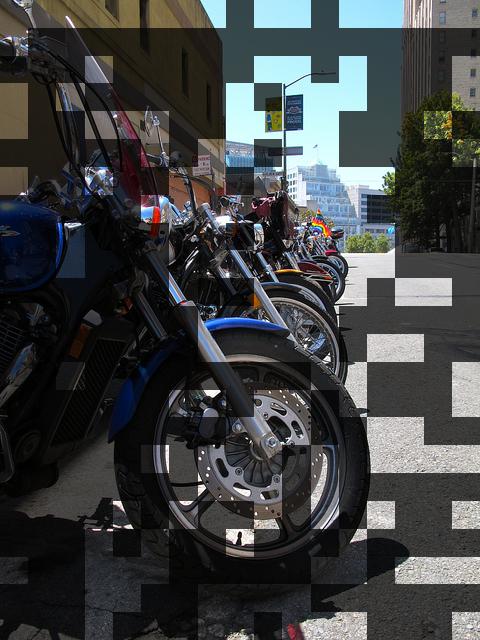}
    \end{subfigure}\hfill
    \begin{subfigure}{0.16\textwidth}
        \centering
        \includegraphics[width=\linewidth]{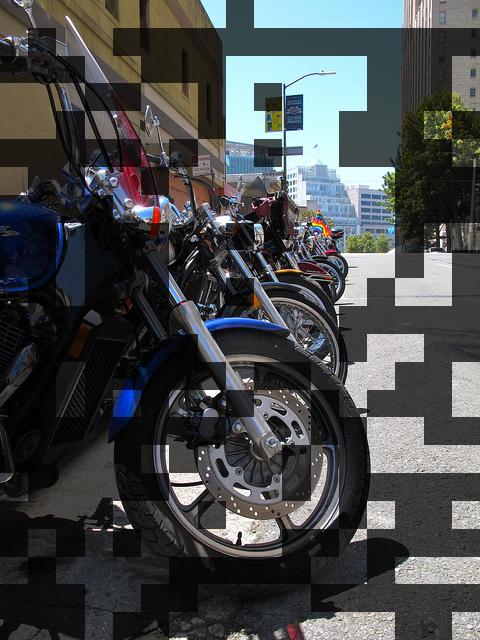}
    \end{subfigure}

    \vspace{1.5mm}

    \begin{subfigure}{0.16\textwidth}
        \centering
        \includegraphics[width=\linewidth]{figs/vis_hpe/input_images/COCO_val2014_000000581829.jpg}
    \end{subfigure}\hfill
    \begin{subfigure}{0.16\textwidth}
        \centering
        \includegraphics[width=\linewidth]{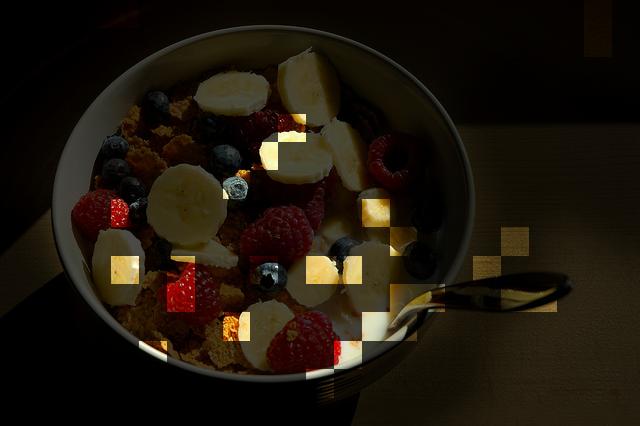}
    \end{subfigure}\hfill
    \begin{subfigure}{0.16\textwidth}
        \centering
        \includegraphics[width=\linewidth]{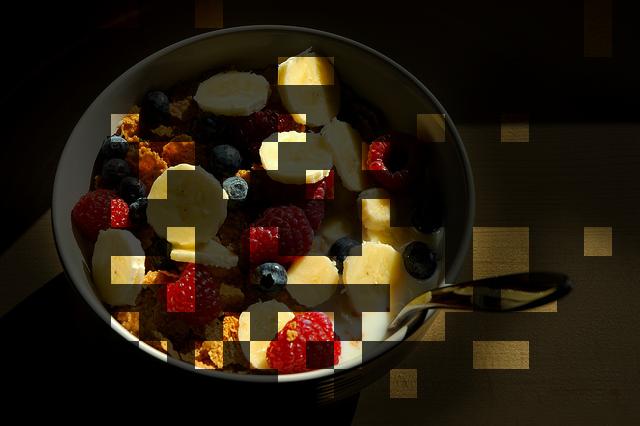}
    \end{subfigure}\hfill
    \begin{subfigure}{0.16\textwidth}
        \centering
        \includegraphics[width=\linewidth]{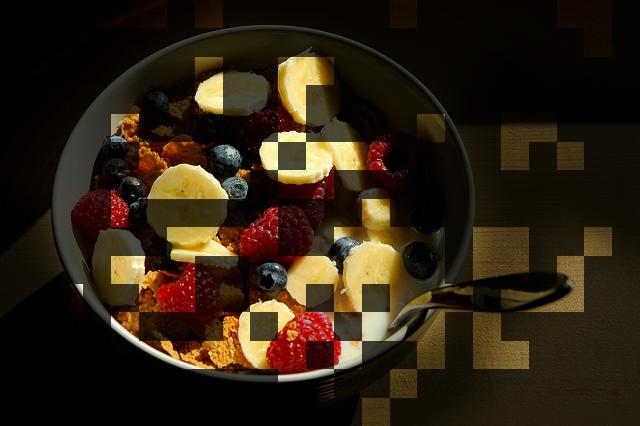}
    \end{subfigure}\hfill
    \begin{subfigure}{0.16\textwidth}
        \centering
        \includegraphics[width=\linewidth]{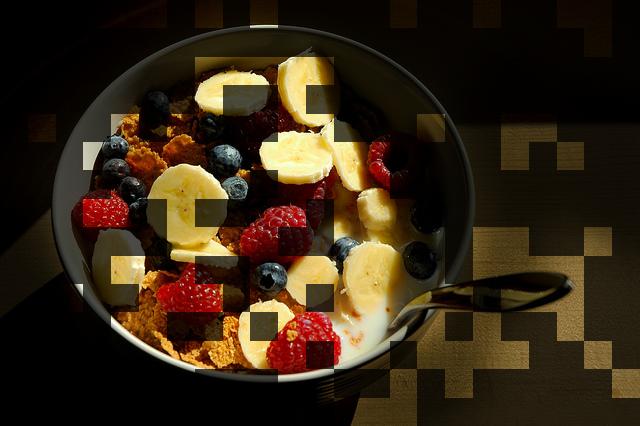}
    \end{subfigure}\hfill
    \begin{subfigure}{0.16\textwidth}
        \centering
        \includegraphics[width=\linewidth]{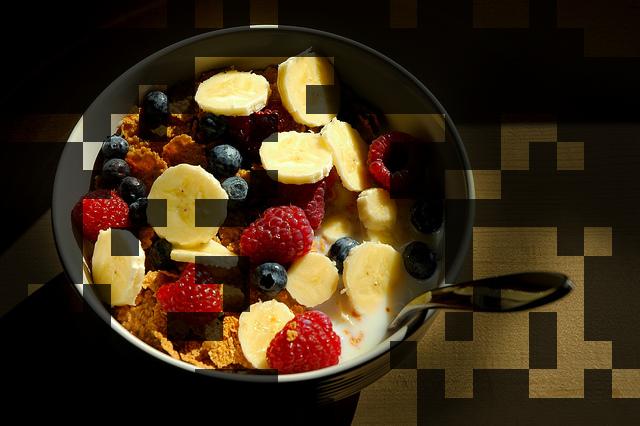}
    \end{subfigure}

    \vspace{1.5mm}

    \begin{subfigure}{0.16\textwidth}
        \centering
        \includegraphics[width=\linewidth]{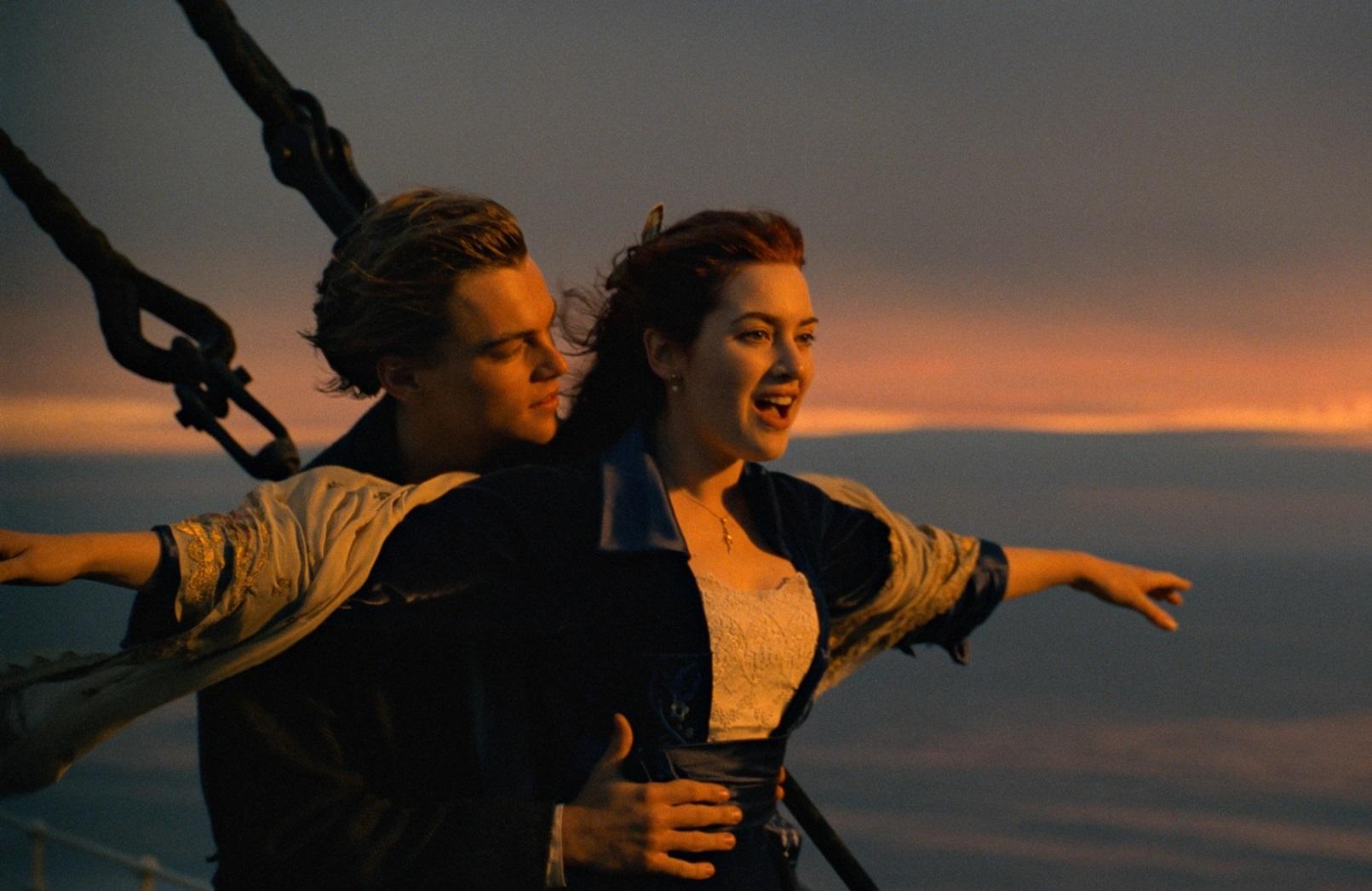}
    \end{subfigure}\hfill
    \begin{subfigure}{0.16\textwidth}
        \centering
        \includegraphics[width=\linewidth]{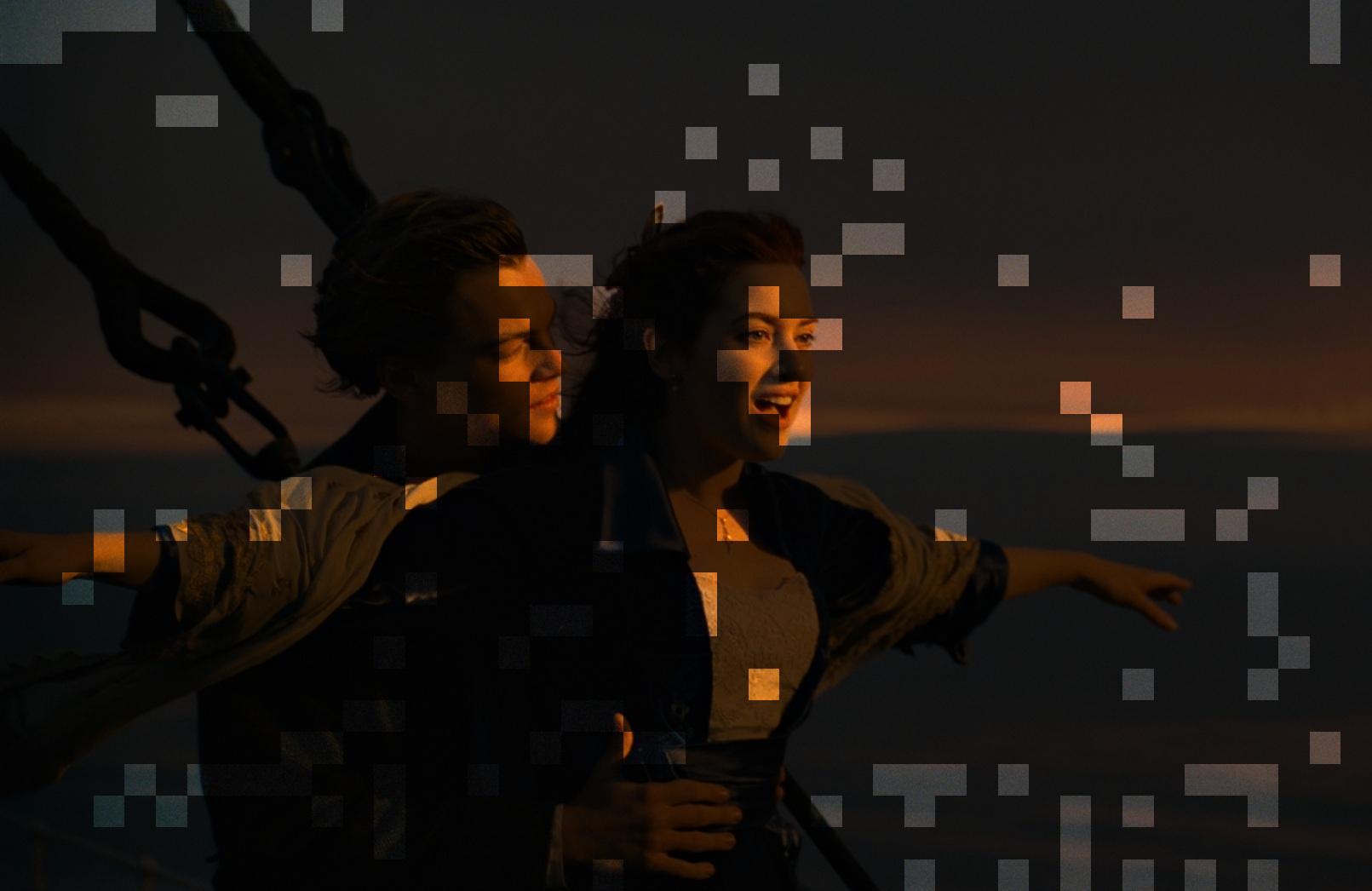}
    \end{subfigure}\hfill
    \begin{subfigure}{0.16\textwidth}
        \centering
        \includegraphics[width=\linewidth]{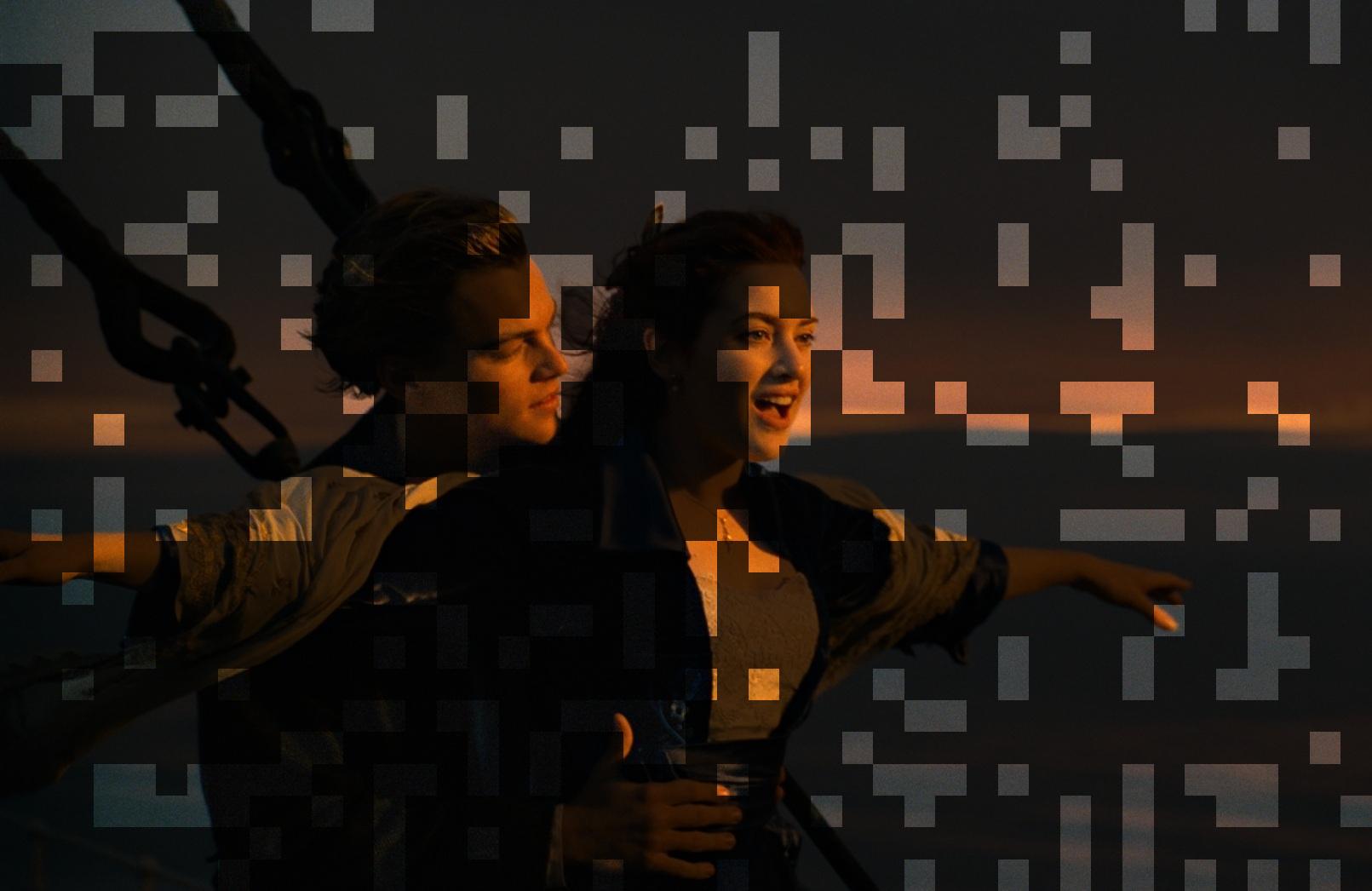}
    \end{subfigure}\hfill
    \begin{subfigure}{0.16\textwidth}
        \centering
        \includegraphics[width=\linewidth]{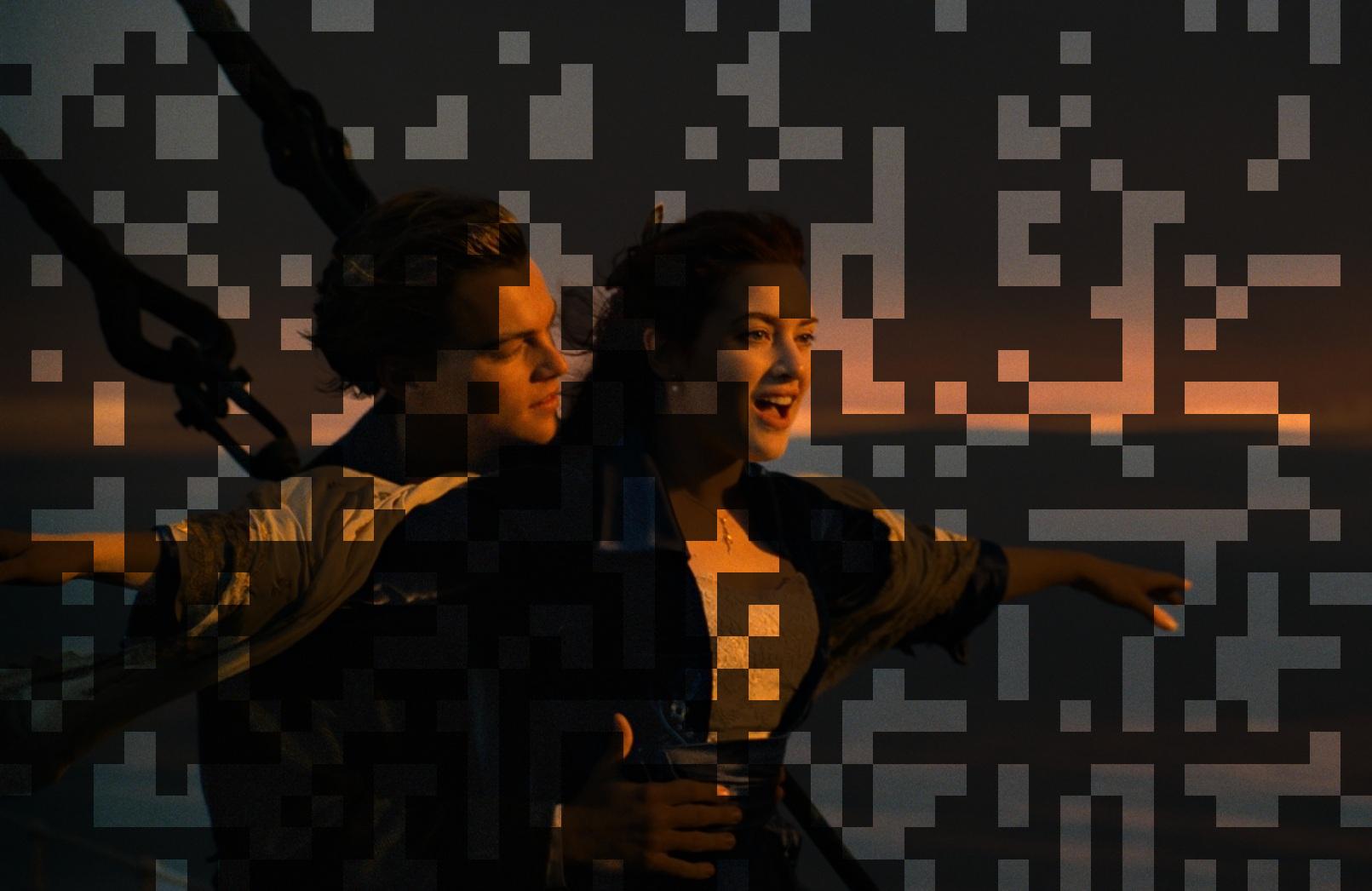}
    \end{subfigure}\hfill
    \begin{subfigure}{0.16\textwidth}
        \centering
        \includegraphics[width=\linewidth]{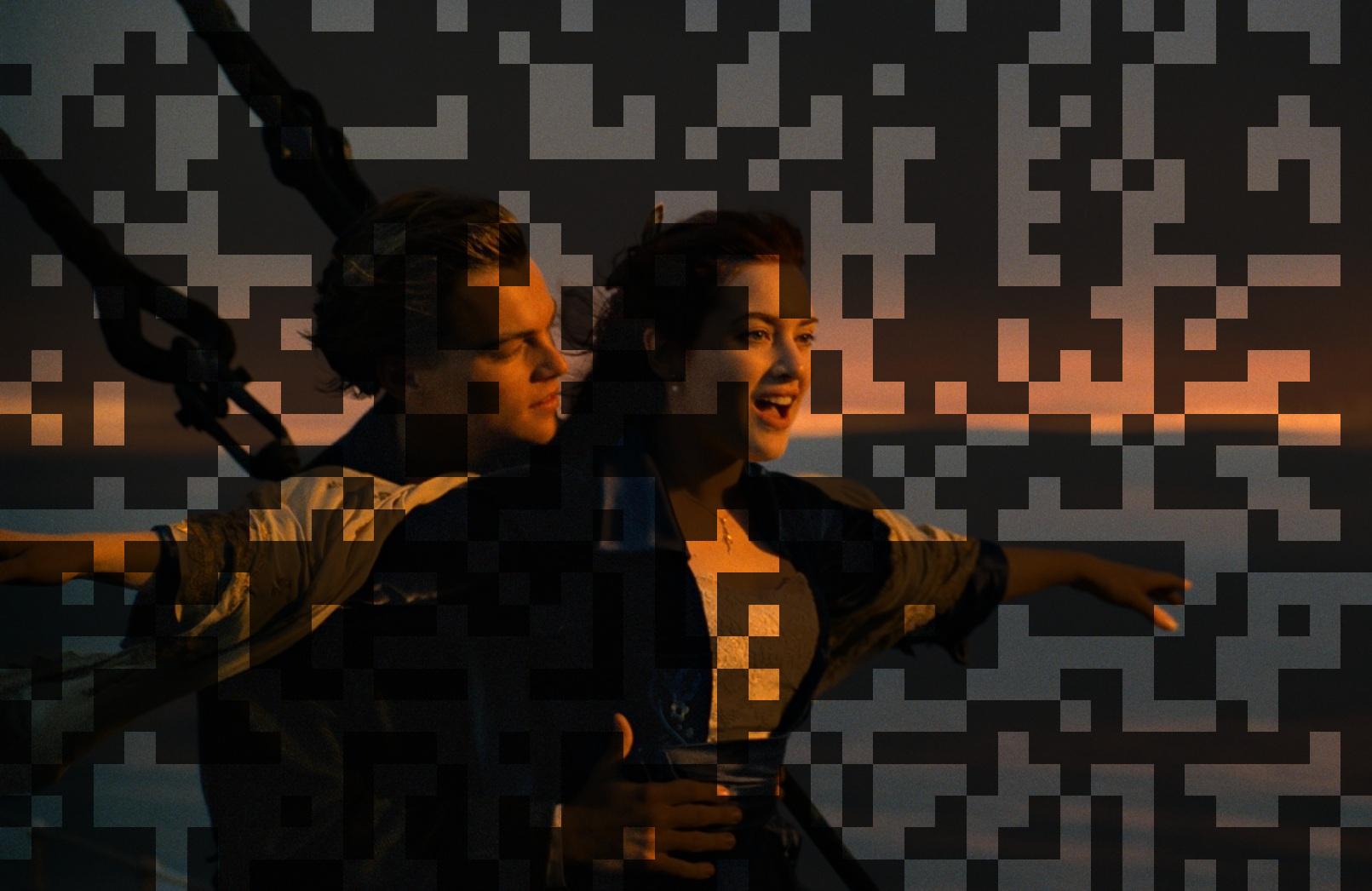}
    \end{subfigure}\hfill
    \begin{subfigure}{0.16\textwidth}
        \centering
        \includegraphics[width=\linewidth]{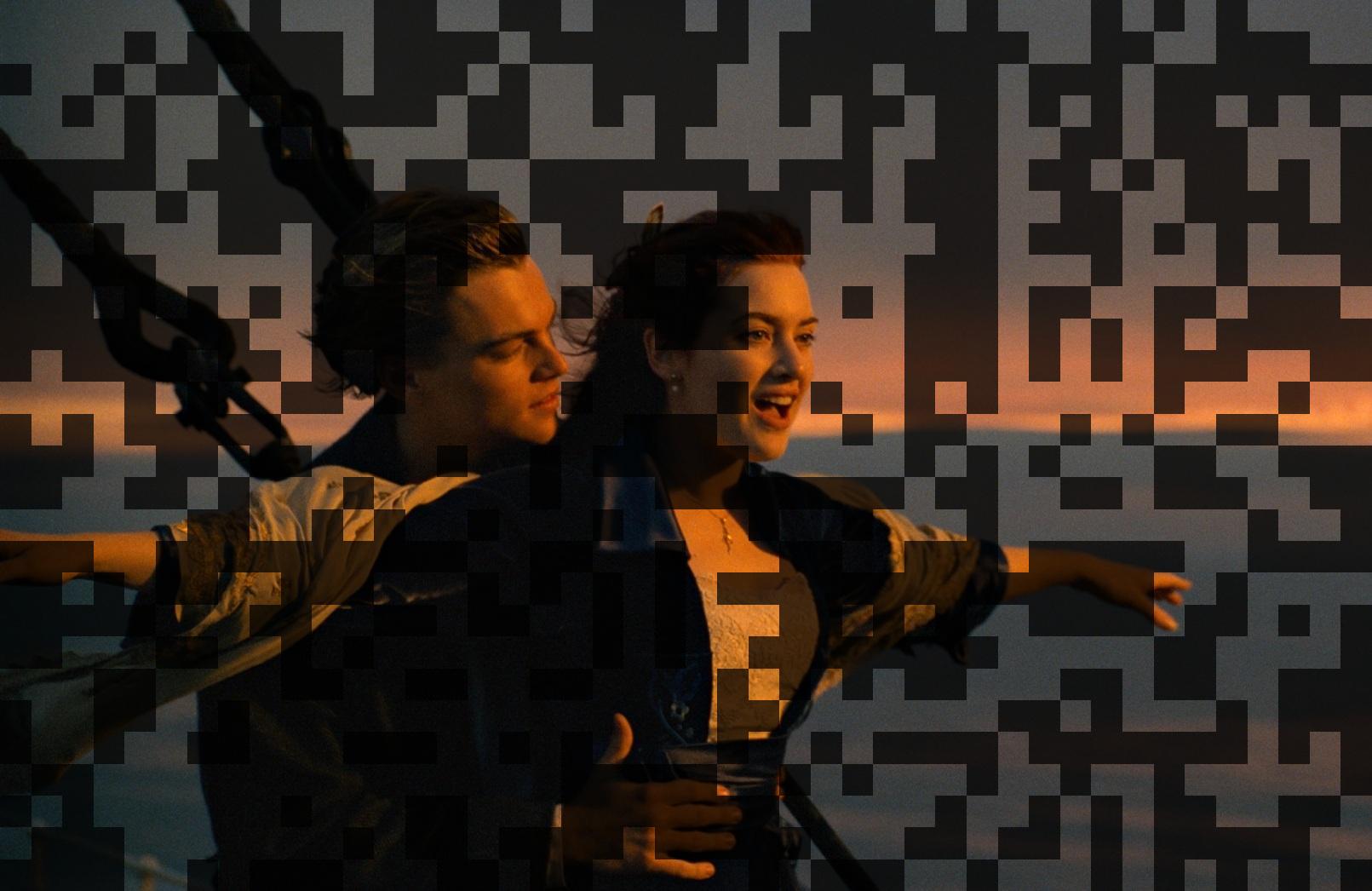}
    \end{subfigure}

    \vspace{1.5mm}

    \begin{subfigure}{0.16\textwidth}
        \centering
        \includegraphics[width=\linewidth]{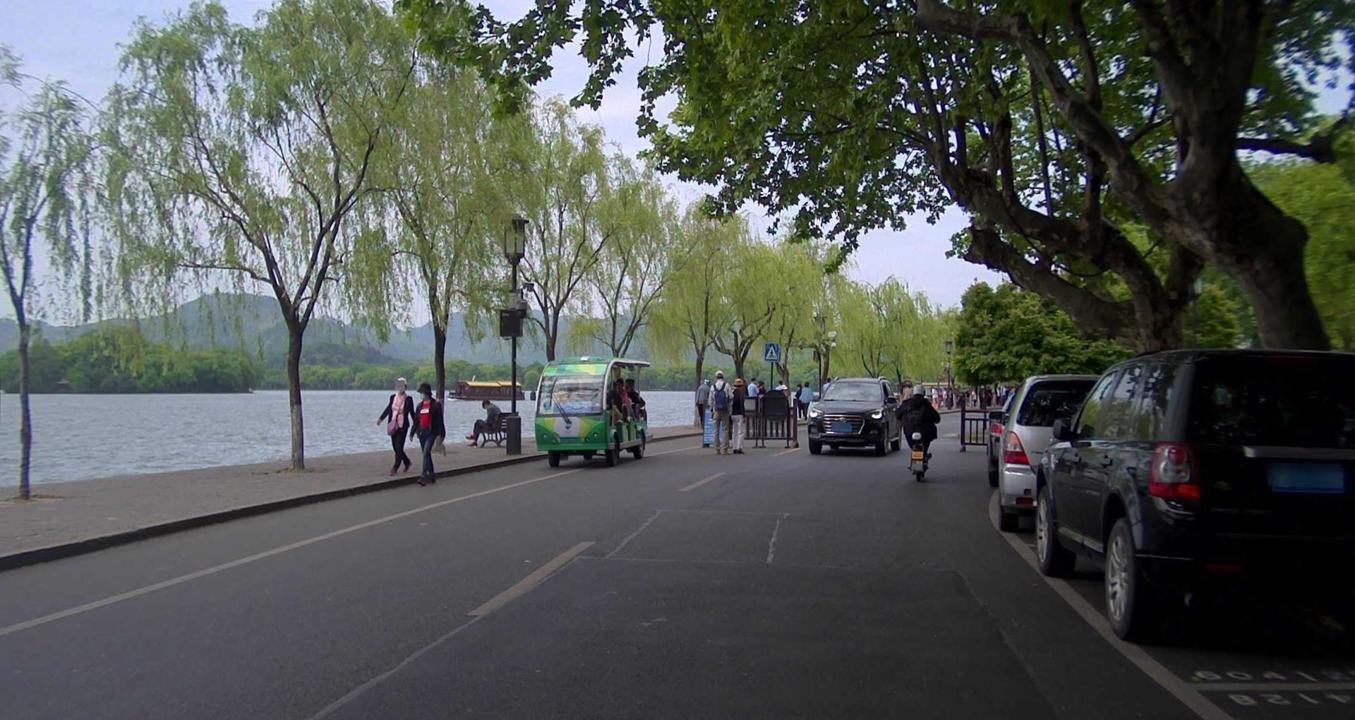}
    \end{subfigure}\hfill
    \begin{subfigure}{0.16\textwidth}
        \centering
        \includegraphics[width=\linewidth]{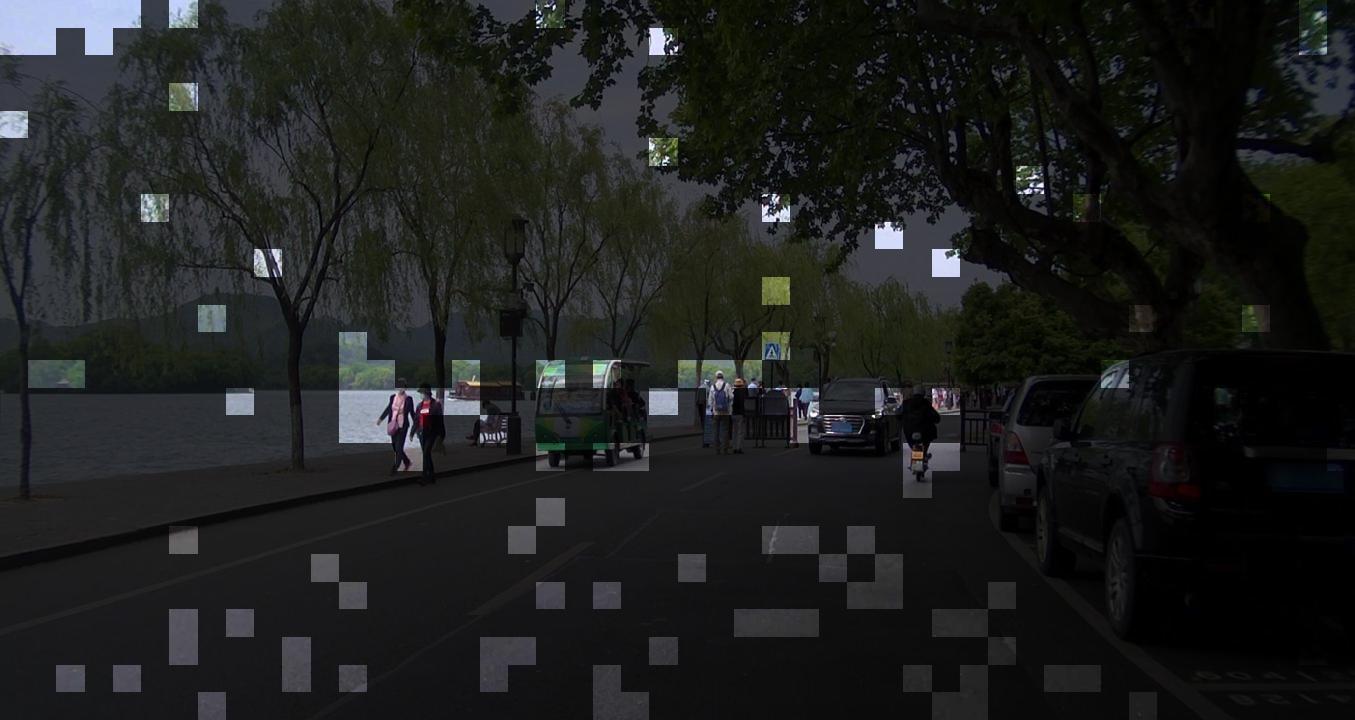}
    \end{subfigure}\hfill
    \begin{subfigure}{0.16\textwidth}
        \centering
        \includegraphics[width=\linewidth]{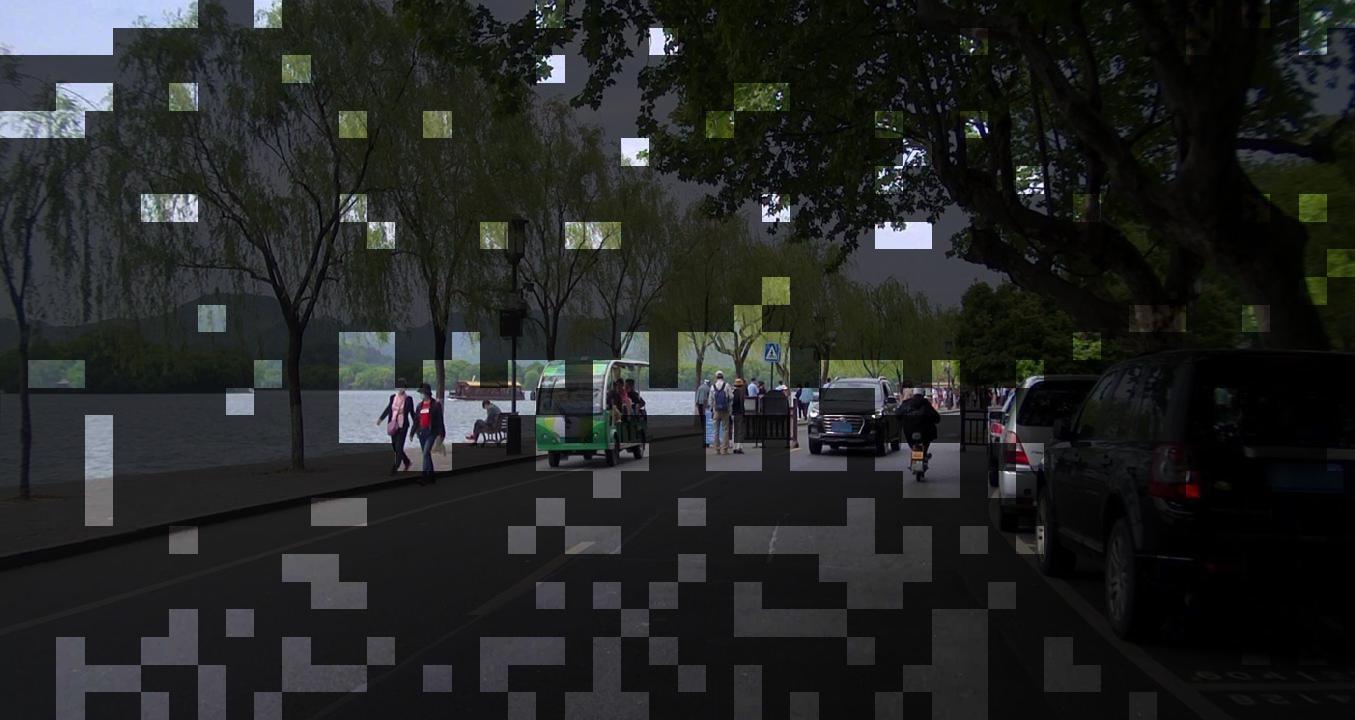}
    \end{subfigure}\hfill
    \begin{subfigure}{0.16\textwidth}
        \centering
        \includegraphics[width=\linewidth]{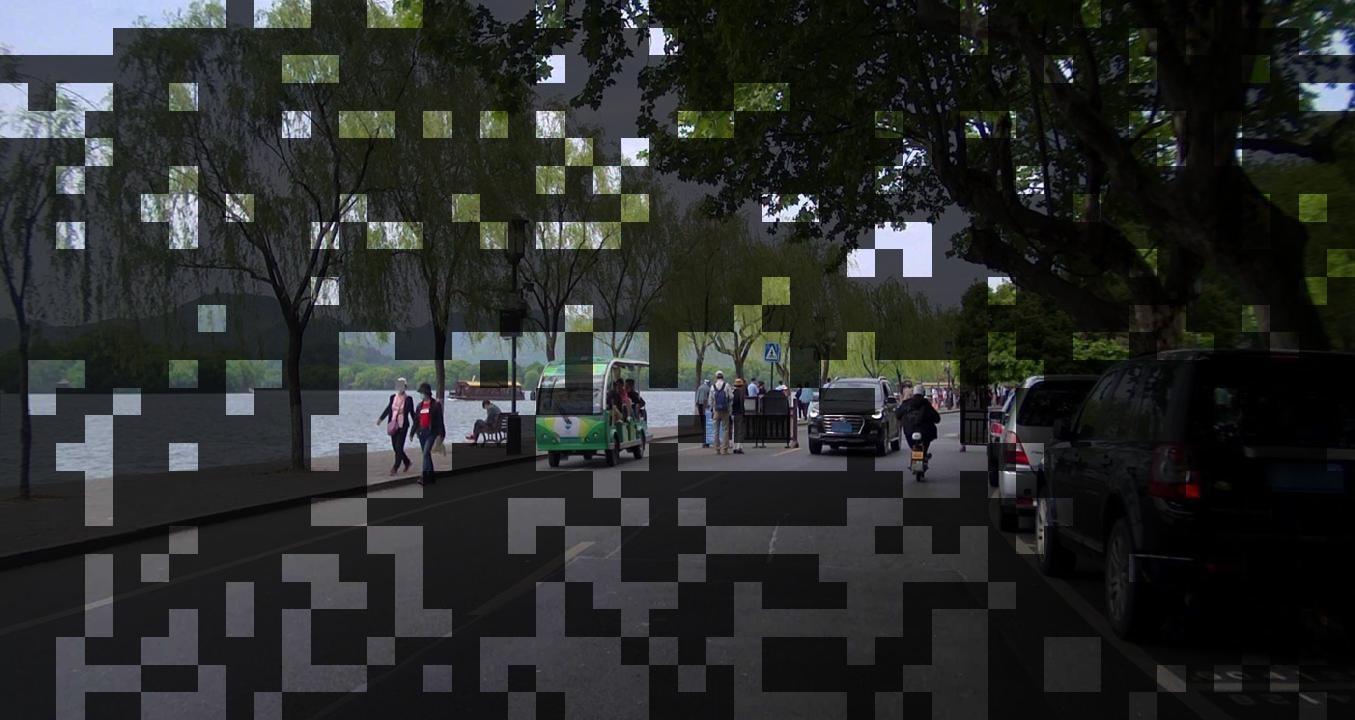}
    \end{subfigure}\hfill
    \begin{subfigure}{0.16\textwidth}
        \centering
        \includegraphics[width=\linewidth]{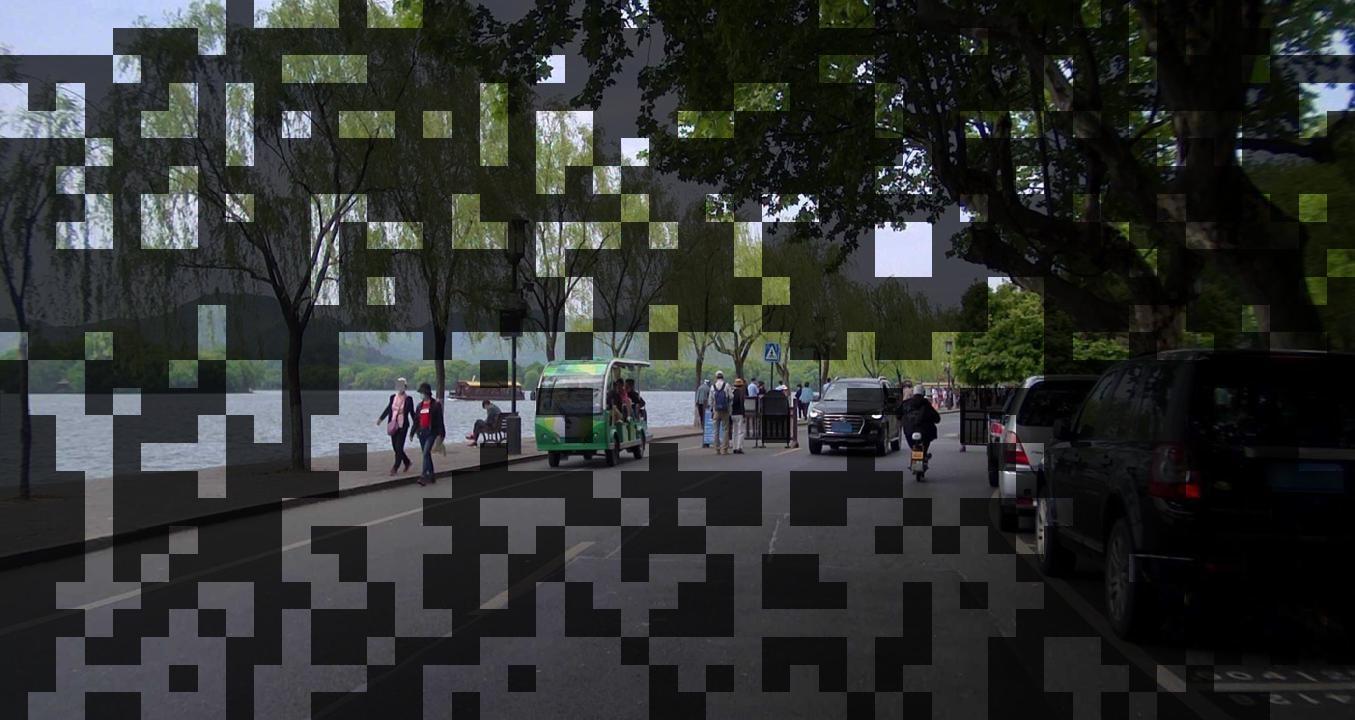}
    \end{subfigure}\hfill
    \begin{subfigure}{0.16\textwidth}
        \centering
        \includegraphics[width=\linewidth]{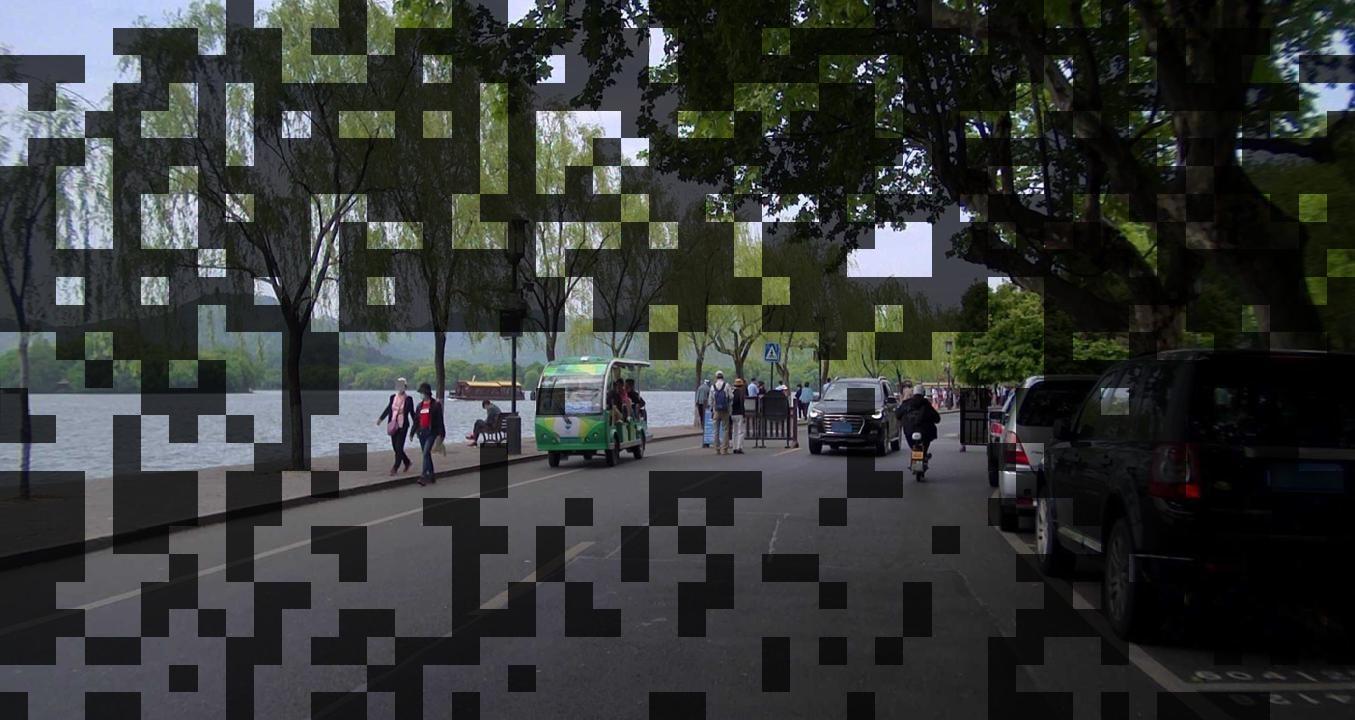}
    \end{subfigure}

    \caption{\textbf{Visualization of pivot buffer construction under different sparsity levels.}
    As the token retention ratio increases from 10\% to 50\%, the buffer progressively covers more visual regions while consistently preserving core semantic objects, local structures, and distributed context. This balanced receiver set supports later backtracking-aware context injection.}
    \label{fig:qualitative_vis_full}
\end{figure*}

To better understand the pivot buffer used in \textbf{ReTrace}, we visualize the retained token maps under different sparsity levels. As shown in Fig.~\ref{fig:qualitative_vis_full}, the constructed buffer consistently covers the main visual subjects while preserving spatially distributed context regions, even under aggressive compression. This is important for ReTrace: the buffer is not intended to be the final reasoning core, but to provide active receivers for later dormant-context restoration.

Our buffer combines global saliency ($T_g$), local saliency ($T_l$), and spatial diversity ($T_d$). These components serve different roles. $T_g$ anchors dominant semantic regions, $T_l$ preserves local structures, and $T_d$ improves scene coverage. Together, they reduce the risk that dormant context will be routed into an overly narrow set of receivers during the injection stage.

\begin{figure*}[p]
    \centering
    \newcommand{\subfigwidth}{0.155\textwidth}

    \begin{subfigure}[c]{0.03\textwidth} \centering \rotatebox{90}{\small \textbf{\textit{001}}} \end{subfigure}\hfill
    \begin{subfigure}[c]{\subfigwidth} \centering \small \textit{Original}\\[1mm] \includegraphics[width=\linewidth]{figs/vis_hpe/input_images/2490.jpg} \end{subfigure}\hfill
    \begin{subfigure}[c]{\subfigwidth} \centering \small \textit{Retain 10\%}\\[1mm] \includegraphics[width=\linewidth]{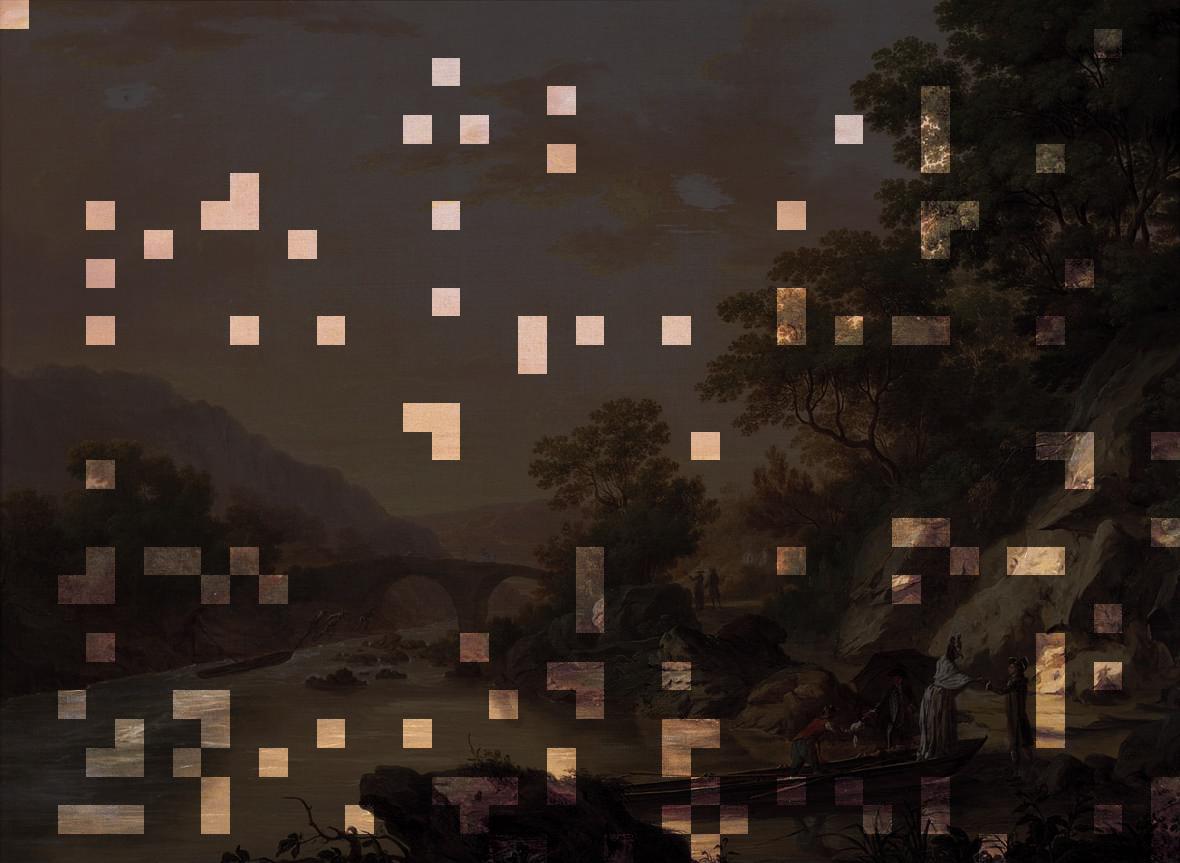} \end{subfigure}\hfill
    \begin{subfigure}[c]{\subfigwidth} \centering \small \textit{Retain 20\%}\\[1mm] \includegraphics[width=\linewidth]{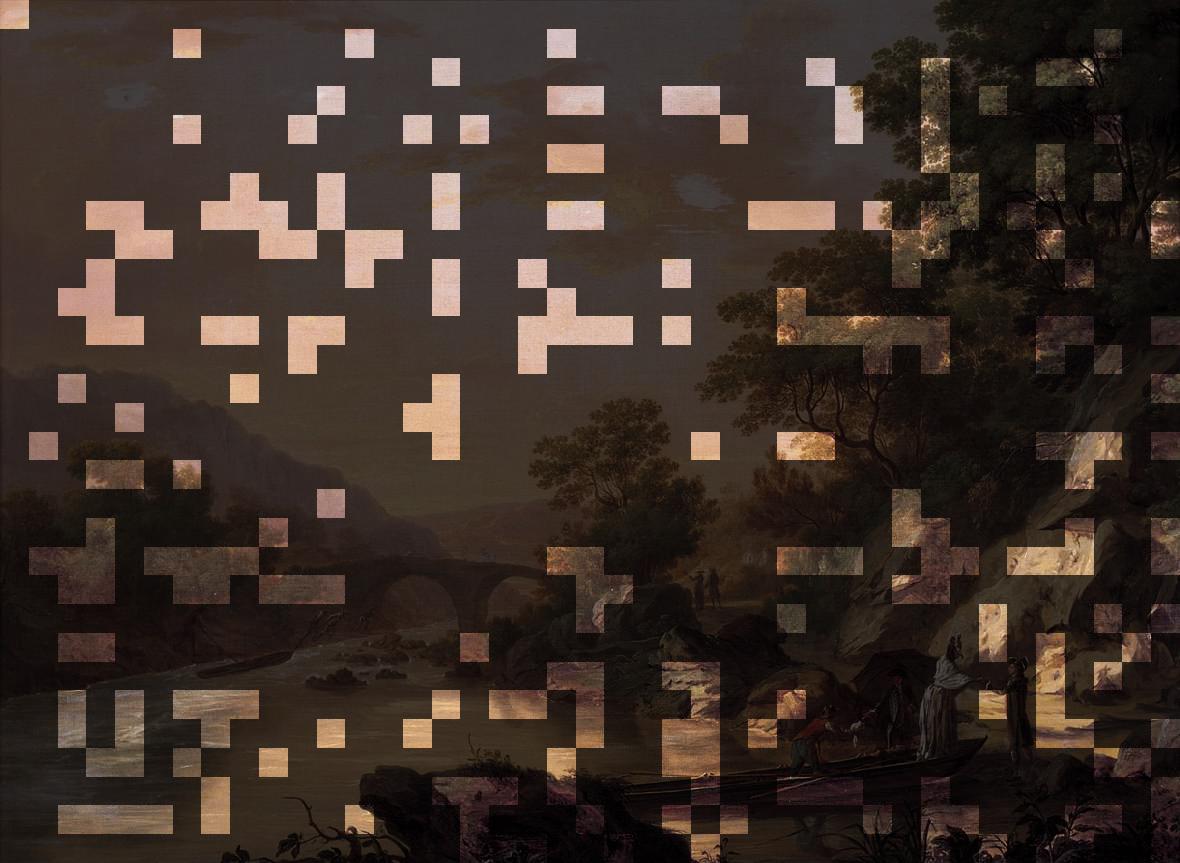} \end{subfigure}\hfill
    \begin{subfigure}[c]{\subfigwidth} \centering \small \textit{Retain 30\%}\\[1mm] \includegraphics[width=\linewidth]{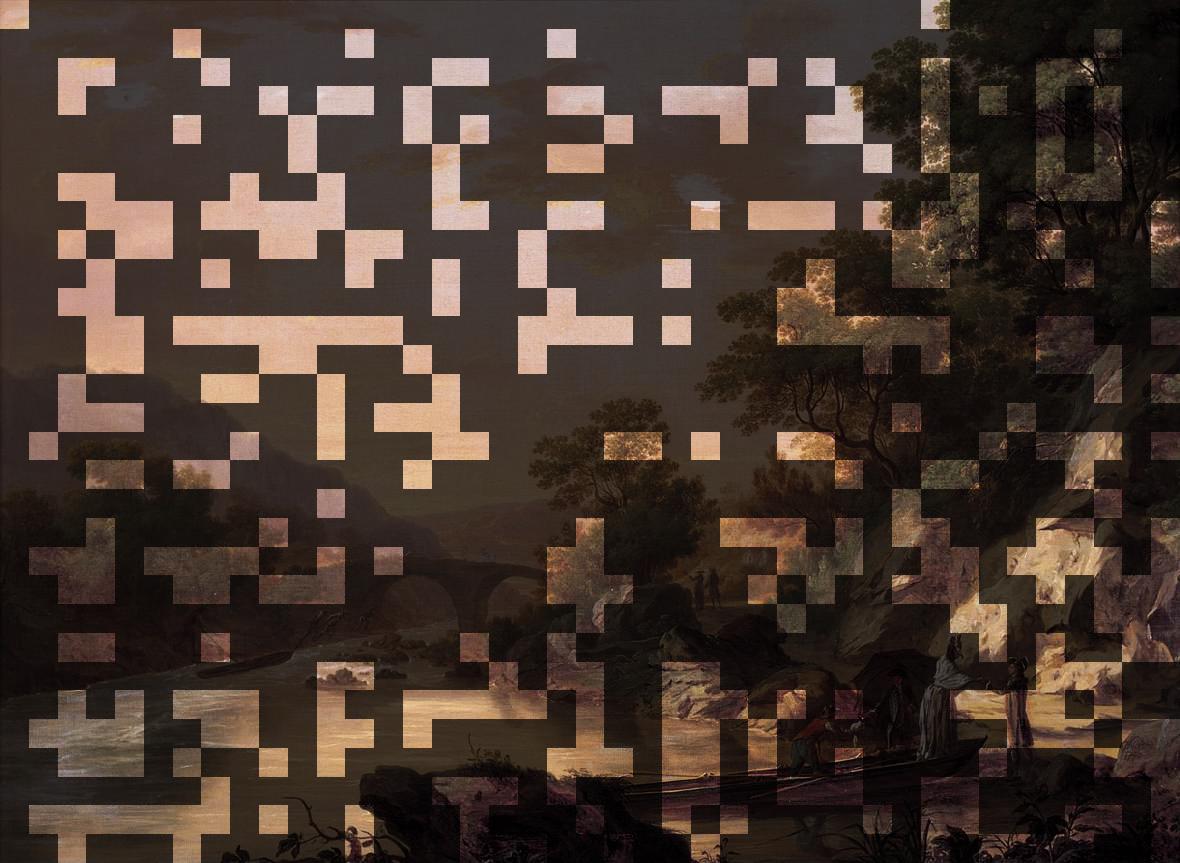} \end{subfigure}\hfill
    \begin{subfigure}[c]{\subfigwidth} \centering \small \textit{Retain 40\%}\\[1mm] \includegraphics[width=\linewidth]{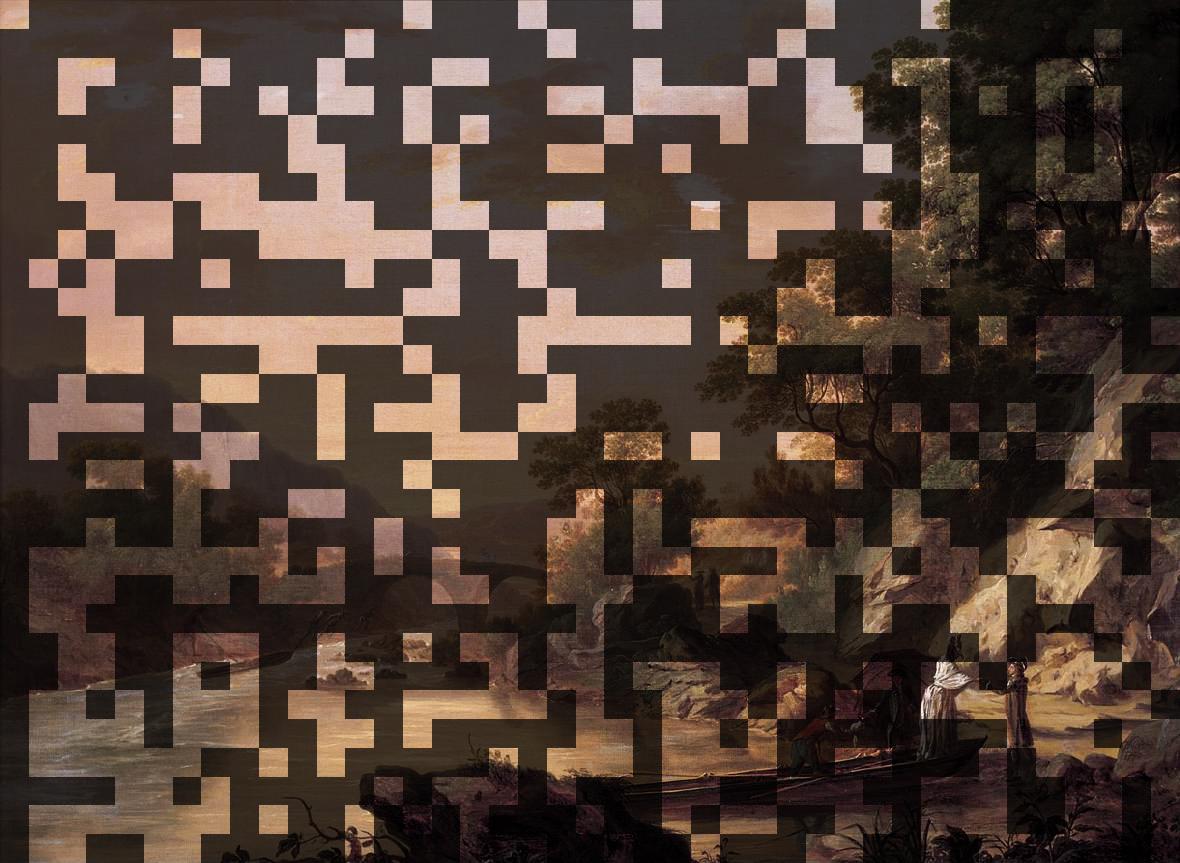} \end{subfigure}\hfill
    \begin{subfigure}[c]{\subfigwidth} \centering \small \textit{Retain 50\%}\\[1mm] \includegraphics[width=\linewidth]{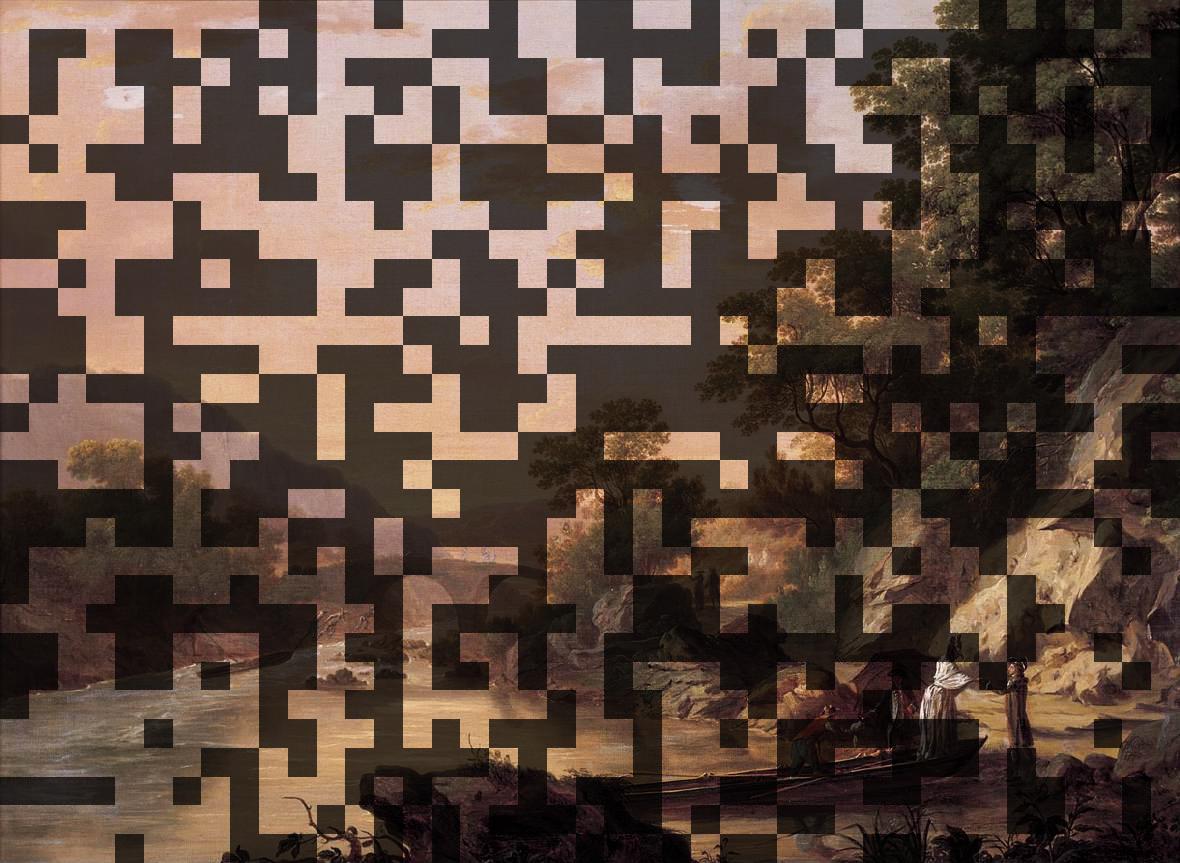} \end{subfigure}\\[1.5mm]

    \begin{subfigure}[c]{0.03\textwidth} \centering \rotatebox{90}{\small \textbf{\textit{100}}} \end{subfigure}\hfill
    \begin{subfigure}[c]{\subfigwidth} \centering \includegraphics[width=\linewidth]{figs/vis_hpe/input_images/2490.jpg} \end{subfigure}\hfill
    \begin{subfigure}[c]{\subfigwidth} \centering \includegraphics[width=\linewidth]{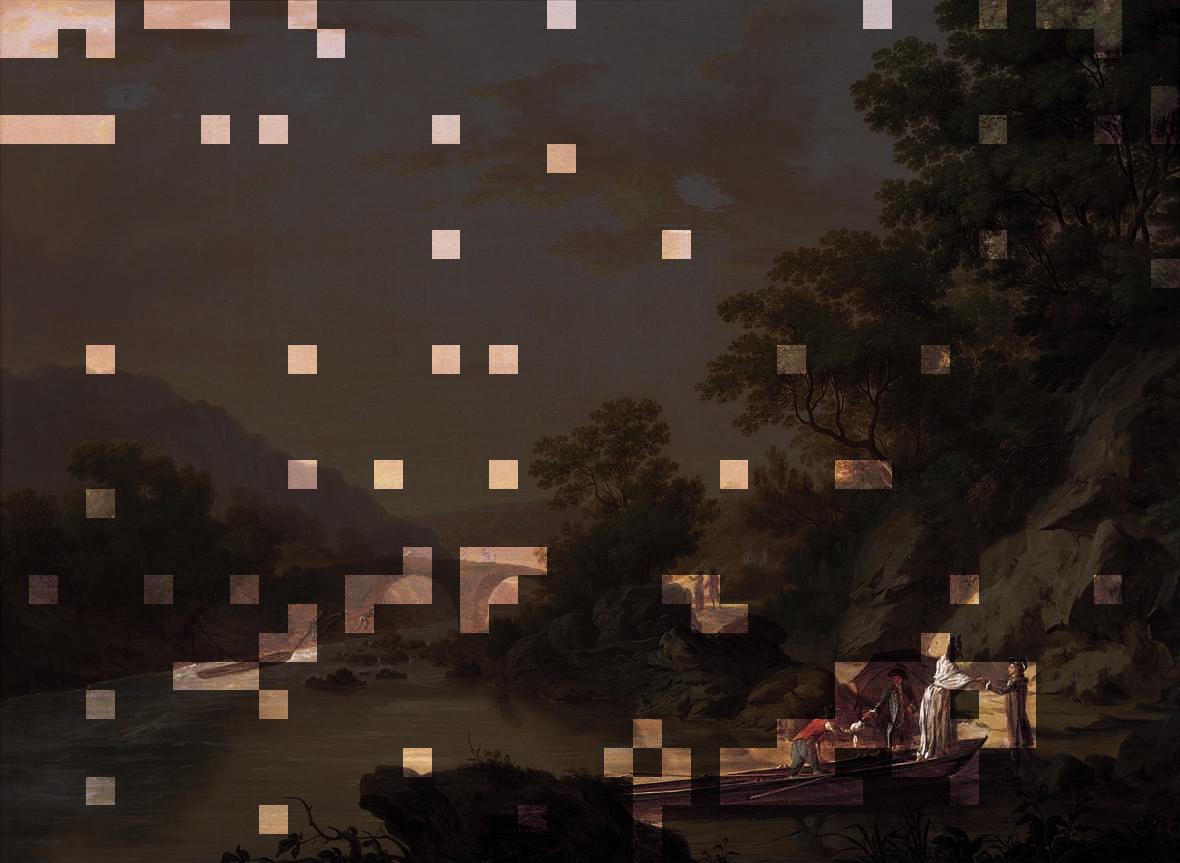} \end{subfigure}\hfill
    \begin{subfigure}[c]{\subfigwidth} \centering \includegraphics[width=\linewidth]{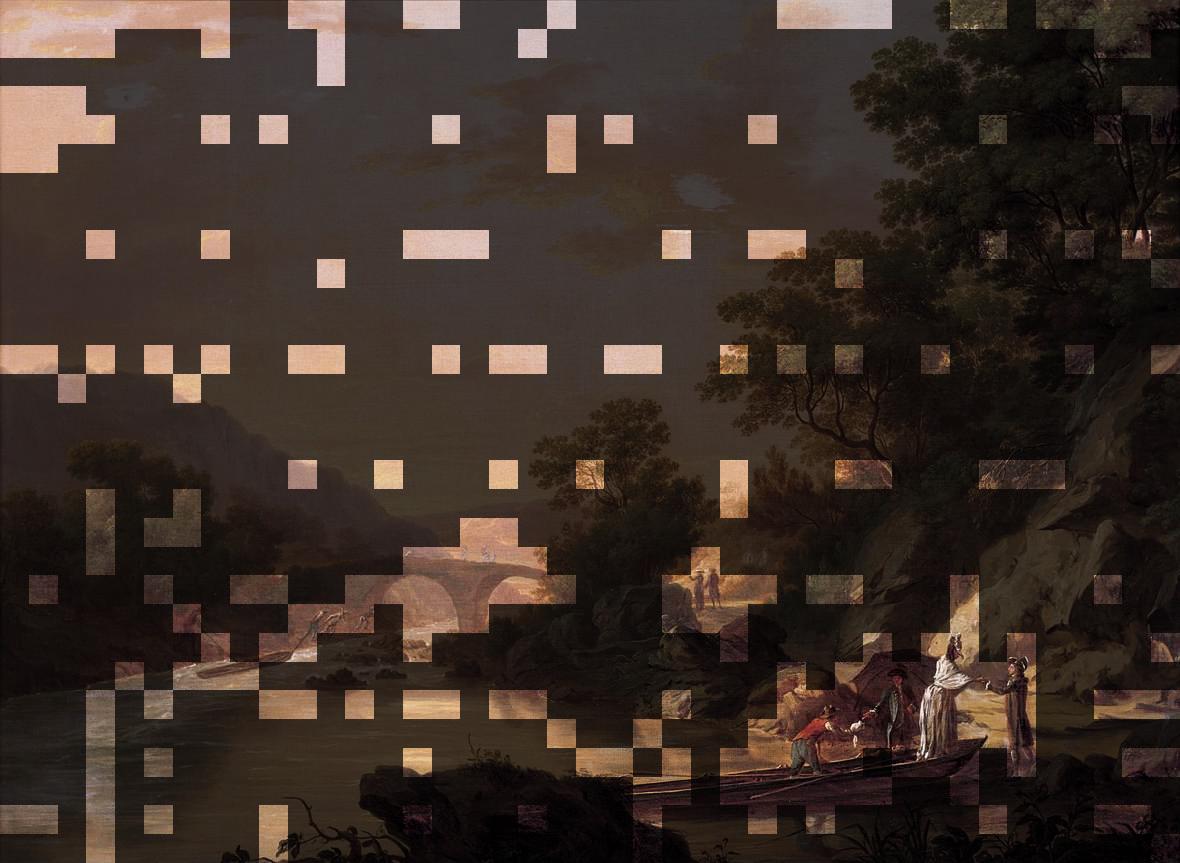} \end{subfigure}\hfill
    \begin{subfigure}[c]{\subfigwidth} \centering \includegraphics[width=\linewidth]{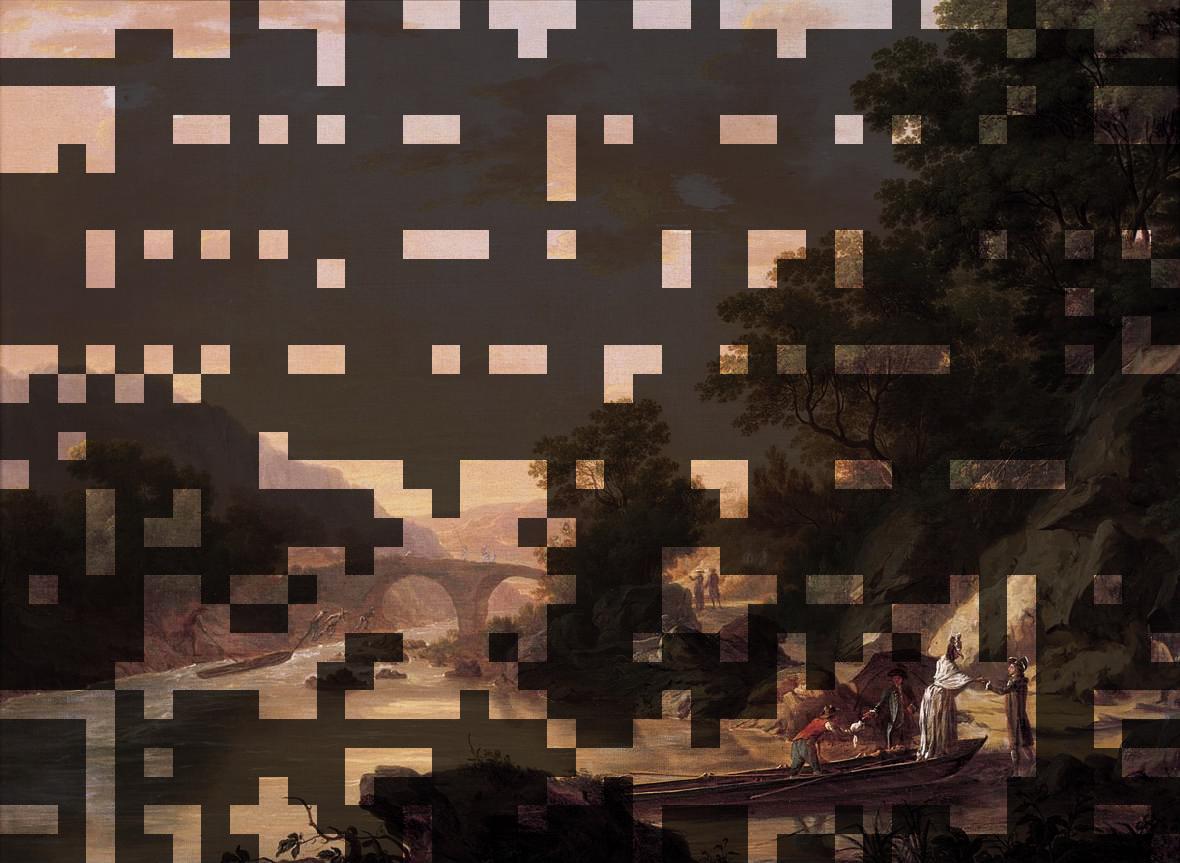} \end{subfigure}\hfill
    \begin{subfigure}[c]{\subfigwidth} \centering \includegraphics[width=\linewidth]{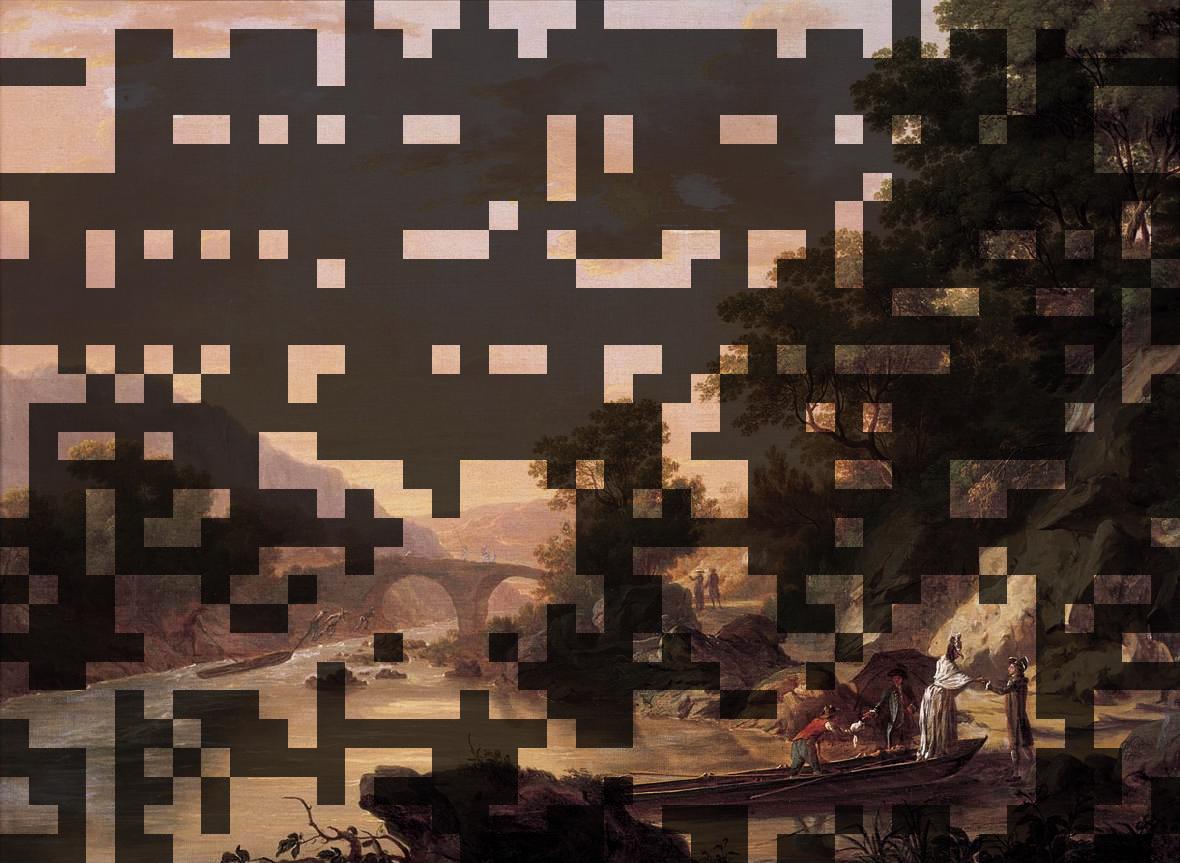} \end{subfigure}\hfill
    \begin{subfigure}[c]{\subfigwidth} \centering \includegraphics[width=\linewidth]{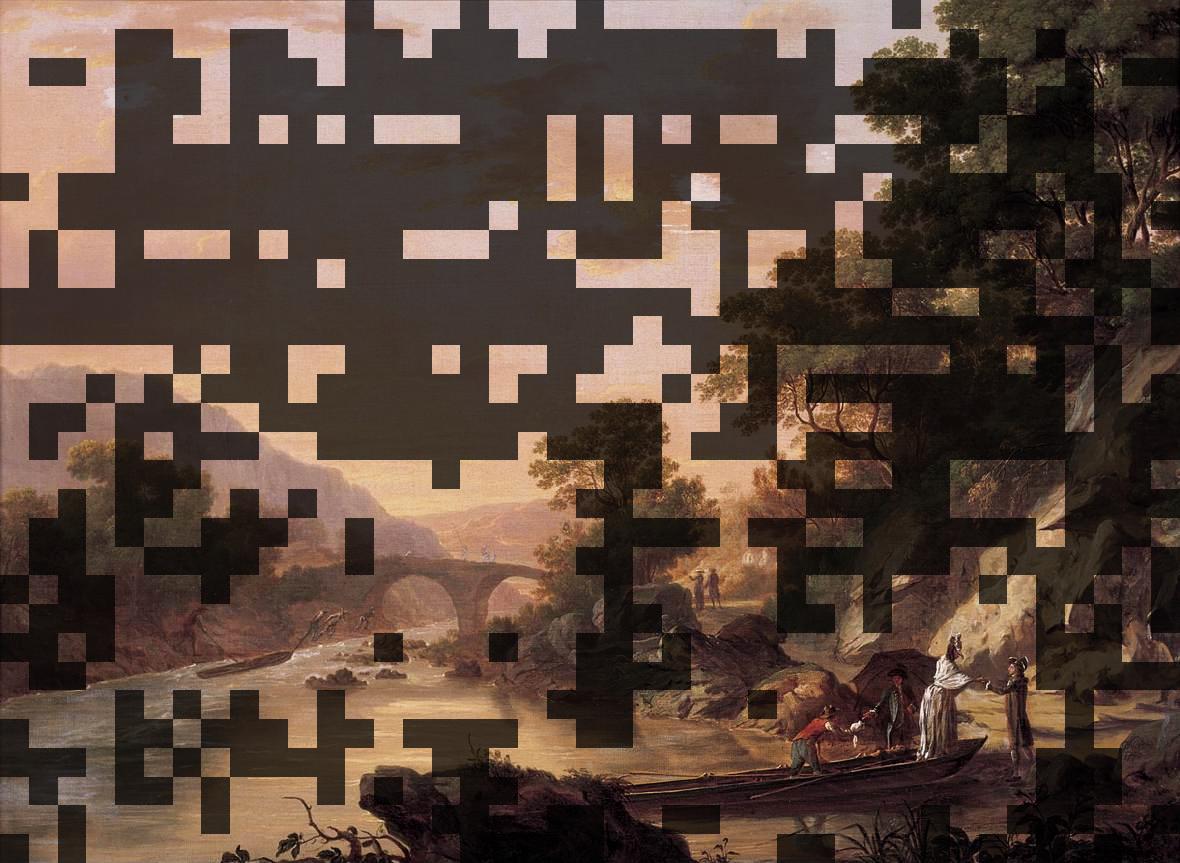} \end{subfigure}\\[1.5mm]

    \begin{subfigure}[c]{0.03\textwidth} \centering \rotatebox{90}{\small \textbf{\textit{101}}} \end{subfigure}\hfill
    \begin{subfigure}[c]{\subfigwidth} \centering \includegraphics[width=\linewidth]{figs/vis_hpe/input_images/2490.jpg} \end{subfigure}\hfill
    \begin{subfigure}[c]{\subfigwidth} \centering \includegraphics[width=\linewidth]{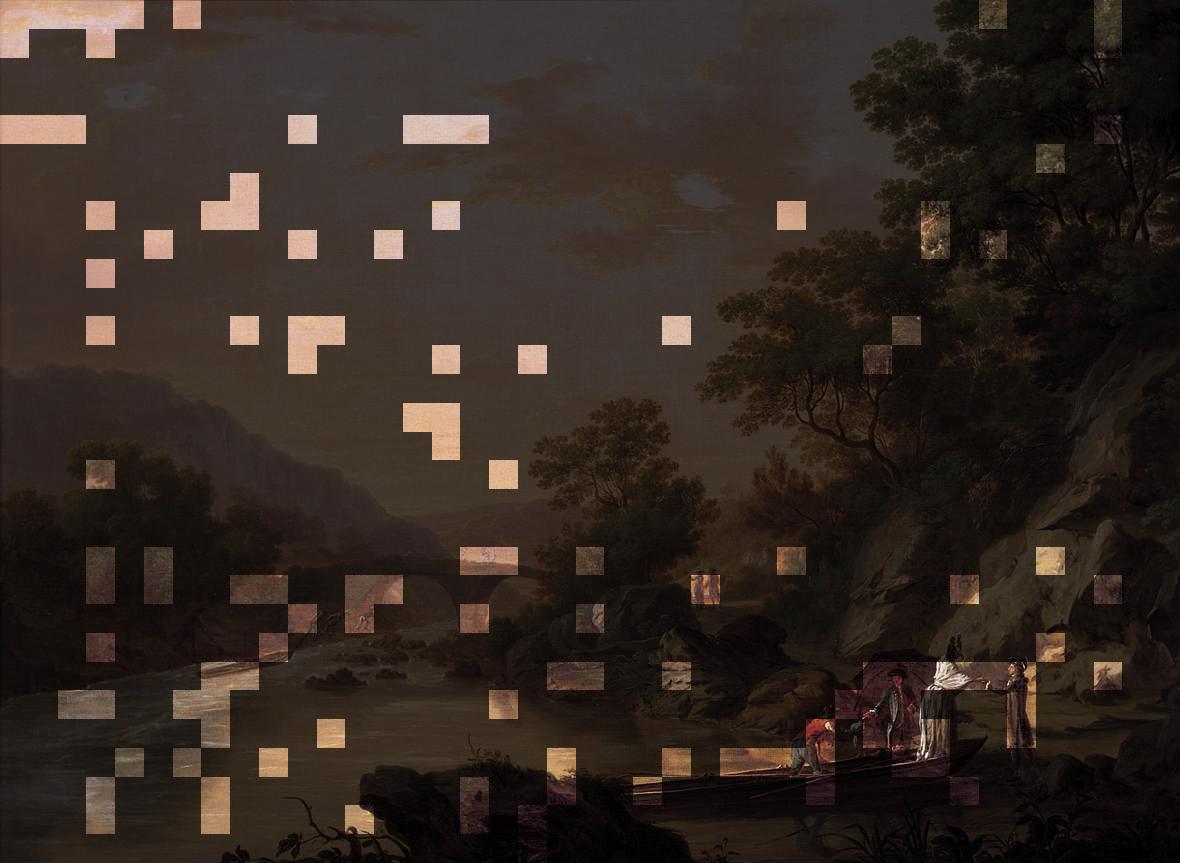} \end{subfigure}\hfill
    \begin{subfigure}[c]{\subfigwidth} \centering \includegraphics[width=\linewidth]{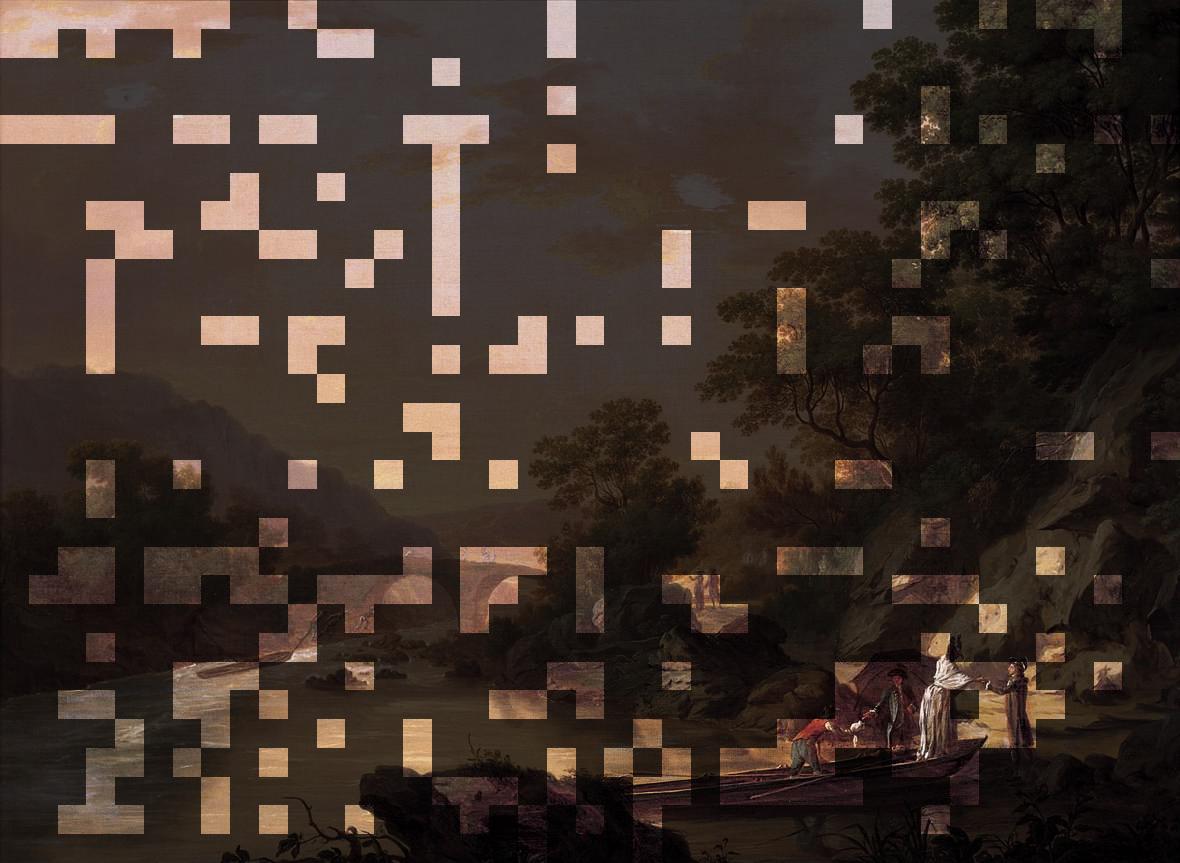} \end{subfigure}\hfill
    \begin{subfigure}[c]{\subfigwidth} \centering \includegraphics[width=\linewidth]{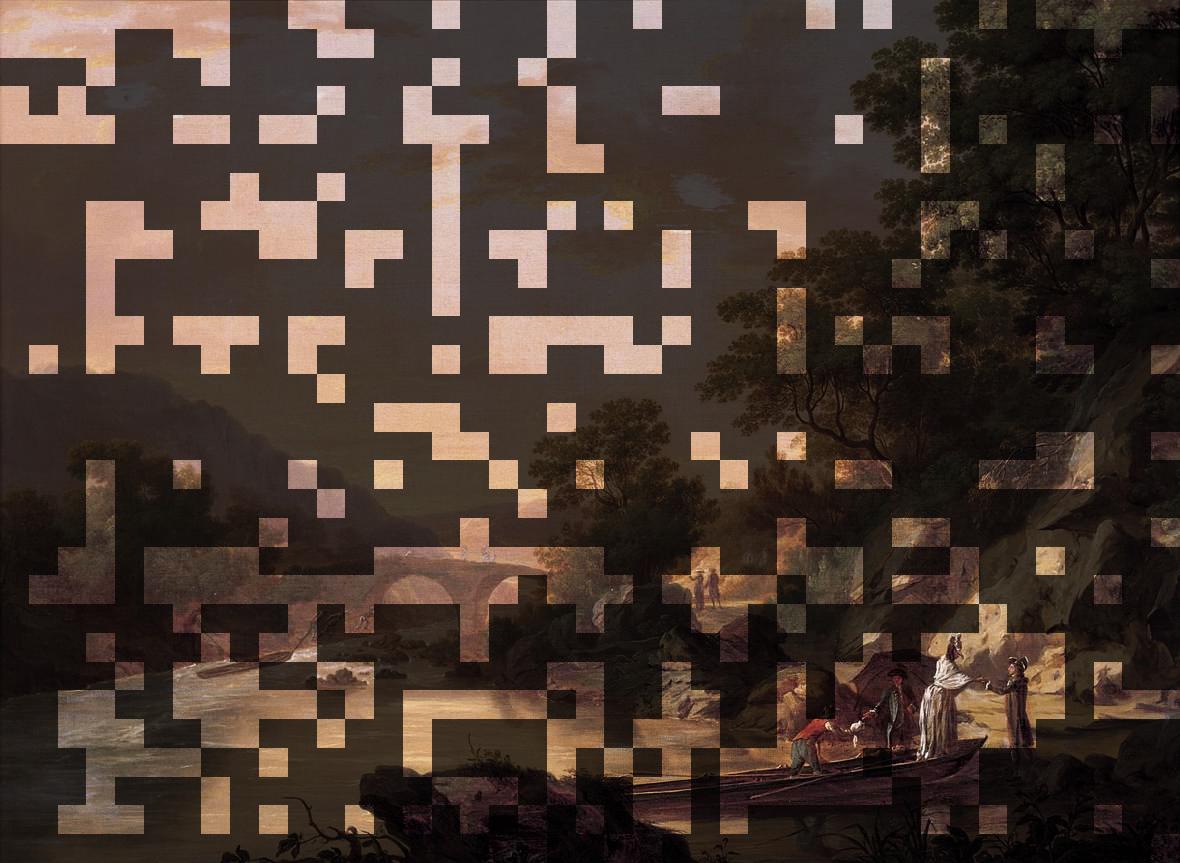} \end{subfigure}\hfill
    \begin{subfigure}[c]{\subfigwidth} \centering \includegraphics[width=\linewidth]{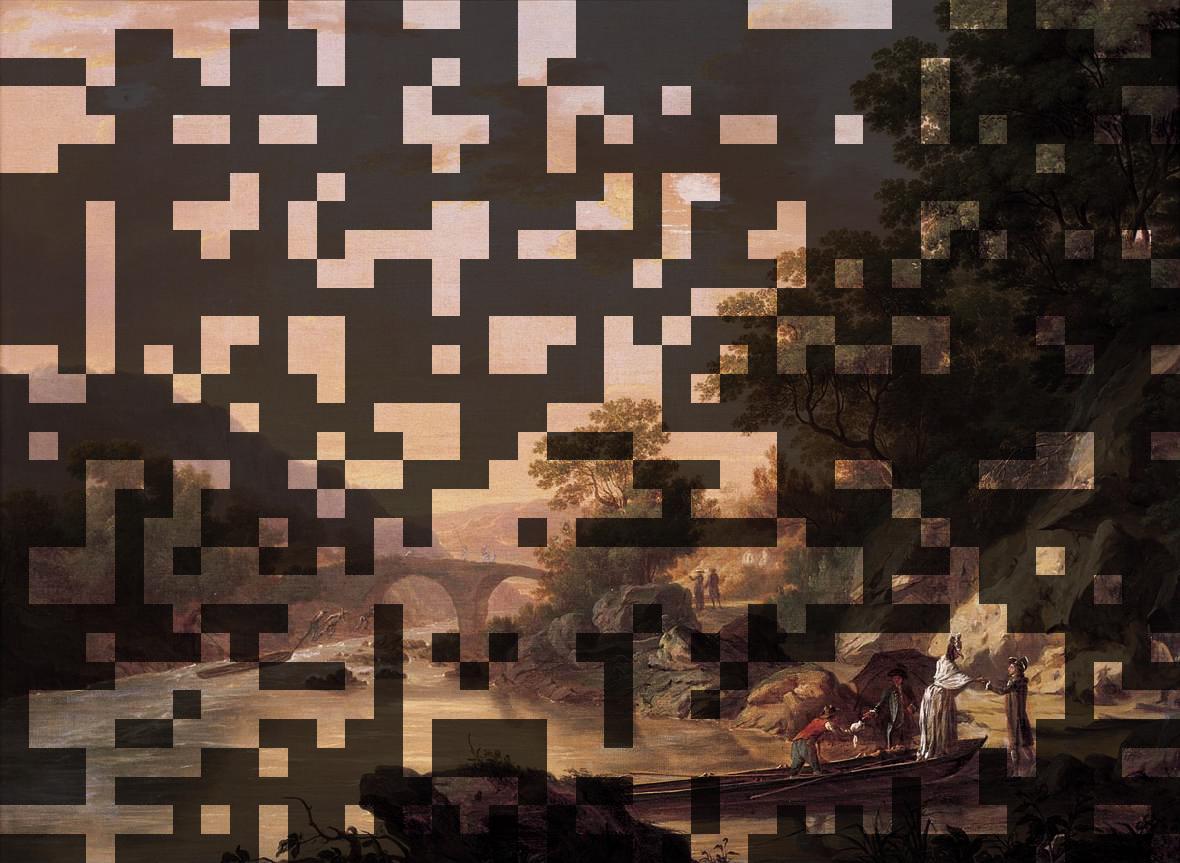} \end{subfigure}\hfill
    \begin{subfigure}[c]{\subfigwidth} \centering \includegraphics[width=\linewidth]{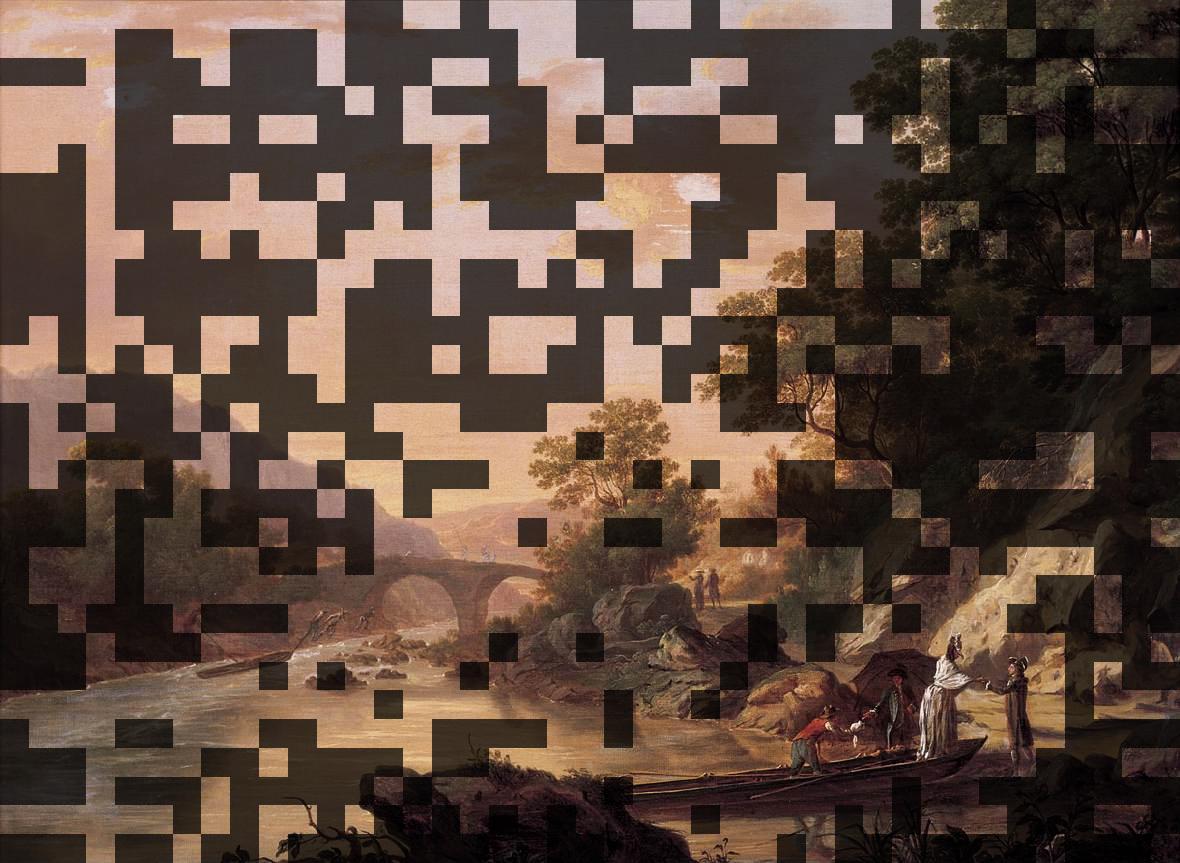} \end{subfigure}\\[1.5mm]

    \begin{subfigure}[c]{0.03\textwidth} \centering \rotatebox{90}{\small \textbf{\textit{110}}} \end{subfigure}\hfill
    \begin{subfigure}[c]{\subfigwidth} \centering \includegraphics[width=\linewidth]{figs/vis_hpe/input_images/2490.jpg} \end{subfigure}\hfill
    \begin{subfigure}[c]{\subfigwidth} \centering \includegraphics[width=\linewidth]{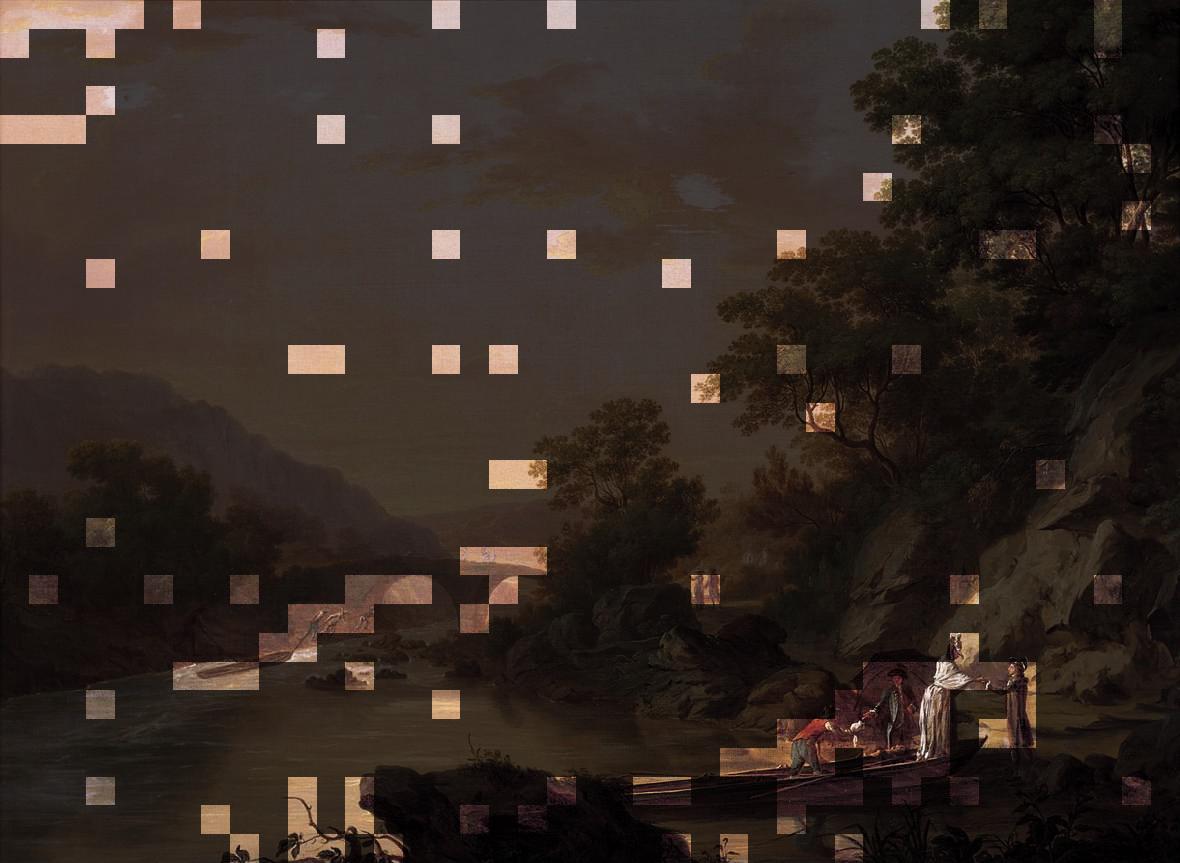} \end{subfigure}\hfill
    \begin{subfigure}[c]{\subfigwidth} \centering \includegraphics[width=\linewidth]{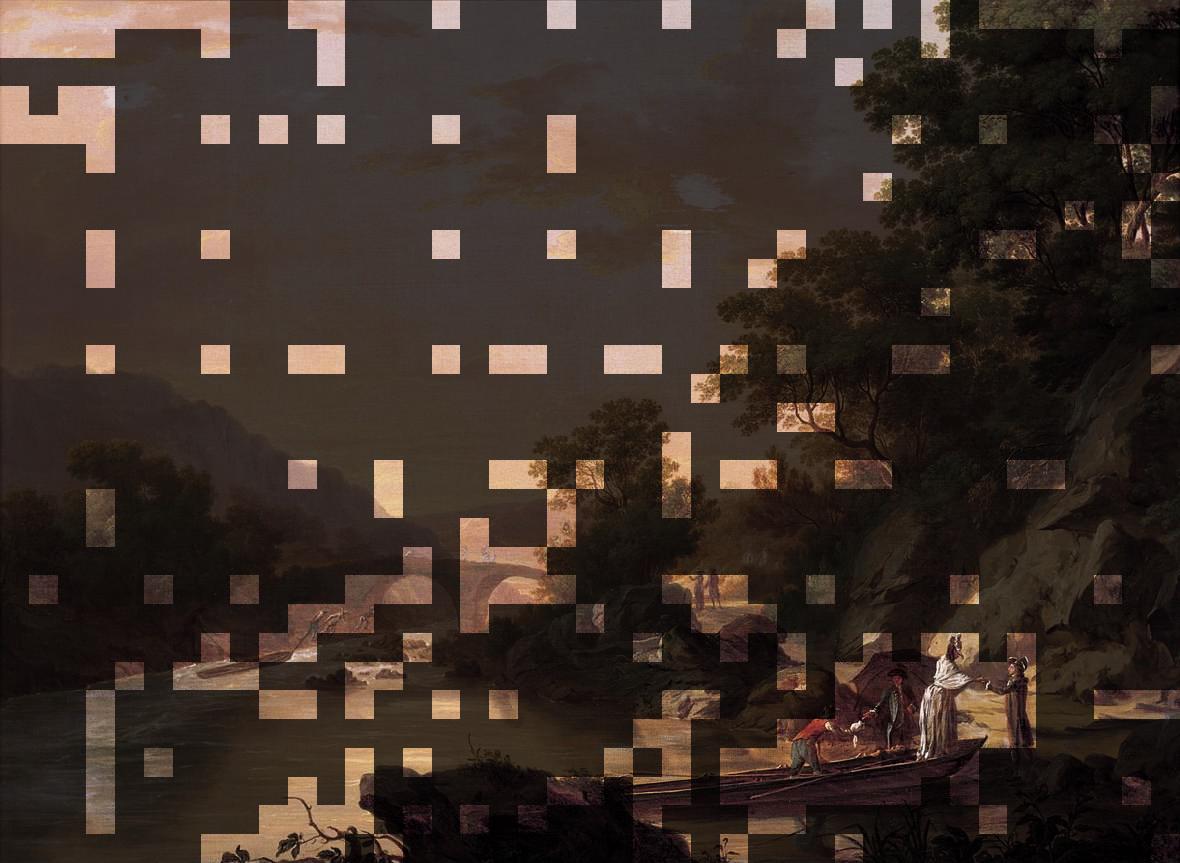} \end{subfigure}\hfill
    \begin{subfigure}[c]{\subfigwidth} \centering \includegraphics[width=\linewidth]{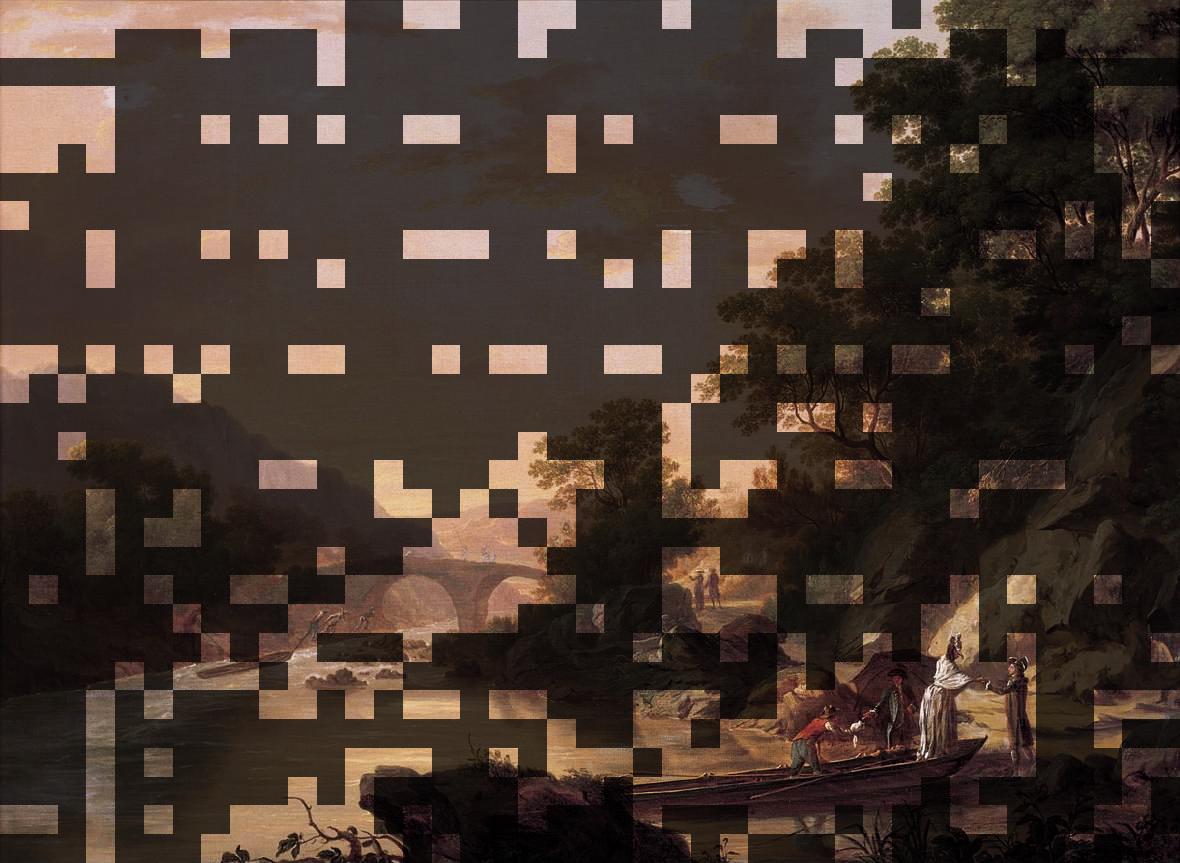} \end{subfigure}\hfill
    \begin{subfigure}[c]{\subfigwidth} \centering \includegraphics[width=\linewidth]{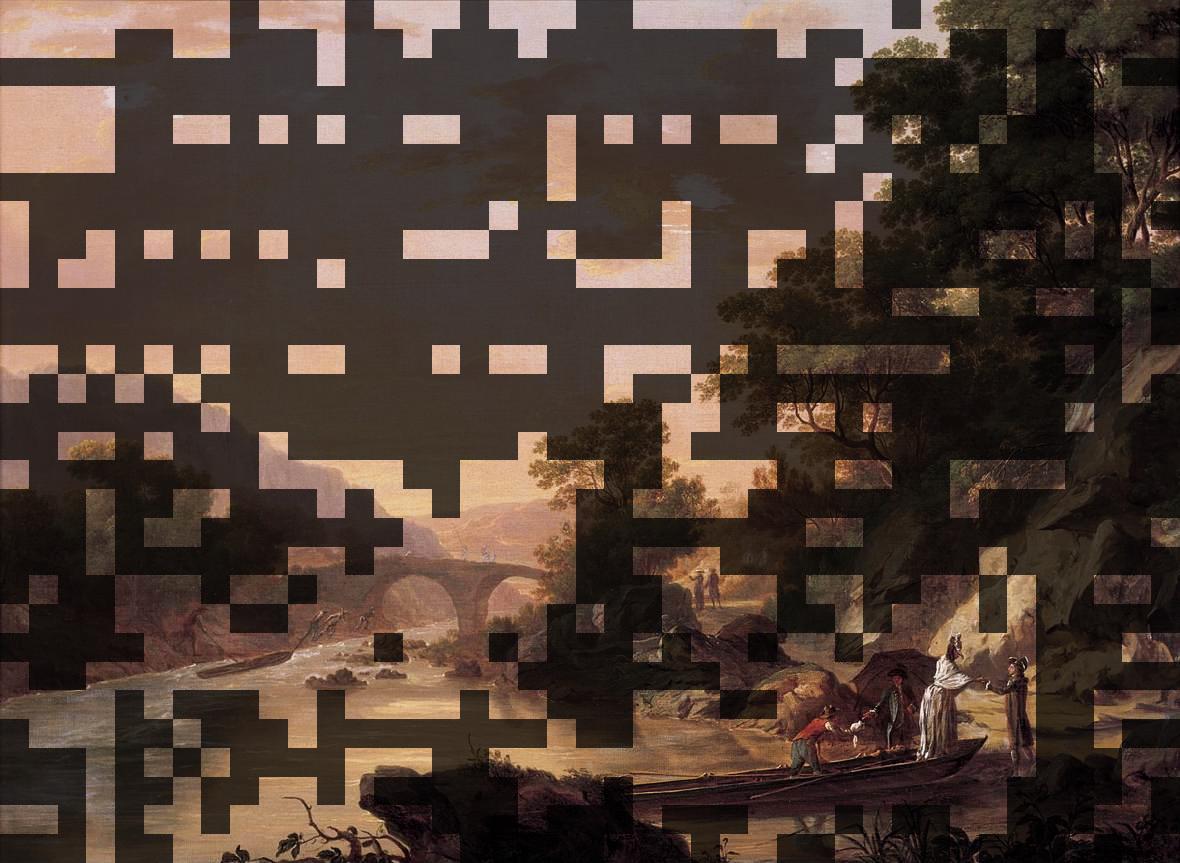} \end{subfigure}\hfill
    \begin{subfigure}[c]{\subfigwidth} \centering \includegraphics[width=\linewidth]{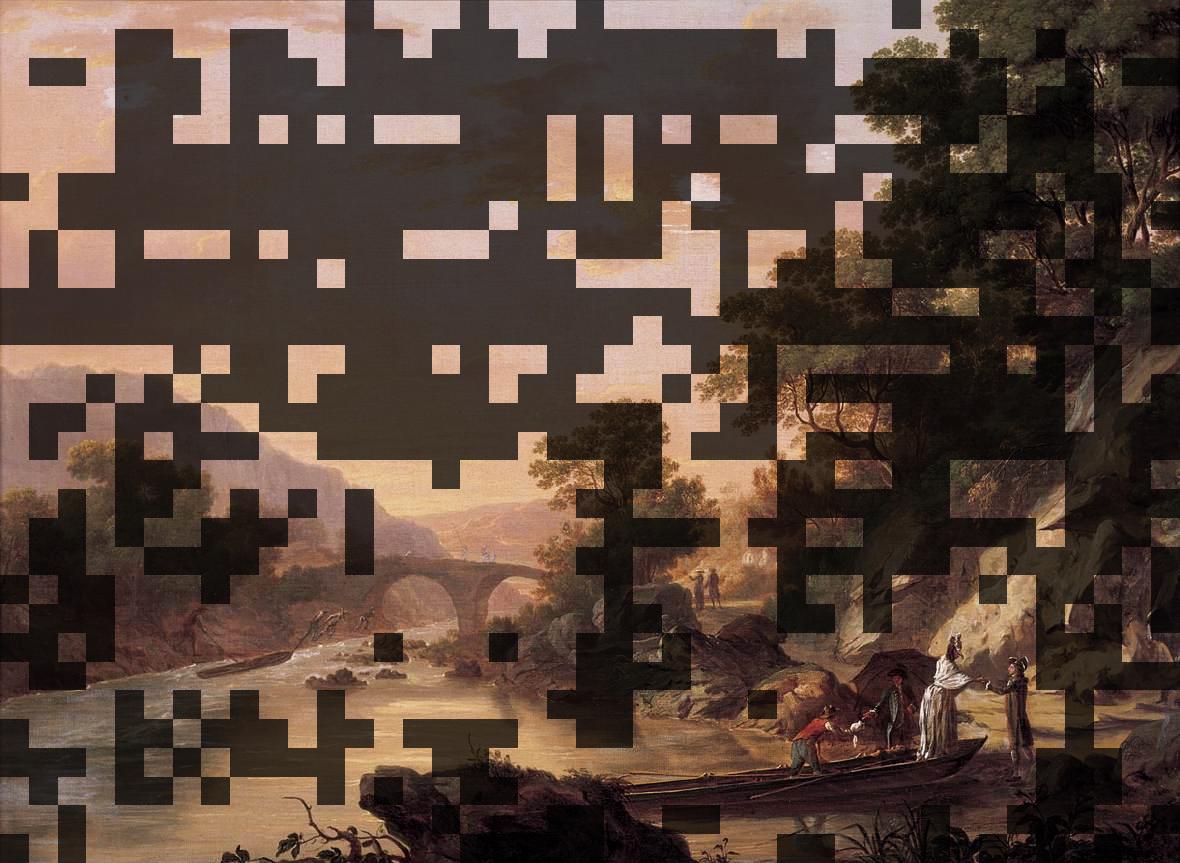} \end{subfigure}\\[1.5mm]

    \begin{subfigure}[c]{0.03\textwidth} \centering \rotatebox{90}{\small \textbf{\textit{ReTrace}}} \end{subfigure}\hfill
    \begin{subfigure}[c]{\subfigwidth} \centering \includegraphics[width=\linewidth]{figs/vis_hpe/input_images/2490.jpg} \end{subfigure}\hfill
    \begin{subfigure}[c]{\subfigwidth} \centering \includegraphics[width=\linewidth]{figs/vis_hpe/vis_keep_0.1/2490.jpg} \end{subfigure}\hfill
    \begin{subfigure}[c]{\subfigwidth} \centering \includegraphics[width=\linewidth]{figs/vis_hpe/vis_keep_0.2/2490.jpg} \end{subfigure}\hfill
    \begin{subfigure}[c]{\subfigwidth} \centering \includegraphics[width=\linewidth]{figs/vis_hpe/vis_keep_0.3/2490.jpg} \end{subfigure}\hfill
    \begin{subfigure}[c]{\subfigwidth} \centering \includegraphics[width=\linewidth]{figs/vis_hpe/vis_keep_0.4/2490.jpg} \end{subfigure}\hfill
    \begin{subfigure}[c]{\subfigwidth} \centering \includegraphics[width=\linewidth]{figs/vis_hpe/vis_keep_0.5/2490.jpg} \end{subfigure}
    \caption{\textbf{Qualitative comparison.}}
    \label{fig:ablation_vis_2490}
\end{figure*}

\begin{figure*}[p]
    \centering
    \newcommand{\subfigwidth}{0.155\textwidth}

    \begin{subfigure}[c]{0.03\textwidth} \centering \rotatebox{90}{\small \textbf{\textit{001}}} \end{subfigure}\hfill
    \begin{subfigure}[c]{\subfigwidth} \centering \small \textit{Original}\\[1mm] \includegraphics[width=\linewidth]{figs/vis_hpe/input_images/COCO_val2014_000000000074.jpg} \end{subfigure}\hfill
    \begin{subfigure}[c]{\subfigwidth} \centering \small \textit{Retain 10\%}\\[1mm] \includegraphics[width=\linewidth]{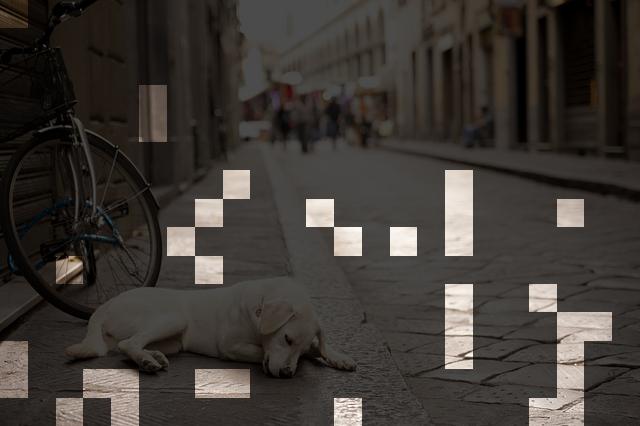} \end{subfigure}\hfill
    \begin{subfigure}[c]{\subfigwidth} \centering \small \textit{Retain 20\%}\\[1mm] \includegraphics[width=\linewidth]{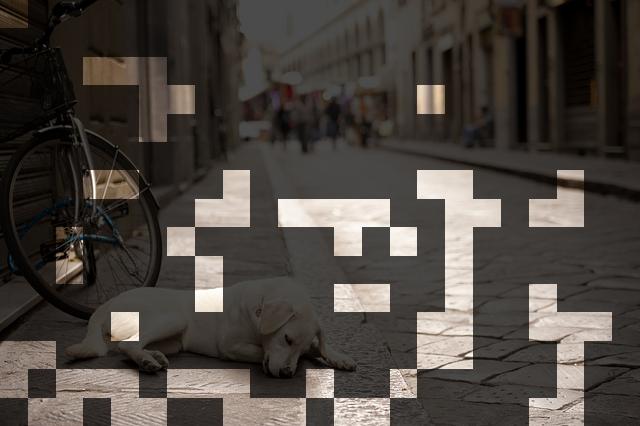} \end{subfigure}\hfill
    \begin{subfigure}[c]{\subfigwidth} \centering \small \textit{Retain 30\%}\\[1mm] \includegraphics[width=\linewidth]{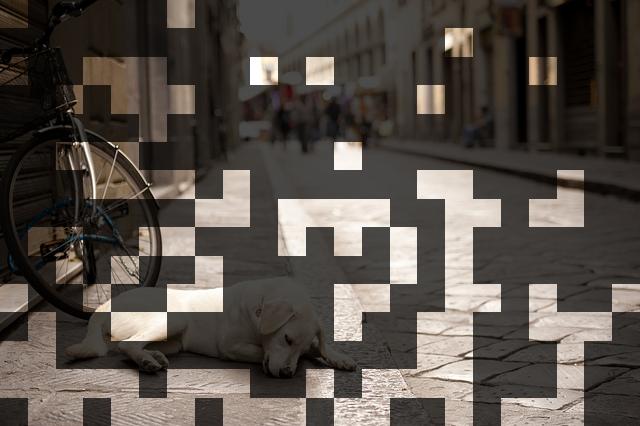} \end{subfigure}\hfill
    \begin{subfigure}[c]{\subfigwidth} \centering \small \textit{Retain 40\%}\\[1mm] \includegraphics[width=\linewidth]{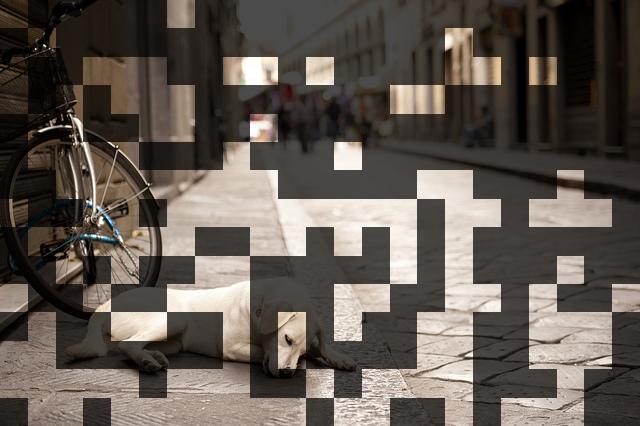} \end{subfigure}\hfill
    \begin{subfigure}[c]{\subfigwidth} \centering \small \textit{Retain 50\%}\\[1mm] \includegraphics[width=\linewidth]{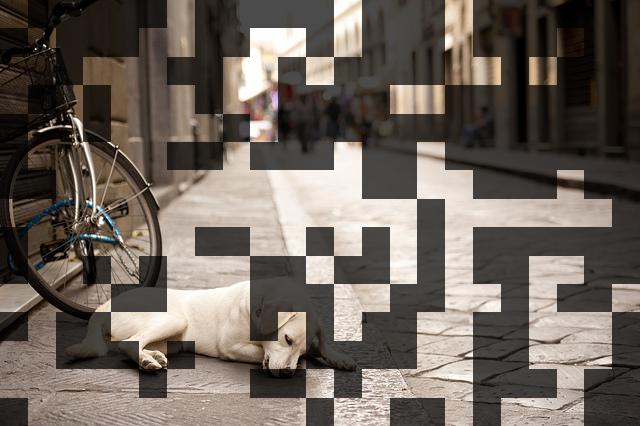} \end{subfigure}\\[1.5mm]

    \begin{subfigure}[c]{0.03\textwidth} \centering \rotatebox{90}{\small \textbf{\textit{100}}} \end{subfigure}\hfill
    \begin{subfigure}[c]{\subfigwidth} \centering \includegraphics[width=\linewidth]{figs/vis_hpe/input_images/COCO_val2014_000000000074.jpg} \end{subfigure}\hfill
    \begin{subfigure}[c]{\subfigwidth} \centering \includegraphics[width=\linewidth]{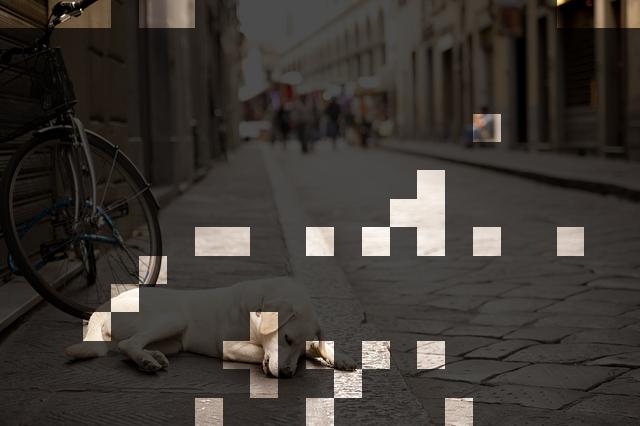} \end{subfigure}\hfill
    \begin{subfigure}[c]{\subfigwidth} \centering \includegraphics[width=\linewidth]{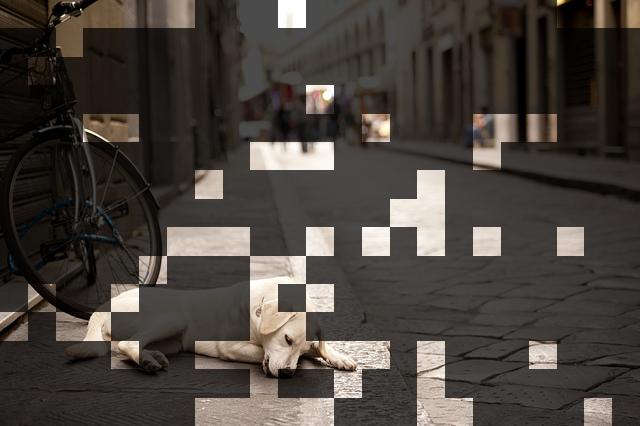} \end{subfigure}\hfill
    \begin{subfigure}[c]{\subfigwidth} \centering \includegraphics[width=\linewidth]{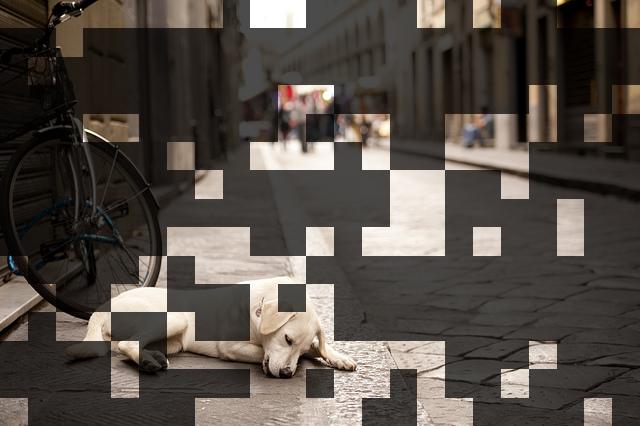} \end{subfigure}\hfill
    \begin{subfigure}[c]{\subfigwidth} \centering \includegraphics[width=\linewidth]{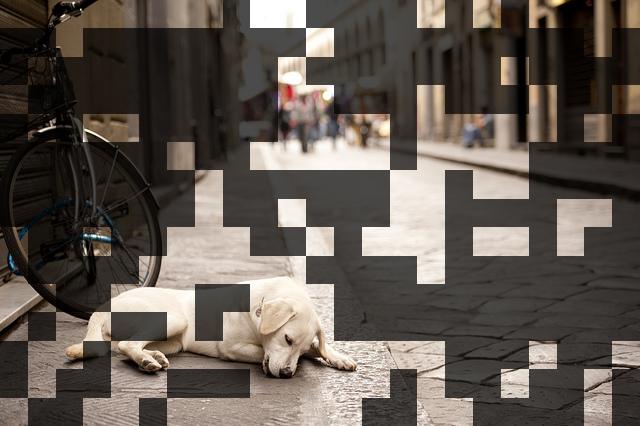} \end{subfigure}\hfill
    \begin{subfigure}[c]{\subfigwidth} \centering \includegraphics[width=\linewidth]{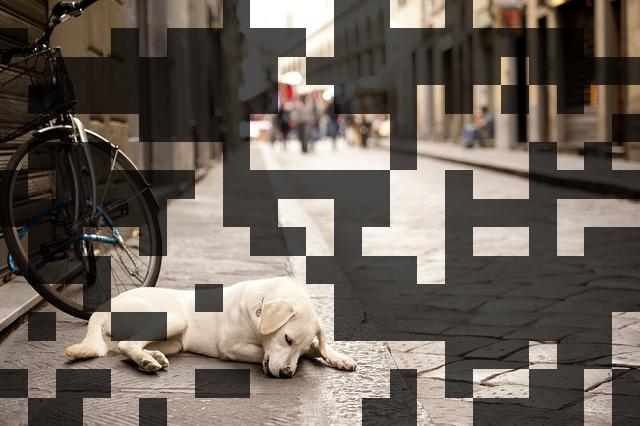} \end{subfigure}\\[1.5mm]

    \begin{subfigure}[c]{0.03\textwidth} \centering \rotatebox{90}{\small \textbf{\textit{101}}} \end{subfigure}\hfill
    \begin{subfigure}[c]{\subfigwidth} \centering \includegraphics[width=\linewidth]{figs/vis_hpe/input_images/COCO_val2014_000000000074.jpg} \end{subfigure}\hfill
    \begin{subfigure}[c]{\subfigwidth} \centering \includegraphics[width=\linewidth]{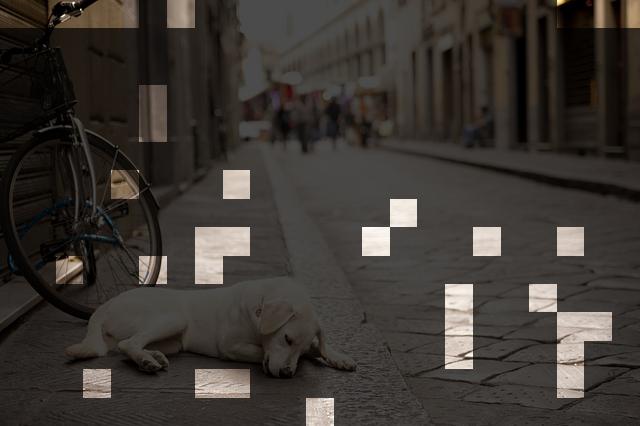} \end{subfigure}\hfill
    \begin{subfigure}[c]{\subfigwidth} \centering \includegraphics[width=\linewidth]{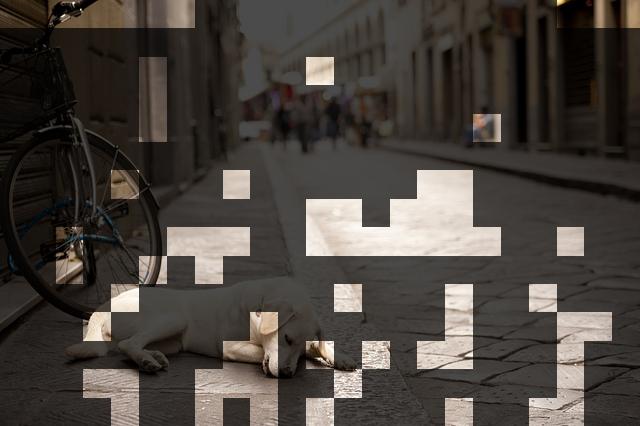} \end{subfigure}\hfill
    \begin{subfigure}[c]{\subfigwidth} \centering \includegraphics[width=\linewidth]{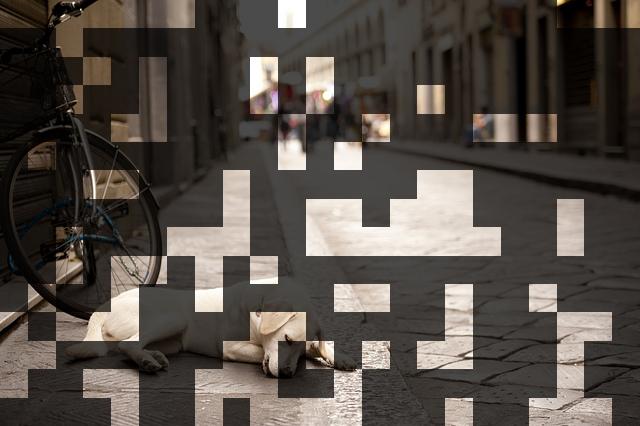} \end{subfigure}\hfill
    \begin{subfigure}[c]{\subfigwidth} \centering \includegraphics[width=\linewidth]{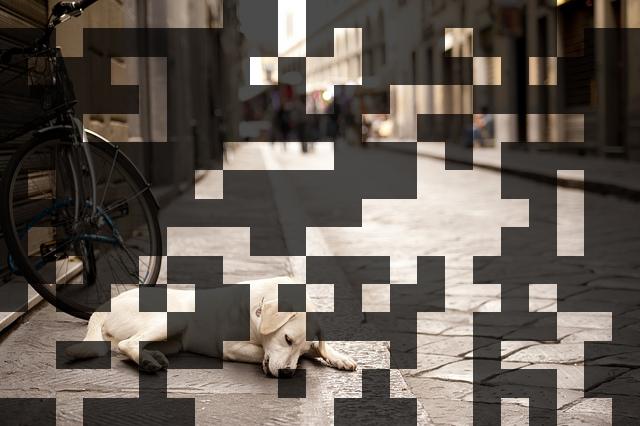} \end{subfigure}\hfill
    \begin{subfigure}[c]{\subfigwidth} \centering \includegraphics[width=\linewidth]{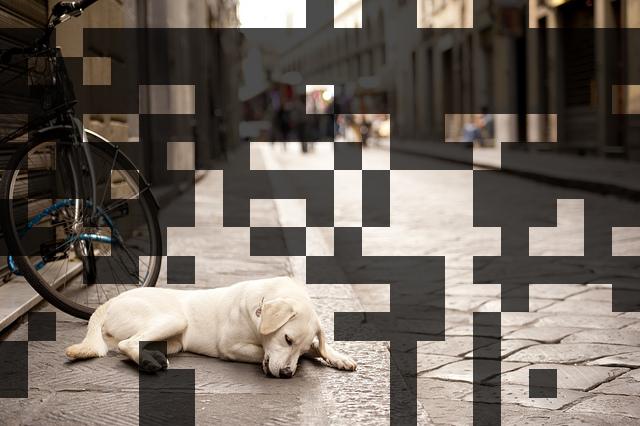} \end{subfigure}\\[1.5mm]

    \begin{subfigure}[c]{0.03\textwidth} \centering \rotatebox{90}{\small \textbf{\textit{110}}} \end{subfigure}\hfill
    \begin{subfigure}[c]{\subfigwidth} \centering \includegraphics[width=\linewidth]{figs/vis_hpe/input_images/COCO_val2014_000000000074.jpg} \end{subfigure}\hfill
    \begin{subfigure}[c]{\subfigwidth} \centering \includegraphics[width=\linewidth]{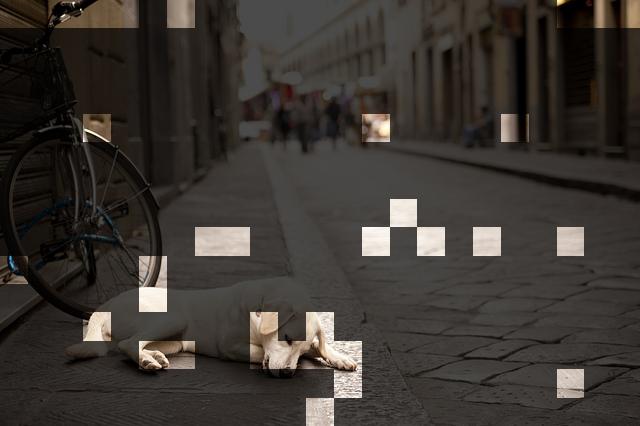} \end{subfigure}\hfill
    \begin{subfigure}[c]{\subfigwidth} \centering \includegraphics[width=\linewidth]{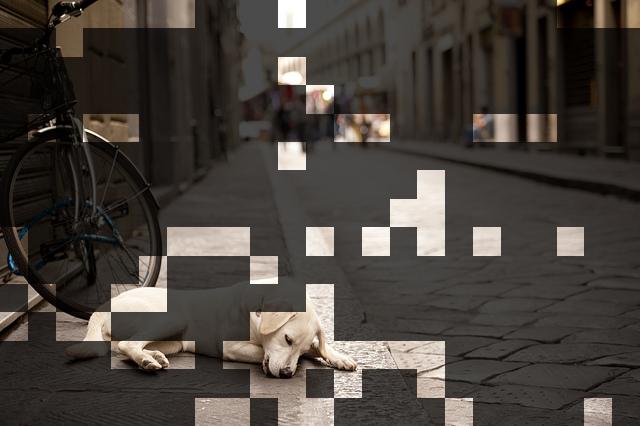} \end{subfigure}\hfill
    \begin{subfigure}[c]{\subfigwidth} \centering \includegraphics[width=\linewidth]{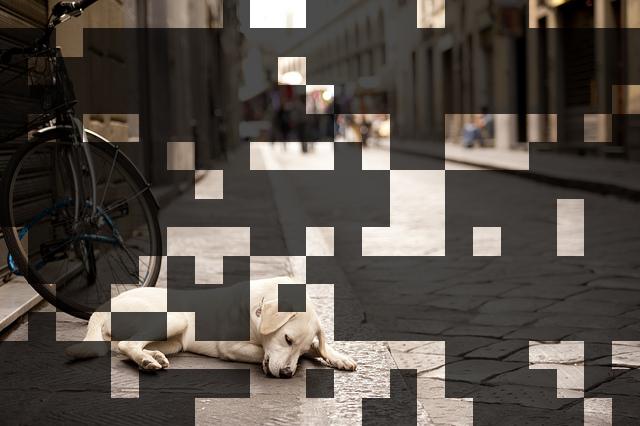} \end{subfigure}\hfill
    \begin{subfigure}[c]{\subfigwidth} \centering \includegraphics[width=\linewidth]{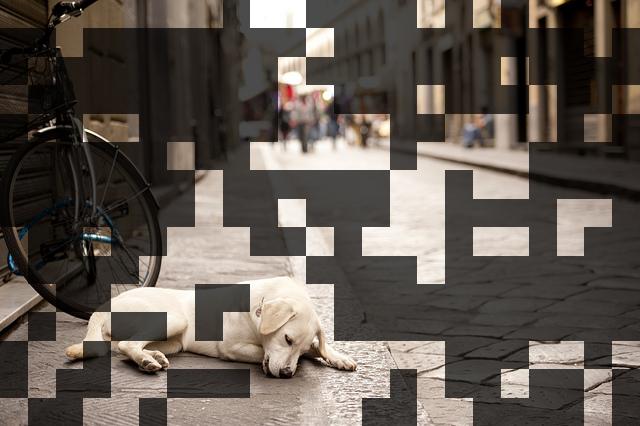} \end{subfigure}\hfill
    \begin{subfigure}[c]{\subfigwidth} \centering \includegraphics[width=\linewidth]{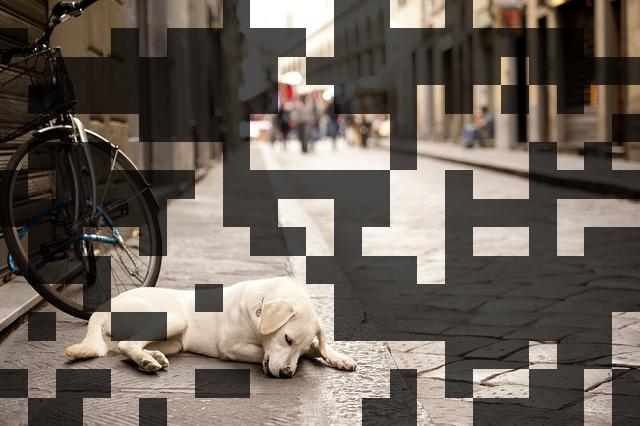} \end{subfigure}\\[1.5mm]

    \begin{subfigure}[c]{0.03\textwidth} \centering \rotatebox{90}{\small \textbf{\textit{ReTrace}}} \end{subfigure}\hfill
    \begin{subfigure}[c]{\subfigwidth} \centering \includegraphics[width=\linewidth]{figs/vis_hpe/input_images/COCO_val2014_000000000074.jpg} \end{subfigure}\hfill
    \begin{subfigure}[c]{\subfigwidth} \centering \includegraphics[width=\linewidth]{figs/vis_hpe/vis_keep_0.1/COCO_val2014_000000000074.jpg} \end{subfigure}\hfill
    \begin{subfigure}[c]{\subfigwidth} \centering \includegraphics[width=\linewidth]{figs/vis_hpe/vis_keep_0.2/COCO_val2014_000000000074.jpg} \end{subfigure}\hfill
    \begin{subfigure}[c]{\subfigwidth} \centering \includegraphics[width=\linewidth]{figs/vis_hpe/vis_keep_0.3/COCO_val2014_000000000074.jpg} \end{subfigure}\hfill
    \begin{subfigure}[c]{\subfigwidth} \centering \includegraphics[width=\linewidth]{figs/vis_hpe/vis_keep_0.4/COCO_val2014_000000000074.jpg} \end{subfigure}\hfill
    \begin{subfigure}[c]{\subfigwidth} \centering \includegraphics[width=\linewidth]{figs/vis_hpe/vis_keep_0.5/COCO_val2014_000000000074.jpg} \end{subfigure}
    \caption{\textbf{Qualitative comparison.}}
    \label{fig:ablation_vis_coco_74}
\end{figure*}

\begin{figure*}[p]
    \centering
    \newcommand{\subfigwidth}{0.155\textwidth}

    \begin{subfigure}[c]{0.03\textwidth} \centering \rotatebox{90}{\small \textbf{\textit{001}}} \end{subfigure}\hfill
    \begin{subfigure}[c]{\subfigwidth} \centering \small \textit{Original}\\[1mm] \includegraphics[width=\linewidth]{figs/vis_hpe/input_images/COCO_val2014_000000000192.jpg} \end{subfigure}\hfill
    \begin{subfigure}[c]{\subfigwidth} \centering \small \textit{Retain 10\%}\\[1mm] \includegraphics[width=\linewidth]{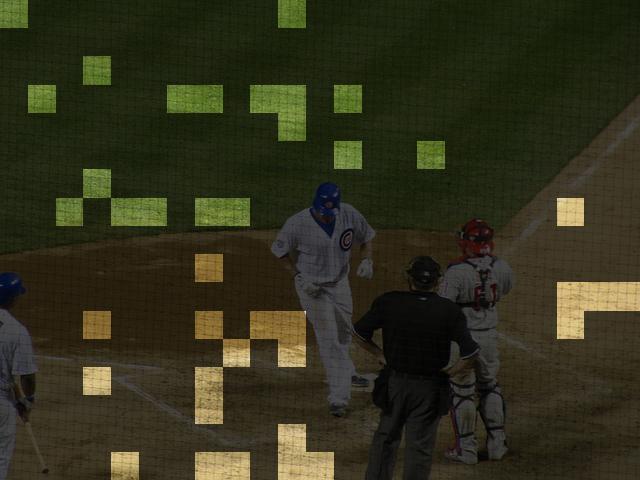} \end{subfigure}\hfill
    \begin{subfigure}[c]{\subfigwidth} \centering \small \textit{Retain 20\%}\\[1mm] \includegraphics[width=\linewidth]{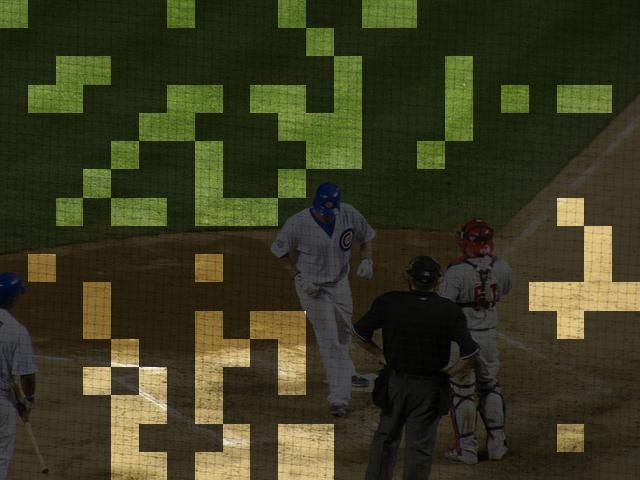} \end{subfigure}\hfill
    \begin{subfigure}[c]{\subfigwidth} \centering \small \textit{Retain 30\%}\\[1mm] \includegraphics[width=\linewidth]{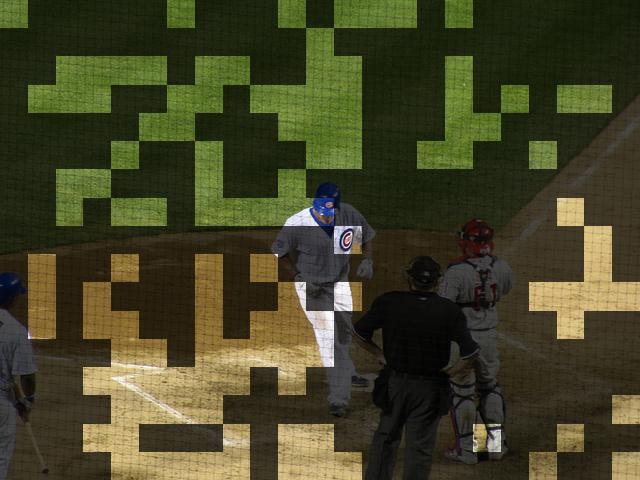} \end{subfigure}\hfill
    \begin{subfigure}[c]{\subfigwidth} \centering \small \textit{Retain 40\%}\\[1mm] \includegraphics[width=\linewidth]{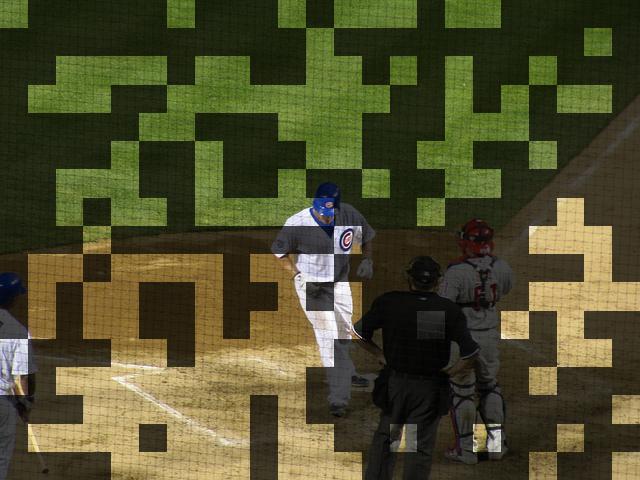} \end{subfigure}\hfill
    \begin{subfigure}[c]{\subfigwidth} \centering \small \textit{Retain 50\%}\\[1mm] \includegraphics[width=\linewidth]{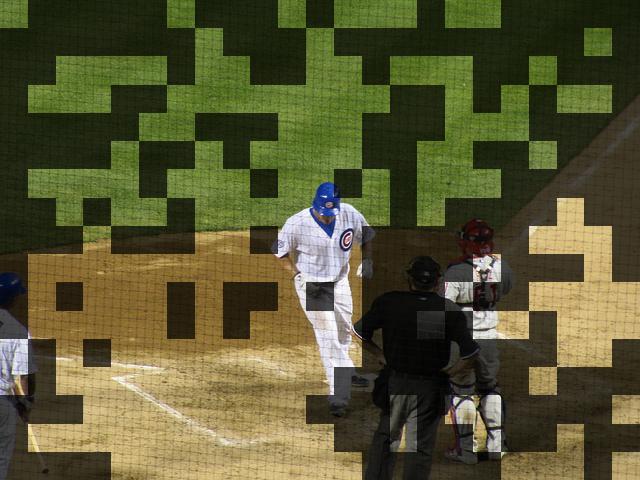} \end{subfigure}\\[1.5mm]

    \begin{subfigure}[c]{0.03\textwidth} \centering \rotatebox{90}{\small \textbf{\textit{100}}} \end{subfigure}\hfill
    \begin{subfigure}[c]{\subfigwidth} \centering \includegraphics[width=\linewidth]{figs/vis_hpe/input_images/COCO_val2014_000000000192.jpg} \end{subfigure}\hfill
    \begin{subfigure}[c]{\subfigwidth} \centering \includegraphics[width=\linewidth]{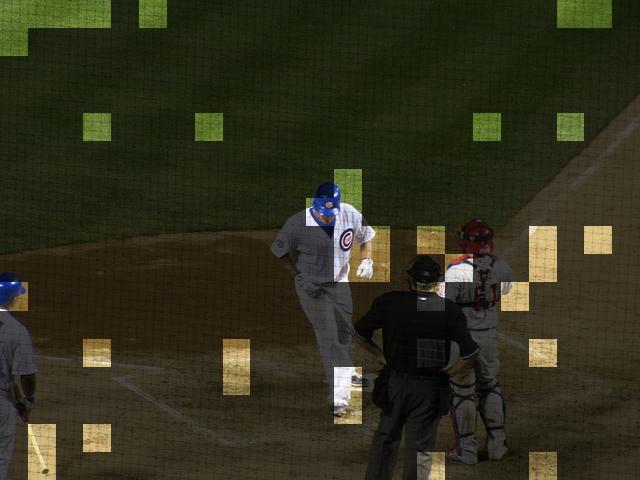} \end{subfigure}\hfill
    \begin{subfigure}[c]{\subfigwidth} \centering \includegraphics[width=\linewidth]{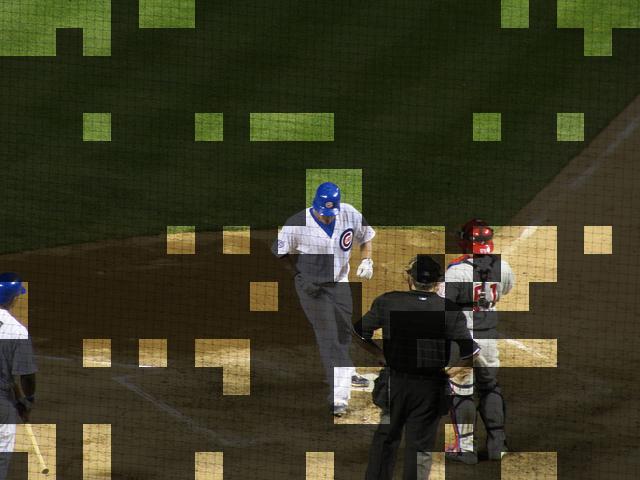} \end{subfigure}\hfill
    \begin{subfigure}[c]{\subfigwidth} \centering \includegraphics[width=\linewidth]{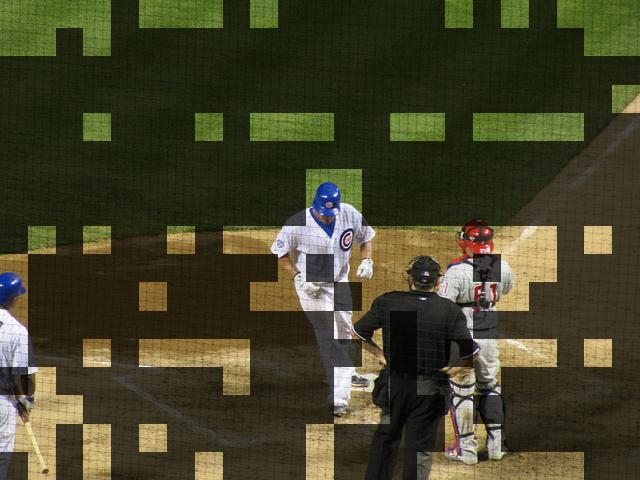} \end{subfigure}\hfill
    \begin{subfigure}[c]{\subfigwidth} \centering \includegraphics[width=\linewidth]{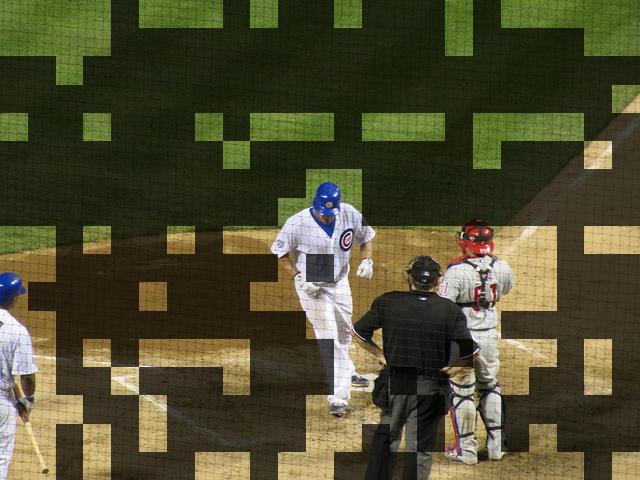} \end{subfigure}\hfill
    \begin{subfigure}[c]{\subfigwidth} \centering \includegraphics[width=\linewidth]{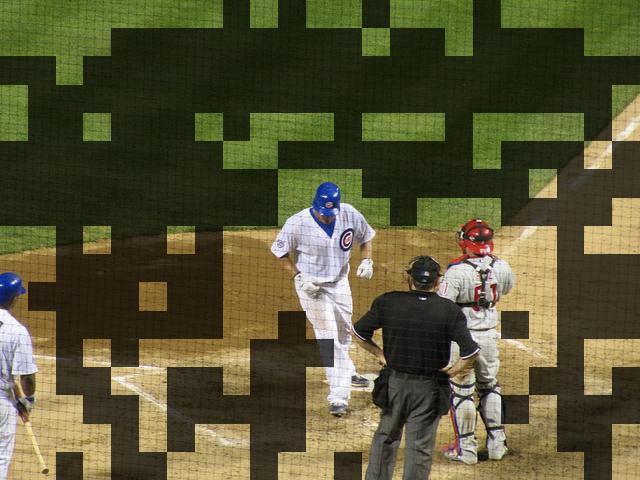} \end{subfigure}\\[1.5mm]

    \begin{subfigure}[c]{0.03\textwidth} \centering \rotatebox{90}{\small \textbf{\textit{101}}} \end{subfigure}\hfill
    \begin{subfigure}[c]{\subfigwidth} \centering \includegraphics[width=\linewidth]{figs/vis_hpe/input_images/COCO_val2014_000000000192.jpg} \end{subfigure}\hfill
    \begin{subfigure}[c]{\subfigwidth} \centering \includegraphics[width=\linewidth]{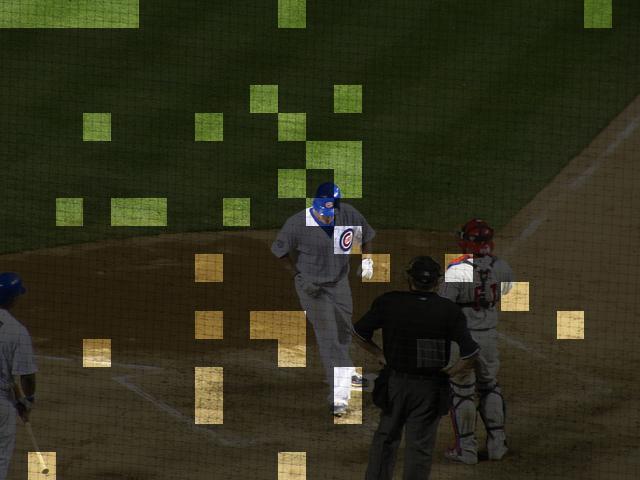} \end{subfigure}\hfill
    \begin{subfigure}[c]{\subfigwidth} \centering \includegraphics[width=\linewidth]{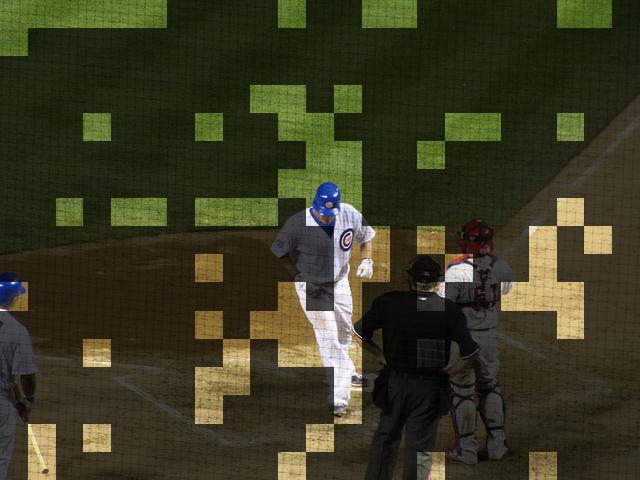} \end{subfigure}\hfill
    \begin{subfigure}[c]{\subfigwidth} \centering \includegraphics[width=\linewidth]{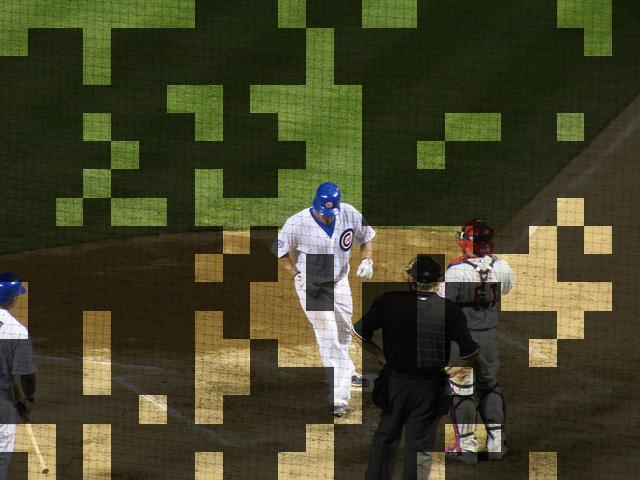} \end{subfigure}\hfill
    \begin{subfigure}[c]{\subfigwidth} \centering \includegraphics[width=\linewidth]{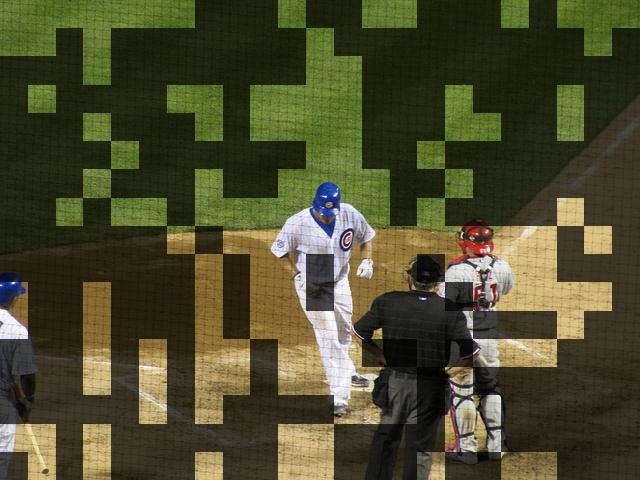} \end{subfigure}\hfill
    \begin{subfigure}[c]{\subfigwidth} \centering \includegraphics[width=\linewidth]{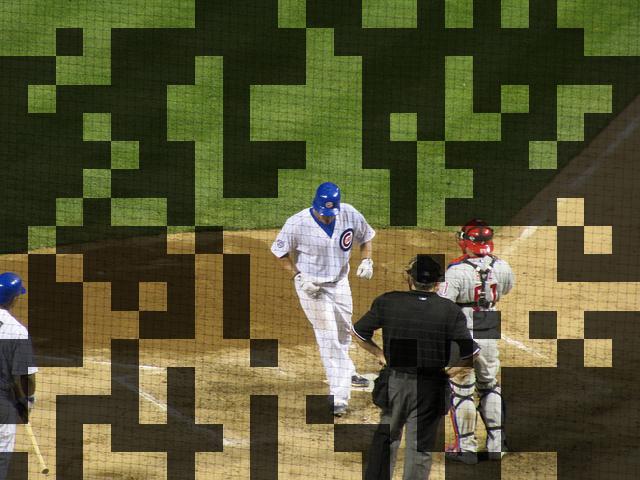} \end{subfigure}\\[1.5mm]

    \begin{subfigure}[c]{0.03\textwidth} \centering \rotatebox{90}{\small \textbf{\textit{110}}} \end{subfigure}\hfill
    \begin{subfigure}[c]{\subfigwidth} \centering \includegraphics[width=\linewidth]{figs/vis_hpe/input_images/COCO_val2014_000000000192.jpg} \end{subfigure}\hfill
    \begin{subfigure}[c]{\subfigwidth} \centering \includegraphics[width=\linewidth]{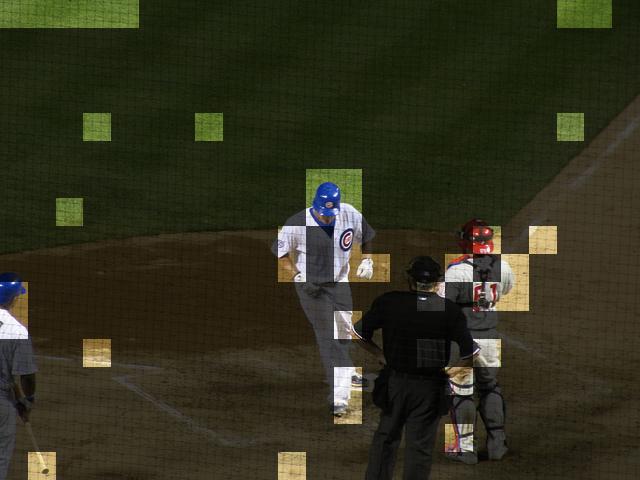} \end{subfigure}\hfill
    \begin{subfigure}[c]{\subfigwidth} \centering \includegraphics[width=\linewidth]{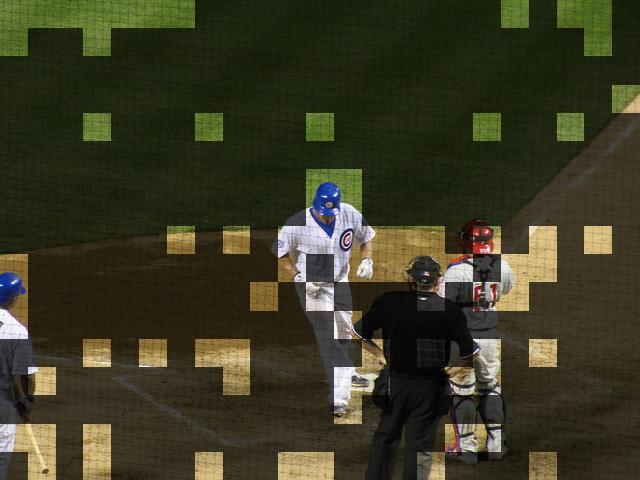} \end{subfigure}\hfill
    \begin{subfigure}[c]{\subfigwidth} \centering \includegraphics[width=\linewidth]{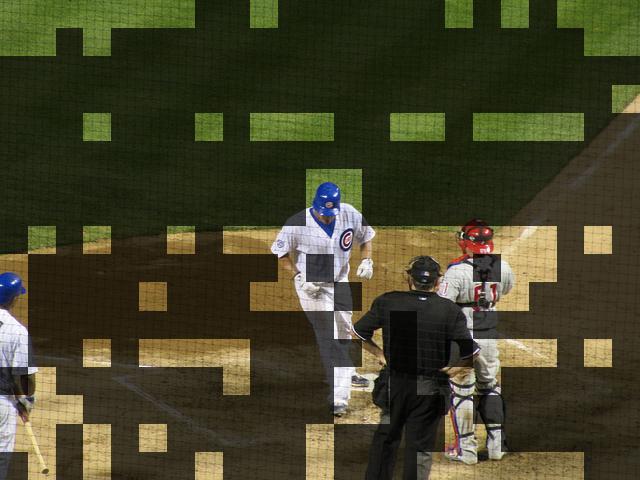} \end{subfigure}\hfill
    \begin{subfigure}[c]{\subfigwidth} \centering \includegraphics[width=\linewidth]{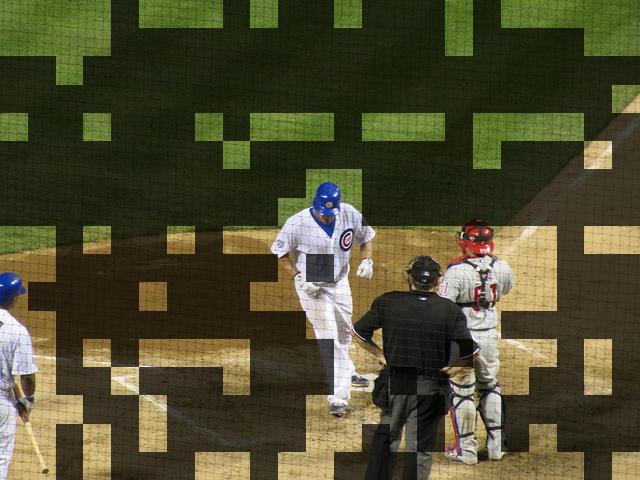} \end{subfigure}\hfill
    \begin{subfigure}[c]{\subfigwidth} \centering \includegraphics[width=\linewidth]{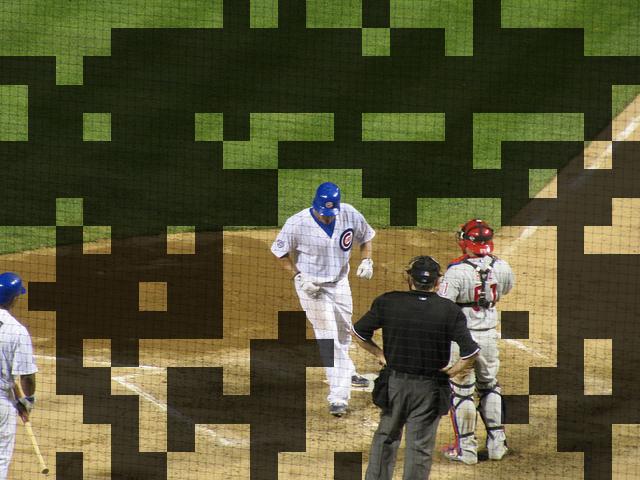} \end{subfigure}\\[1.5mm]

    \begin{subfigure}[c]{0.03\textwidth} \centering \rotatebox{90}{\small \textbf{\textit{ReTrace}}} \end{subfigure}\hfill
    \begin{subfigure}[c]{\subfigwidth} \centering \includegraphics[width=\linewidth]{figs/vis_hpe/input_images/COCO_val2014_000000000192.jpg} \end{subfigure}\hfill
    \begin{subfigure}[c]{\subfigwidth} \centering \includegraphics[width=\linewidth]{figs/vis_hpe/vis_keep_0.1/COCO_val2014_000000000192.jpg} \end{subfigure}\hfill
    \begin{subfigure}[c]{\subfigwidth} \centering \includegraphics[width=\linewidth]{figs/vis_hpe/vis_keep_0.2/COCO_val2014_000000000192.jpg} \end{subfigure}\hfill
    \begin{subfigure}[c]{\subfigwidth} \centering \includegraphics[width=\linewidth]{figs/vis_hpe/vis_keep_0.3/COCO_val2014_000000000192.jpg} \end{subfigure}\hfill
    \begin{subfigure}[c]{\subfigwidth} \centering \includegraphics[width=\linewidth]{figs/vis_hpe/vis_keep_0.4/COCO_val2014_000000000192.jpg} \end{subfigure}\hfill
    \begin{subfigure}[c]{\subfigwidth} \centering \includegraphics[width=\linewidth]{figs/vis_hpe/vis_keep_0.5/COCO_val2014_000000000192.jpg} \end{subfigure}
    \caption{\textbf{Qualitative comparison.}}
    \label{fig:ablation_vis_coco_192}
\end{figure*}

\begin{figure*}[p]
    \centering
    \newcommand{\subfigwidth}{0.155\textwidth}

    \begin{subfigure}[c]{0.03\textwidth} \centering \rotatebox{90}{\small \textbf{\textit{001}}} \end{subfigure}\hfill
    \begin{subfigure}[c]{\subfigwidth} \centering \small \textit{Original}\\[1mm] \includegraphics[width=\linewidth]{figs/vis_hpe/input_images/COCO_val2014_000000580975.jpg} \end{subfigure}\hfill
    \begin{subfigure}[c]{\subfigwidth} \centering \small \textit{Retain 10\%}\\[1mm] \includegraphics[width=\linewidth]{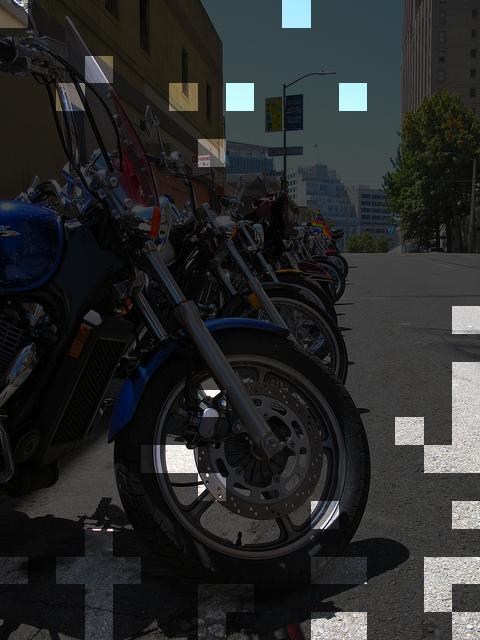} \end{subfigure}\hfill
    \begin{subfigure}[c]{\subfigwidth} \centering \small \textit{Retain 20\%}\\[1mm] \includegraphics[width=\linewidth]{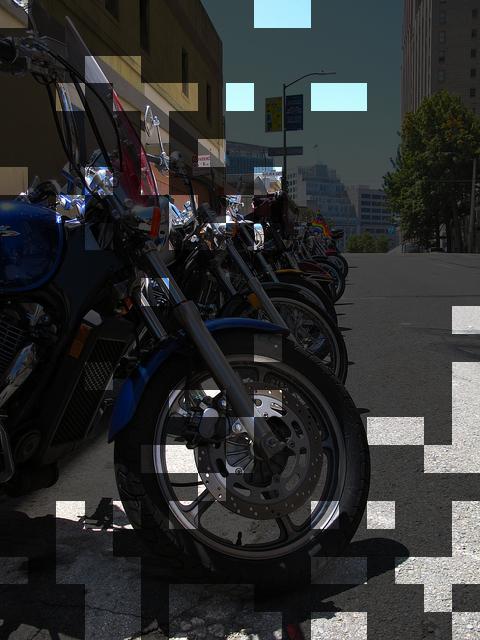} \end{subfigure}\hfill
    \begin{subfigure}[c]{\subfigwidth} \centering \small \textit{Retain 30\%}\\[1mm] \includegraphics[width=\linewidth]{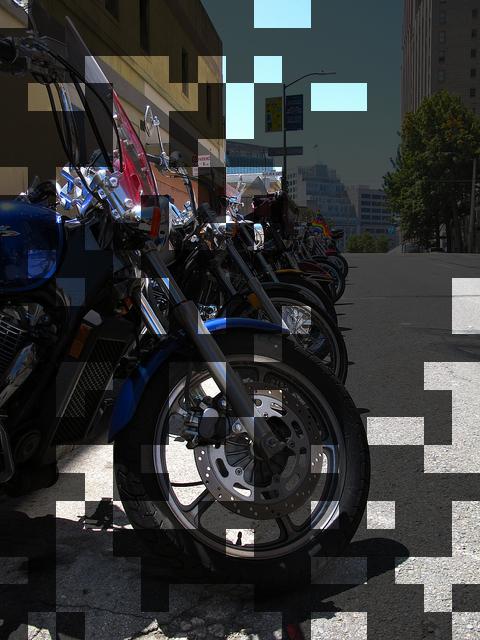} \end{subfigure}\hfill
    \begin{subfigure}[c]{\subfigwidth} \centering \small \textit{Retain 40\%}\\[1mm] \includegraphics[width=\linewidth]{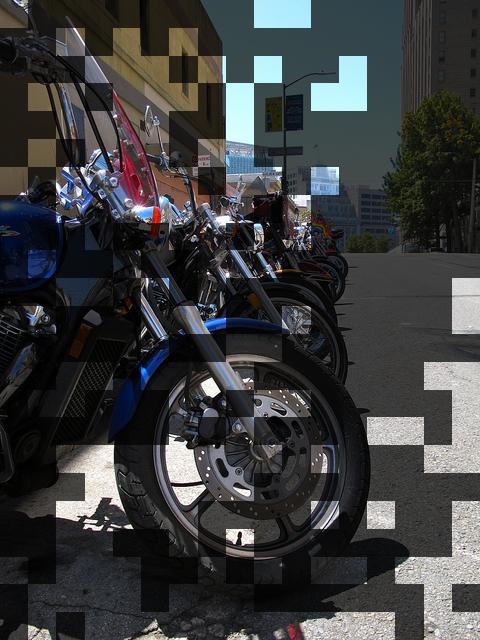} \end{subfigure}\hfill
    \begin{subfigure}[c]{\subfigwidth} \centering \small \textit{Retain 50\%}\\[1mm] \includegraphics[width=\linewidth]{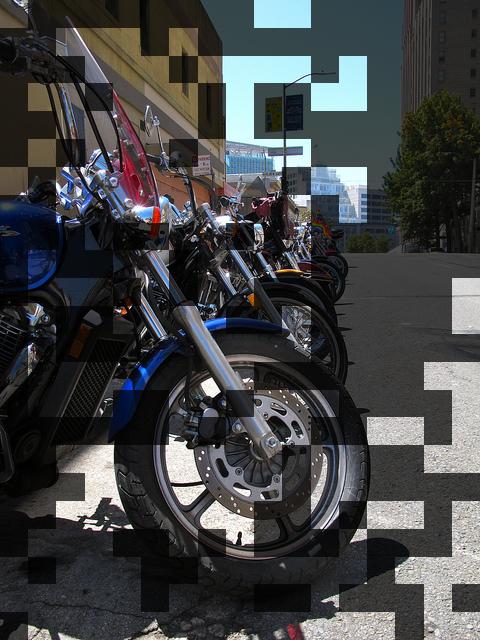} \end{subfigure}\\[1.5mm]

    \begin{subfigure}[c]{0.03\textwidth} \centering \rotatebox{90}{\small \textbf{\textit{100}}} \end{subfigure}\hfill
    \begin{subfigure}[c]{\subfigwidth} \centering \includegraphics[width=\linewidth]{figs/vis_hpe/input_images/COCO_val2014_000000580975.jpg} \end{subfigure}\hfill
    \begin{subfigure}[c]{\subfigwidth} \centering \includegraphics[width=\linewidth]{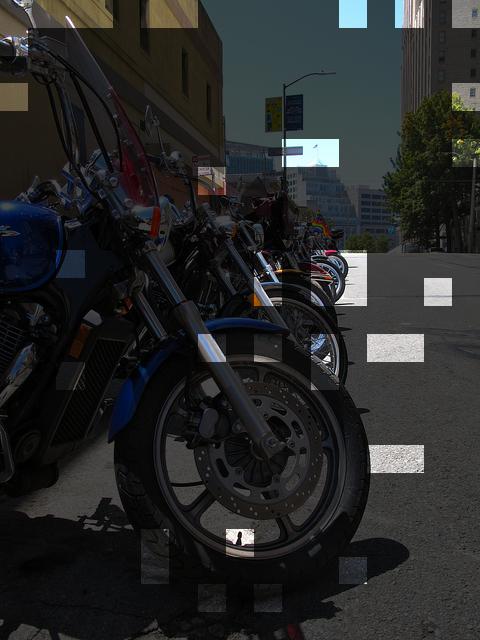} \end{subfigure}\hfill
    \begin{subfigure}[c]{\subfigwidth} \centering \includegraphics[width=\linewidth]{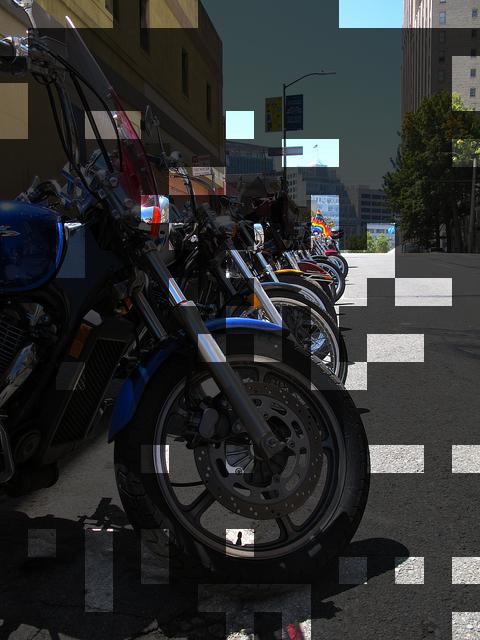} \end{subfigure}\hfill
    \begin{subfigure}[c]{\subfigwidth} \centering \includegraphics[width=\linewidth]{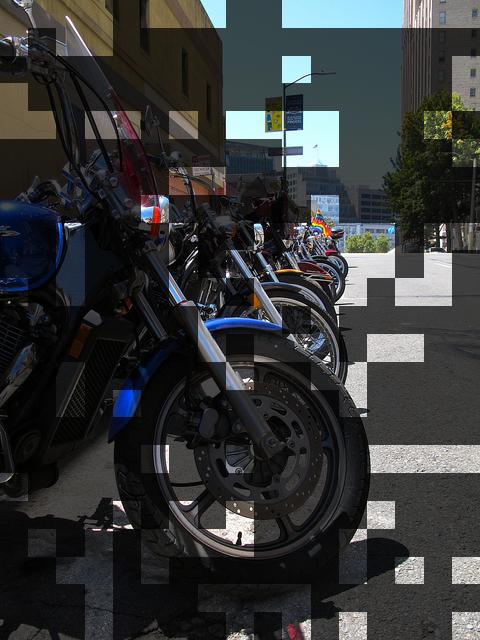} \end{subfigure}\hfill
    \begin{subfigure}[c]{\subfigwidth} \centering \includegraphics[width=\linewidth]{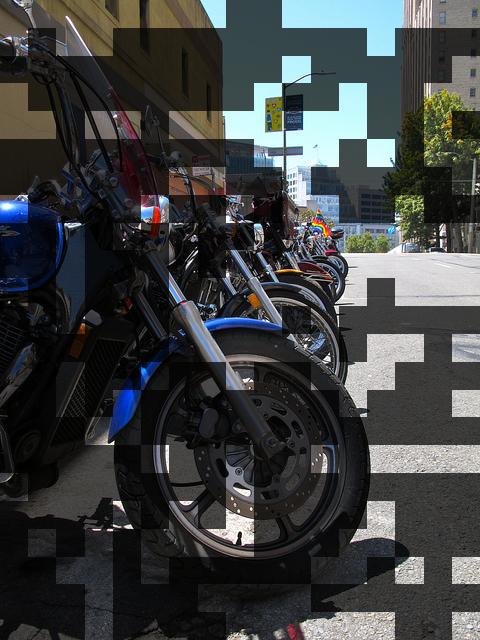} \end{subfigure}\hfill
    \begin{subfigure}[c]{\subfigwidth} \centering \includegraphics[width=\linewidth]{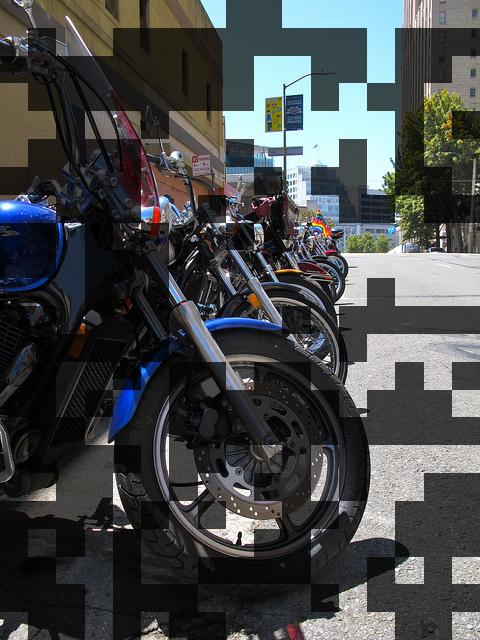} \end{subfigure}\\[1.5mm]

    \begin{subfigure}[c]{0.03\textwidth} \centering \rotatebox{90}{\small \textbf{\textit{101}}} \end{subfigure}\hfill
    \begin{subfigure}[c]{\subfigwidth} \centering \includegraphics[width=\linewidth]{figs/vis_hpe/input_images/COCO_val2014_000000580975.jpg} \end{subfigure}\hfill
    \begin{subfigure}[c]{\subfigwidth} \centering \includegraphics[width=\linewidth]{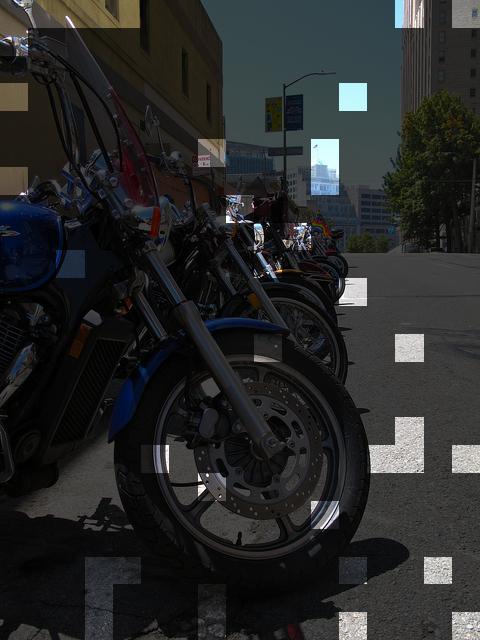} \end{subfigure}\hfill
    \begin{subfigure}[c]{\subfigwidth} \centering \includegraphics[width=\linewidth]{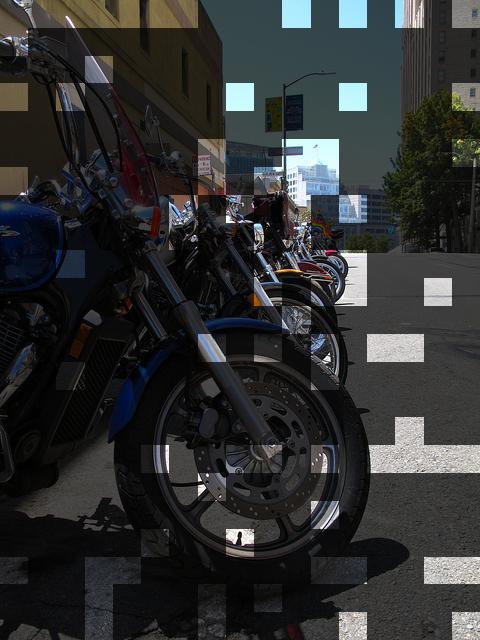} \end{subfigure}\hfill
    \begin{subfigure}[c]{\subfigwidth} \centering \includegraphics[width=\linewidth]{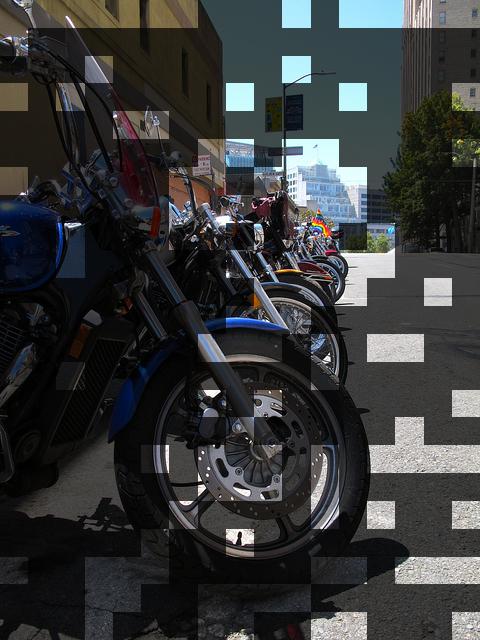} \end{subfigure}\hfill
    \begin{subfigure}[c]{\subfigwidth} \centering \includegraphics[width=\linewidth]{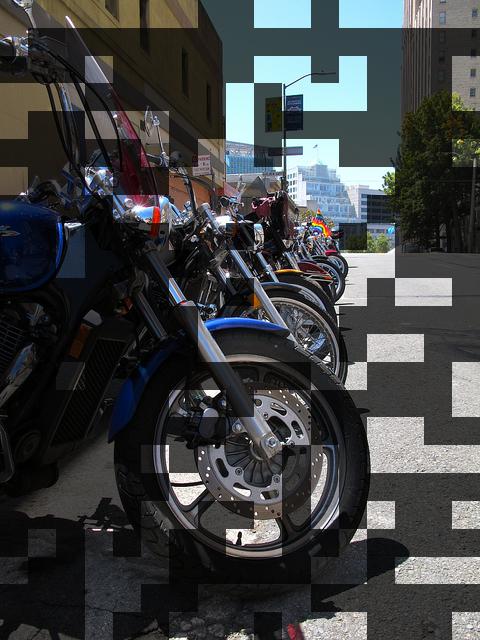} \end{subfigure}\hfill
    \begin{subfigure}[c]{\subfigwidth} \centering \includegraphics[width=\linewidth]{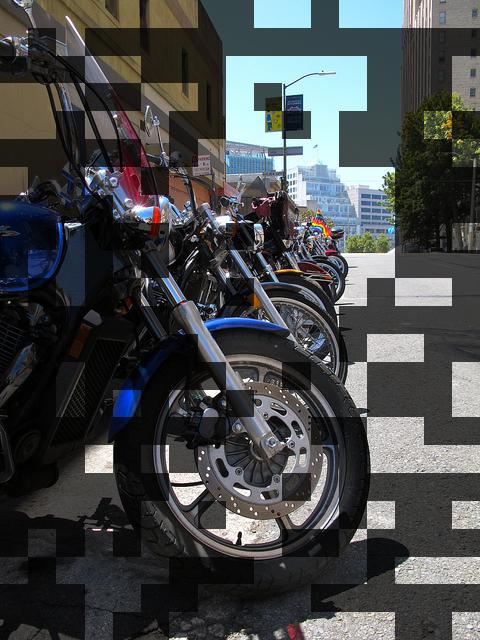} \end{subfigure}\\[1.5mm]

    \begin{subfigure}[c]{0.03\textwidth} \centering \rotatebox{90}{\small \textbf{\textit{110}}} \end{subfigure}\hfill
    \begin{subfigure}[c]{\subfigwidth} \centering \includegraphics[width=\linewidth]{figs/vis_hpe/input_images/COCO_val2014_000000580975.jpg} \end{subfigure}\hfill
    \begin{subfigure}[c]{\subfigwidth} \centering \includegraphics[width=\linewidth]{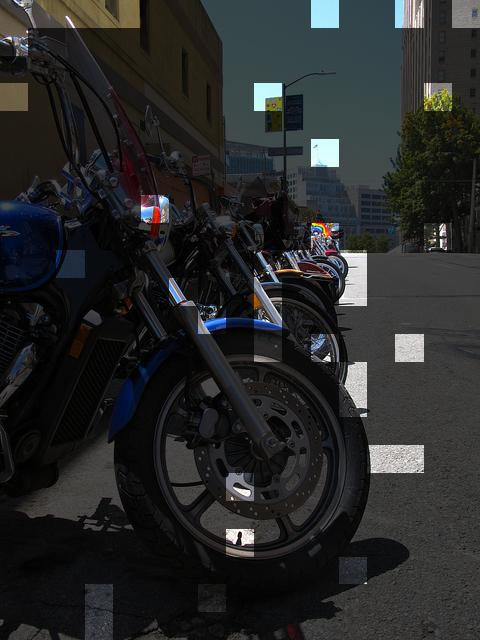} \end{subfigure}\hfill
    \begin{subfigure}[c]{\subfigwidth} \centering \includegraphics[width=\linewidth]{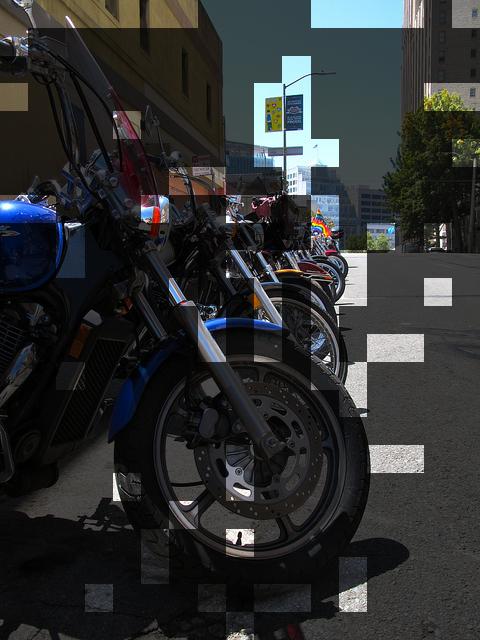} \end{subfigure}\hfill
    \begin{subfigure}[c]{\subfigwidth} \centering \includegraphics[width=\linewidth]{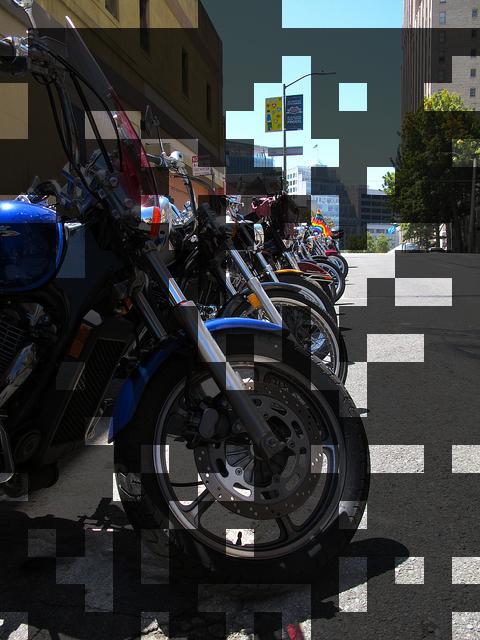} \end{subfigure}\hfill
    \begin{subfigure}[c]{\subfigwidth} \centering \includegraphics[width=\linewidth]{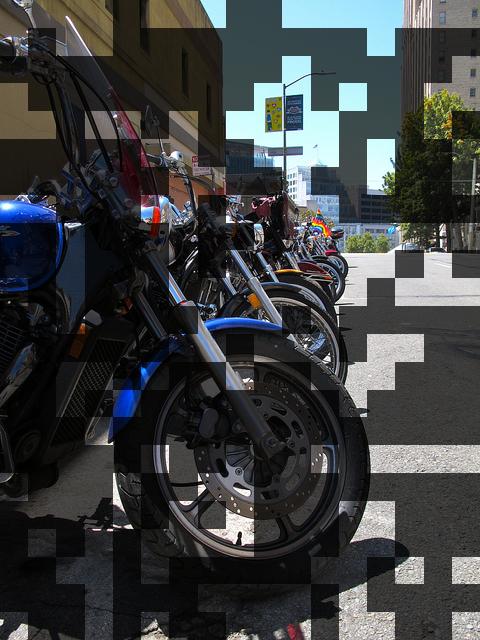} \end{subfigure}\hfill
    \begin{subfigure}[c]{\subfigwidth} \centering \includegraphics[width=\linewidth]{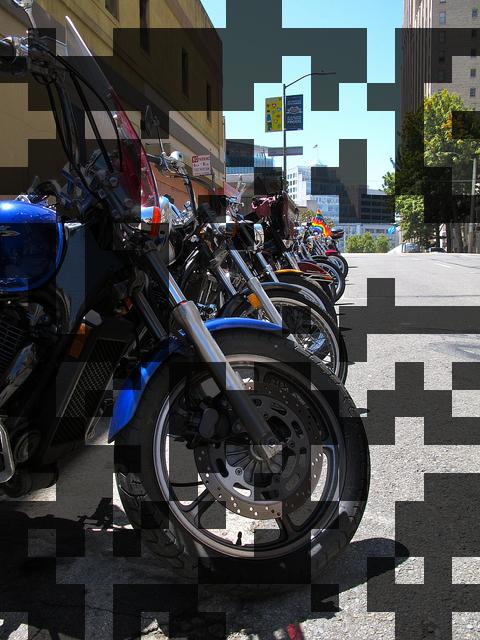} \end{subfigure}\\[1.5mm]

    \begin{subfigure}[c]{0.03\textwidth} \centering \rotatebox{90}{\small \textbf{\textit{ReTrace}}} \end{subfigure}\hfill
    \begin{subfigure}[c]{\subfigwidth} \centering \includegraphics[width=\linewidth]{figs/vis_hpe/input_images/COCO_val2014_000000580975.jpg} \end{subfigure}\hfill
    \begin{subfigure}[c]{\subfigwidth} \centering \includegraphics[width=\linewidth]{figs/vis_hpe/vis_keep_0.1/COCO_val2014_000000580975.jpg} \end{subfigure}\hfill
    \begin{subfigure}[c]{\subfigwidth} \centering \includegraphics[width=\linewidth]{figs/vis_hpe/vis_keep_0.2/COCO_val2014_000000580975.jpg} \end{subfigure}\hfill
    \begin{subfigure}[c]{\subfigwidth} \centering \includegraphics[width=\linewidth]{figs/vis_hpe/vis_keep_0.3/COCO_val2014_000000580975.jpg} \end{subfigure}\hfill
    \begin{subfigure}[c]{\subfigwidth} \centering \includegraphics[width=\linewidth]{figs/vis_hpe/vis_keep_0.4/COCO_val2014_000000580975.jpg} \end{subfigure}\hfill
    \begin{subfigure}[c]{\subfigwidth} \centering \includegraphics[width=\linewidth]{figs/vis_hpe/vis_keep_0.5/COCO_val2014_000000580975.jpg} \end{subfigure}
    \caption{\textbf{Qualitative comparison.}}
    \label{fig:ablation_vis_coco_580975}
\end{figure*}

\begin{figure*}[p]
    \centering
    \newcommand{\subfigwidth}{0.155\textwidth}

    \begin{subfigure}[c]{0.03\textwidth} \centering \rotatebox{90}{\small \textbf{\textit{001}}} \end{subfigure}\hfill
    \begin{subfigure}[c]{\subfigwidth} \centering \small \textit{Original}\\[1mm] \includegraphics[width=\linewidth]{figs/vis_hpe/input_images/COCO_val2014_000000581829.jpg} \end{subfigure}\hfill
    \begin{subfigure}[c]{\subfigwidth} \centering \small \textit{Retain 10\%}\\[1mm] \includegraphics[width=\linewidth]{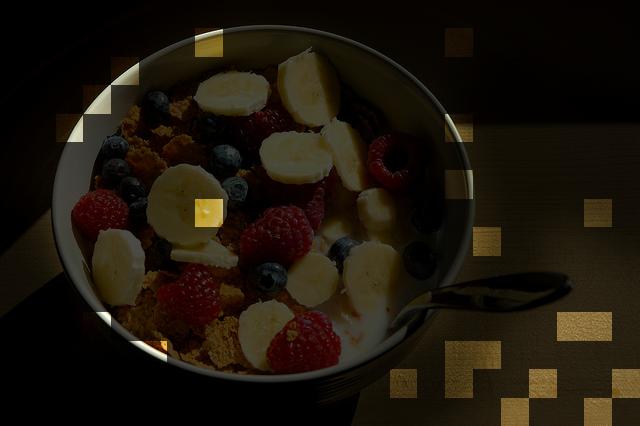} \end{subfigure}\hfill
    \begin{subfigure}[c]{\subfigwidth} \centering \small \textit{Retain 20\%}\\[1mm] \includegraphics[width=\linewidth]{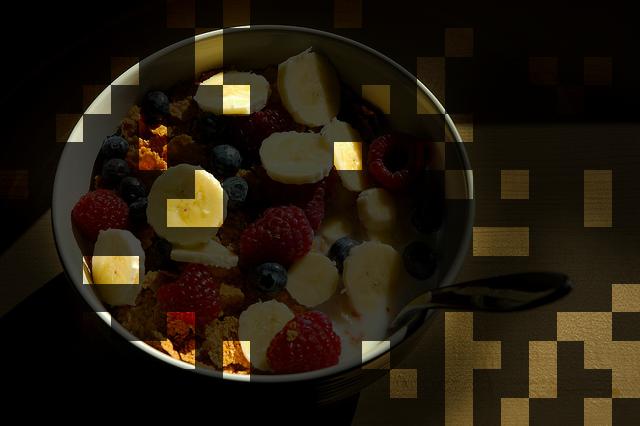} \end{subfigure}\hfill
    \begin{subfigure}[c]{\subfigwidth} \centering \small \textit{Retain 30\%}\\[1mm] \includegraphics[width=\linewidth]{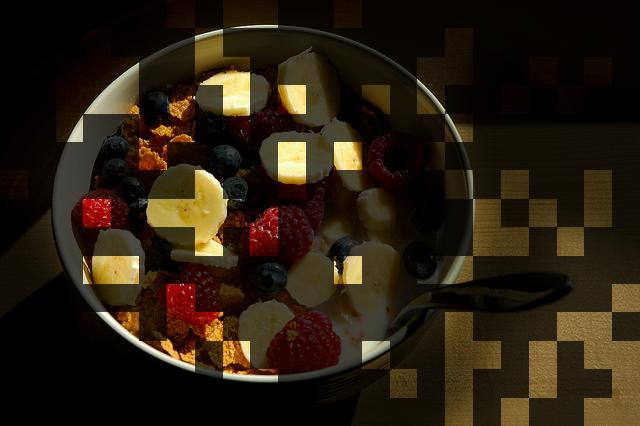} \end{subfigure}\hfill
    \begin{subfigure}[c]{\subfigwidth} \centering \small \textit{Retain 40\%}\\[1mm] \includegraphics[width=\linewidth]{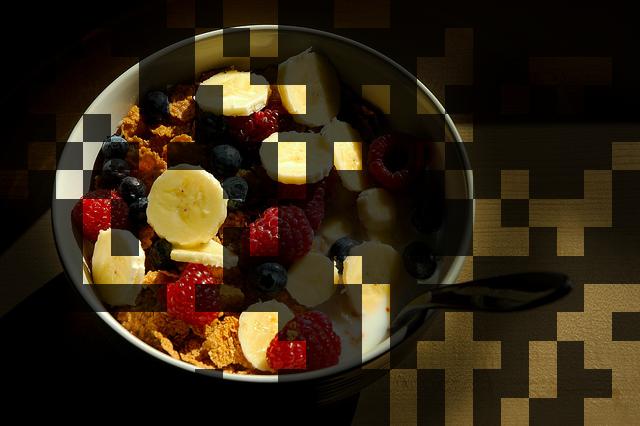} \end{subfigure}\hfill
    \begin{subfigure}[c]{\subfigwidth} \centering \small \textit{Retain 50\%}\\[1mm] \includegraphics[width=\linewidth]{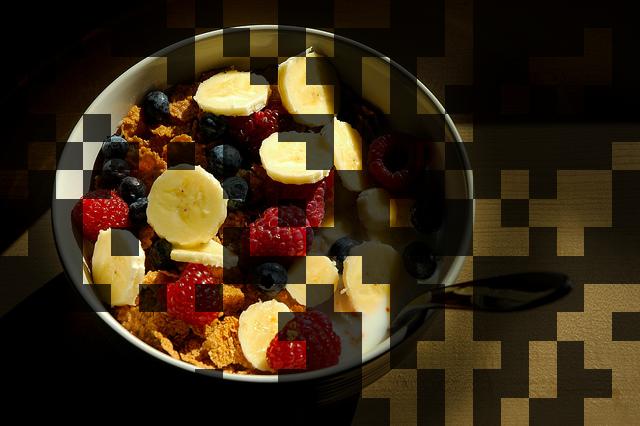} \end{subfigure}\\[1.5mm]

    \begin{subfigure}[c]{0.03\textwidth} \centering \rotatebox{90}{\small \textbf{\textit{100}}} \end{subfigure}\hfill
    \begin{subfigure}[c]{\subfigwidth} \centering \includegraphics[width=\linewidth]{figs/vis_hpe/input_images/COCO_val2014_000000581829.jpg} \end{subfigure}\hfill
    \begin{subfigure}[c]{\subfigwidth} \centering \includegraphics[width=\linewidth]{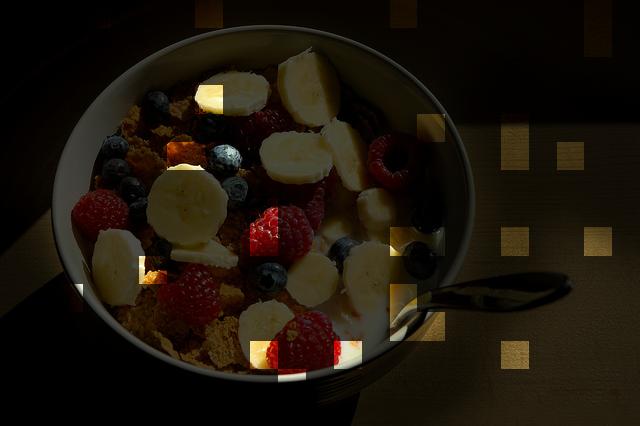} \end{subfigure}\hfill
    \begin{subfigure}[c]{\subfigwidth} \centering \includegraphics[width=\linewidth]{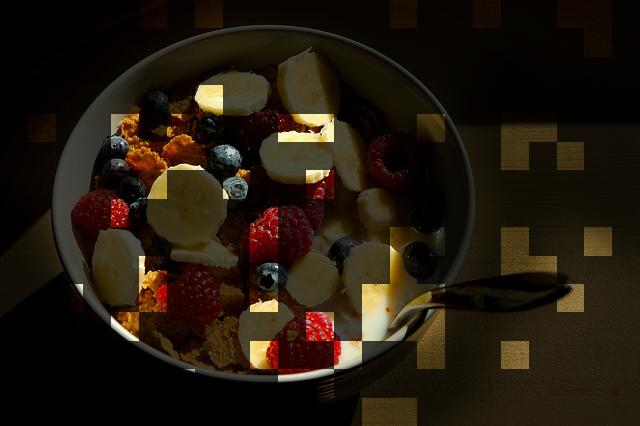} \end{subfigure}\hfill
    \begin{subfigure}[c]{\subfigwidth} \centering \includegraphics[width=\linewidth]{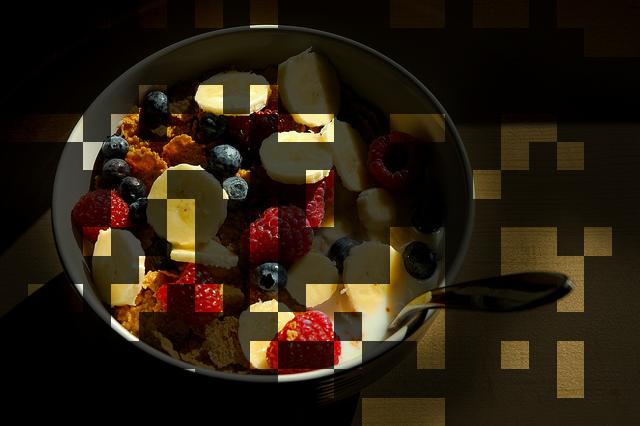} \end{subfigure}\hfill
    \begin{subfigure}[c]{\subfigwidth} \centering \includegraphics[width=\linewidth]{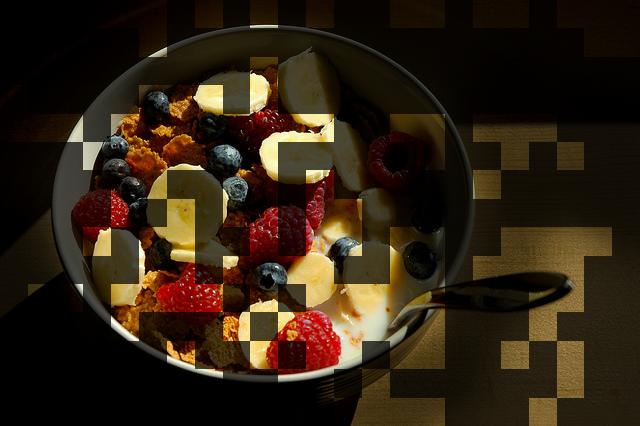} \end{subfigure}\hfill
    \begin{subfigure}[c]{\subfigwidth} \centering \includegraphics[width=\linewidth]{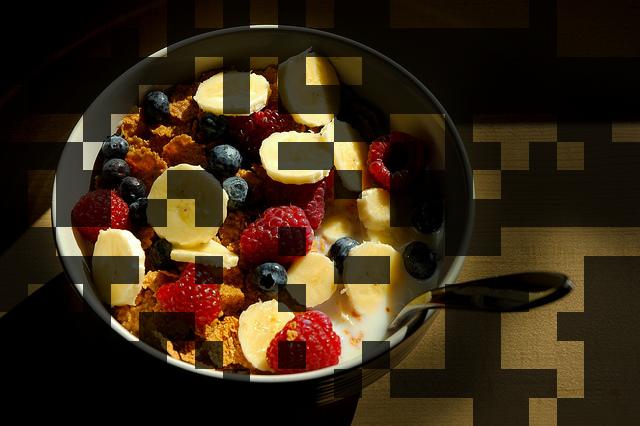} \end{subfigure}\\[1.5mm]

    \begin{subfigure}[c]{0.03\textwidth} \centering \rotatebox{90}{\small \textbf{\textit{101}}} \end{subfigure}\hfill
    \begin{subfigure}[c]{\subfigwidth} \centering \includegraphics[width=\linewidth]{figs/vis_hpe/input_images/COCO_val2014_000000581829.jpg} \end{subfigure}\hfill
    \begin{subfigure}[c]{\subfigwidth} \centering \includegraphics[width=\linewidth]{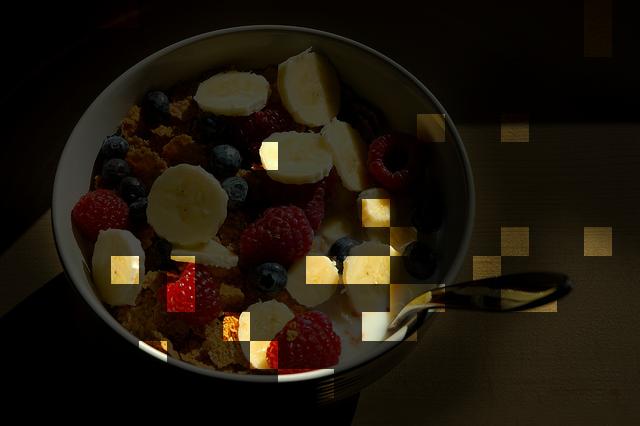} \end{subfigure}\hfill
    \begin{subfigure}[c]{\subfigwidth} \centering \includegraphics[width=\linewidth]{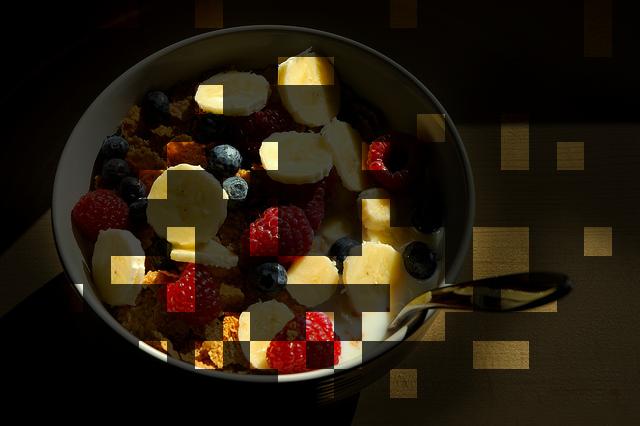} \end{subfigure}\hfill
    \begin{subfigure}[c]{\subfigwidth} \centering \includegraphics[width=\linewidth]{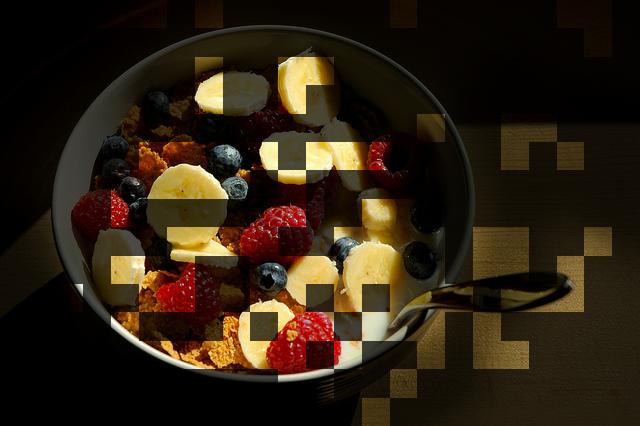} \end{subfigure}\hfill
    \begin{subfigure}[c]{\subfigwidth} \centering \includegraphics[width=\linewidth]{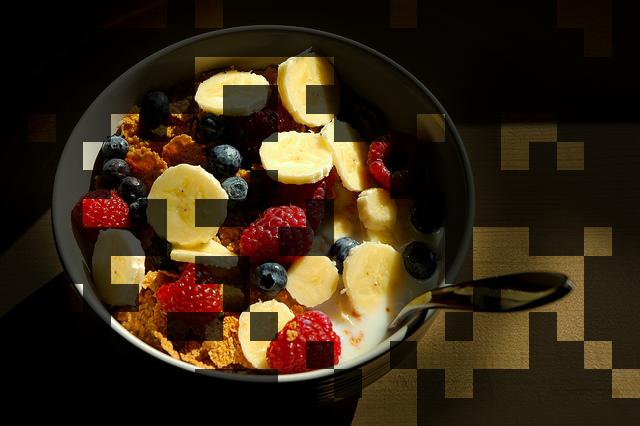} \end{subfigure}\hfill
    \begin{subfigure}[c]{\subfigwidth} \centering \includegraphics[width=\linewidth]{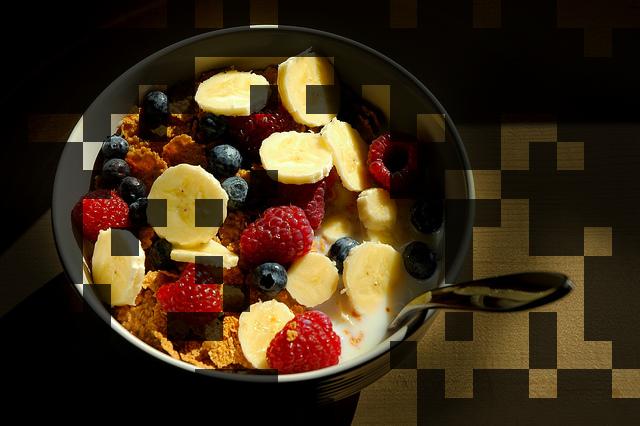} \end{subfigure}\\[1.5mm]

    \begin{subfigure}[c]{0.03\textwidth} \centering \rotatebox{90}{\small \textbf{\textit{110}}} \end{subfigure}\hfill
    \begin{subfigure}[c]{\subfigwidth} \centering \includegraphics[width=\linewidth]{figs/vis_hpe/input_images/COCO_val2014_000000581829.jpg} \end{subfigure}\hfill
    \begin{subfigure}[c]{\subfigwidth} \centering \includegraphics[width=\linewidth]{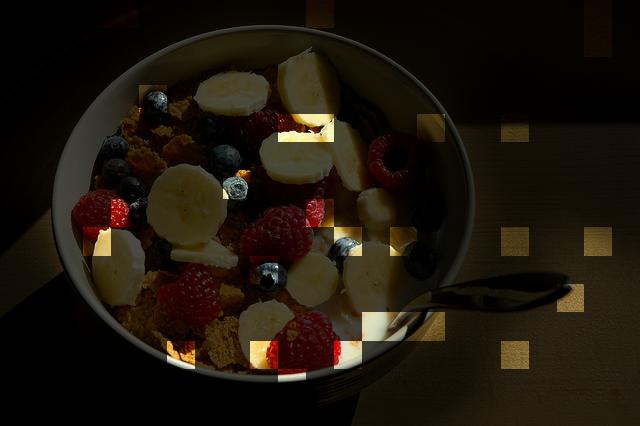} \end{subfigure}\hfill
    \begin{subfigure}[c]{\subfigwidth} \centering \includegraphics[width=\linewidth]{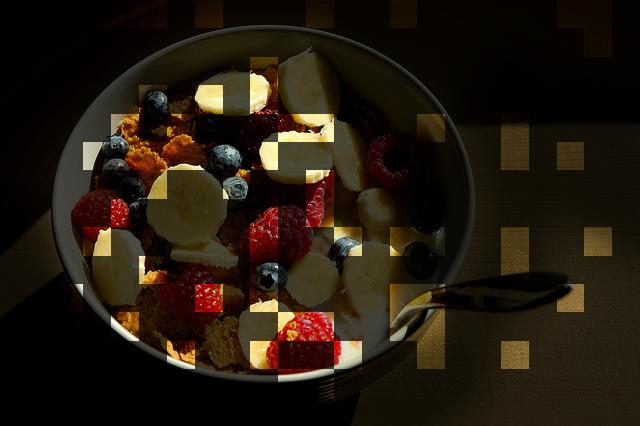} \end{subfigure}\hfill
    \begin{subfigure}[c]{\subfigwidth} \centering \includegraphics[width=\linewidth]{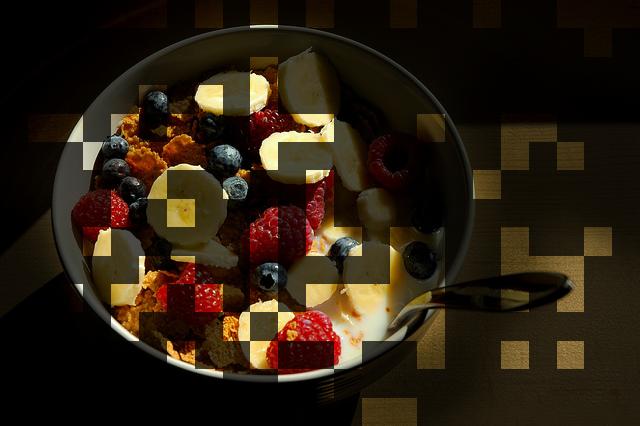} \end{subfigure}\hfill
    \begin{subfigure}[c]{\subfigwidth} \centering \includegraphics[width=\linewidth]{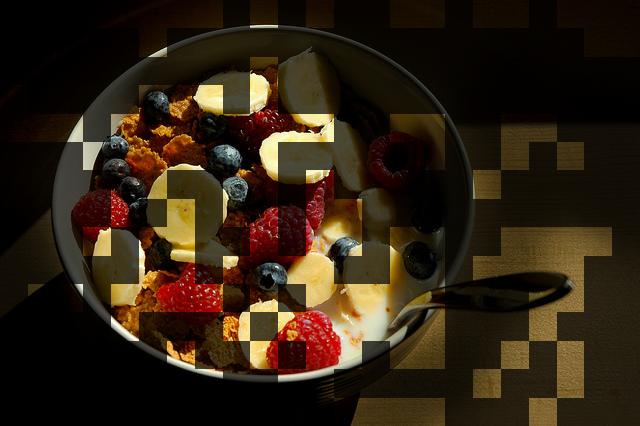} \end{subfigure}\hfill
    \begin{subfigure}[c]{\subfigwidth} \centering \includegraphics[width=\linewidth]{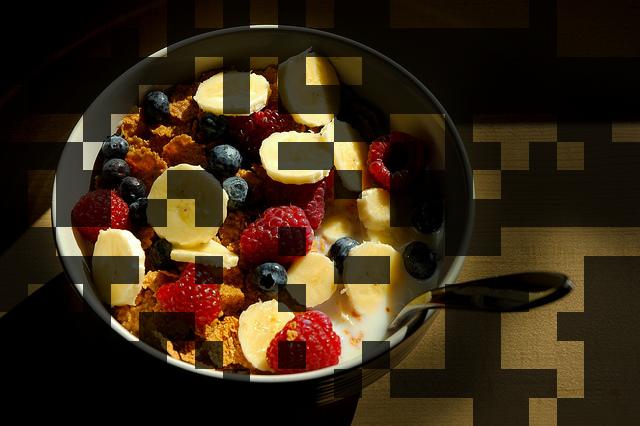} \end{subfigure}\\[1.5mm]

    \begin{subfigure}[c]{0.03\textwidth} \centering \rotatebox{90}{\small \textbf{\textit{ReTrace}}} \end{subfigure}\hfill
    \begin{subfigure}[c]{\subfigwidth} \centering \includegraphics[width=\linewidth]{figs/vis_hpe/input_images/COCO_val2014_000000581829.jpg} \end{subfigure}\hfill
    \begin{subfigure}[c]{\subfigwidth} \centering \includegraphics[width=\linewidth]{figs/vis_hpe/vis_keep_0.1/COCO_val2014_000000581829.jpg} \end{subfigure}\hfill
    \begin{subfigure}[c]{\subfigwidth} \centering \includegraphics[width=\linewidth]{figs/vis_hpe/vis_keep_0.2/COCO_val2014_000000581829.jpg} \end{subfigure}\hfill
    \begin{subfigure}[c]{\subfigwidth} \centering \includegraphics[width=\linewidth]{figs/vis_hpe/vis_keep_0.3/COCO_val2014_000000581829.jpg} \end{subfigure}\hfill
    \begin{subfigure}[c]{\subfigwidth} \centering \includegraphics[width=\linewidth]{figs/vis_hpe/vis_keep_0.4/COCO_val2014_000000581829.jpg} \end{subfigure}\hfill
    \begin{subfigure}[c]{\subfigwidth} \centering \includegraphics[width=\linewidth]{figs/vis_hpe/vis_keep_0.5/COCO_val2014_000000581829.jpg} \end{subfigure}
    \caption{\textbf{Qualitative comparison.}}
    \label{fig:ablation_vis_coco_581829}
\end{figure*}

\begin{figure*}[p]
    \centering
    \newcommand{\subfigwidth}{0.155\textwidth}

    \begin{subfigure}[c]{0.03\textwidth} \centering \rotatebox{90}{\small \textbf{\textit{001}}} \end{subfigure}\hfill
    \begin{subfigure}[c]{\subfigwidth} \centering \small \textit{Original}\\[1mm] \includegraphics[width=\linewidth]{figs/vis_hpe/input_images/image_1.jpg} \end{subfigure}\hfill
    \begin{subfigure}[c]{\subfigwidth} \centering \small \textit{Retain 10\%}\\[1mm] \includegraphics[width=\linewidth]{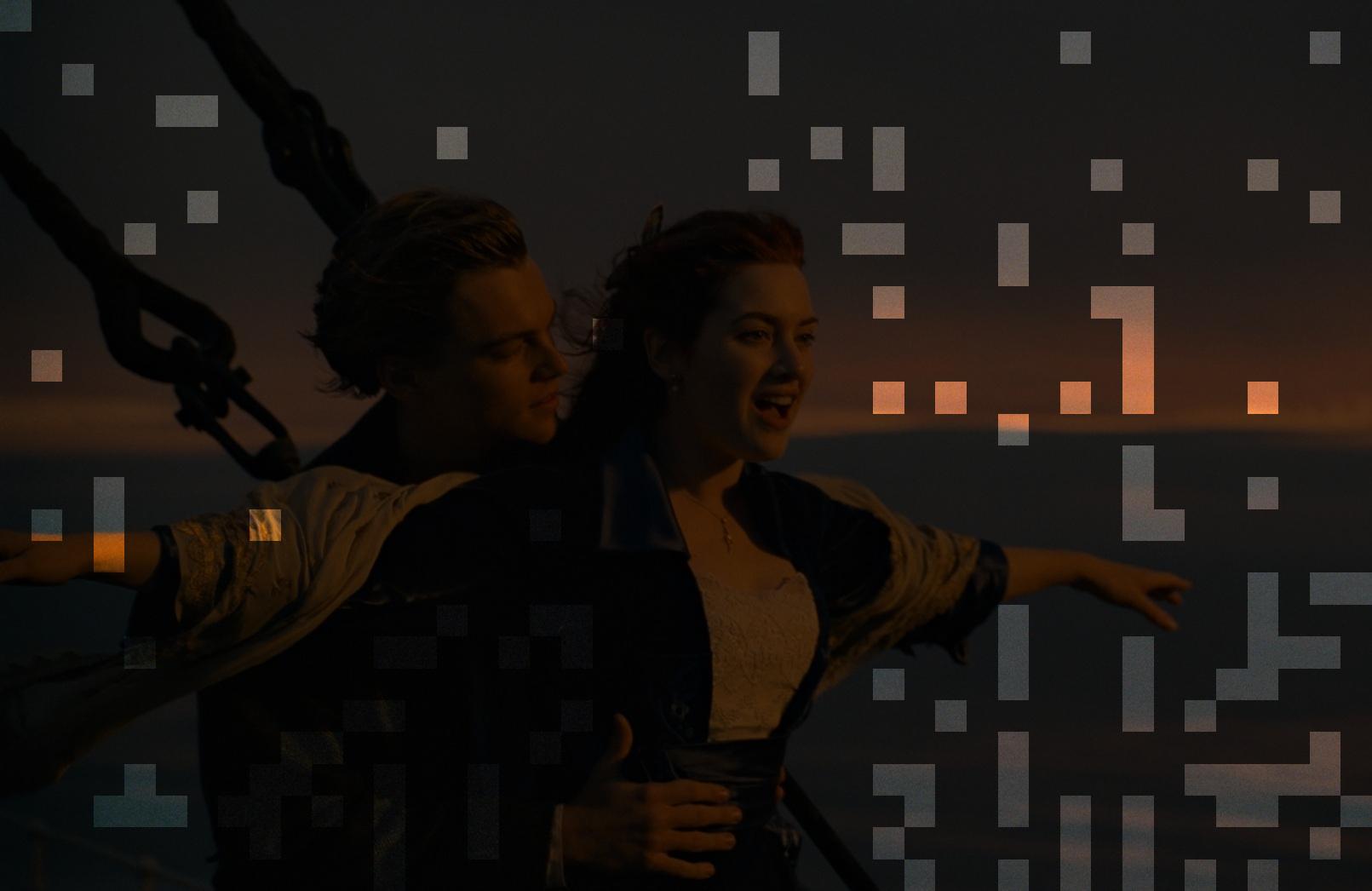} \end{subfigure}\hfill
    \begin{subfigure}[c]{\subfigwidth} \centering \small \textit{Retain 20\%}\\[1mm] \includegraphics[width=\linewidth]{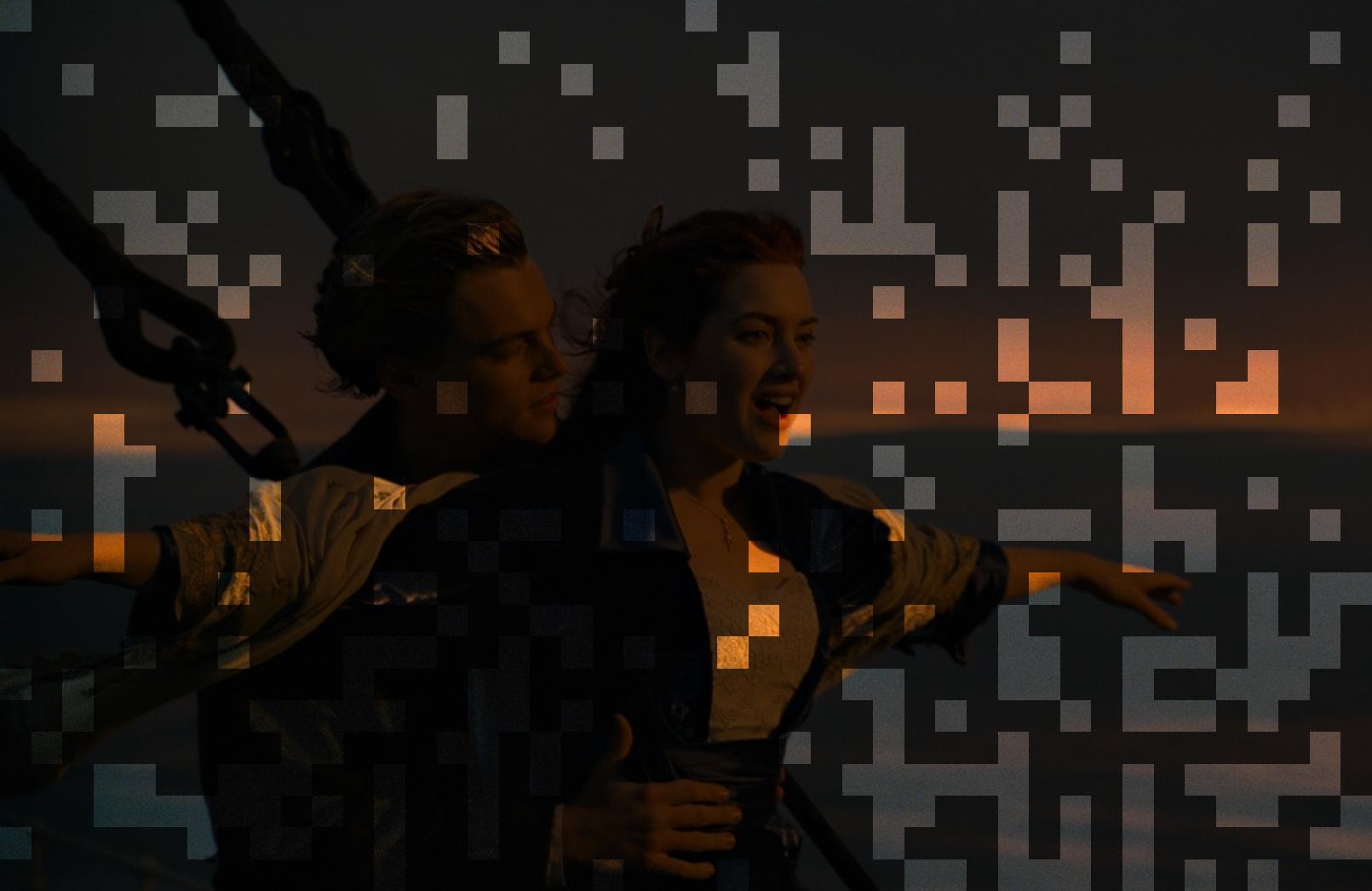} \end{subfigure}\hfill
    \begin{subfigure}[c]{\subfigwidth} \centering \small \textit{Retain 30\%}\\[1mm] \includegraphics[width=\linewidth]{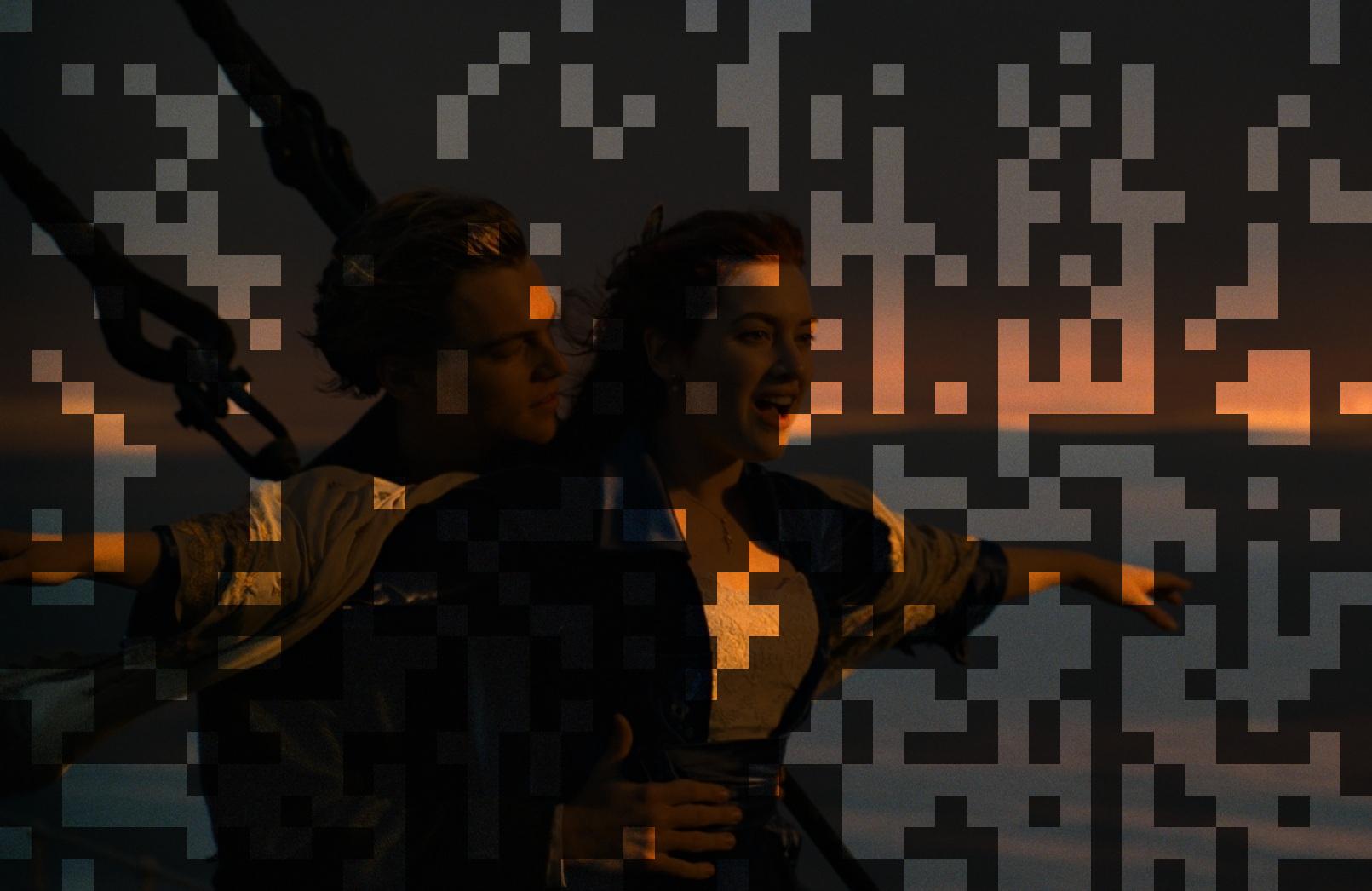} \end{subfigure}\hfill
    \begin{subfigure}[c]{\subfigwidth} \centering \small \textit{Retain 40\%}\\[1mm] \includegraphics[width=\linewidth]{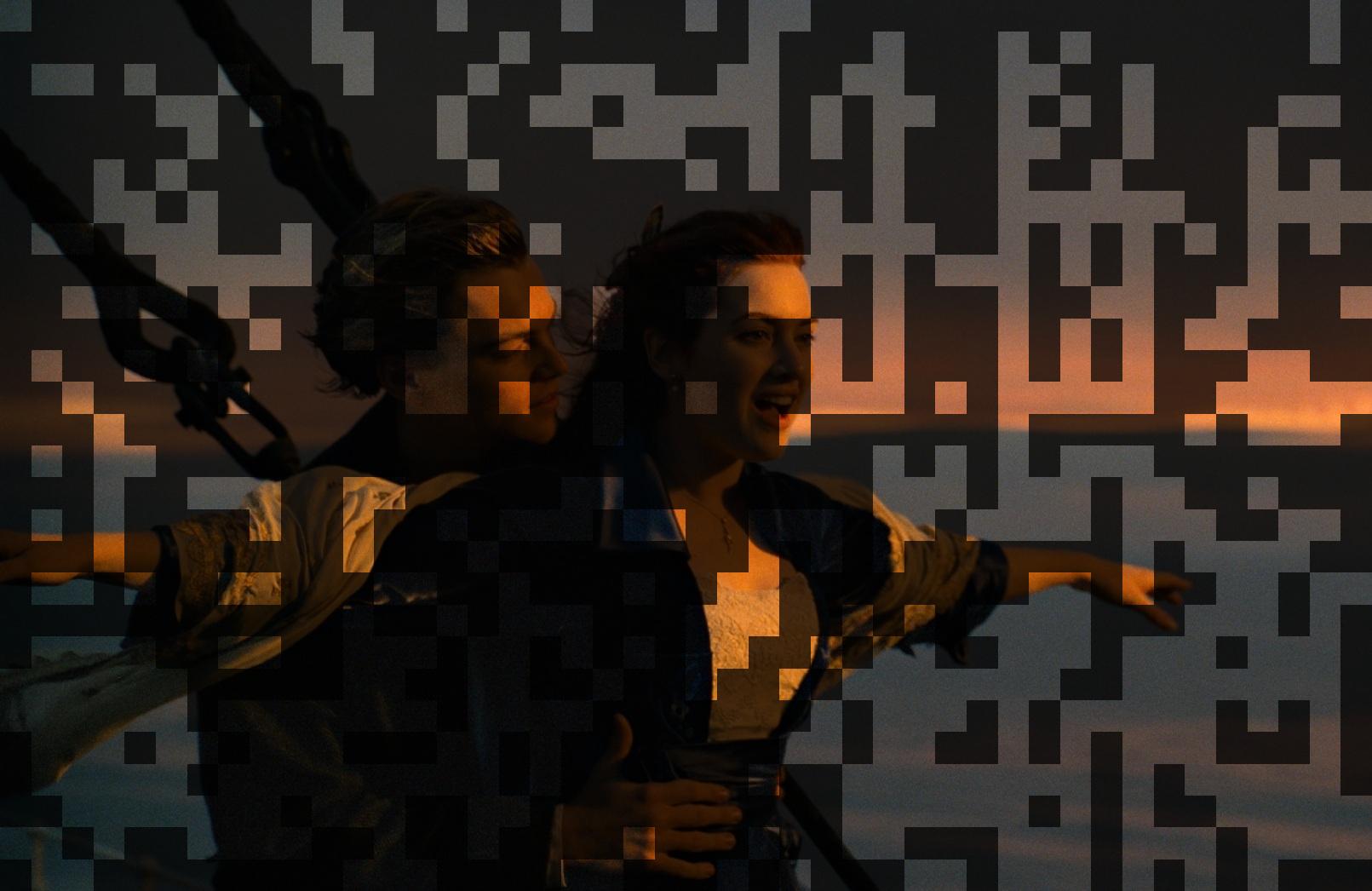} \end{subfigure}\hfill
    \begin{subfigure}[c]{\subfigwidth} \centering \small \textit{Retain 50\%}\\[1mm] \includegraphics[width=\linewidth]{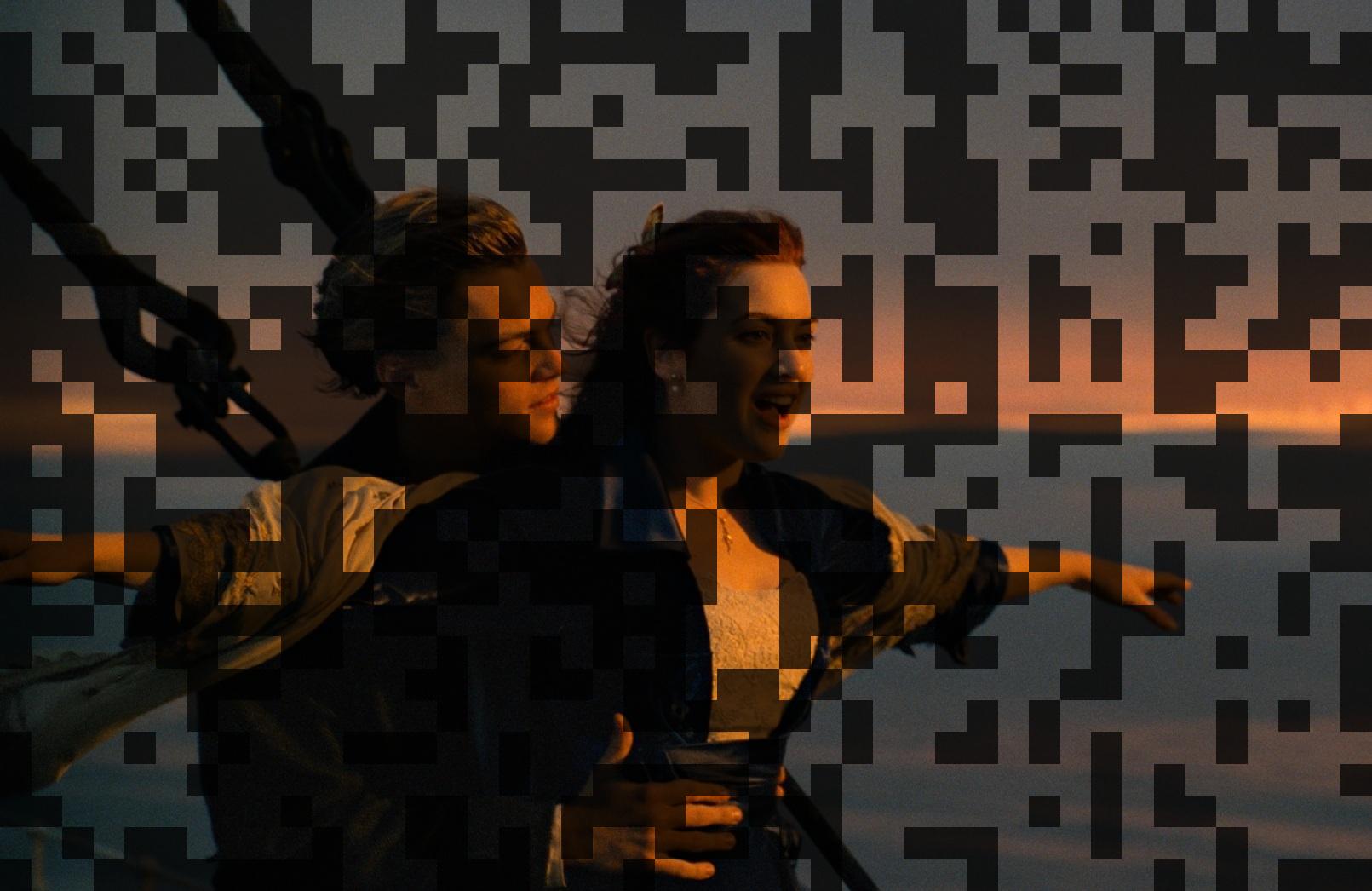} \end{subfigure}\\[1.5mm]

    \begin{subfigure}[c]{0.03\textwidth} \centering \rotatebox{90}{\small \textbf{\textit{100}}} \end{subfigure}\hfill
    \begin{subfigure}[c]{\subfigwidth} \centering \includegraphics[width=\linewidth]{figs/vis_hpe/input_images/image_1.jpg} \end{subfigure}\hfill
    \begin{subfigure}[c]{\subfigwidth} \centering \includegraphics[width=\linewidth]{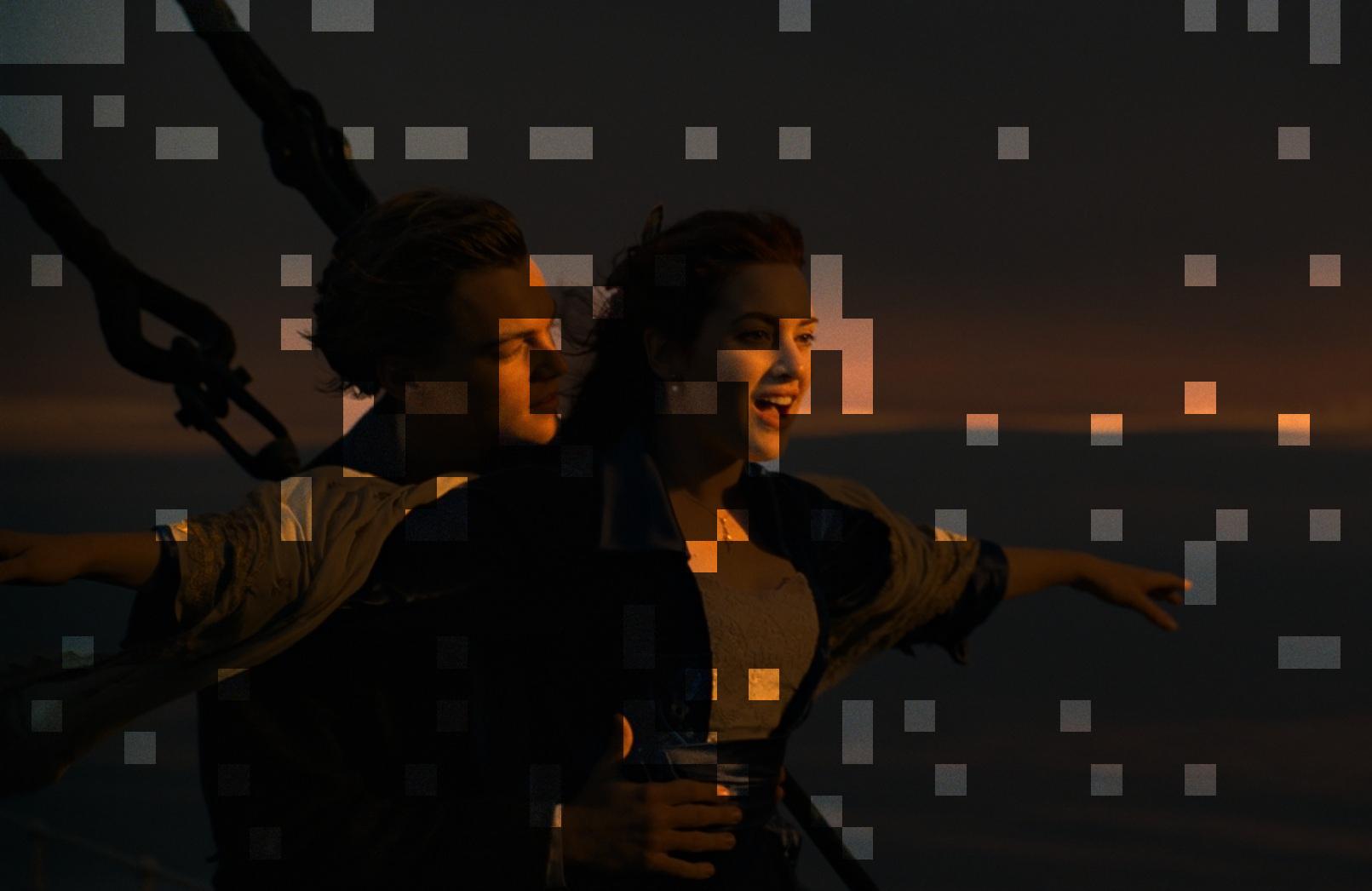} \end{subfigure}\hfill
    \begin{subfigure}[c]{\subfigwidth} \centering \includegraphics[width=\linewidth]{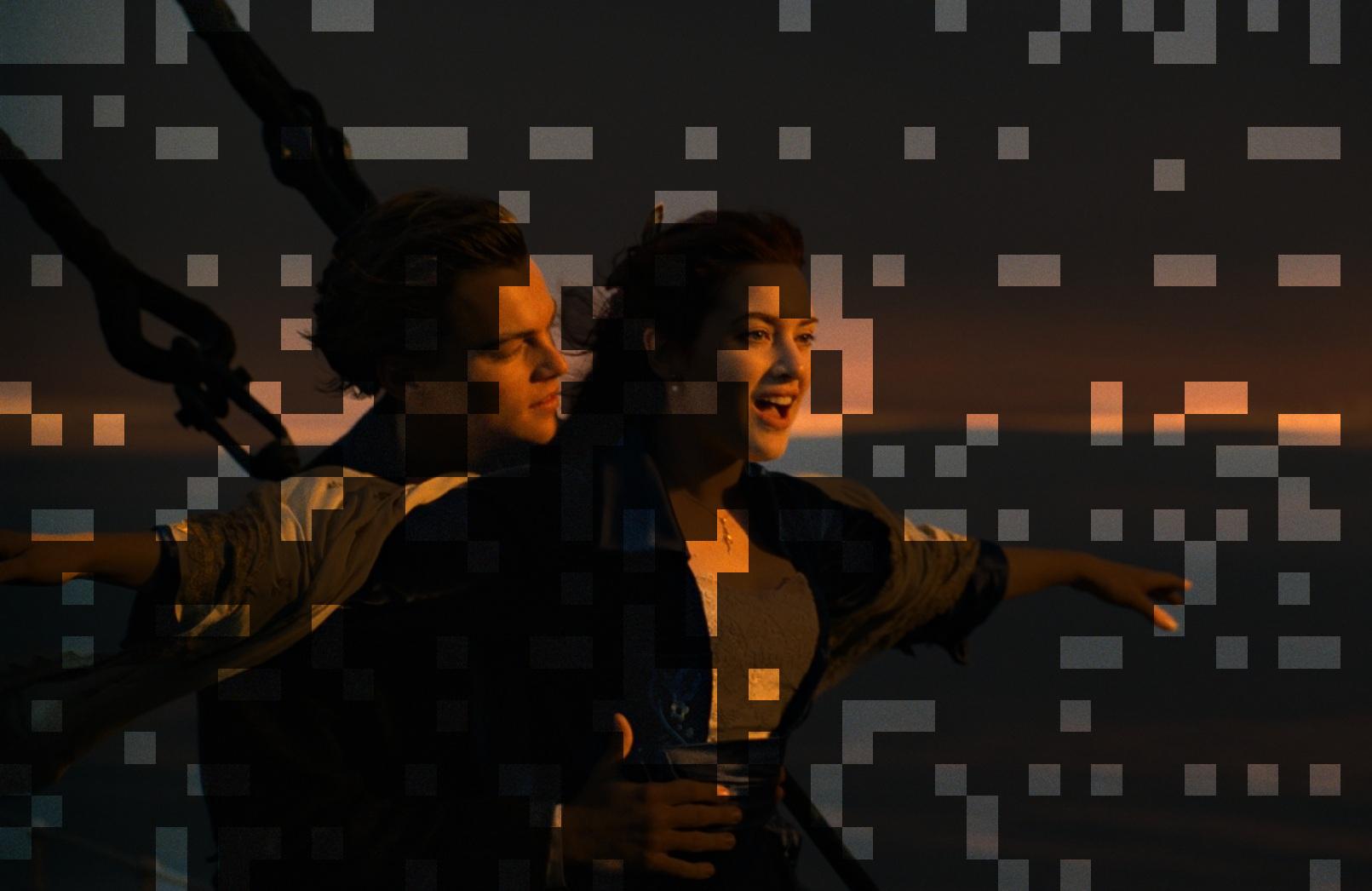} \end{subfigure}\hfill
    \begin{subfigure}[c]{\subfigwidth} \centering \includegraphics[width=\linewidth]{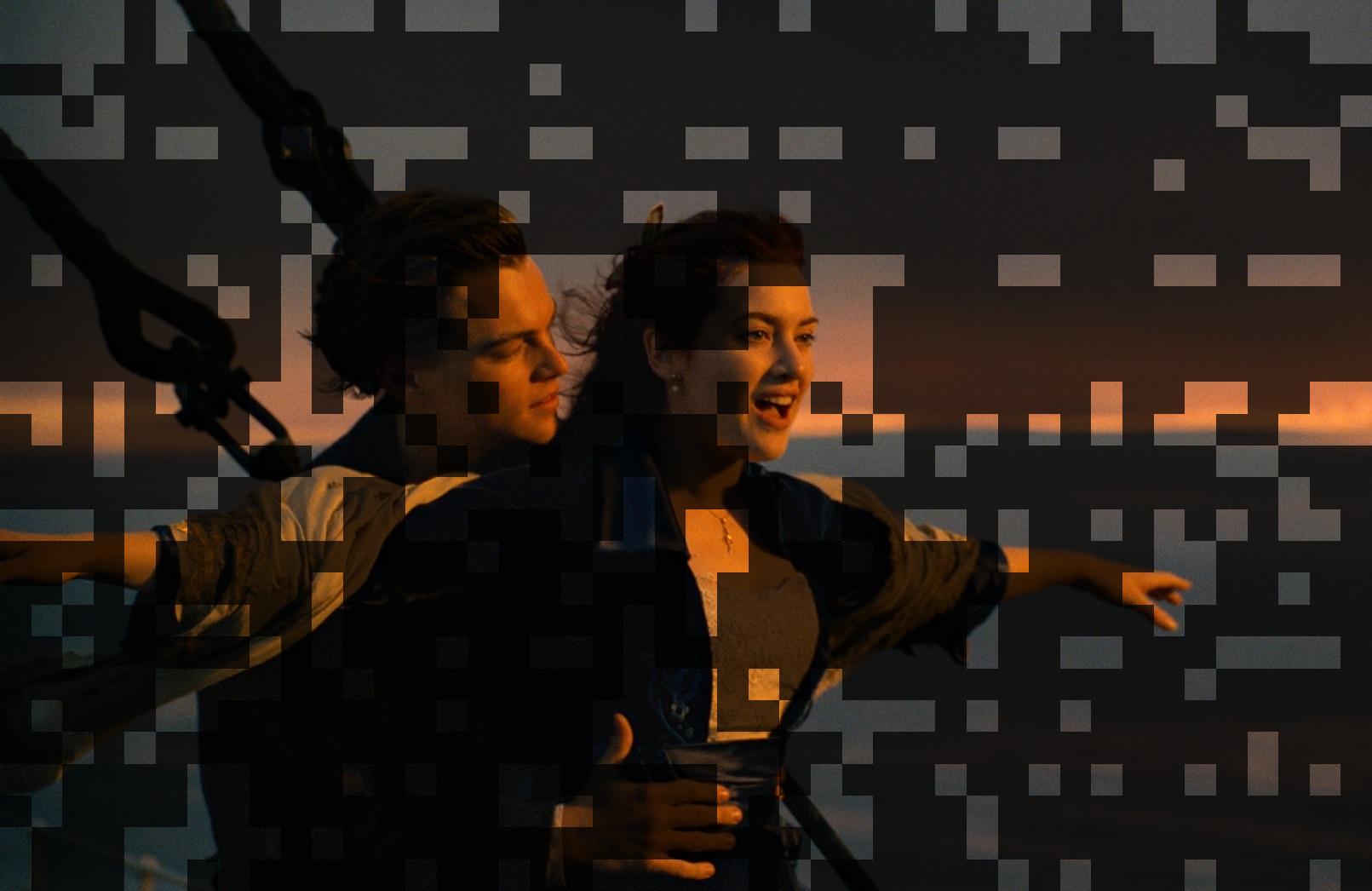} \end{subfigure}\hfill
    \begin{subfigure}[c]{\subfigwidth} \centering \includegraphics[width=\linewidth]{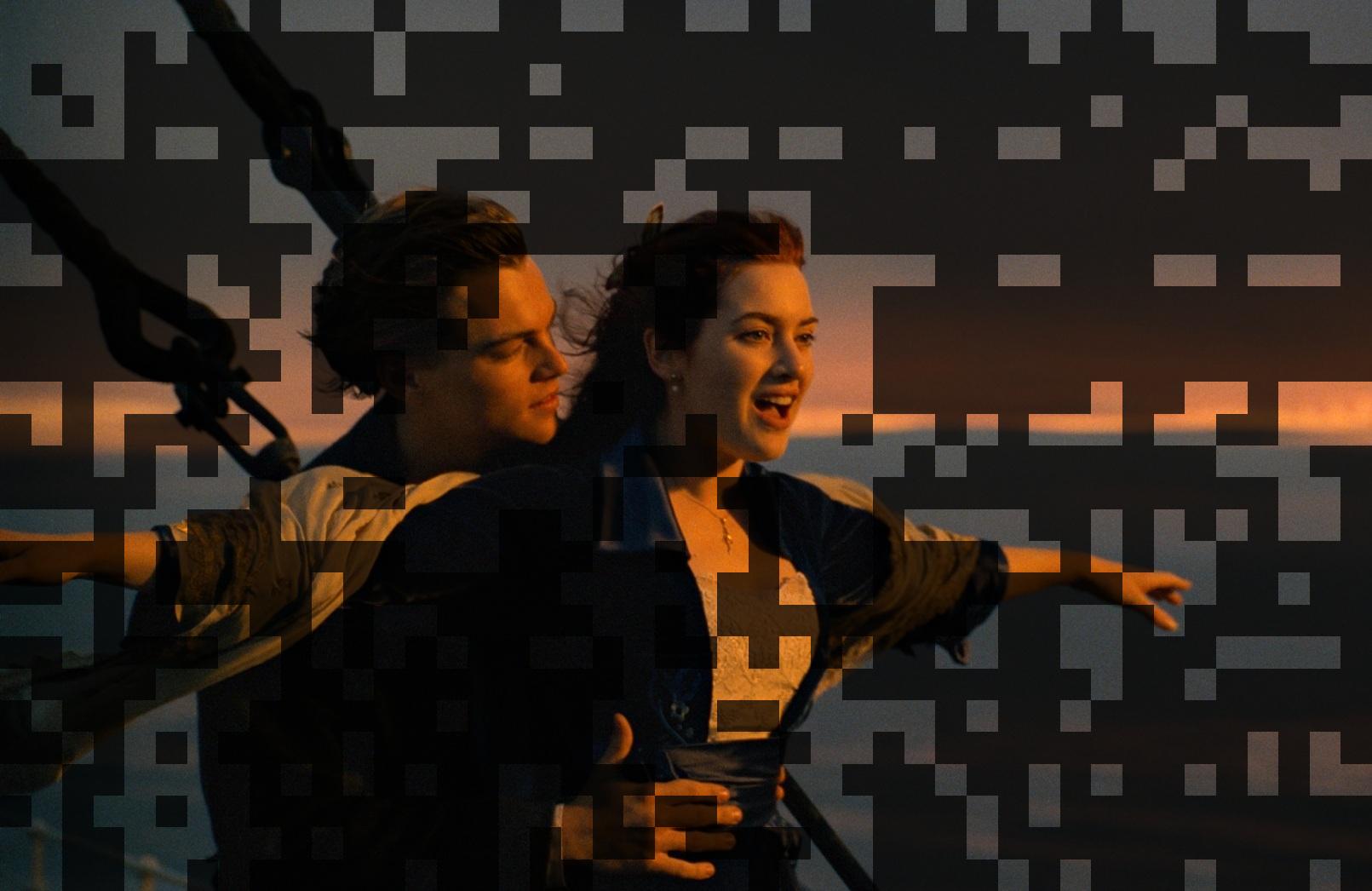} \end{subfigure}\hfill
    \begin{subfigure}[c]{\subfigwidth} \centering \includegraphics[width=\linewidth]{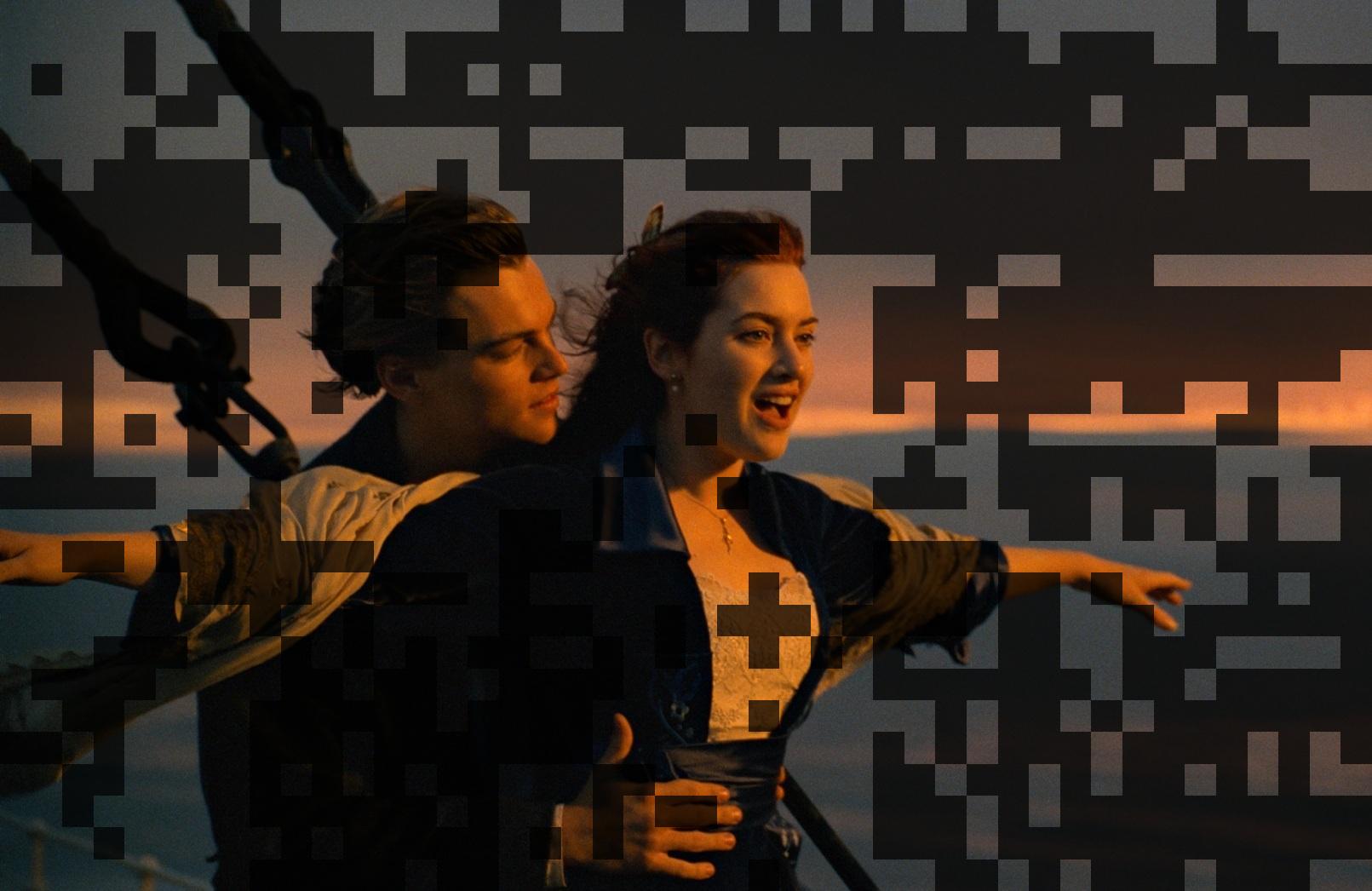} \end{subfigure}\\[1.5mm]

    \begin{subfigure}[c]{0.03\textwidth} \centering \rotatebox{90}{\small \textbf{\textit{101}}} \end{subfigure}\hfill
    \begin{subfigure}[c]{\subfigwidth} \centering \includegraphics[width=\linewidth]{figs/vis_hpe/input_images/image_1.jpg} \end{subfigure}\hfill
    \begin{subfigure}[c]{\subfigwidth} \centering \includegraphics[width=\linewidth]{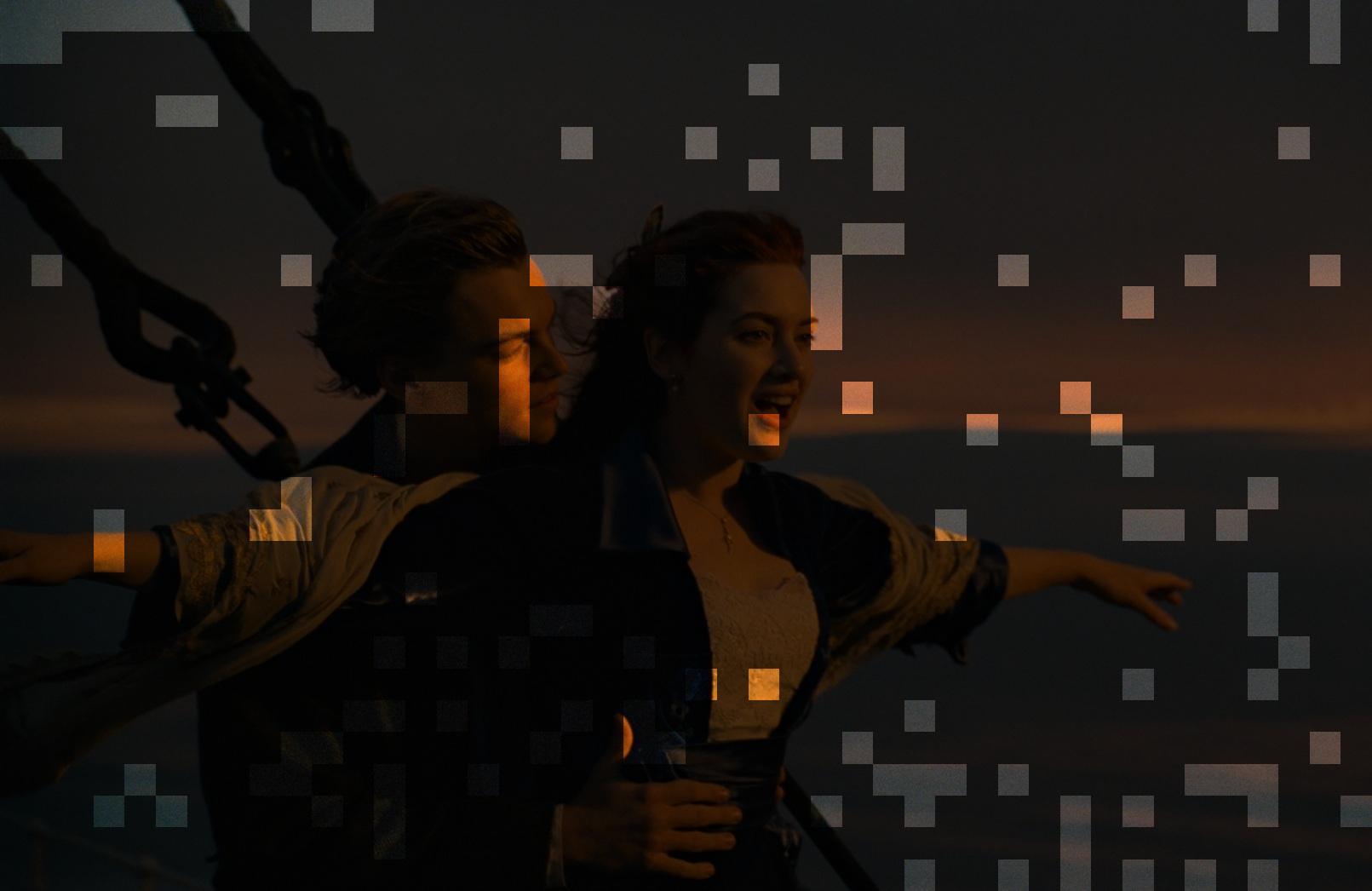} \end{subfigure}\hfill
    \begin{subfigure}[c]{\subfigwidth} \centering \includegraphics[width=\linewidth]{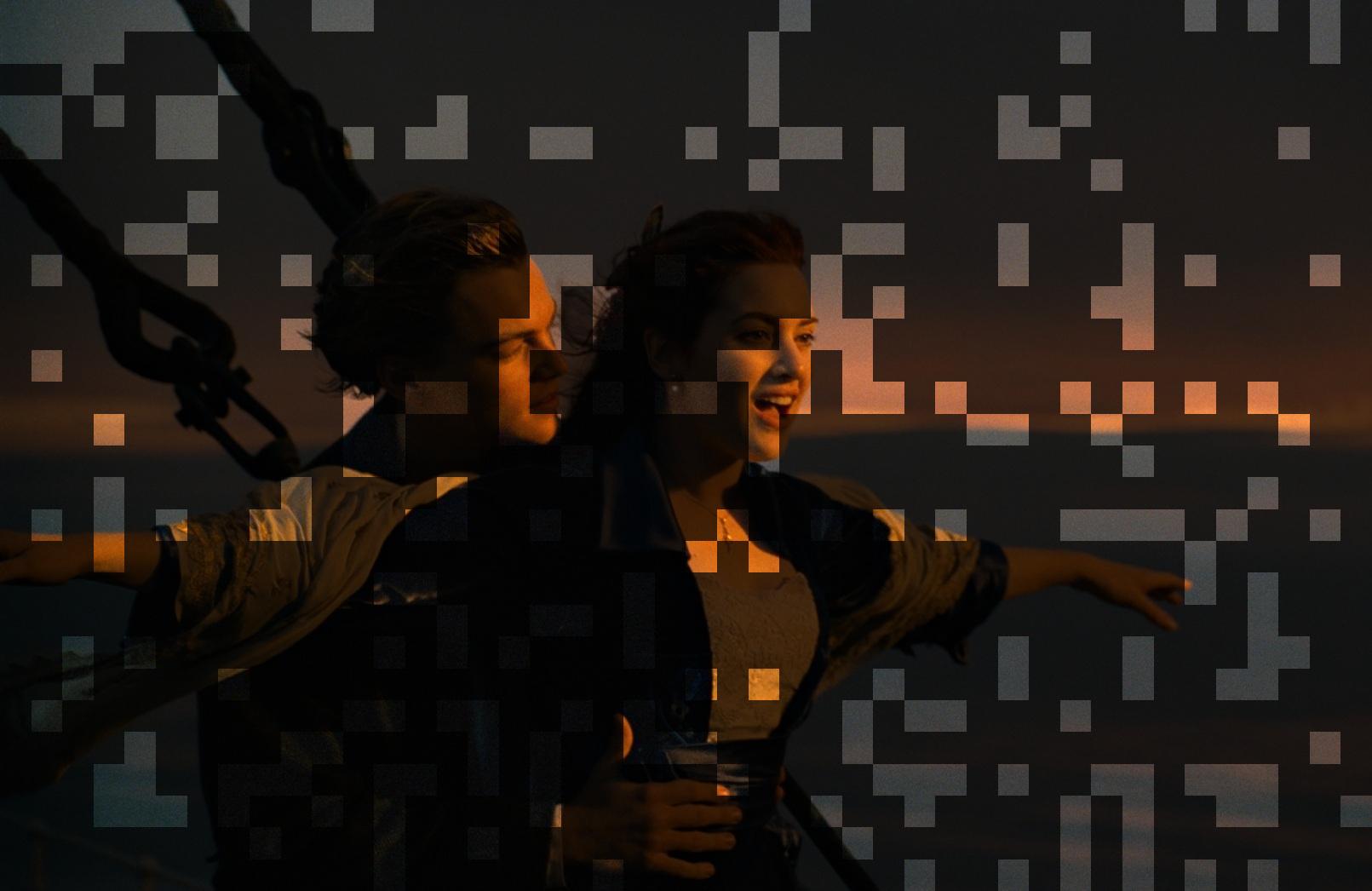} \end{subfigure}\hfill
    \begin{subfigure}[c]{\subfigwidth} \centering \includegraphics[width=\linewidth]{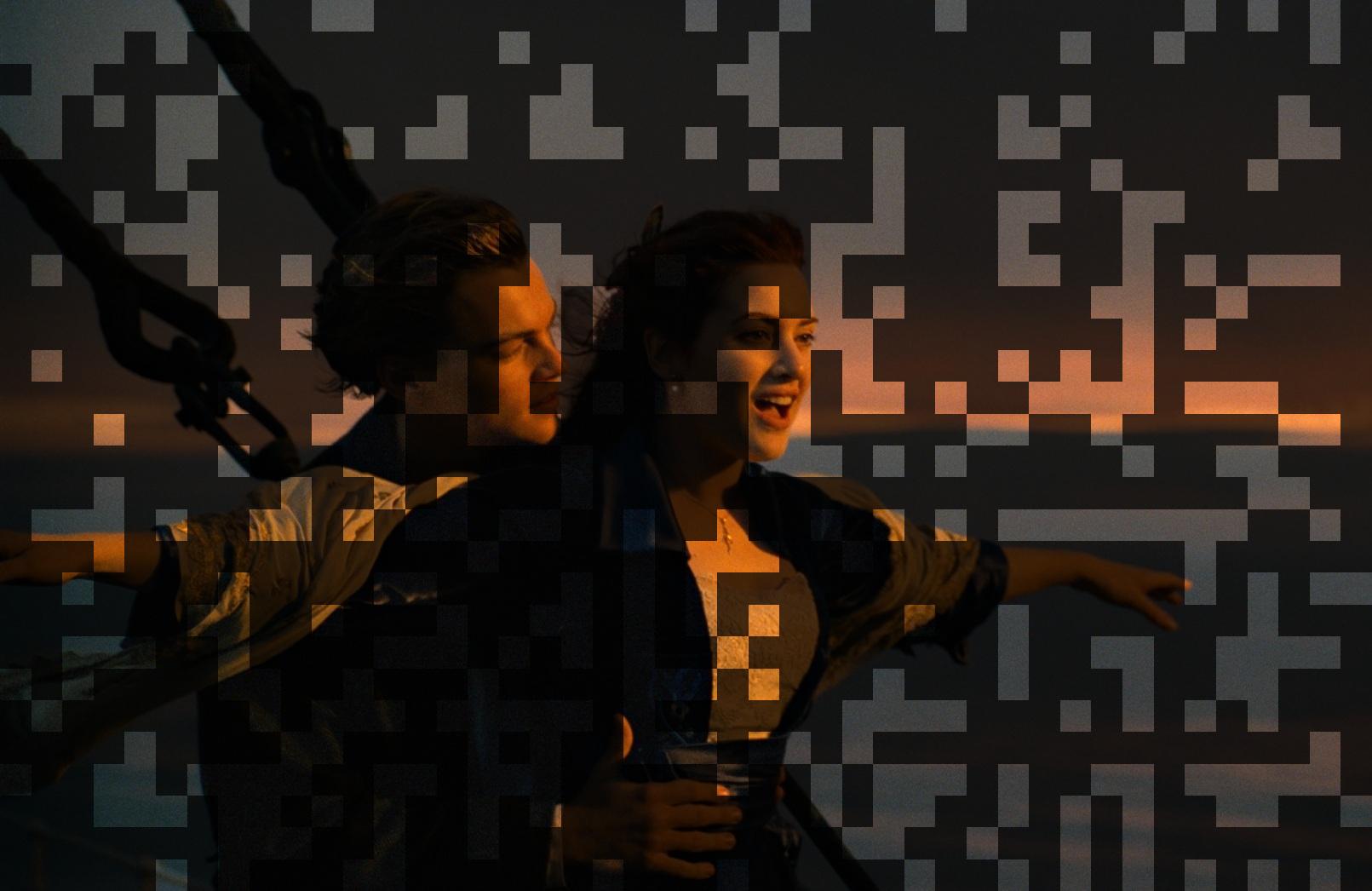} \end{subfigure}\hfill
    \begin{subfigure}[c]{\subfigwidth} \centering \includegraphics[width=\linewidth]{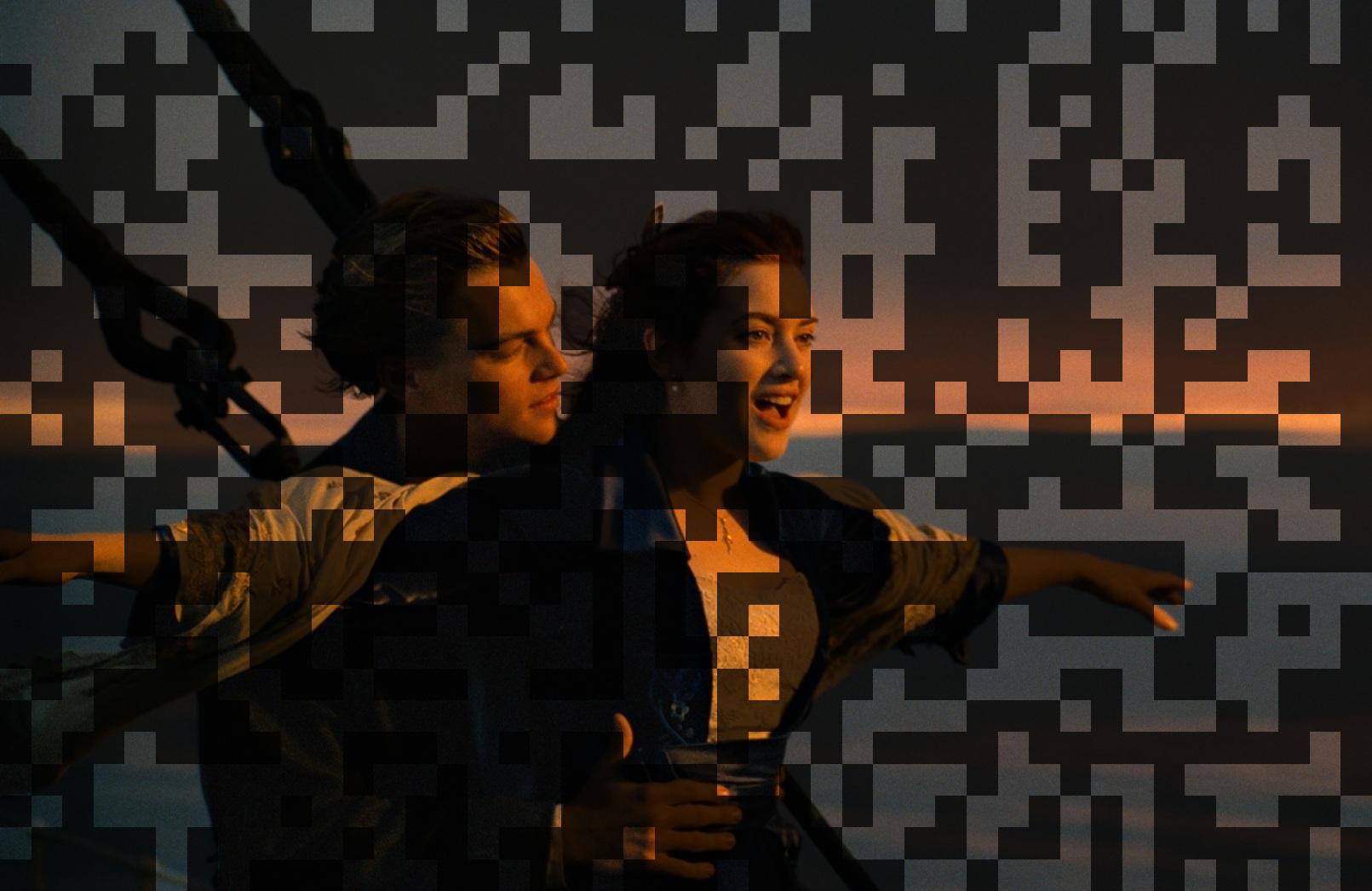} \end{subfigure}\hfill
    \begin{subfigure}[c]{\subfigwidth} \centering \includegraphics[width=\linewidth]{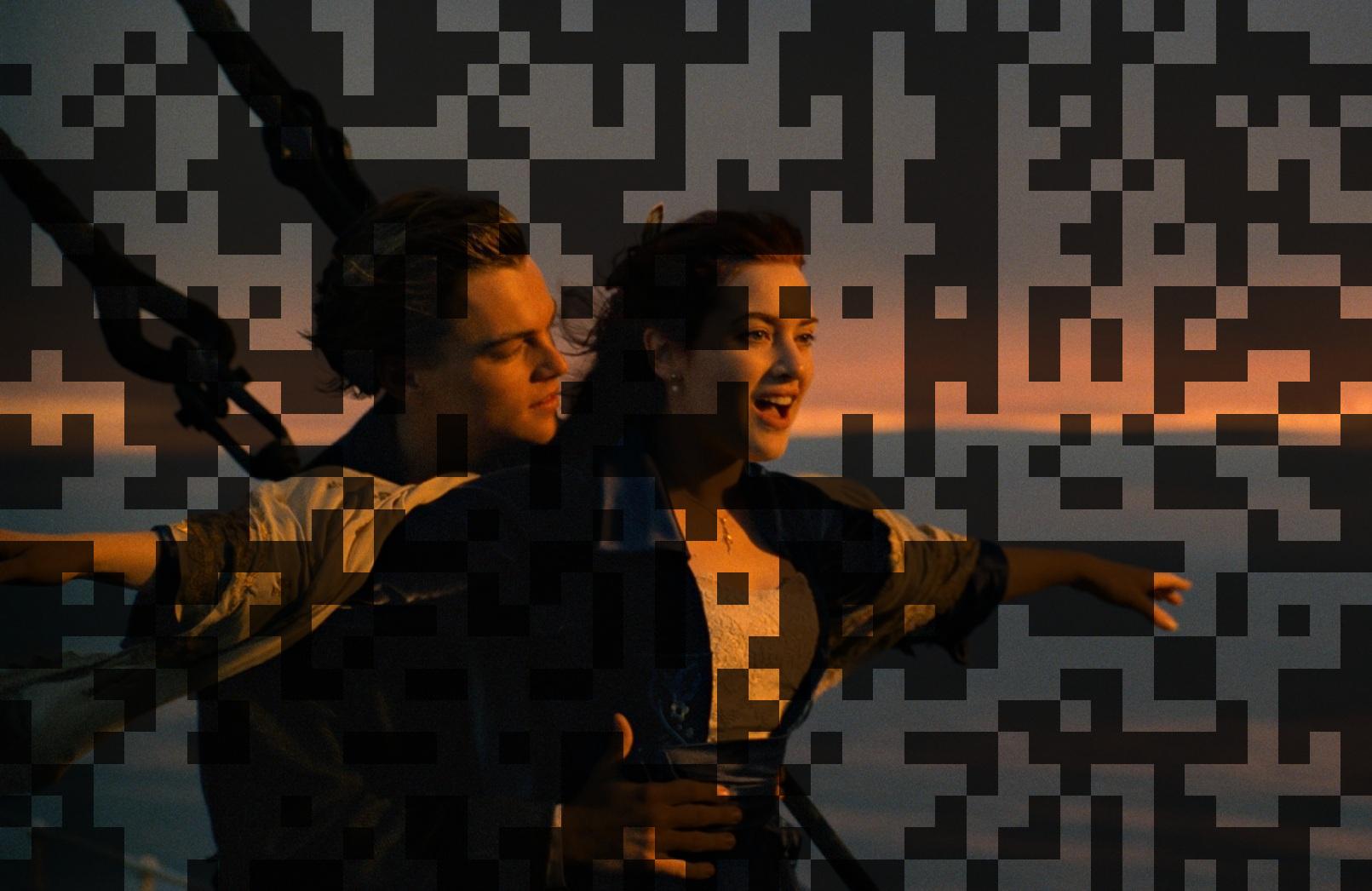} \end{subfigure}\\[1.5mm]

    \begin{subfigure}[c]{0.03\textwidth} \centering \rotatebox{90}{\small \textbf{\textit{110}}} \end{subfigure}\hfill
    \begin{subfigure}[c]{\subfigwidth} \centering \includegraphics[width=\linewidth]{figs/vis_hpe/input_images/image_1.jpg} \end{subfigure}\hfill
    \begin{subfigure}[c]{\subfigwidth} \centering \includegraphics[width=\linewidth]{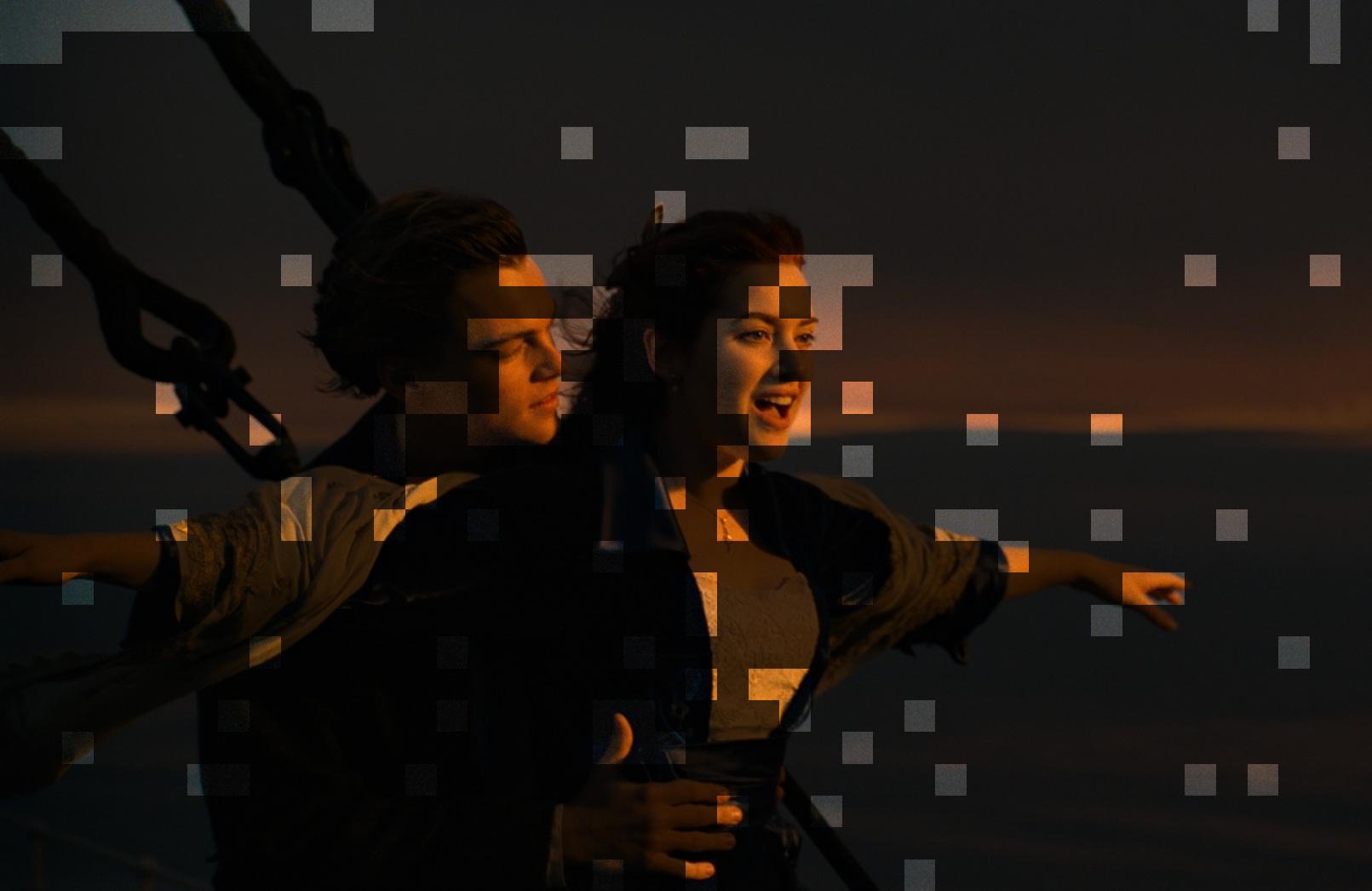} \end{subfigure}\hfill
    \begin{subfigure}[c]{\subfigwidth} \centering \includegraphics[width=\linewidth]{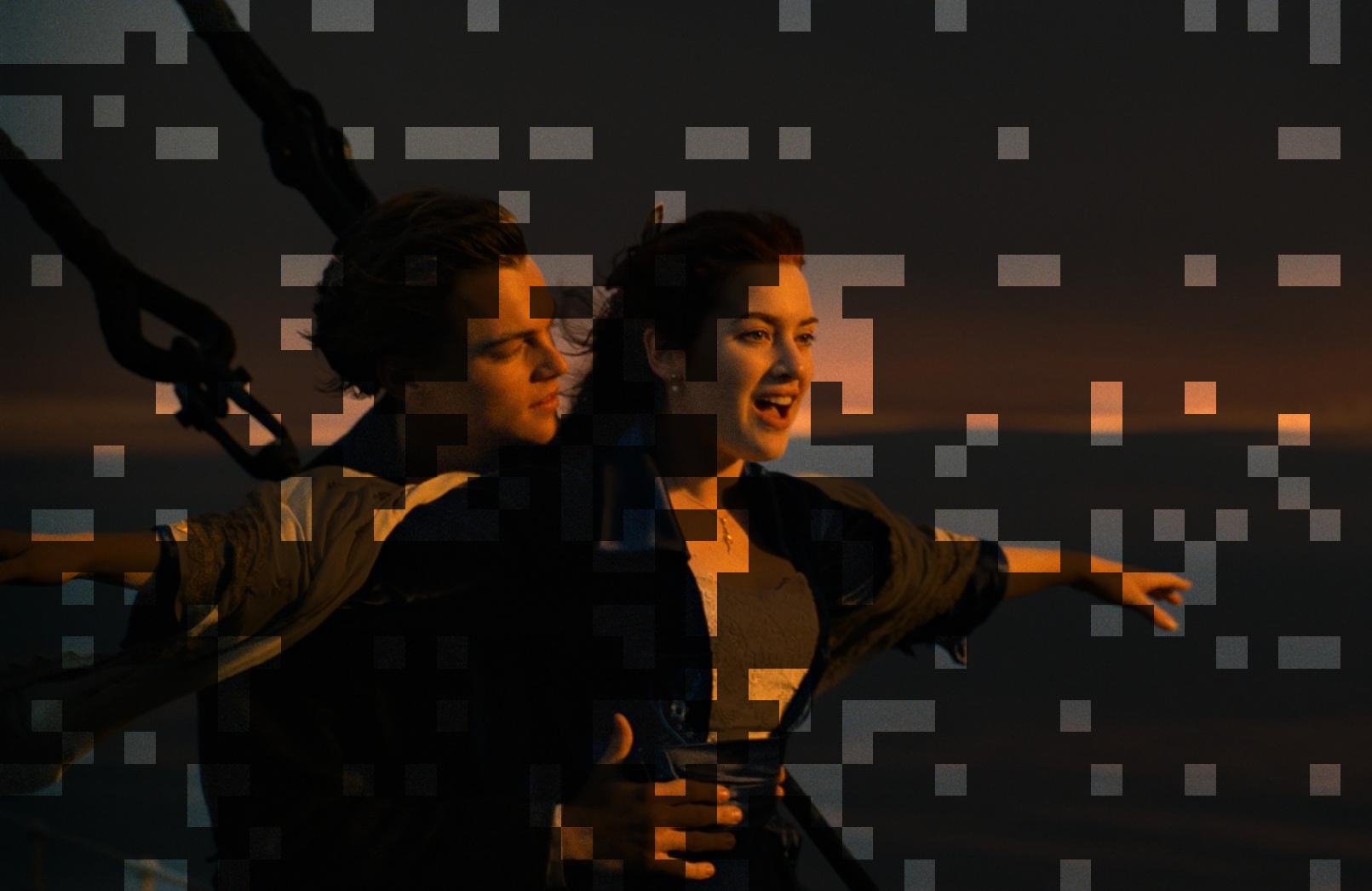} \end{subfigure}\hfill
    \begin{subfigure}[c]{\subfigwidth} \centering \includegraphics[width=\linewidth]{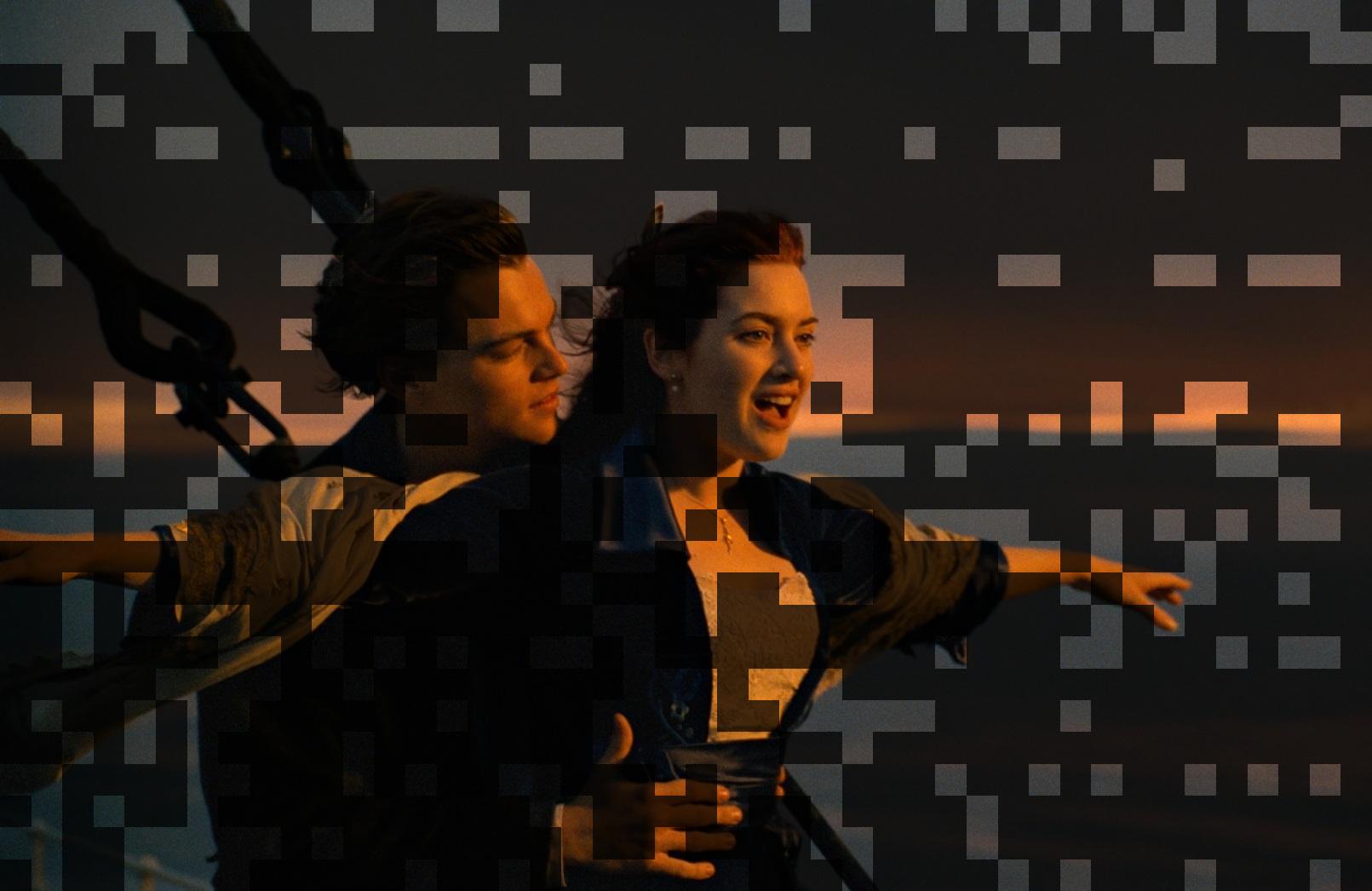} \end{subfigure}\hfill
    \begin{subfigure}[c]{\subfigwidth} \centering \includegraphics[width=\linewidth]{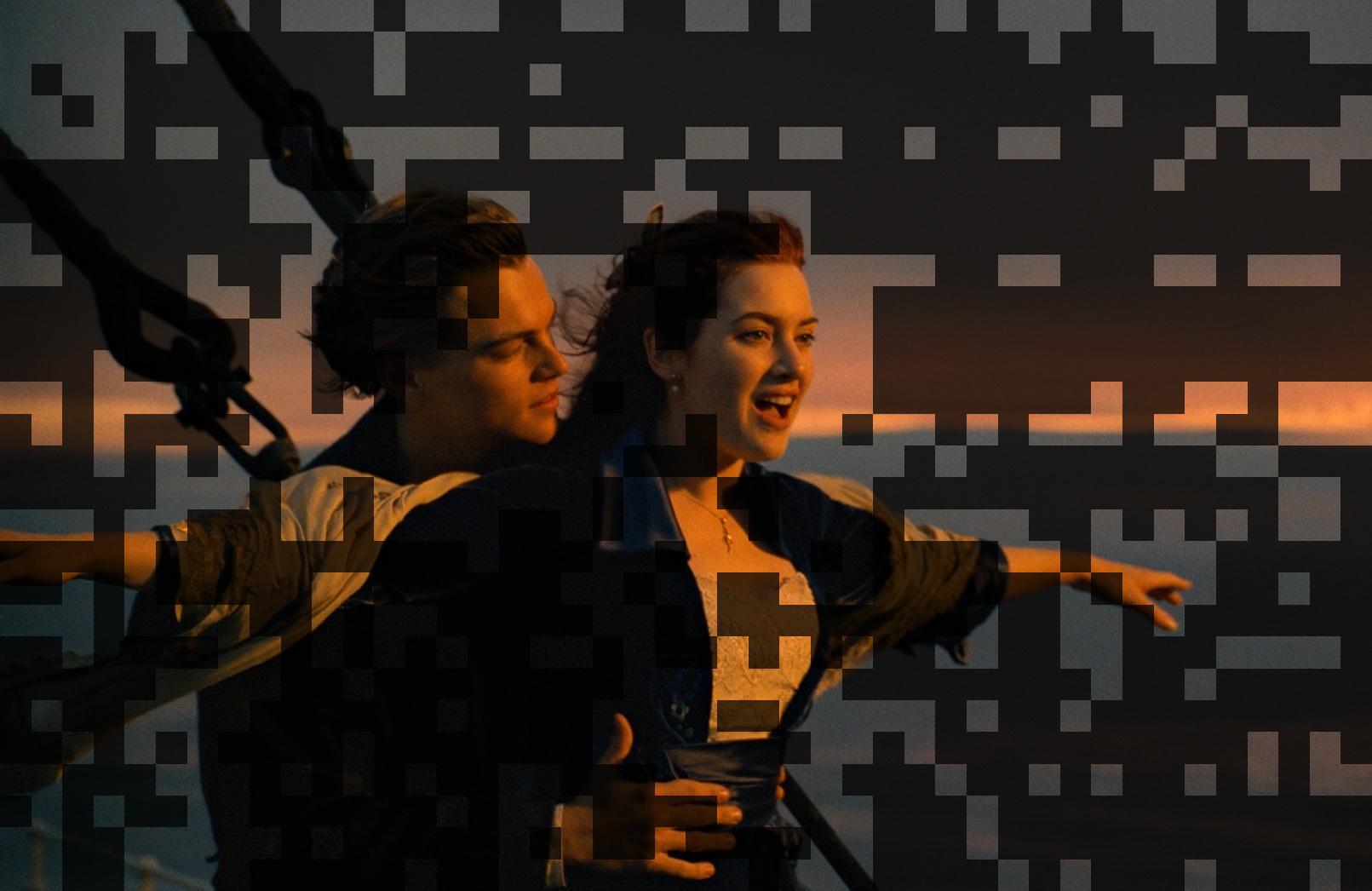} \end{subfigure}\hfill
    \begin{subfigure}[c]{\subfigwidth} \centering \includegraphics[width=\linewidth]{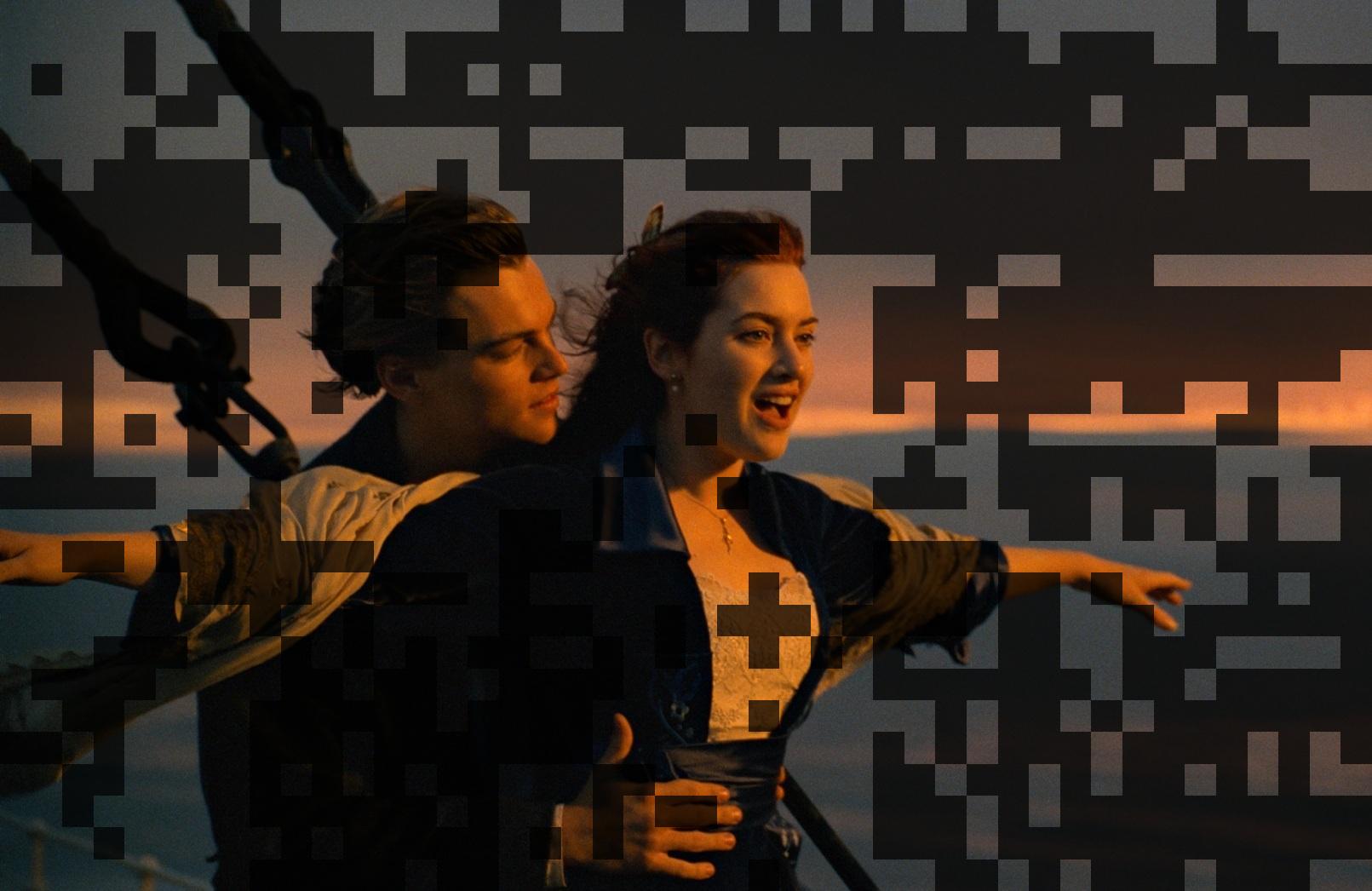} \end{subfigure}\\[1.5mm]

    \begin{subfigure}[c]{0.03\textwidth} \centering \rotatebox{90}{\small \textbf{\textit{ReTrace}}} \end{subfigure}\hfill
    \begin{subfigure}[c]{\subfigwidth} \centering \includegraphics[width=\linewidth]{figs/vis_hpe/input_images/image_1.jpg} \end{subfigure}\hfill
    \begin{subfigure}[c]{\subfigwidth} \centering \includegraphics[width=\linewidth]{figs/vis_hpe/vis_keep_0.1/image_1.jpg} \end{subfigure}\hfill
    \begin{subfigure}[c]{\subfigwidth} \centering \includegraphics[width=\linewidth]{figs/vis_hpe/vis_keep_0.2/image_1.jpg} \end{subfigure}\hfill
    \begin{subfigure}[c]{\subfigwidth} \centering \includegraphics[width=\linewidth]{figs/vis_hpe/vis_keep_0.3/image_1.jpg} \end{subfigure}\hfill
    \begin{subfigure}[c]{\subfigwidth} \centering \includegraphics[width=\linewidth]{figs/vis_hpe/vis_keep_0.4/image_1.jpg} \end{subfigure}\hfill
    \begin{subfigure}[c]{\subfigwidth} \centering \includegraphics[width=\linewidth]{figs/vis_hpe/vis_keep_0.5/image_1.jpg} \end{subfigure}
    \caption{\textbf{Qualitative comparison.}}
    \label{fig:ablation_vis_image_1}
\end{figure*}

\begin{figure*}[p]
    \centering
    \newcommand{\subfigwidth}{0.155\textwidth}

    \begin{subfigure}[c]{0.03\textwidth} \centering \rotatebox{90}{\small \textbf{\textit{001}}} \end{subfigure}\hfill
    \begin{subfigure}[c]{\subfigwidth} \centering \small \textit{Original}\\[1mm] \includegraphics[width=\linewidth]{figs/vis_hpe/input_images/image.jpg} \end{subfigure}\hfill
    \begin{subfigure}[c]{\subfigwidth} \centering \small \textit{Retain 10\%}\\[1mm] \includegraphics[width=\linewidth]{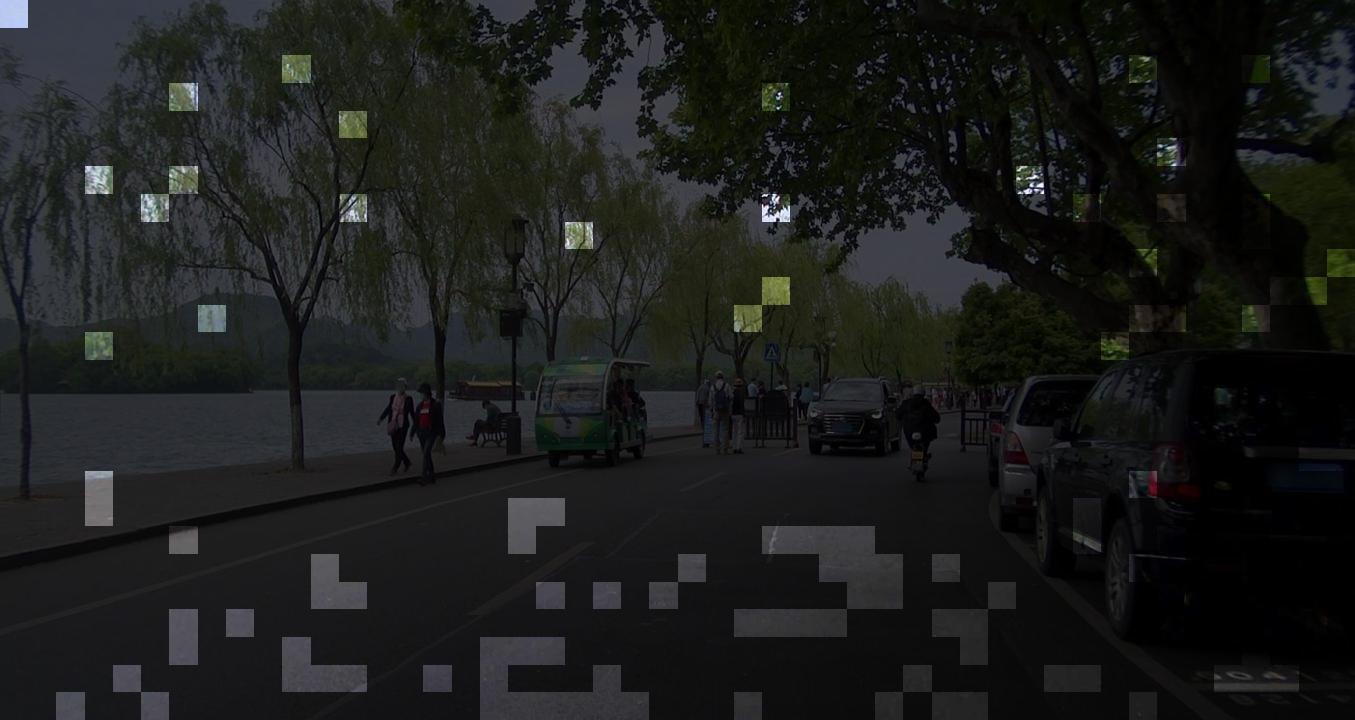} \end{subfigure}\hfill
    \begin{subfigure}[c]{\subfigwidth} \centering \small \textit{Retain 20\%}\\[1mm] \includegraphics[width=\linewidth]{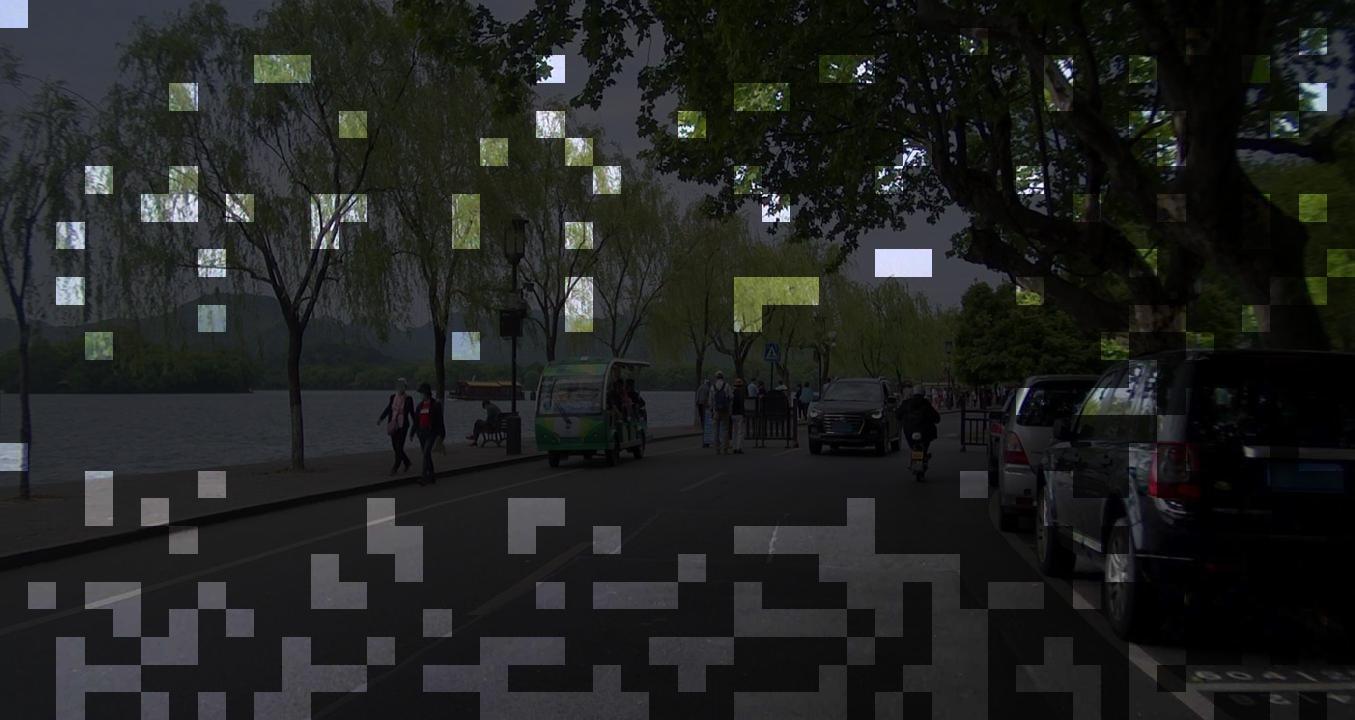} \end{subfigure}\hfill
    \begin{subfigure}[c]{\subfigwidth} \centering \small \textit{Retain 30\%}\\[1mm] \includegraphics[width=\linewidth]{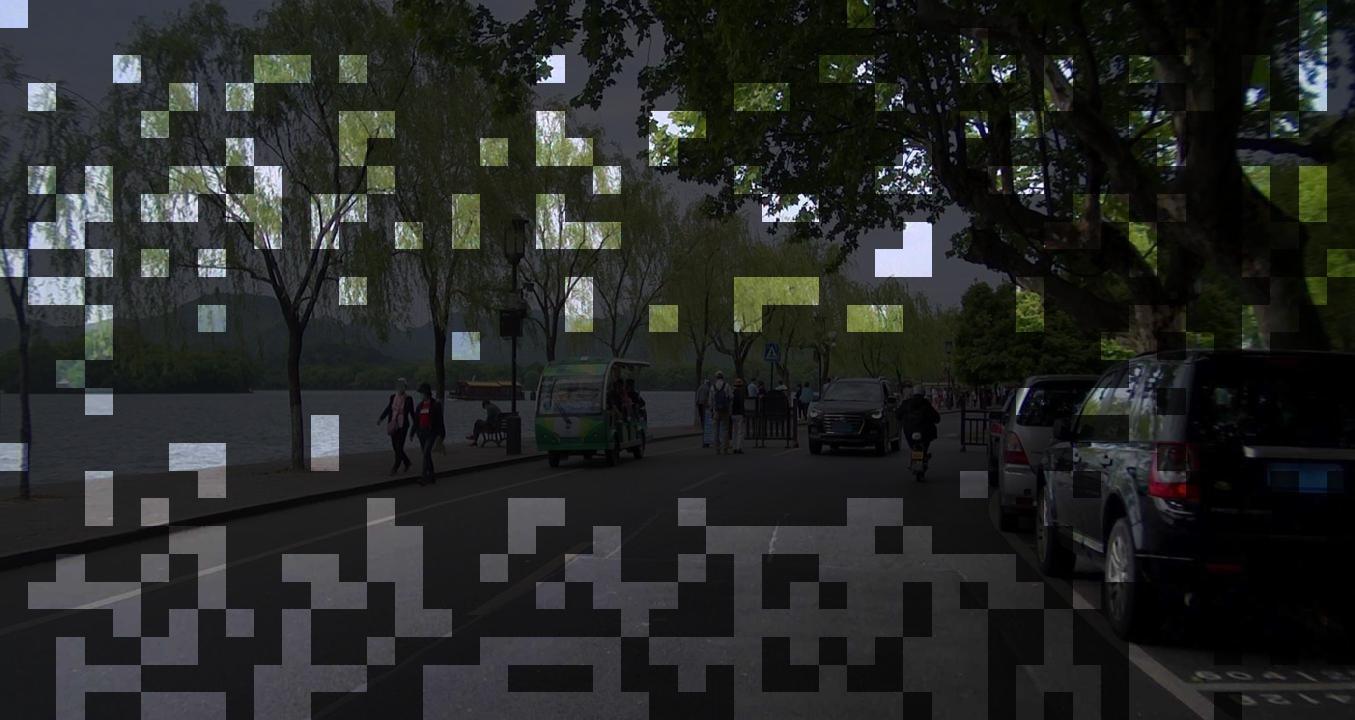} \end{subfigure}\hfill
    \begin{subfigure}[c]{\subfigwidth} \centering \small \textit{Retain 40\%}\\[1mm] \includegraphics[width=\linewidth]{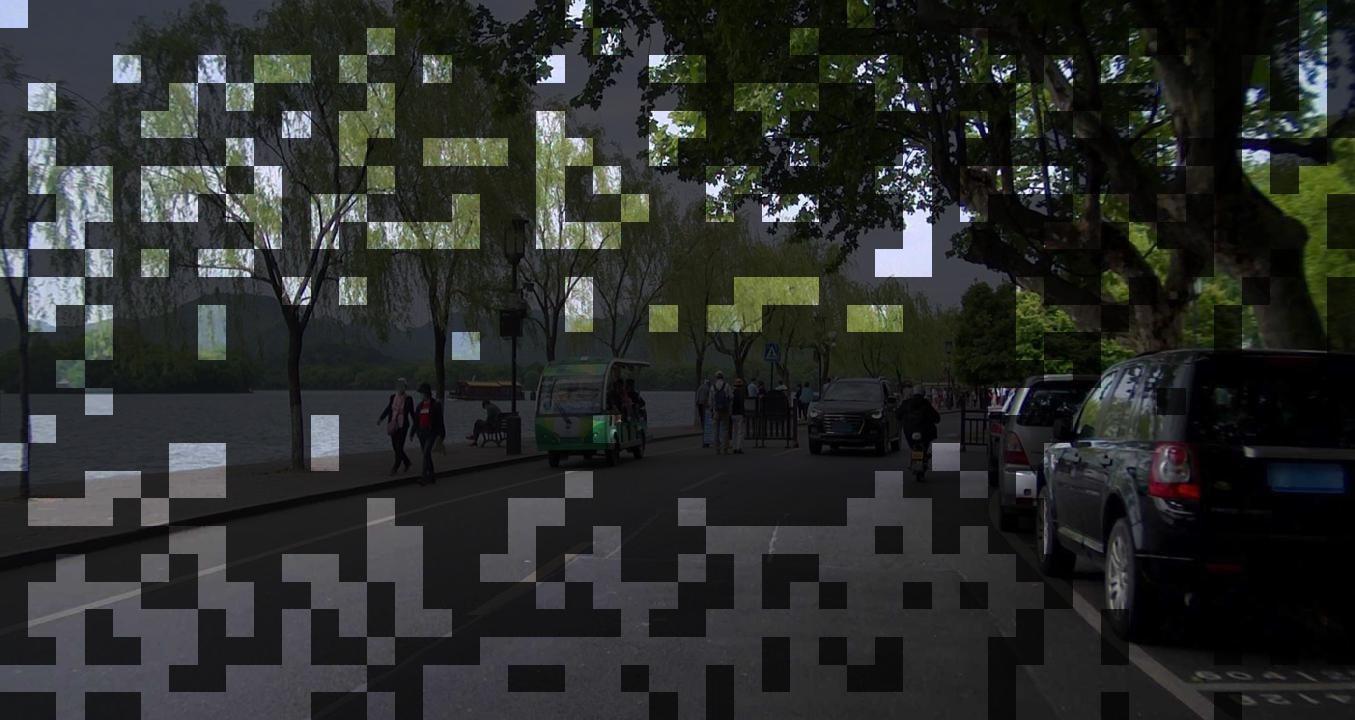} \end{subfigure}\hfill
    \begin{subfigure}[c]{\subfigwidth} \centering \small \textit{Retain 50\%}\\[1mm] \includegraphics[width=\linewidth]{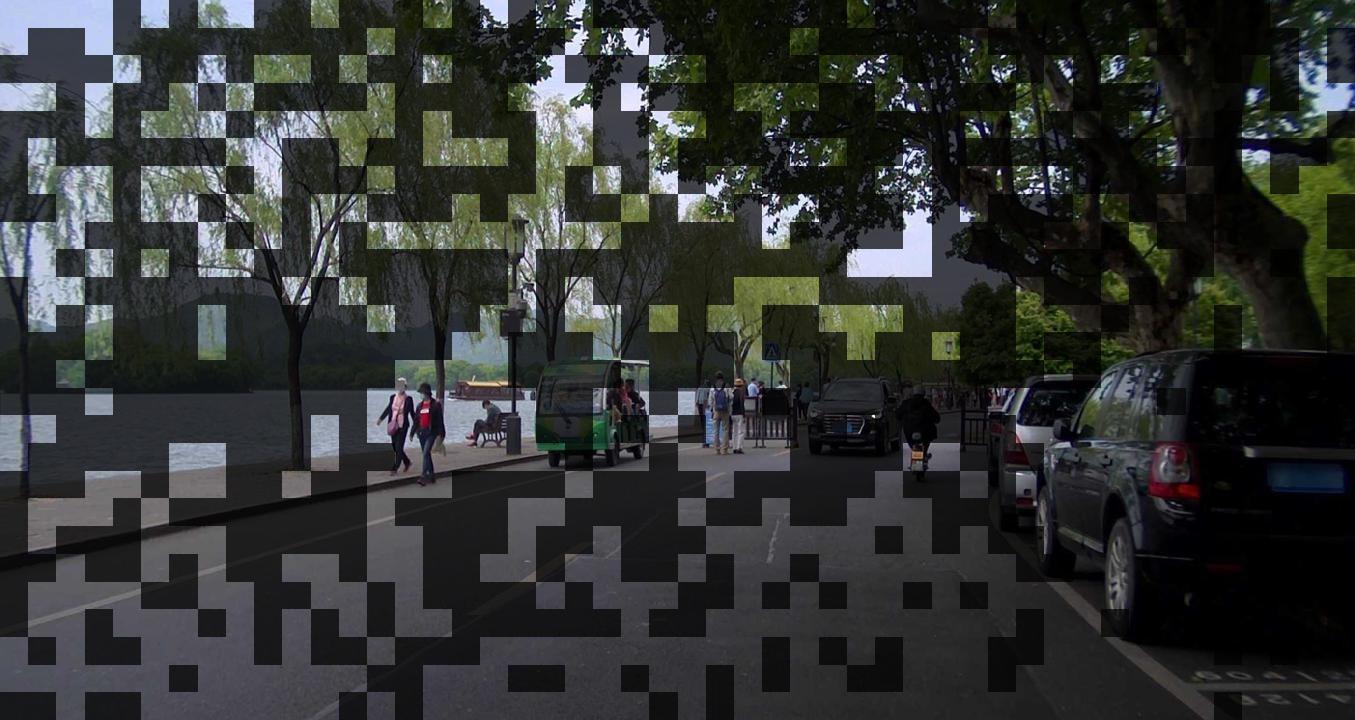} \end{subfigure}\\[1.5mm]

    \begin{subfigure}[c]{0.03\textwidth} \centering \rotatebox{90}{\small \textbf{\textit{100}}} \end{subfigure}\hfill
    \begin{subfigure}[c]{\subfigwidth} \centering \includegraphics[width=\linewidth]{figs/vis_hpe/input_images/image.jpg} \end{subfigure}\hfill
    \begin{subfigure}[c]{\subfigwidth} \centering \includegraphics[width=\linewidth]{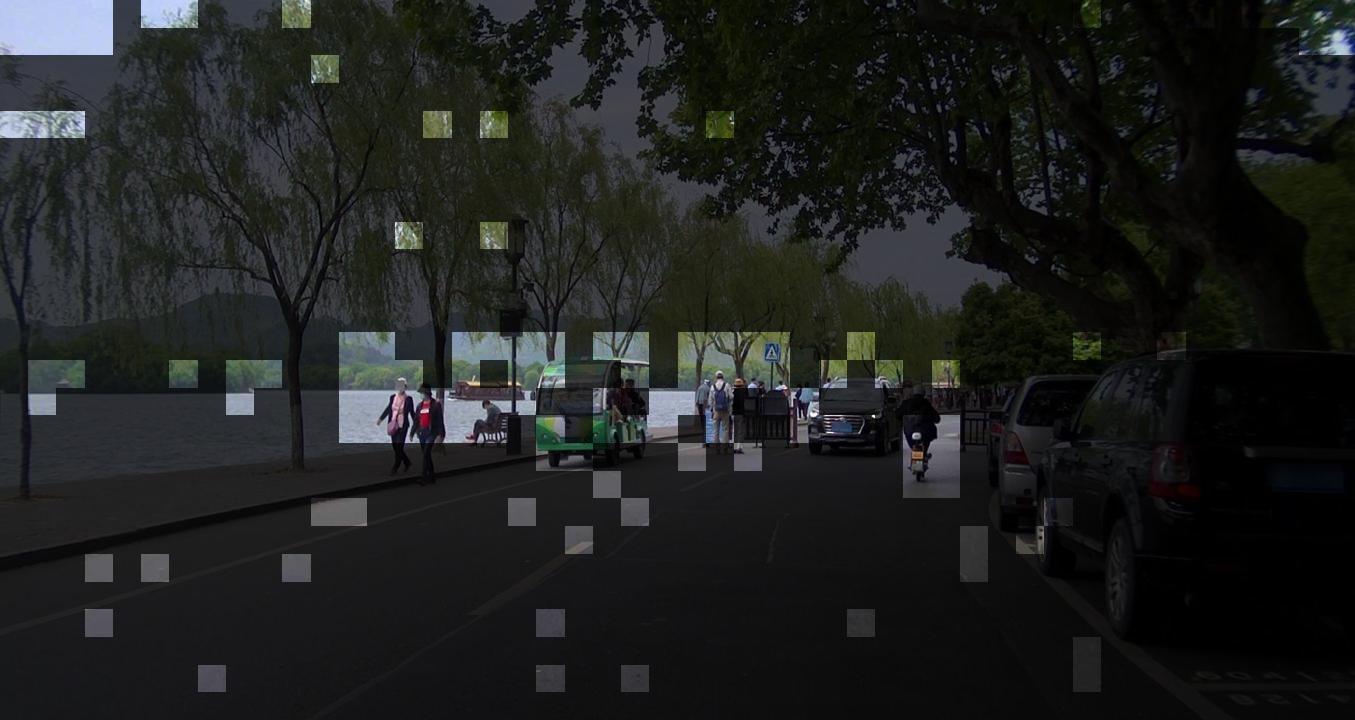} \end{subfigure}\hfill
    \begin{subfigure}[c]{\subfigwidth} \centering \includegraphics[width=\linewidth]{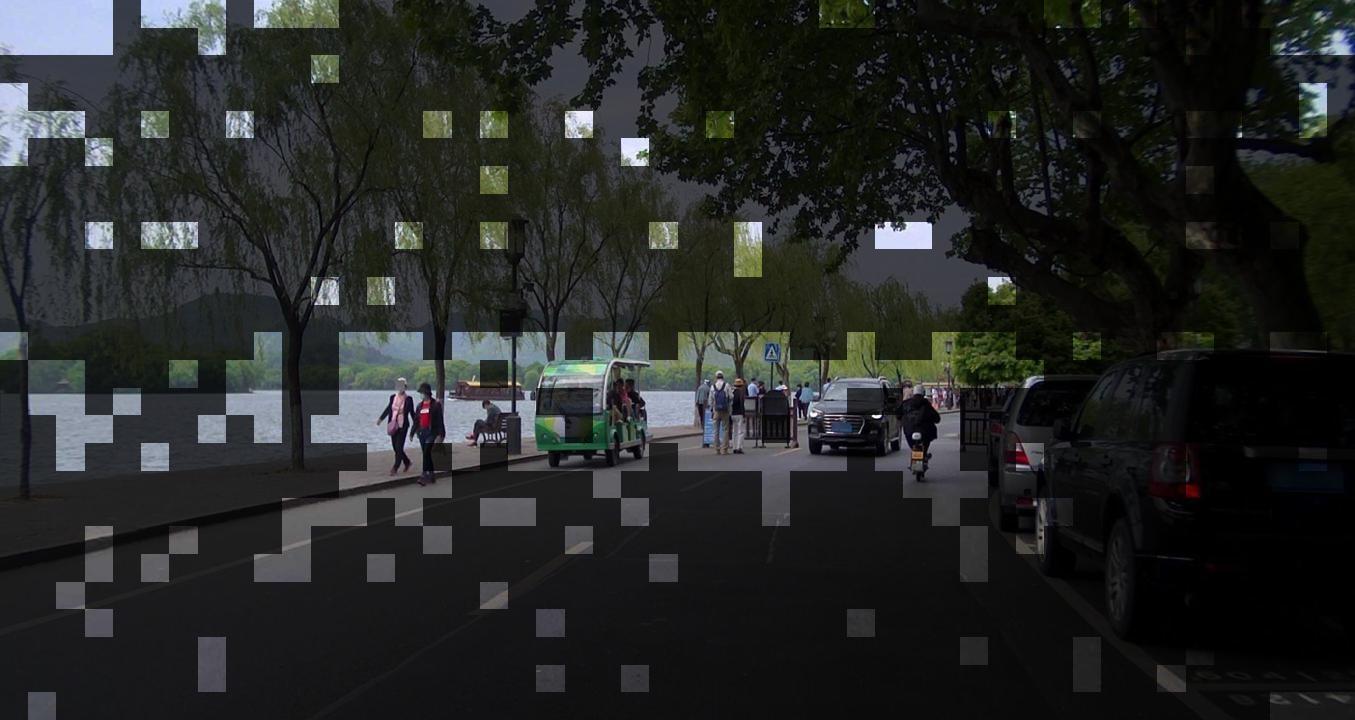} \end{subfigure}\hfill
    \begin{subfigure}[c]{\subfigwidth} \centering \includegraphics[width=\linewidth]{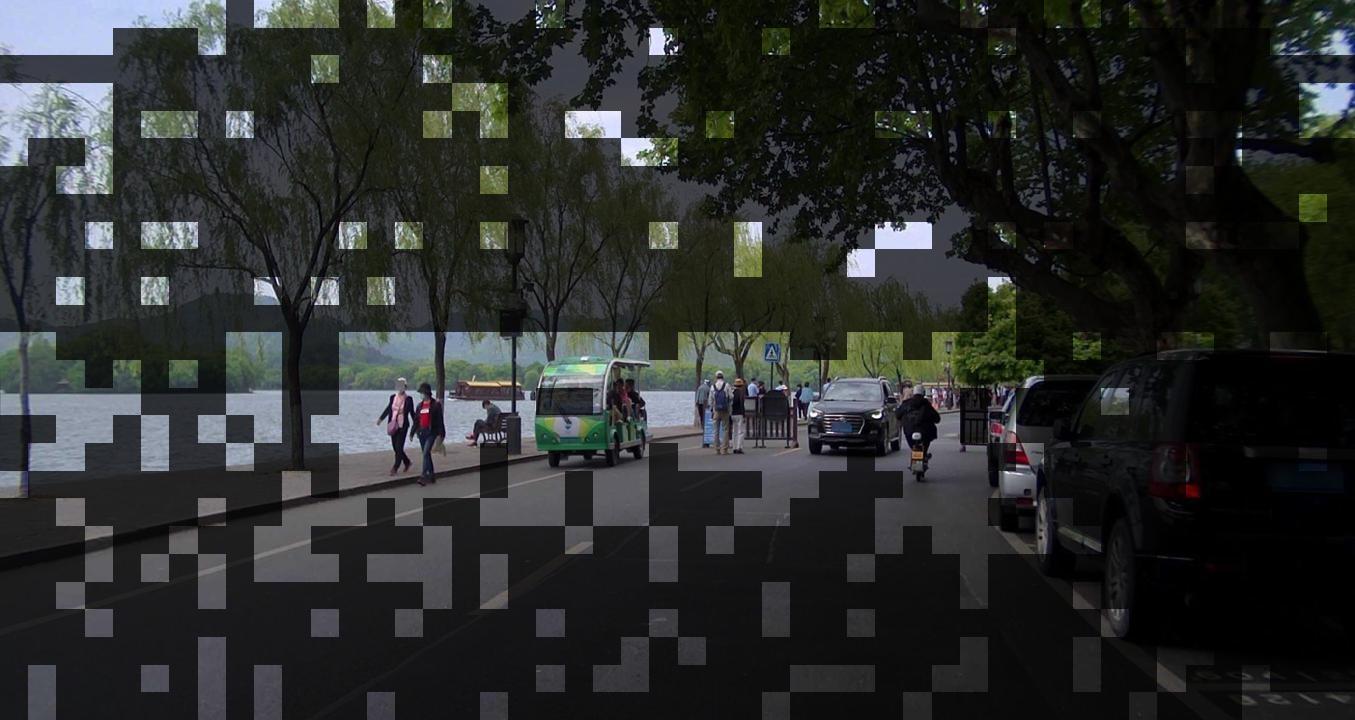} \end{subfigure}\hfill
    \begin{subfigure}[c]{\subfigwidth} \centering \includegraphics[width=\linewidth]{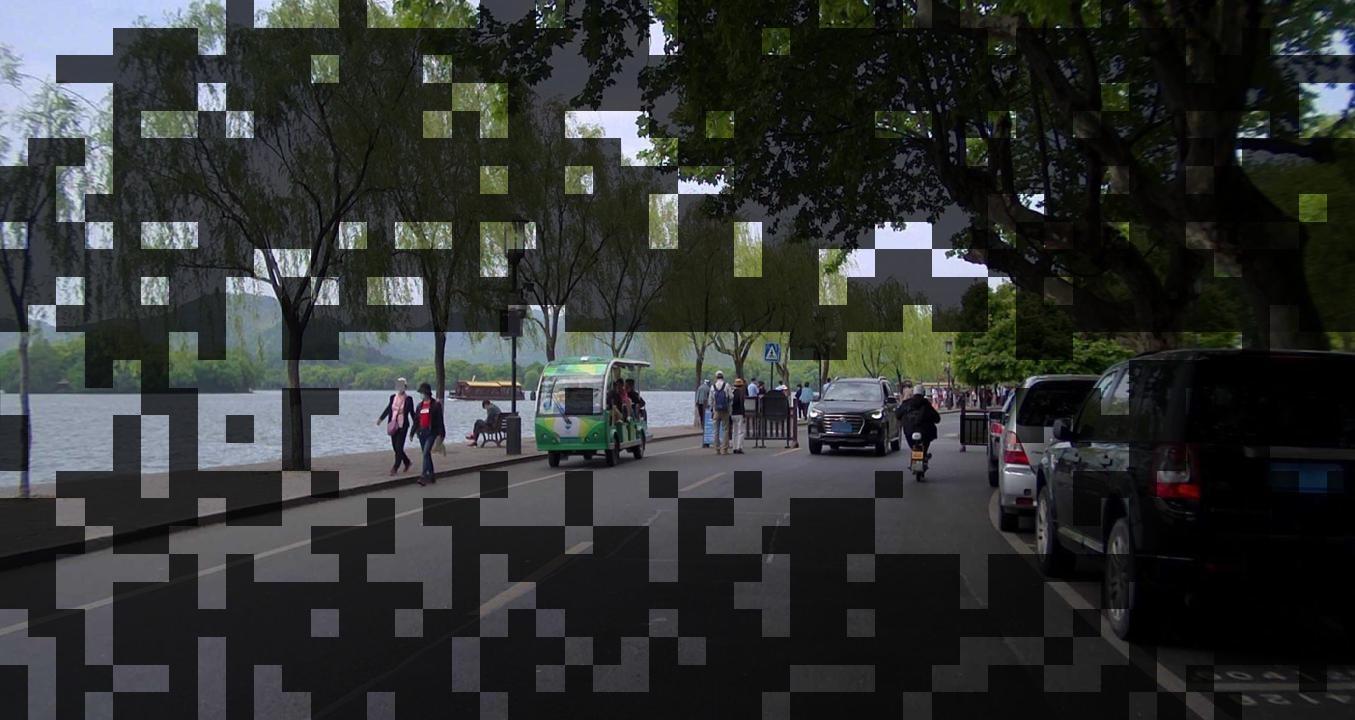} \end{subfigure}\hfill
    \begin{subfigure}[c]{\subfigwidth} \centering \includegraphics[width=\linewidth]{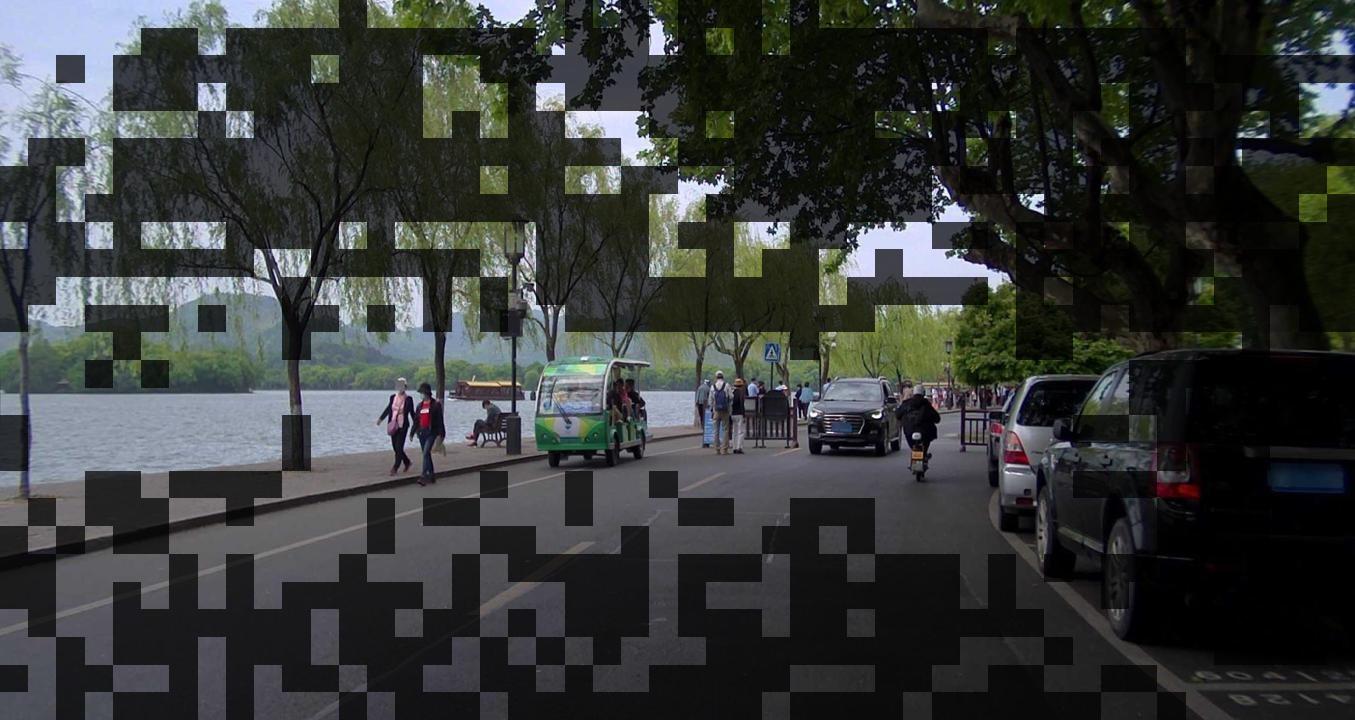} \end{subfigure}\\[1.5mm]

    \begin{subfigure}[c]{0.03\textwidth} \centering \rotatebox{90}{\small \textbf{\textit{101}}} \end{subfigure}\hfill
    \begin{subfigure}[c]{\subfigwidth} \centering \includegraphics[width=\linewidth]{figs/vis_hpe/input_images/image.jpg} \end{subfigure}\hfill
    \begin{subfigure}[c]{\subfigwidth} \centering \includegraphics[width=\linewidth]{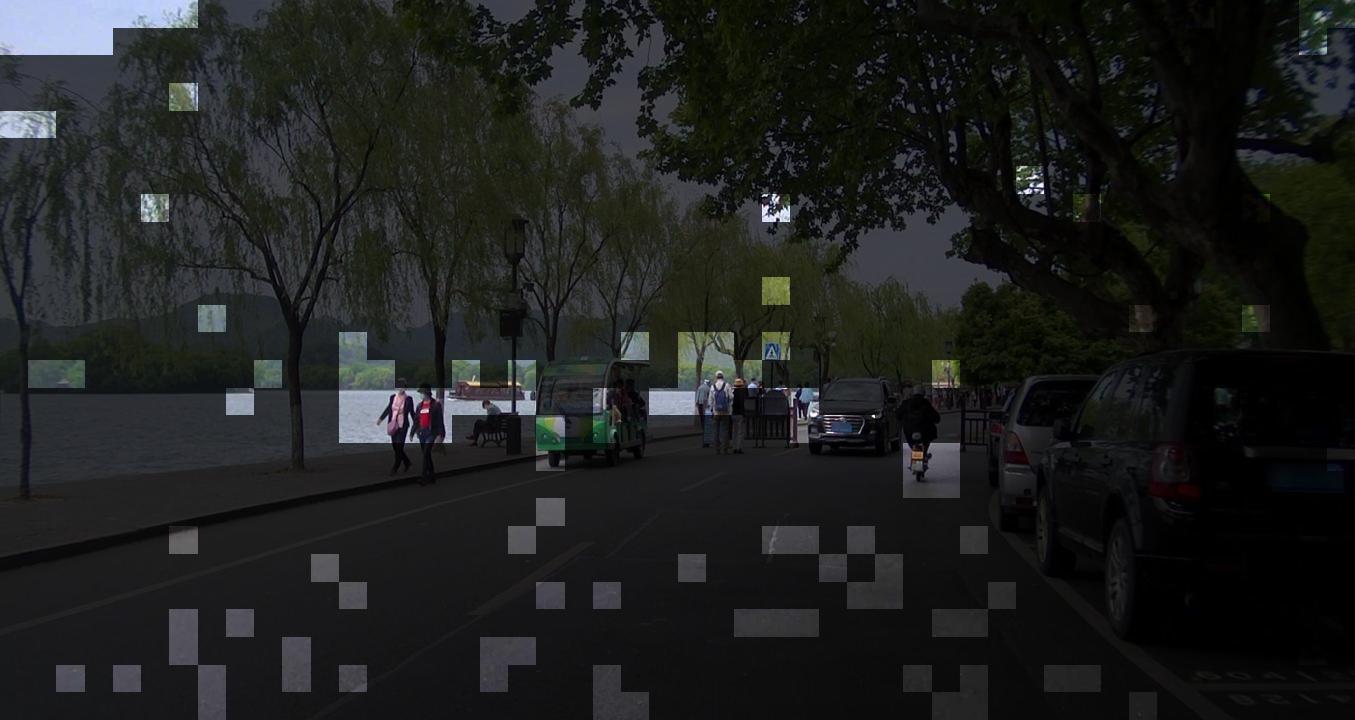} \end{subfigure}\hfill
    \begin{subfigure}[c]{\subfigwidth} \centering \includegraphics[width=\linewidth]{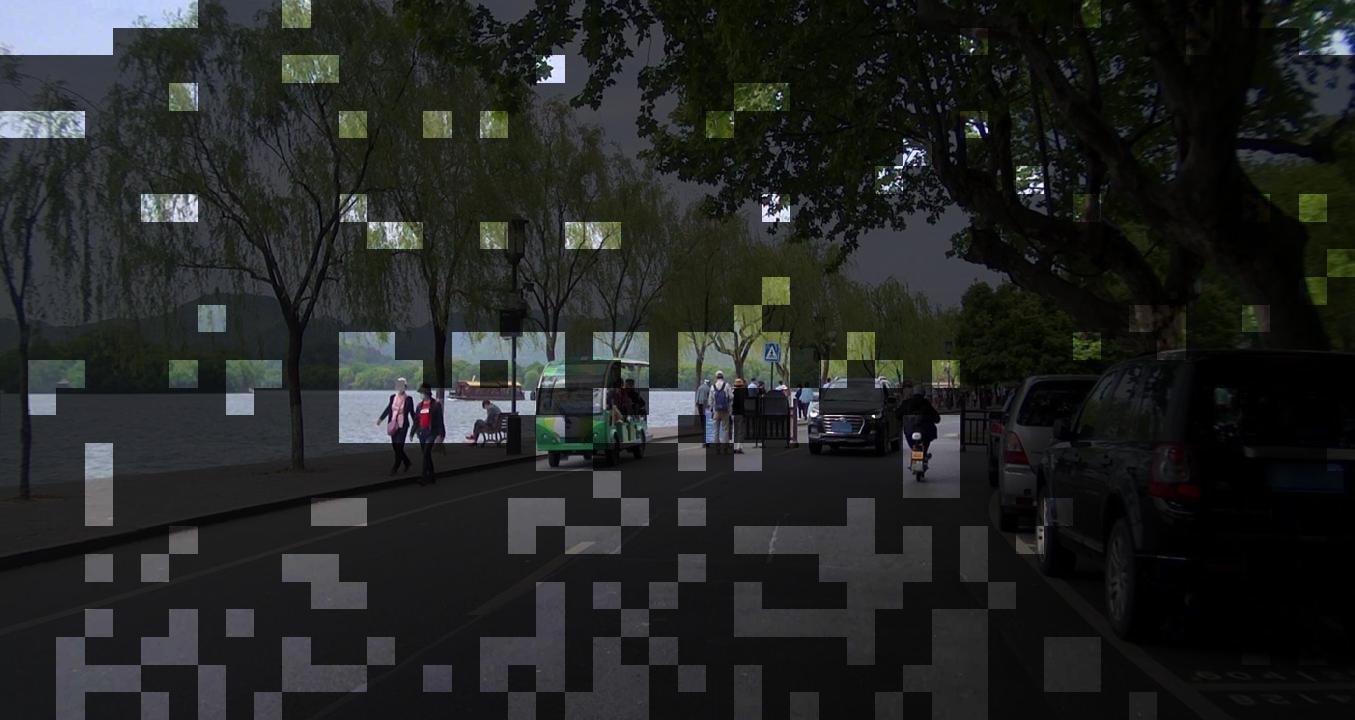} \end{subfigure}\hfill
    \begin{subfigure}[c]{\subfigwidth} \centering \includegraphics[width=\linewidth]{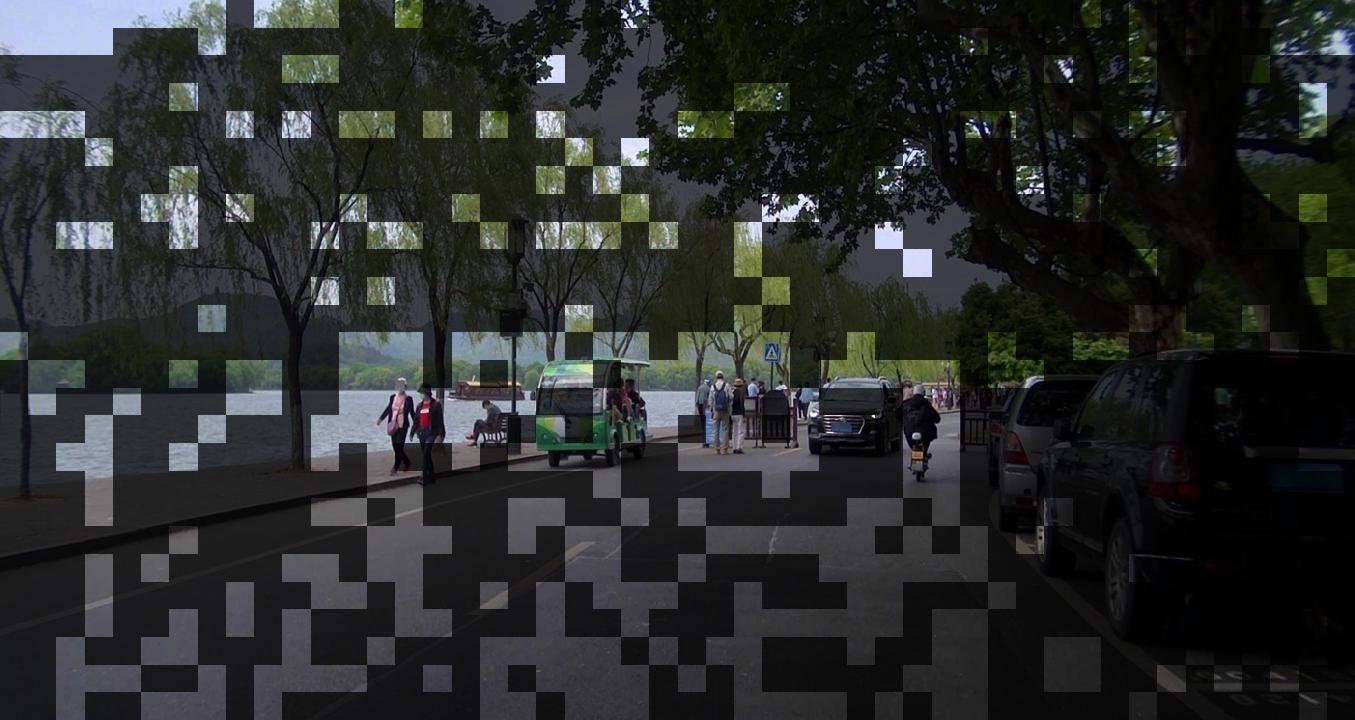} \end{subfigure}\hfill
    \begin{subfigure}[c]{\subfigwidth} \centering \includegraphics[width=\linewidth]{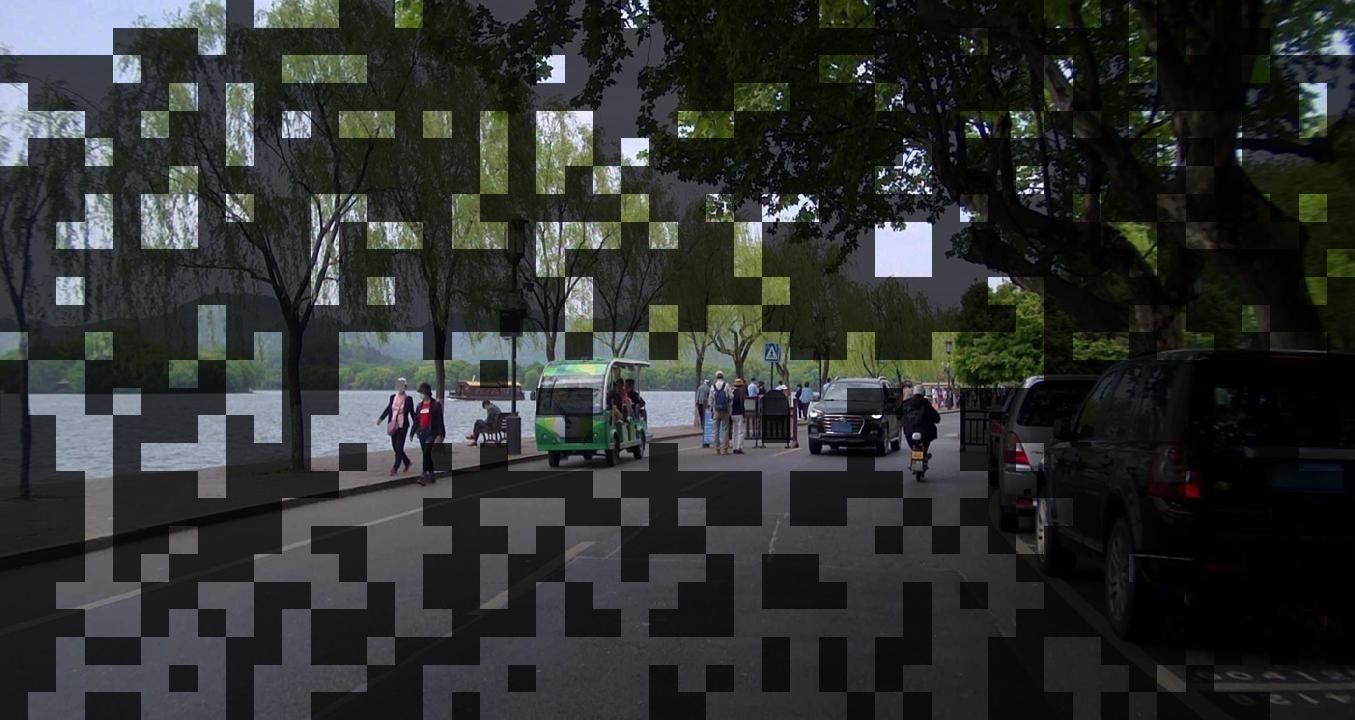} \end{subfigure}\hfill
    \begin{subfigure}[c]{\subfigwidth} \centering \includegraphics[width=\linewidth]{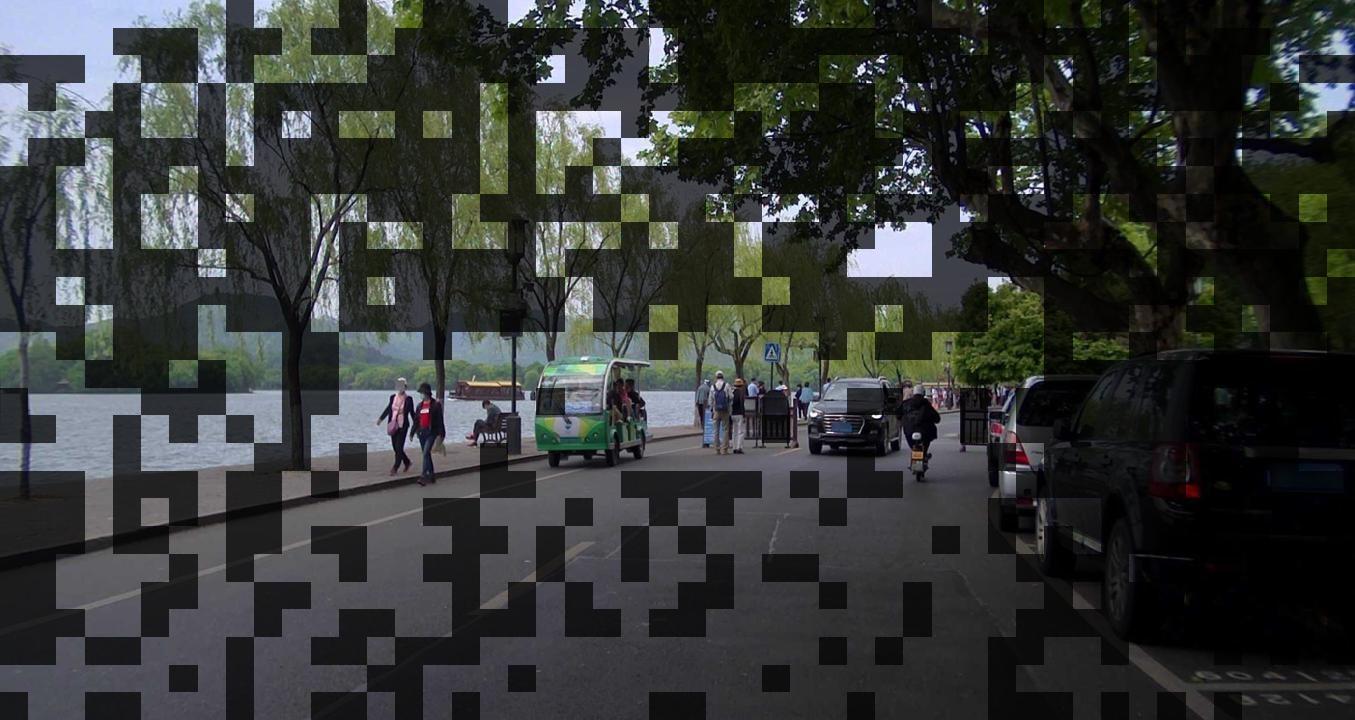} \end{subfigure}\\[1.5mm]

    \begin{subfigure}[c]{0.03\textwidth} \centering \rotatebox{90}{\small \textbf{\textit{110}}} \end{subfigure}\hfill
    \begin{subfigure}[c]{\subfigwidth} \centering \includegraphics[width=\linewidth]{figs/vis_hpe/input_images/image.jpg} \end{subfigure}\hfill
    \begin{subfigure}[c]{\subfigwidth} \centering \includegraphics[width=\linewidth]{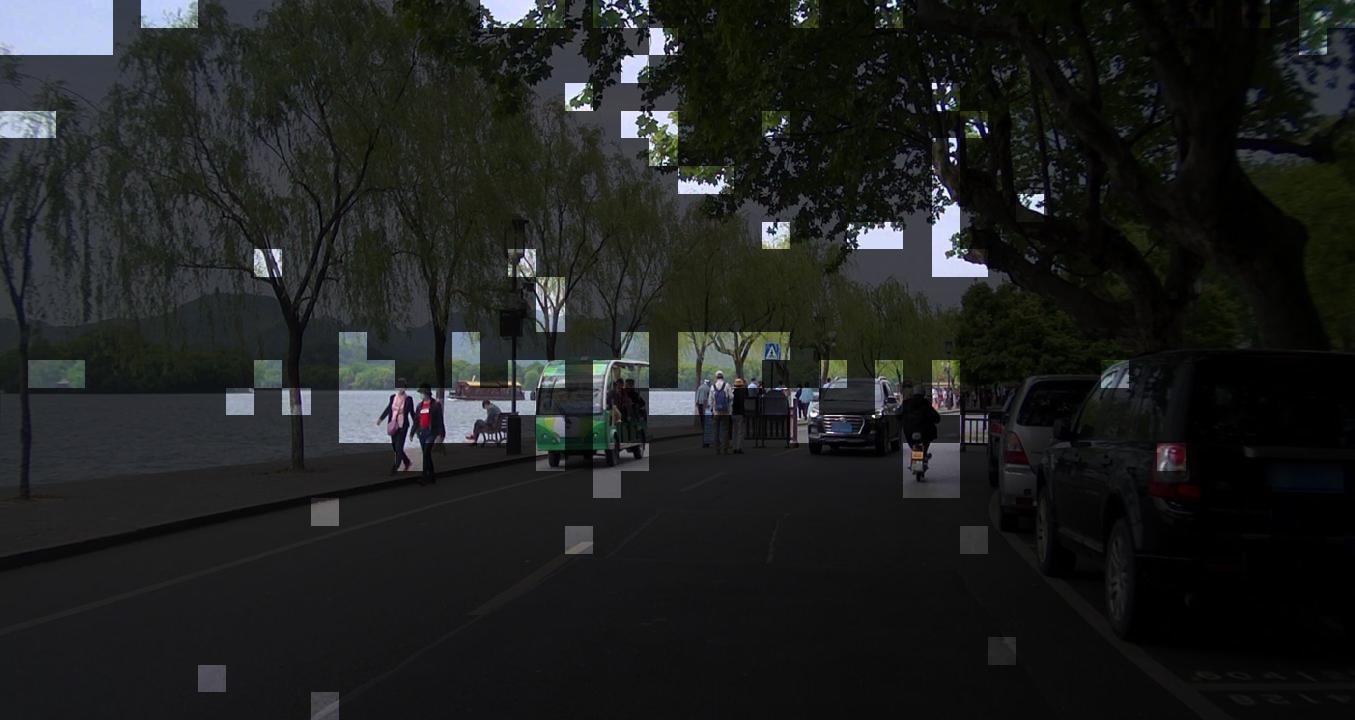} \end{subfigure}\hfill
    \begin{subfigure}[c]{\subfigwidth} \centering \includegraphics[width=\linewidth]{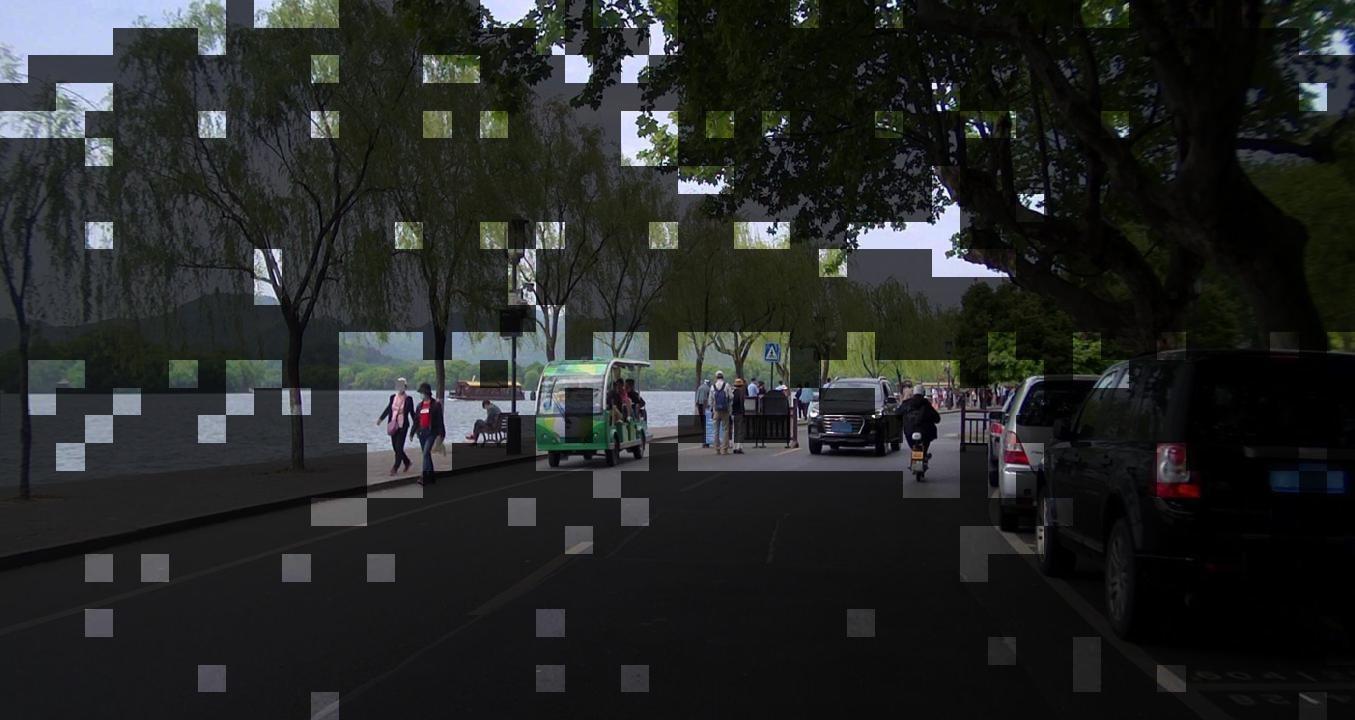} \end{subfigure}\hfill
    \begin{subfigure}[c]{\subfigwidth} \centering \includegraphics[width=\linewidth]{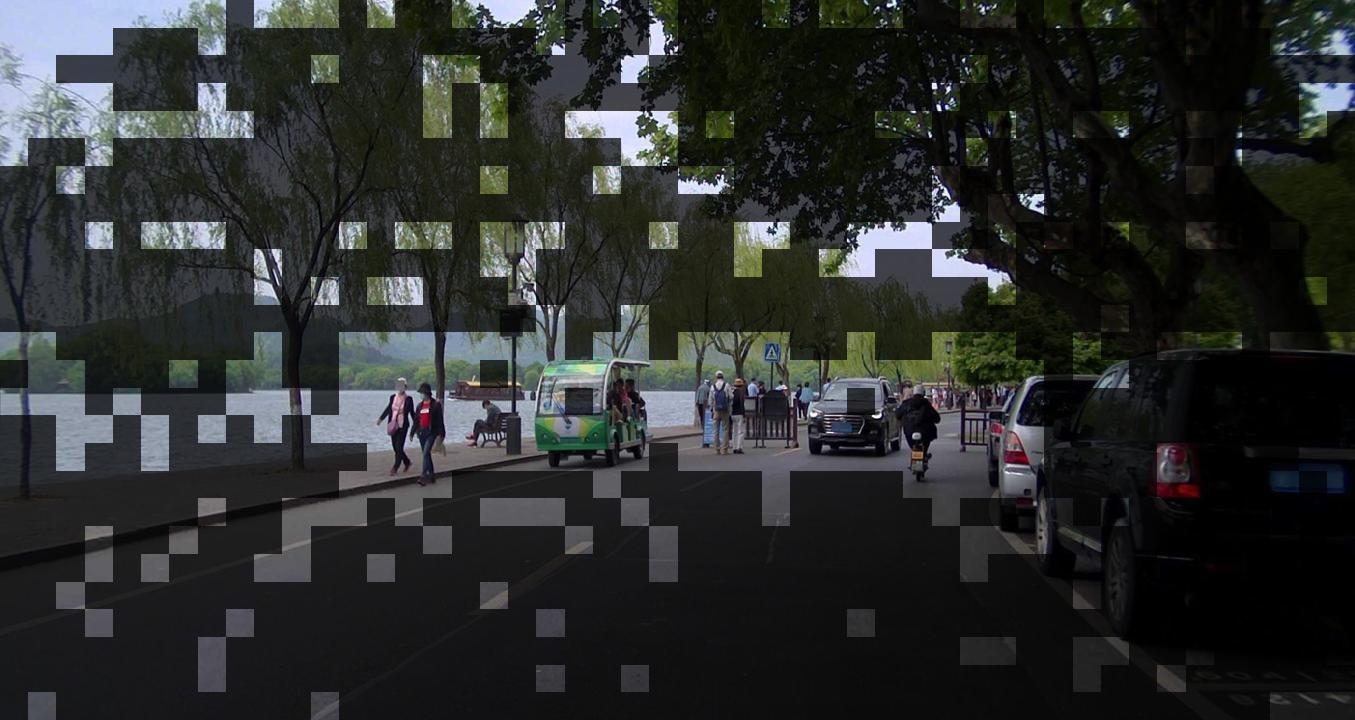} \end{subfigure}\hfill
    \begin{subfigure}[c]{\subfigwidth} \centering \includegraphics[width=\linewidth]{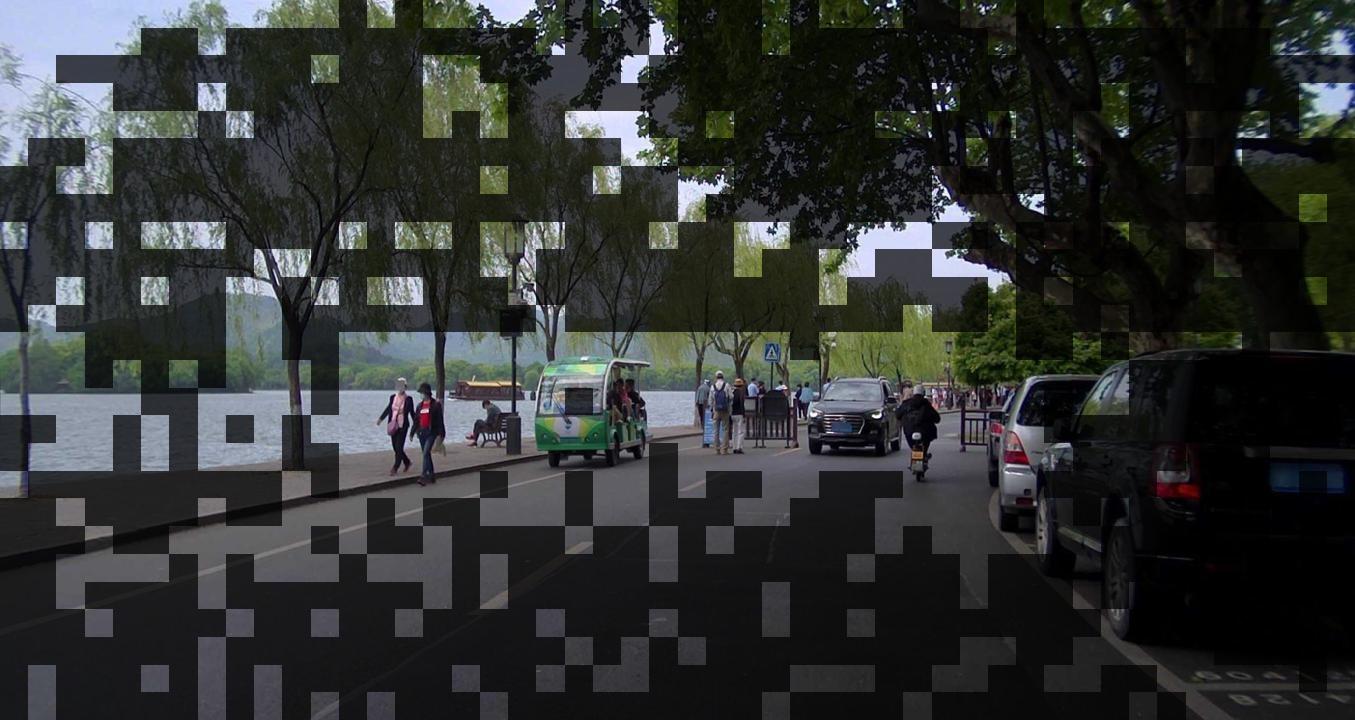} \end{subfigure}\hfill
    \begin{subfigure}[c]{\subfigwidth} \centering \includegraphics[width=\linewidth]{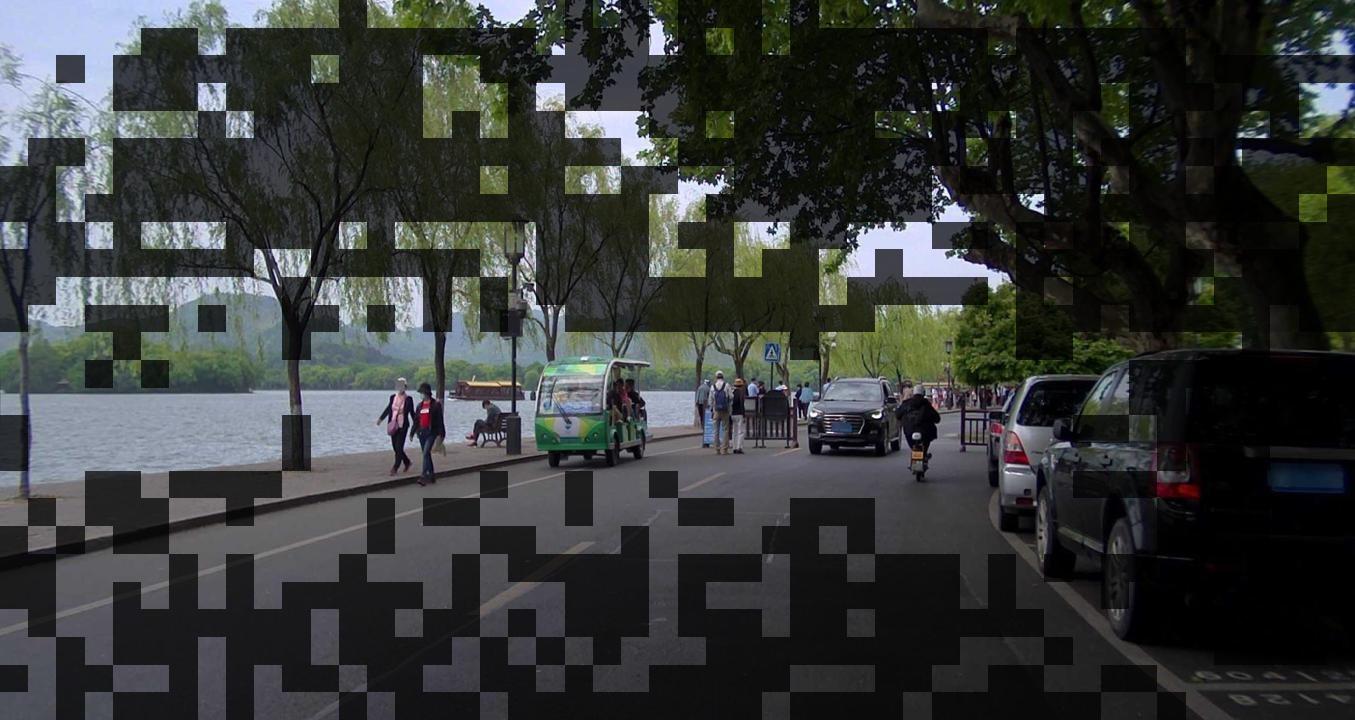} \end{subfigure}\\[1.5mm]

    \begin{subfigure}[c]{0.03\textwidth} \centering \rotatebox{90}{\small \textbf{\textit{ReTrace}}} \end{subfigure}\hfill
    \begin{subfigure}[c]{\subfigwidth} \centering \includegraphics[width=\linewidth]{figs/vis_hpe/input_images/image.jpg} \end{subfigure}\hfill
    \begin{subfigure}[c]{\subfigwidth} \centering \includegraphics[width=\linewidth]{figs/vis_hpe/vis_keep_0.1/image.jpg} \end{subfigure}\hfill
    \begin{subfigure}[c]{\subfigwidth} \centering \includegraphics[width=\linewidth]{figs/vis_hpe/vis_keep_0.2/image.jpg} \end{subfigure}\hfill
    \begin{subfigure}[c]{\subfigwidth} \centering \includegraphics[width=\linewidth]{figs/vis_hpe/vis_keep_0.3/image.jpg} \end{subfigure}\hfill
    \begin{subfigure}[c]{\subfigwidth} \centering \includegraphics[width=\linewidth]{figs/vis_hpe/vis_keep_0.4/image.jpg} \end{subfigure}\hfill
    \begin{subfigure}[c]{\subfigwidth} \centering \includegraphics[width=\linewidth]{figs/vis_hpe/vis_keep_0.5/image.jpg} \end{subfigure}
    \caption{\textbf{Qualitative comparison.}}
    \label{fig:ablation_vis_image}
\end{figure*}

We further compare the default allocation ($T_g:T_l:T_d = 1:1:2$) with several ablation variants. As shown in Fig.~\ref{fig:ablation_vis_2490} - Fig.~\ref{fig:ablation_vis_image}, isolated strategies show clear failure modes. Using only global saliency (1:0:0) over-concentrates tokens on salient foreground objects and loses background layout. Using only spatial diversity (0:0:1) spreads tokens widely but misses core semantic subjects. Partial combinations, such as removing spatial diversity (1:1:0) or local saliency (1:0:1), still lead to weaker coverage or less stable local detail preservation. The default 1:1:2 allocation offers a more balanced spatial-semantic buffer, which supports the later backtracking-aware injection step.

\section{More Qualitative Results}
\label{sec:more_qualitative}

\begin{figure*}[t]
\centering
\setlength{\tabcolsep}{10pt}
\scriptsize

\begin{tabular}{m{0.16\textwidth} m{0.74\textwidth}}
\includegraphics[width=\linewidth]{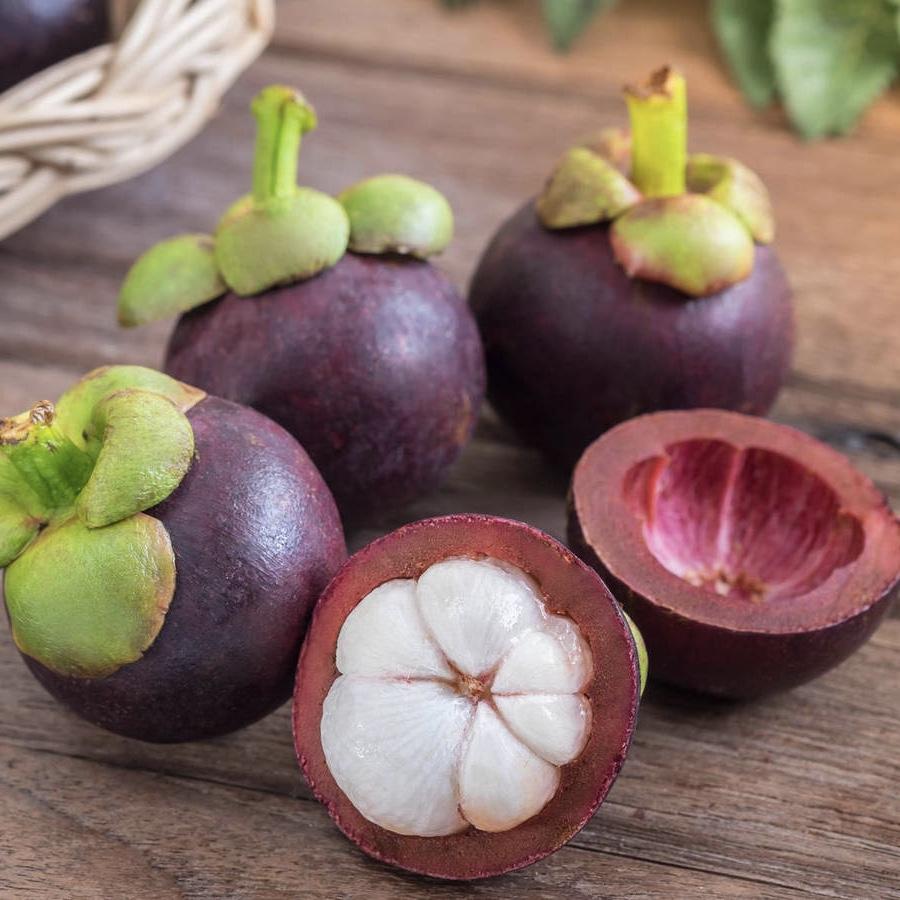}
&
\begin{tcolorbox}[
    enhanced,
    colback=gray!2,
    colframe=gray!40,
    boxrule=0.2pt,
    arc=2pt,
    left=4pt,right=4pt,top=4pt,bottom=4pt
]
\textbf{Q:} Describe this photo in detail. \\[4pt]
\textbf{A:} The image shows a group of mangosteen fruits, which are known for their distinctive appearance and flavor. The outer skin of the mangosteen is a deep purple color, while the interior is segmented with a white flesh that has a unique star-like pattern when cut open. The fruit in the foreground is cut in half, revealing its inner structure, which consists of several segments with a white, fleshy texture. The outer skin appears to be smooth and slightly bumpy, typical of mangosteens. \\[4pt]
\textbf{No Hallucination.}
\end{tcolorbox}
\\[6pt]

\includegraphics[width=\linewidth]{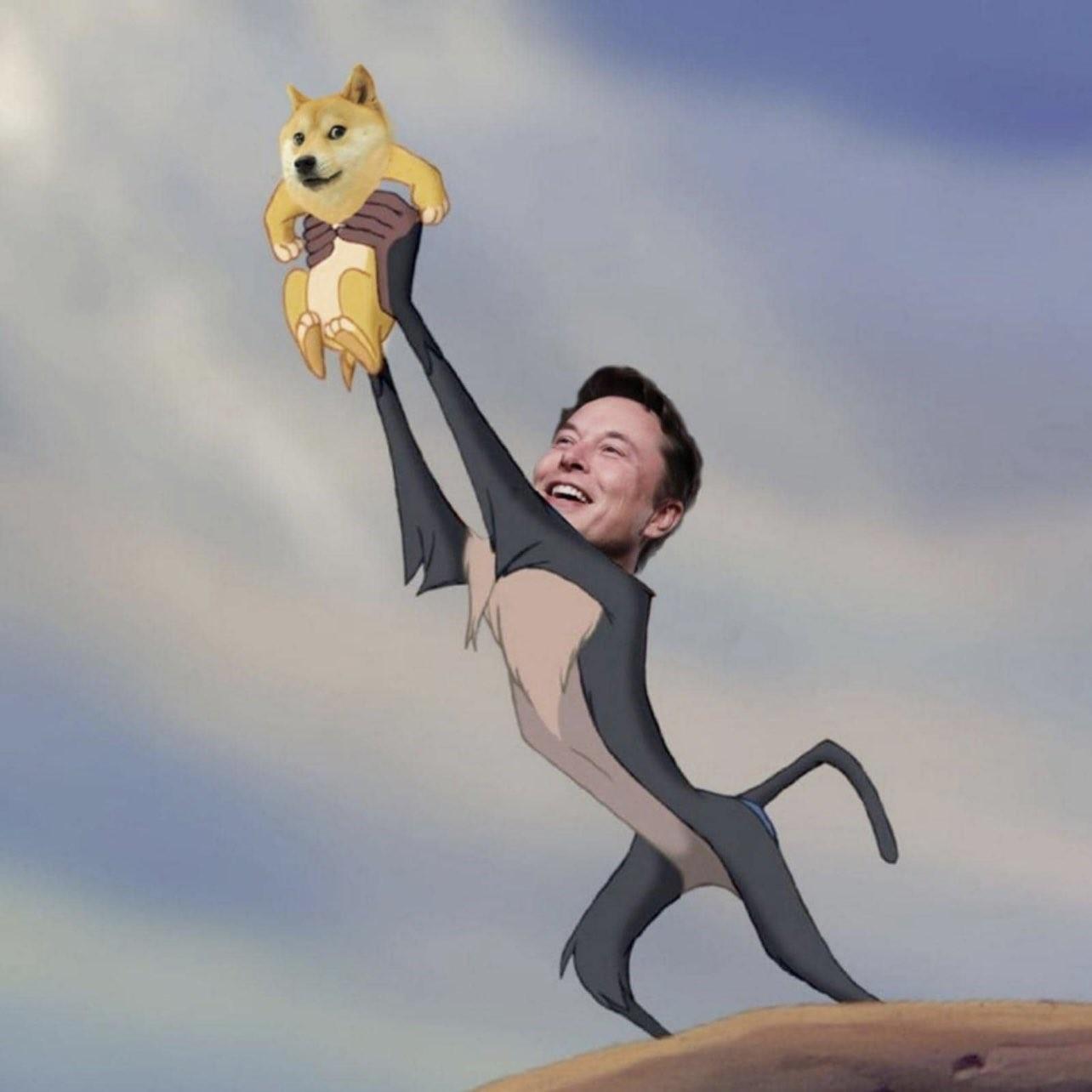}
&
\begin{tcolorbox}[
    enhanced,
    colback=gray!2,
    colframe=gray!40,
    boxrule=0.2pt,
    arc=2pt,
    left=4pt,right=4pt,top=4pt,bottom=4pt
]
\textbf{Q:} Which iconic movie scene is being parodied in the meme? \\[4pt]
\textbf{A:} The meme appears to parody the iconic presentation scene from Disney's "The Lion King," in which Rafiki lifts the young Simba into the air. In this version, Elon Musk is shown holding up a dog in a similar pose, creating a humorous visual reference to the original scene. \\[4pt]
\textbf{No Hallucination.}
\end{tcolorbox}
\\[6pt]

\includegraphics[width=\linewidth]{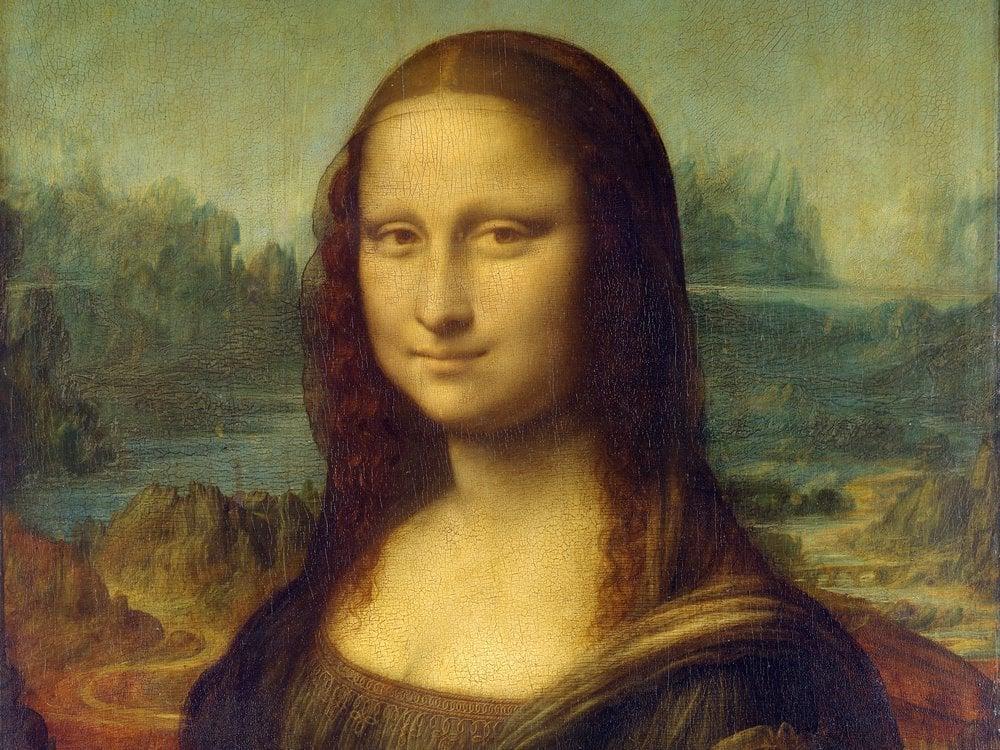}
&
\begin{tcolorbox}[
    enhanced,
    colback=gray!2,
    colframe=gray!40,
    boxrule=0.2pt,
    arc=2pt,
    left=4pt,right=4pt,top=4pt,bottom=4pt
]
\textbf{Q:} Describe this painting in detail. \\[4pt]
\textbf{A:} This is a famous painting titled "Mona Lisa" by Leonardo da Vinci, one of the most recognizable and enigmatic works of art in the world. The painting depicts a woman who is believed to be Mona Lisa di Giocondo, though her identity remains uncertain. The subject is shown in a three-quarter view, with her head slightly tilted to the right and her gaze directed slightly upwards and to the left, creating an air of mystery and intrigue. The woman's expression is often described as enigmatic, with a slight smile that some interpret as a hint of amusement or even a knowing look. \\[4pt]
\textbf{No Hallucination.}
\end{tcolorbox}
\\[6pt]

\includegraphics[width=\linewidth]{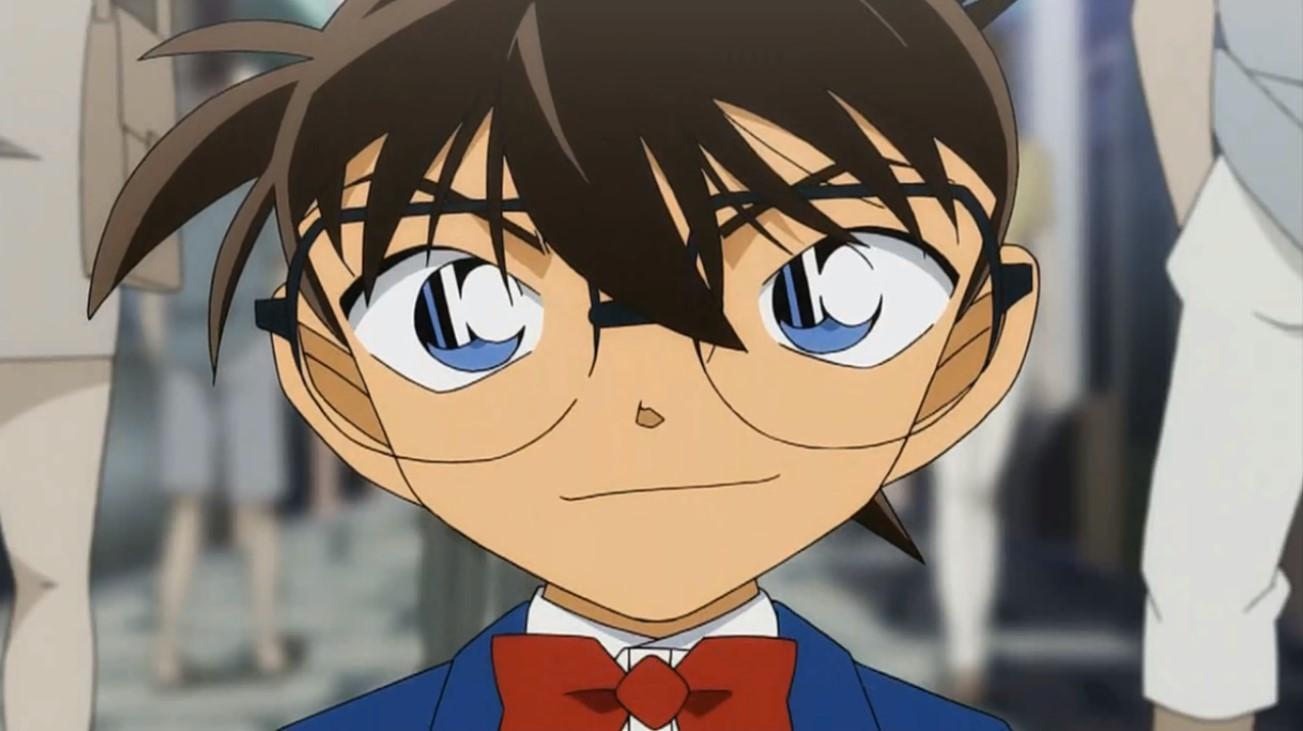}
&
\begin{tcolorbox}[
    enhanced,
    colback=gray!2,
    colframe=gray!40,
    boxrule=0.2pt,
    arc=2pt,
    left=4pt,right=4pt,top=4pt,bottom=4pt
]
\textbf{Q:} What is the name of the character in the image? \\[4pt]
\textbf{A:} The character in the image is Conan Edogawa, also known as Shinichi Kudo, from the anime series "Detective Conan" (also known as "Case Closed"). \\[4pt]
\textbf{No Hallucination.}
\end{tcolorbox}
\\[6pt]

\end{tabular}

\caption{\textbf{More qualitative results on retain 22.2\% visual tokens.} Using LLaVA-bench, we have demonstrated the effectiveness of our method on OpenQA, where the model's hallucination-free responses fully showcase the validity of our method.}
\label{fig:morequal}
\end{figure*}

We provide additional qualitative examples from LLaVA-Bench to evaluate the generation quality in open-ended scenarios. Aggressive token reduction often discards critical visual details, leading to object omissions or fabricated hallucinations in the generated text. As illustrated in Fig.~\ref{fig:morequal}, ReTrace effectively mitigates this challenge by preserving spatial diversity and mid-layer structural features, thereby preventing the degradation of visual grounding. The generated responses demonstrate accurate comprehension of complex scenes without fabricating ungrounded contents. These results indicate that our method achieves significant inference acceleration while maintaining a text-visual consistency strictly comparable to the unpruned baseline, without exacerbating hallucination issues.

\section{Benchmark Details}
\label{sec:bench}
We provide a detailed description of benchmarks here.

ScienceQA \cite{sqa} is a benchmark focused on multimodal question answering in the scientific domain. Code use MIT License, Datasets use CC BY-NC-SA 4.0.

POPE \cite{pope} is designed to quantitatively evaluate the object hallucination phenomenon in MLLMs. The testing task is strictly formulated as a binary "Yes/No" classification, forcing the model to determine whether a specific object exists in the image. In terms of scale, thousands of test samples are typically constructed for different source datasets. The samples are generated using three negative sampling strategies: random, popular, and adversarial, to accurately detect whether the model tends to fabricate non-existent visual content based on linguistic priors. MIT License.

MME \cite{mme} is a comprehensive benchmark for evaluating the perception and cognition abilities of MLLMs. The task format is uniformly designed as concise binary "Yes/No" questions to minimize the interference of prompt engineering. The benchmark includes 14 independent subtasks, categorized into perception (covering object detection, OCR, color recognition, etc.) and cognition (covering commonsense reasoning, code reasoning, mathematical calculation, etc). To prevent training data leakage, its instruction-answer pairs are constructed and verified entirely through rigorous manual annotation. Academic research only.

MMBench \cite{mmbench} aims to evaluate the all-around capabilities of MLLMs at a fine-grained level. The task format consists of single-choice questions, covering 20 specific visual and cognitive capability dimensions, such as object localization, spatial relations, and social reasoning. Apache-2.0 license.

MMBench-CN \cite{mmbench} is the Chinese branch of the MMBench evaluation framework. It is specifically used to independently test the cross-modal comprehension and reasoning performance of multimodal models in a Chinese linguistic context. Apache-2.0 license.

MMStar \cite{mmstar} is an evaluation benchmark specifically designed to test the visual dependency of MLLMs. The task format comprises multiple-choice questions, containing 1,500 high-quality samples manually filtered to ensure that the questions cannot be answered using purely text-based instructions. The test categories cover 6 core capabilities and 18 detailed evaluation dimensions. The benchmark introduces two data metrics: multi-modal gain and multi-modal leakage, which are used to quantitatively detect data leakage issues during the pre-training phase of MLLMs. Academic research only.

MMERealWorld \cite{mmerealworld} is an evaluation benchmark targeting high-resolution real-world scenarios. The primary evaluation formats are multiple-choice and objective question-answering, with testing scenarios covering 43 complex real-world categories to challenge the limits of fine-grained visual recognition. The dataset contains 13,366 high-resolution captured images equipped with 29,429 manually annotated question-answer pairs by professionals, focusing on testing the parsing performance of MLLMs in actual, complex, and high-definition environments. MME-RealWorld is only used for academic research.

MMERealWorld-CN \cite{mmerealworld} is the Chinese branch of MMERealWorld. MME-RealWorld-CN is only used for academic research.

\section{Model Card}
\label{sec:model}
We provide a detailed description of MLLMs here.

Qwen2.5-VL \cite{bai2025qwen2.5} adopts dynamic-resolution visual processing with a native Vision Transformer and window attention, enabling it to handle images of varying sizes and aspect ratios. It also supports visual grounding by producing points or bounding boxes for object localization. The model is further optimized for structured document understanding, including forms, tables, charts, diagrams, and layouts, and extends to long-video comprehension with second-level temporal localization. Apache-2.0 License.

InternVL3 \cite{internvl3} follows the ``ViT-MLP-LLM'' paradigm and introduces a native multimodal pre-training strategy, where multimodal data and large-scale textual corpora are jointly optimized within a single pre-training stage. It incorporates Variable Visual Position Encoding (V2PE), which uses smaller and more flexible position increments for visual tokens to better support long multimodal contexts. MIT License.
\section{AI Use}
\label{sec:ai}
During the preparation of this work, the authors used AI tools solely to improve the linguistic expression and grammatical correctness of the manuscript. After using this service, the authors reviewed and edited the content as needed and take full responsibility for the technical accuracy and original contributions of the publication.

\end{document}